\documentclass[acmtog,screen]{acmart}

\usepackage{xspace}
\usepackage{tabularray}
\usepackage{cleveref}

\usepackage{soul,color,xcolor}

\definecolor{gainsboro}{rgb}{0.86, 0.86, 0.86}
\definecolor{myblue}{rgb}{0,0.6902,0.94118}

\newcommand{\matte}{\boldsymbol{\alpha}}%
\newcommand{\latent}{\mathbf{z}}%

\def\Ours{{GVM}\xspace}

\AtBeginDocument{%
  }

\setcopyright{acmlicensed}
\copyrightyear{2025}
\acmYear{2025}
\acmConference[SIGGRAPH Conference Papers '25]{Special Interest Group on Computer Graphics and Interactive Techniques Conference Conference Papers }{August 10--14, 2025}{Vancouver, BC, Canada}
\acmBooktitle{Special Interest Group on Computer Graphics and Interactive Techniques Conference Conference Papers (SIGGRAPH Conference Papers '25), August 10--14, 2025, Vancouver, BC, Canada}
\acmDOI{10.1145/3721238.3730642}
\acmISBN{979-8-4007-1540-2/2025/08}

\acmSubmissionID{411}

\usepackage{booktabs}
\usepackage{longtable}
\usepackage{tocloft}
\usepackage{tabularray}
\usepackage{multicol}
\usepackage{multirow}
\usepackage{graphicx}
\usepackage{amsmath}

\begin{document}

\title[Generative Video Matting]{Generative Video Matting}
\author{Yongtao Ge}
\email{yongtaoo.ge@gmail.com}
\orcid{0000-0003-1265-3204}
\affiliation{%
  \institution{The University of Adelaide}
  \country{Australia}
}
\affiliation{%
  \institution{Zhejiang University}
  \country{China}
}

\author{Kangyang Xie}
\email{felixky@zju.edu.cn}
\orcid{0009-0003-1048-1274}
\affiliation{%
  \institution{Zhejiang University}
  \country{China}
}

\author{Guangkai Xu}
\email{kuangkaixu@zju.edu.cn}
\orcid{0000-0002-1209-9533}
\affiliation{%
  \institution{Zhejiang University}
  \country{China}
}

\author{Mingyu Liu}
\email{mingyuliu@zju.edu.cn}
\orcid{0009-0006-1379-2846}
\affiliation{%
  \institution{Zhejiang University}
  \country{China}
}

\author{Li Ke}
\email{keli.kl@alibaba-inc.com}
\orcid{0000-0002-2350-8224}
\affiliation{%
  \institution{Alibaba Group}
  \country{China}
}

\author{Longtao Huang}
\email{kaiyang.hlt@alibaba-inc.com }
\orcid{0000-0002-0517-1592}
\affiliation{%
  \institution{Alibaba Group}
  \country{China}
}

\author{Hui Xue}
\email{hui.xueh@alibaba-inc.com}
\orcid{0000-0002-2093-2839}
\affiliation{%
  \institution{Alibaba Group}
  \country{China}
}

\author{Hao Chen}
\email{haochen.cad@zju.edu.cn}
\orcid{0000-0003-4417-614X}
\affiliation{%
  \institution{Zhejiang University}
  \country{China}
}

\author{Chunhua Shen}
\email{chunhuashen@zju.edu.cn}
\orcid{0000-0002-8648-8718}
\affiliation{%
  \institution{Zhejiang University of Technology and Zhejiang University}
  \country{China}
}

\renewcommand{\shortauthors}{Ge et al.}
\begin{abstract}
Video matting has traditionally been limited by the lack of high-quality ground-truth data. Most existing video matting datasets provide only human-annotated imperfect alpha and foreground annotations, which must be composited to background images or videos during the training stage. Thus, the generalization capability of previous methods in real-world scenarios is typically poor.
In this work, we propose to solve the problem from two perspectives. First, we emphasize the importance of large-scale pre-training by pursuing diverse synthetic and pseudo-labeled segmentation datasets. We also develop a scalable synthetic data generation pipeline that can render diverse human bodies and fine-grained hairs, yielding around 200 video clips with a 3-second duration for fine-tuning. 
Second, we introduce a novel video matting approach that can effectively leverage the rich priors from pre-trained video diffusion models. This architecture offers two key advantages. First, strong priors play a critical role in bridging the domain gap between synthetic and real-world scenes.
Second, unlike most existing methods that process video matting frame-by-frame and use an independent decoder to aggregate temporal information, our model is inherently designed for video, ensuring strong temporal consistency. We provide a comprehensive quantitative evaluation across three benchmark datasets, demonstrating our approach's superior performance, and present comprehensive qualitative results in diverse real-world scenes, illustrating the strong generalization capability of our method. The code is available at \url{https://github.com/aim-uofa/GVM}.

\end{abstract}


\begin{teaserfigure}
\small
\centering
\newlength{\teaserw}
\setlength{\teaserw}{0.245\linewidth}
\newcommand{\teaserrowup}[5]{
 {\includegraphics[trim={#2 #3 #4 #5}, clip,width=\teaserw]{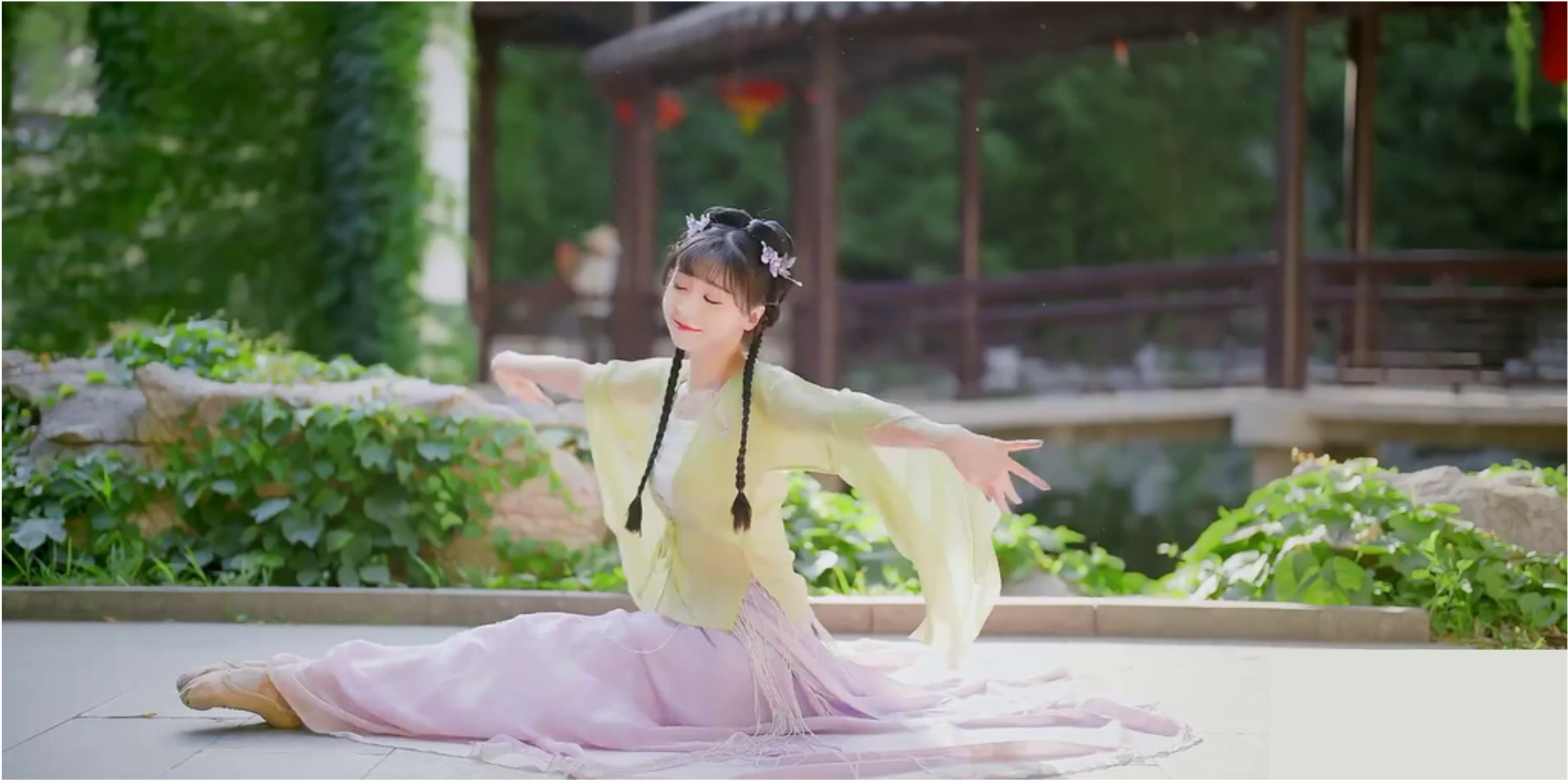}} & 
 {\includegraphics[trim={#2 #3 #4 #5}, clip,width=\teaserw]{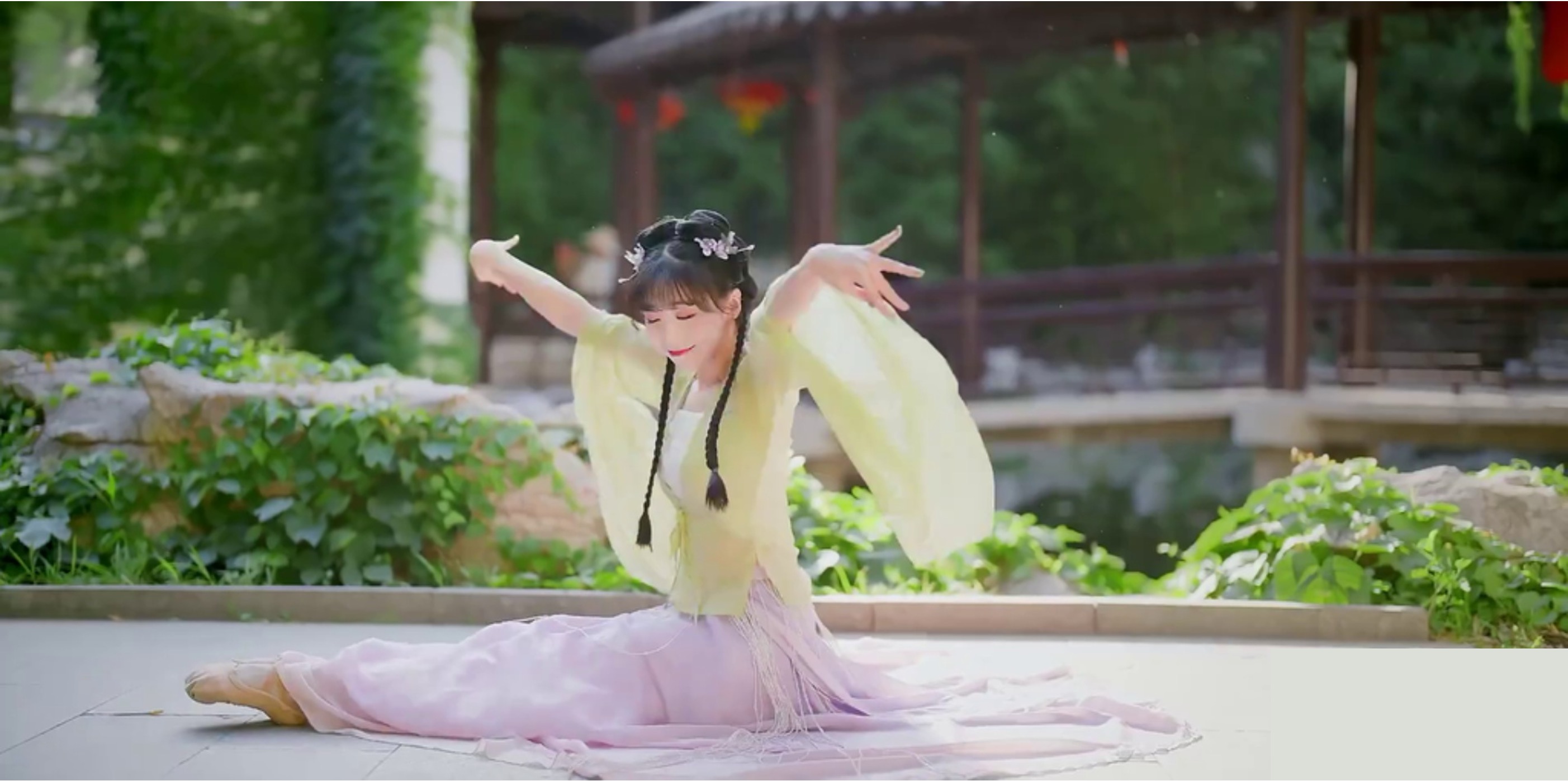}} & 
  {\includegraphics[trim={#2 #3 #4 #5}, clip,width=\teaserw]{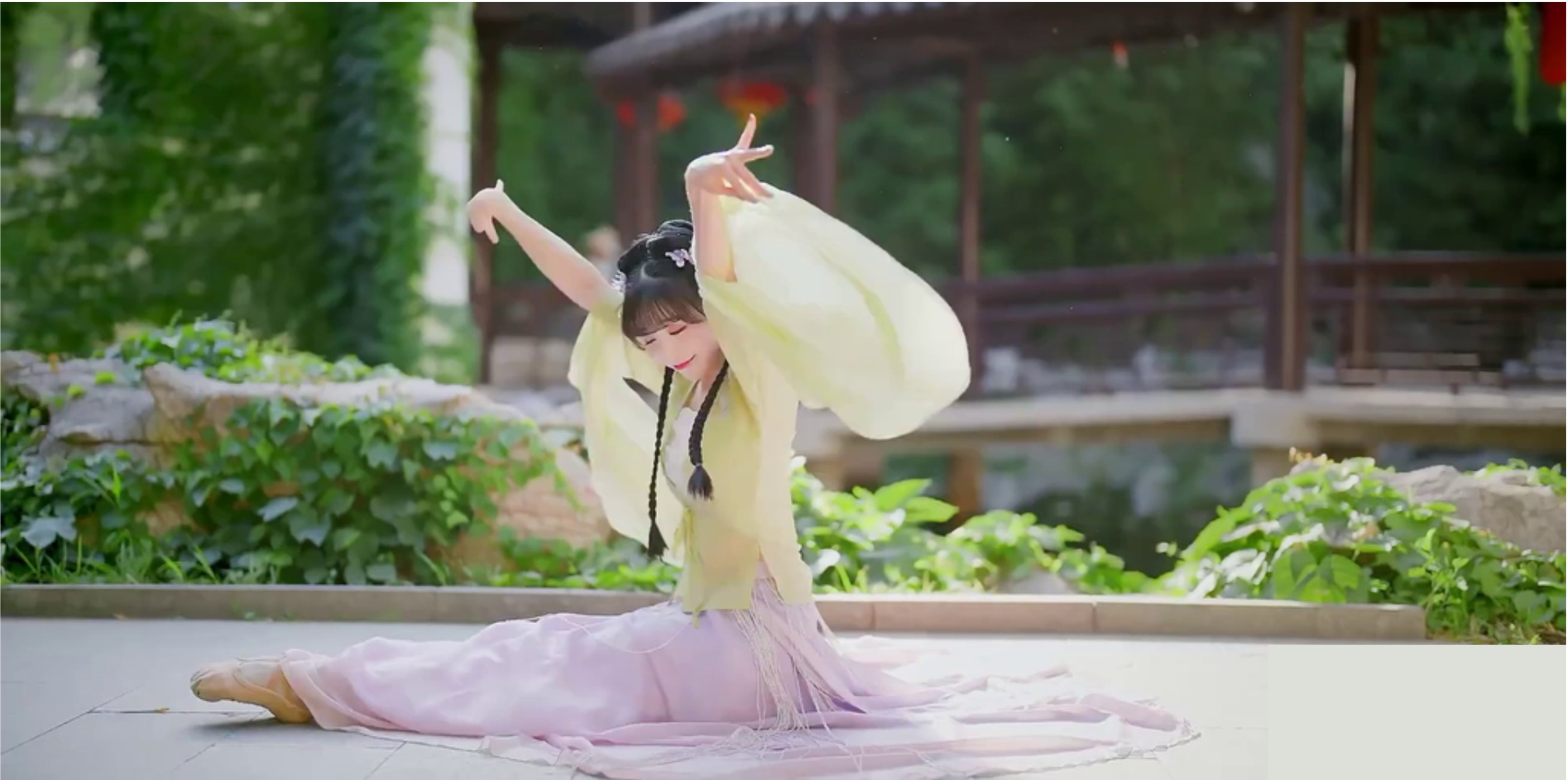}} &
  {\includegraphics[trim={#2 #3 #4 #5}, clip,width=\teaserw]{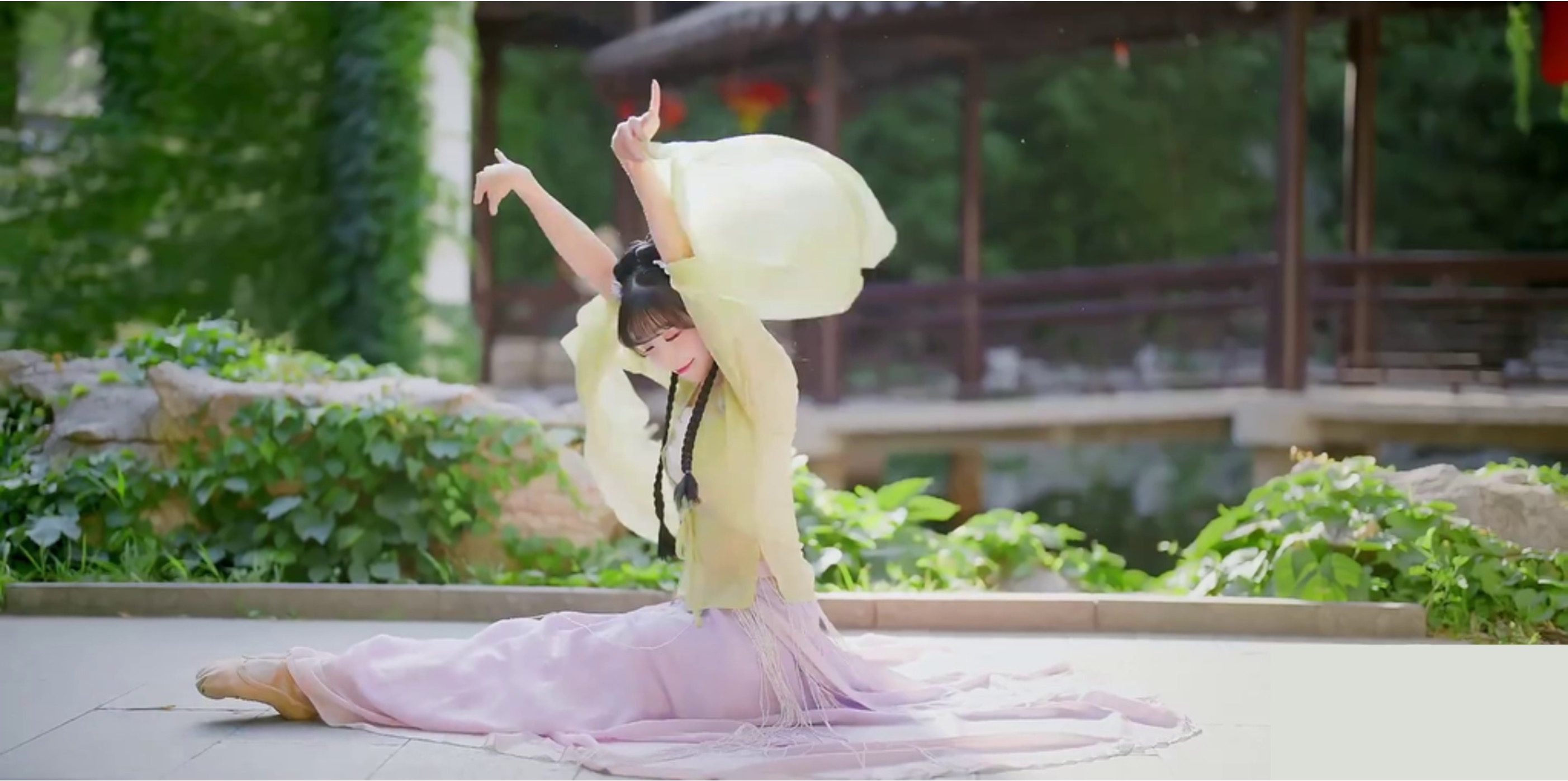}}
}
\newcommand{\teaserrowmiddle}[5]{
{\includegraphics[trim={#2 #3 #4 #5}, clip,width=\teaserw]{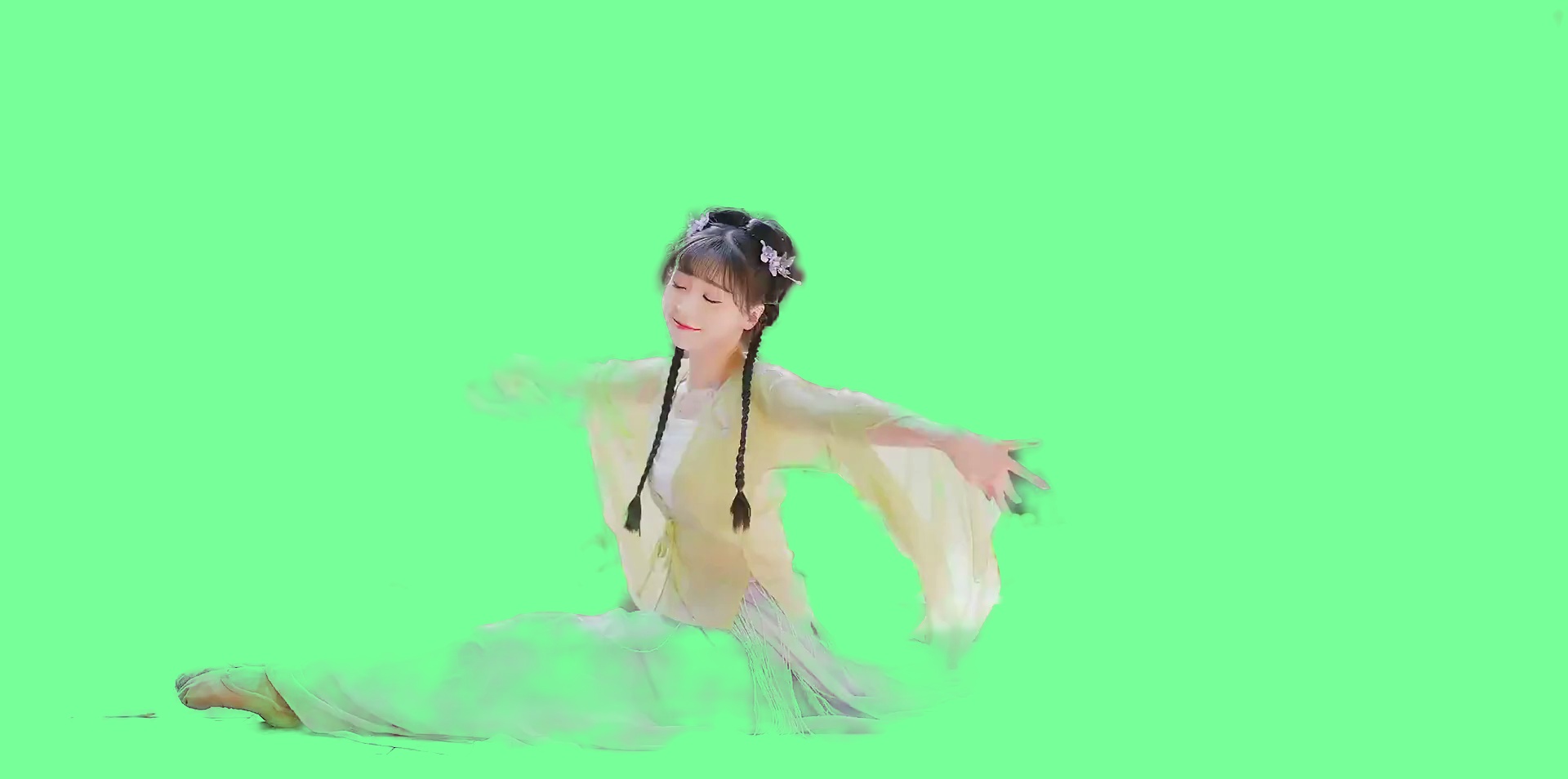}} &
{\includegraphics[trim={#2 #3 #4 #5}, clip,width=\teaserw]{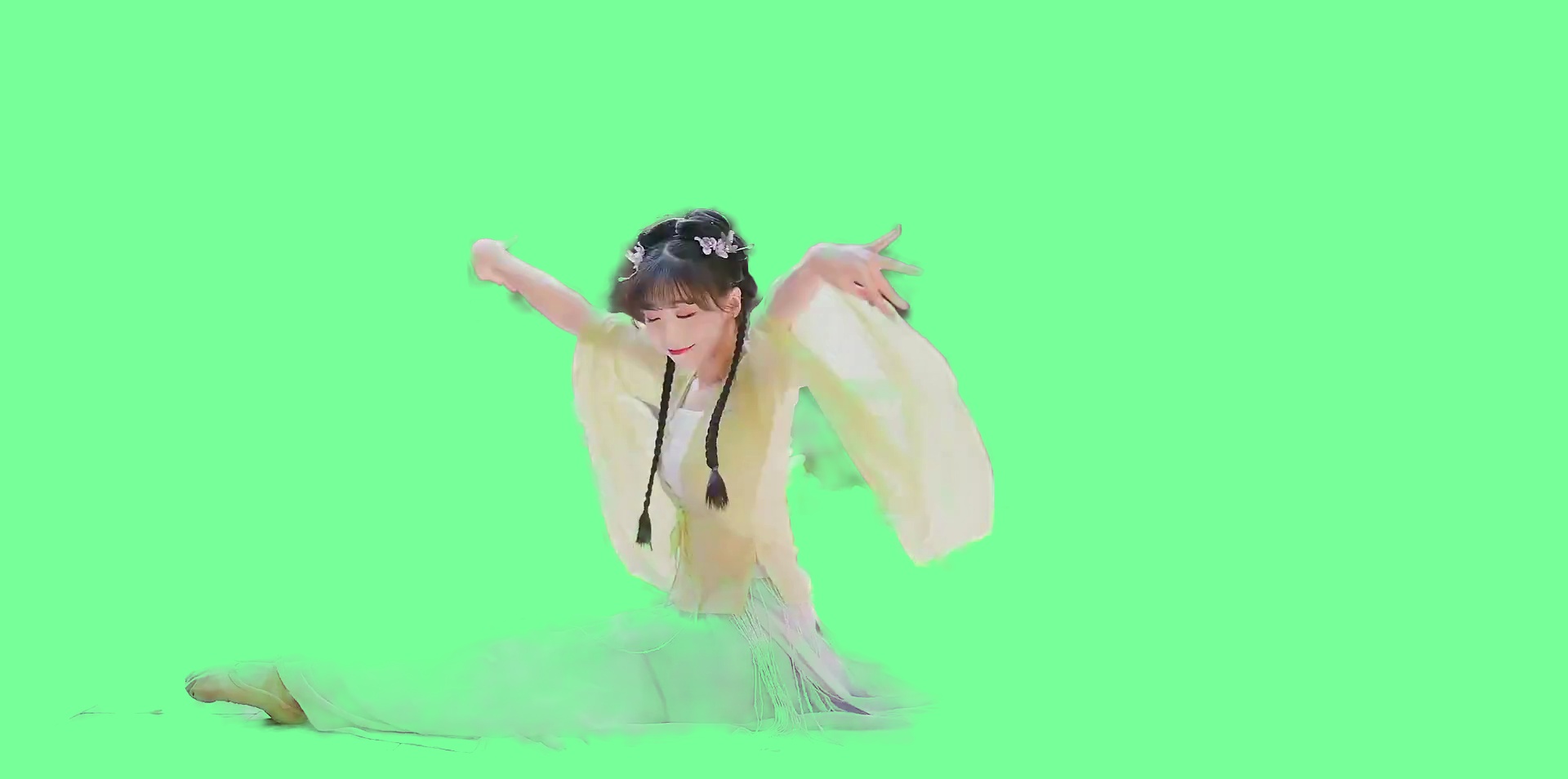}} &
{\includegraphics[trim={#2 #3 #4 #5}, clip,width=\teaserw]{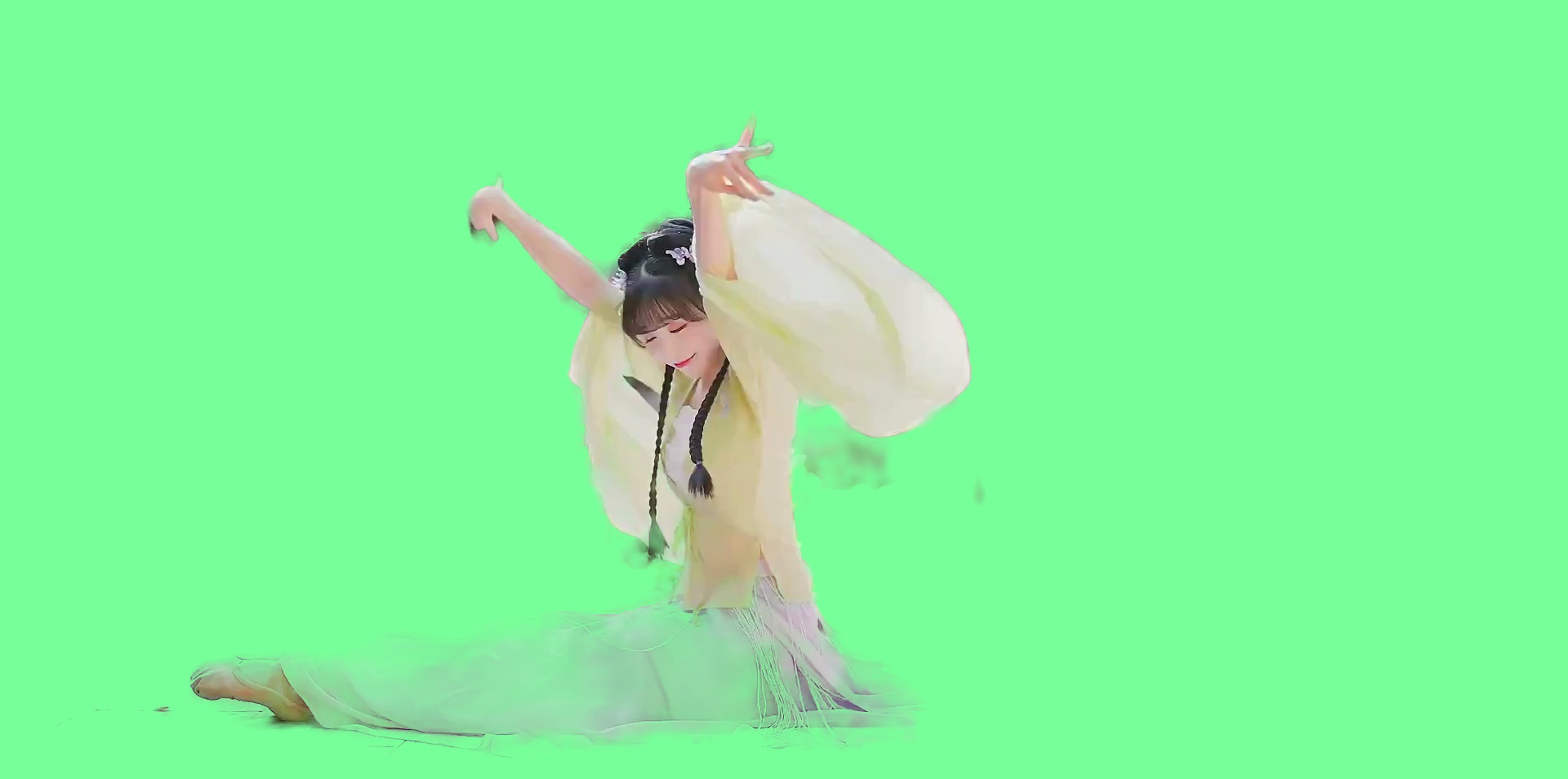}} &
{\includegraphics[trim={#2 #3 #4 #5}, clip,width=\teaserw]{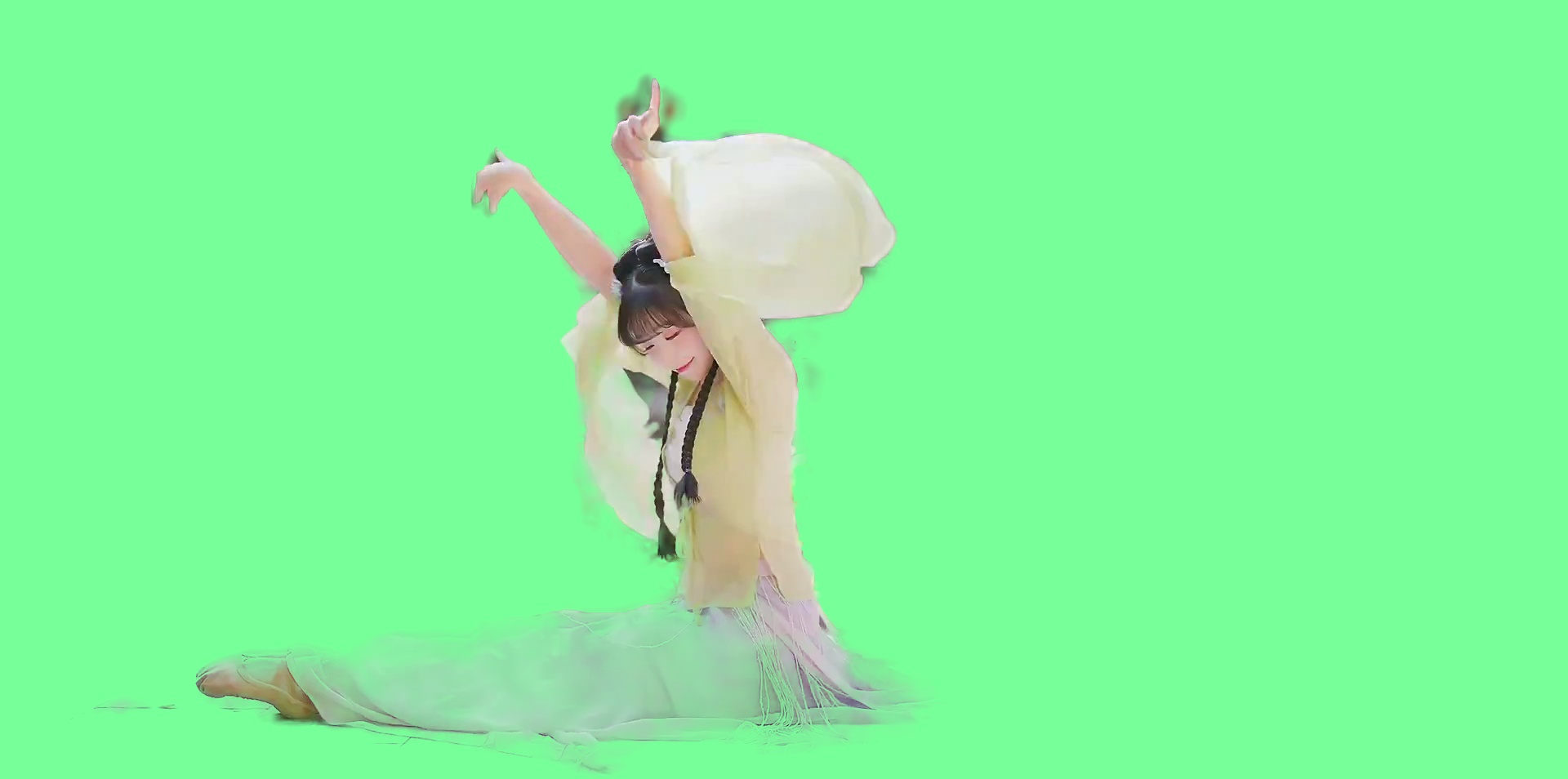}}
}
\newcommand{\teaserrowbottom}[5]{
{\includegraphics[trim={#2 #3 #4 #5}, clip,width=\teaserw]{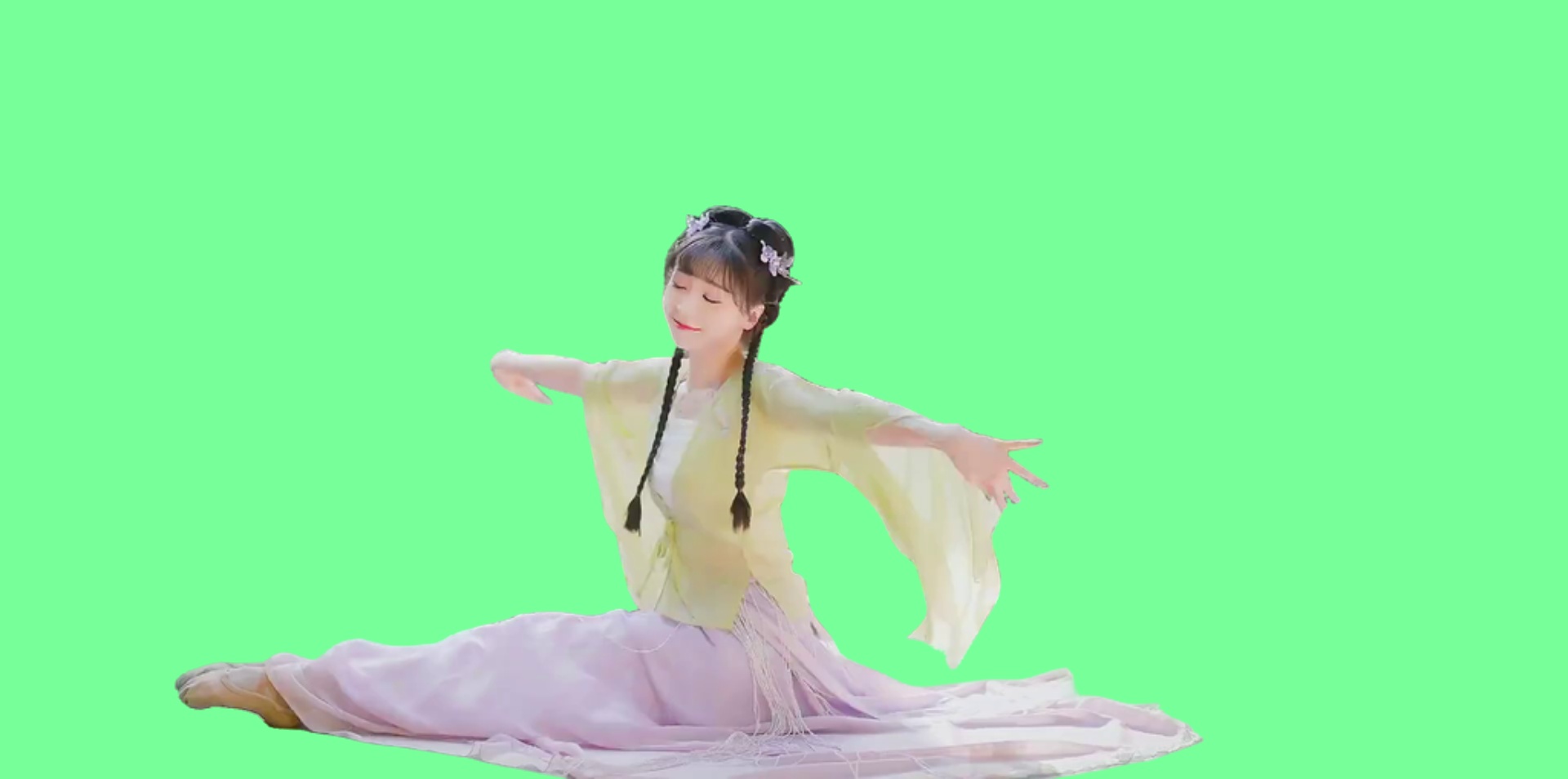}} &
{\includegraphics[trim={#2 #3 #4 #5}, clip,width=\teaserw]{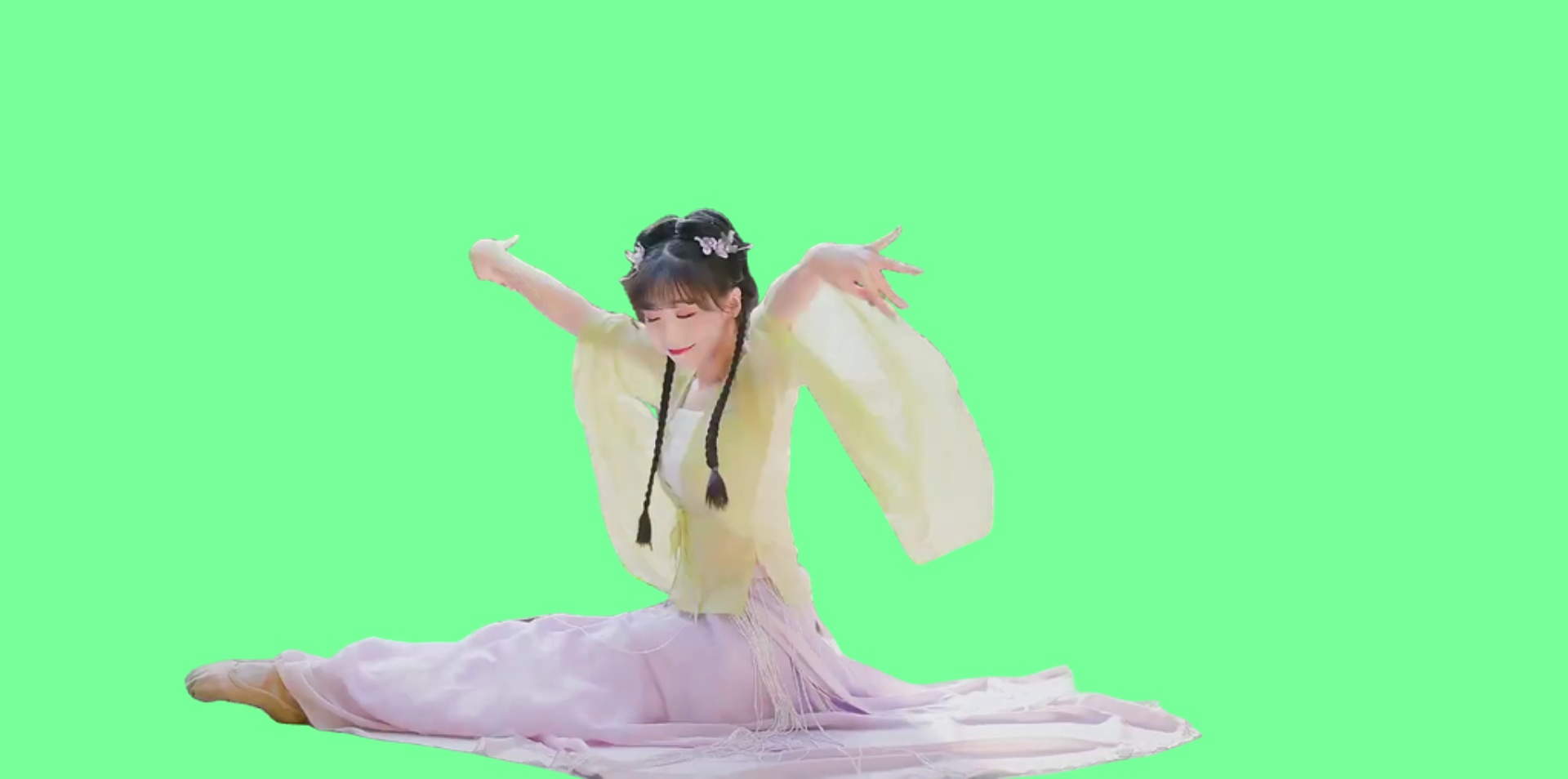}} &
{\includegraphics[trim={#2 #3 #4 #5}, clip,width=\teaserw]{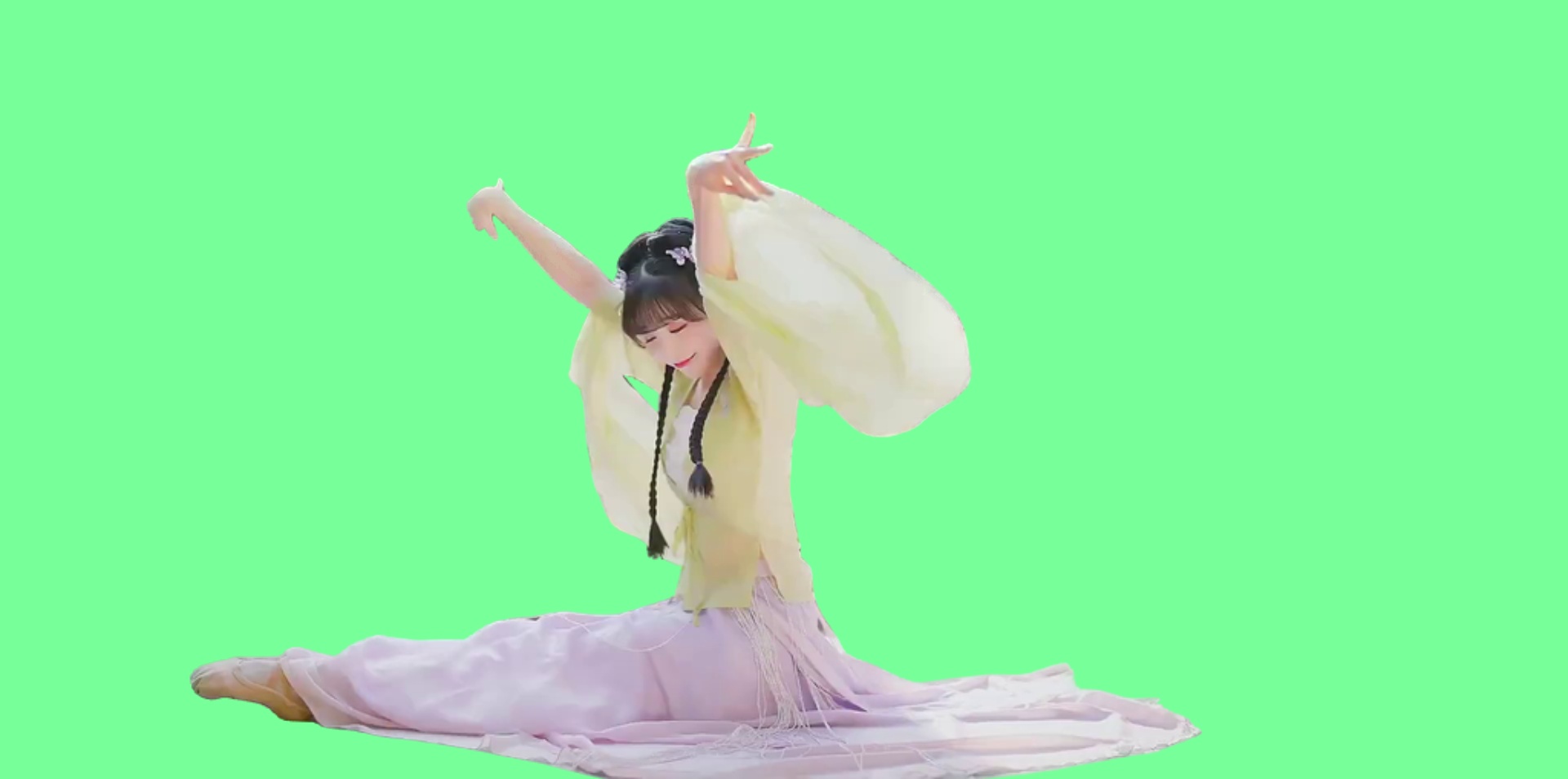}} &
{\includegraphics[trim={#2 #3 #4 #5}, clip,width=\teaserw]{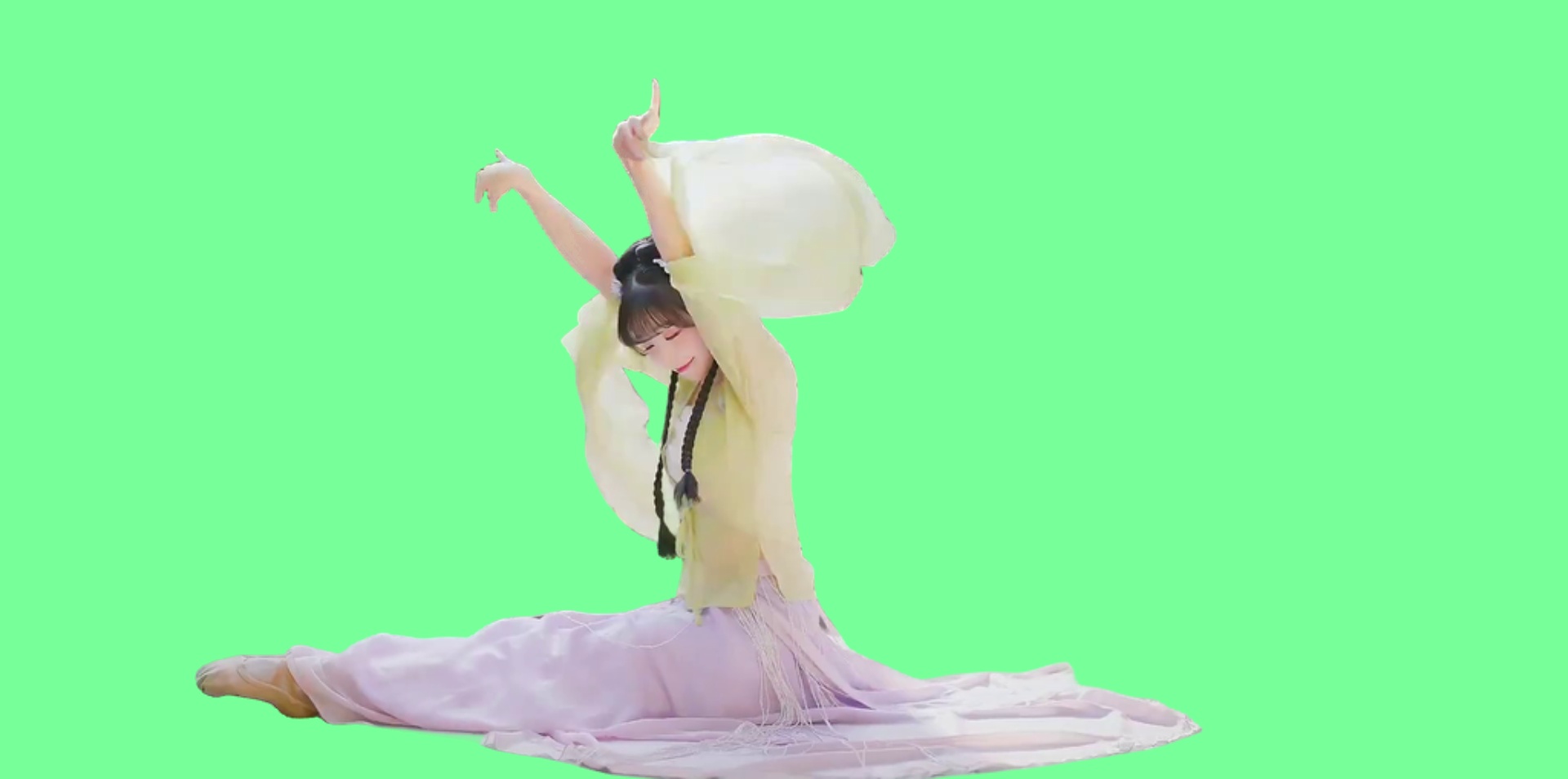}}
}
\small
\begin{tabular}
    {   
        @{\hspace{0mm}}c@{\hspace{0.8mm}} 
        @{\hspace{0mm}}c@{\hspace{0.8mm}}
        @{\hspace{0mm}}c@{\hspace{0.8mm}}
        @{\hspace{0mm}}c@{\hspace{0.8mm}} 
    }
    \centering
    \teaserrowup{input_patch}{0}{0}{0}{0}\\
    \teaserrowmiddle{input_patch}{0}{0}{0}{0}\\
    \teaserrowbottom{input_patch}{0}{0}{0}{0}\\
\end{tabular}
    \caption{We present \textbf{GVM}, a generative video matting model supporting fine-grained video matting for both humans and animals. Derived from Stable Video Diffusion~\cite{svd} and fine-tuned with diverse and high-quality segmentation and matting data, our model achieves remarkably robust generalization across various unseen scenarios. Additionally, it faithfully captures intricate fine-grained details while ensuring temporal consistency throughout the video. From top to down, the teaser shows: the input sequence (first row), RVM~\cite{lin2022robust} prediction (second row), and our prediction (third row).}
  \Description{teaser.}
  \label{fig:teaser}
\end{teaserfigure}

\maketitle
\section{Introduction}

The task of video matting involves estimating the temporal consistent alpha mattes, which define the transparency of each pixel, for a foreground object across all frames in a video sequence. It is a fundamental problem for various video editing applications, such as background replacement, compositing, and visual effects (VFX). Compared to image matting, video matting is considerably more challenging due to two key factors. 

\begin{figure}
\small
\centering
\begin{tabular}{@{\hspace{0mm}}c@{\hspace{1.2mm}} @{\hspace{0mm}}c@{\hspace{1.2mm}}@{\hspace{0mm}}c@{\hspace{1.2mm}}@{\hspace{0mm}}c@{\hspace{1.2mm}}}
\includegraphics[width=0.24\columnwidth]{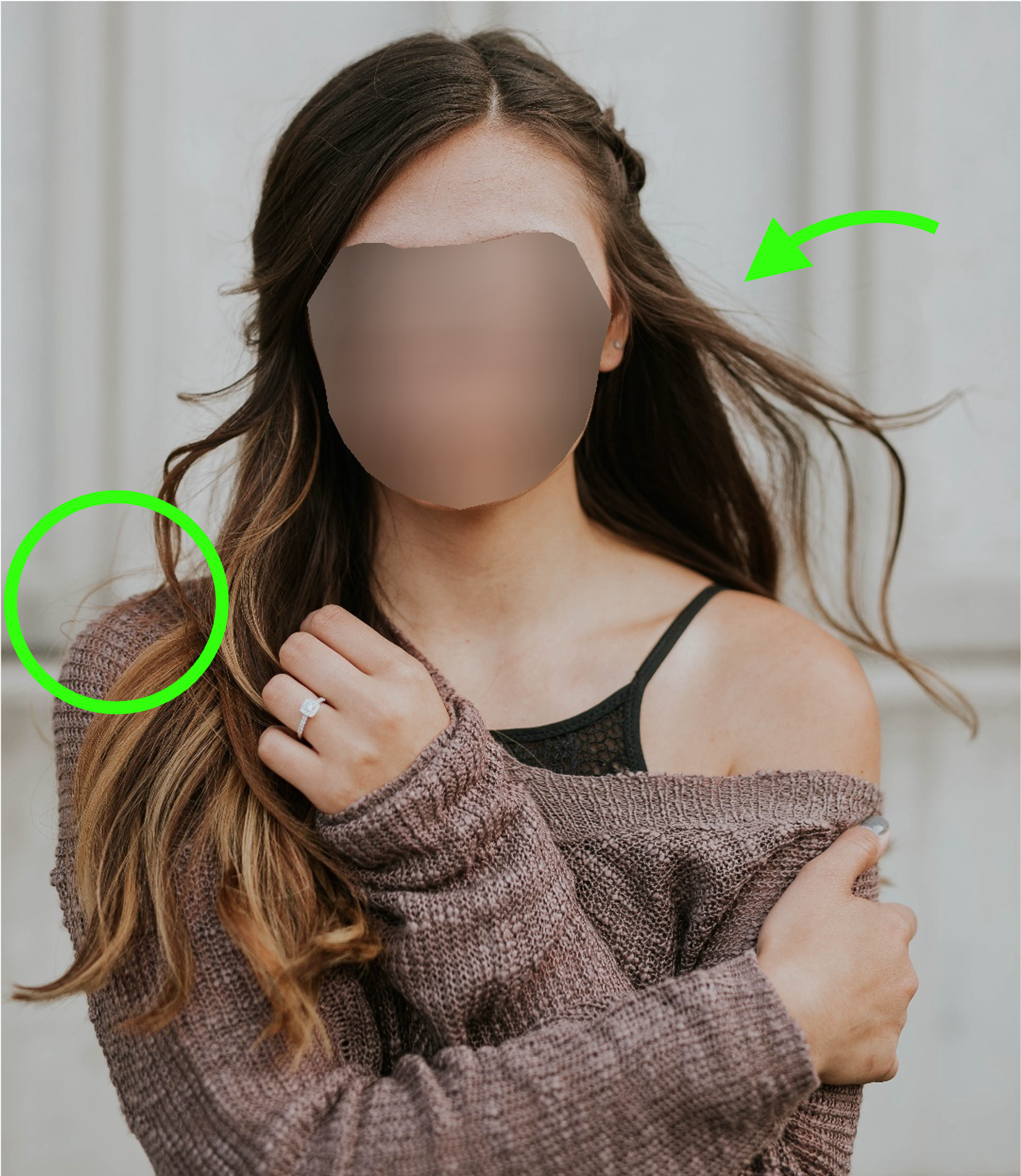} & 
\includegraphics[width=0.24\columnwidth]{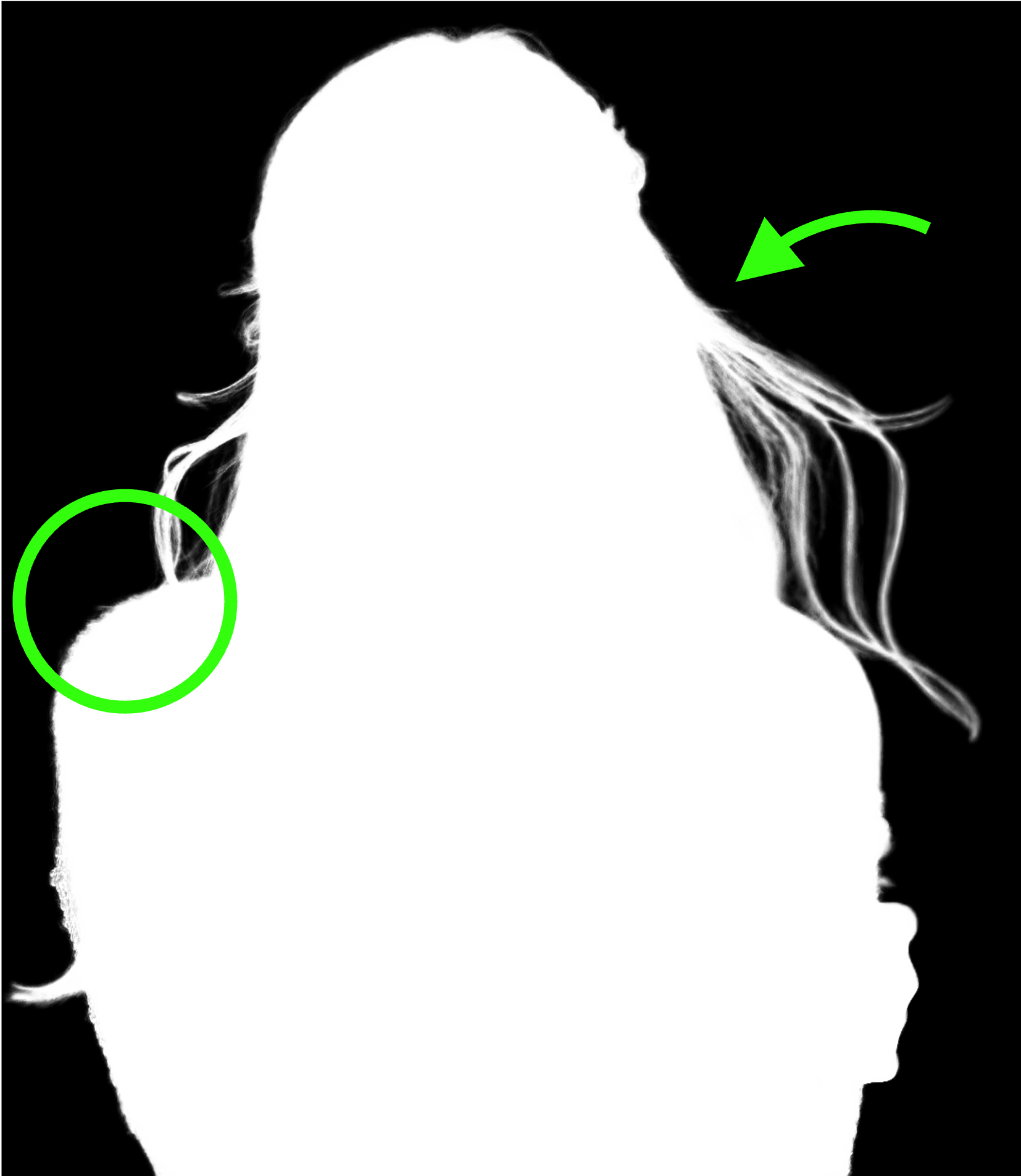} &  
\includegraphics[width=0.24\columnwidth]{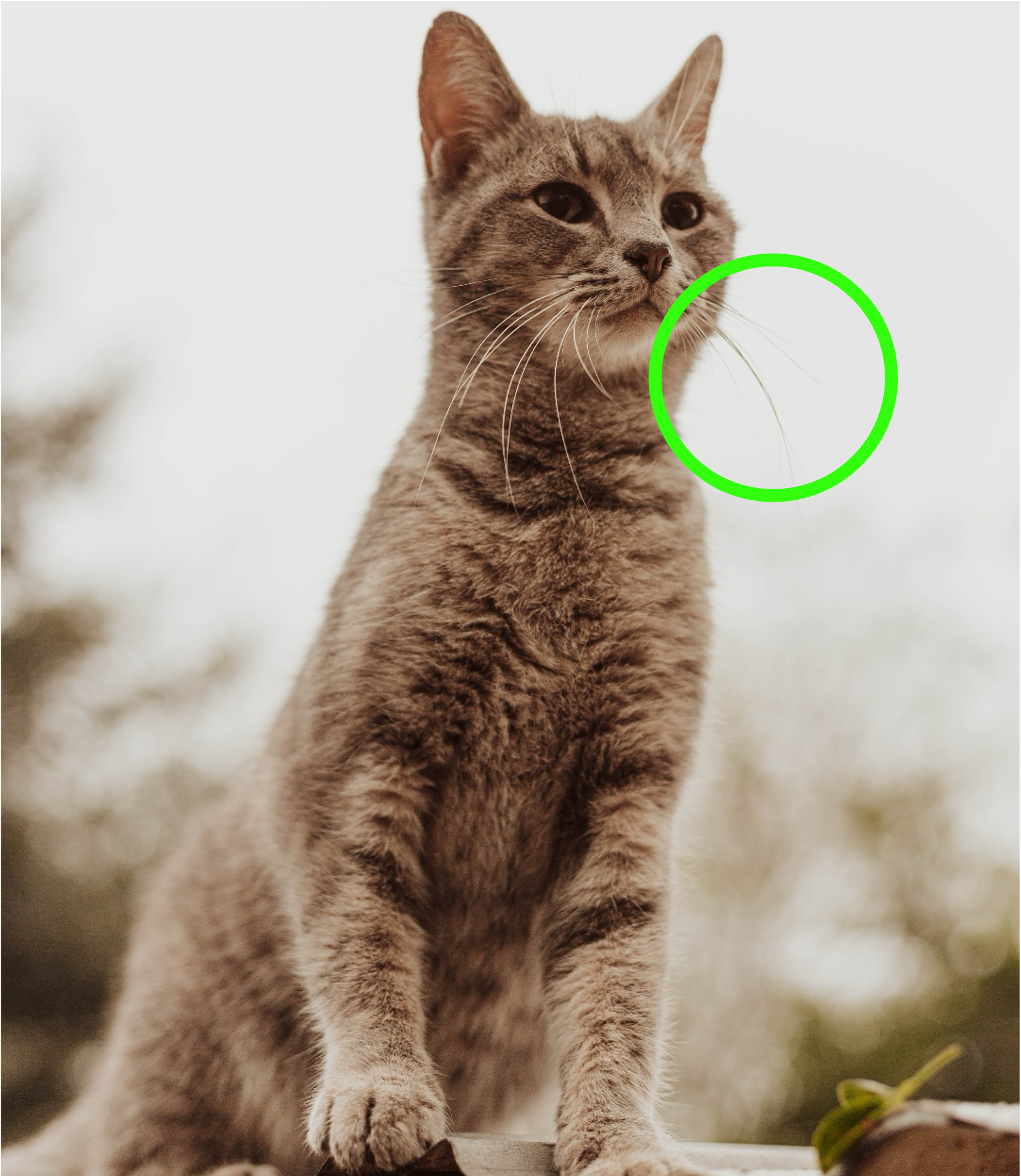} & 
\includegraphics[width=0.24\columnwidth]{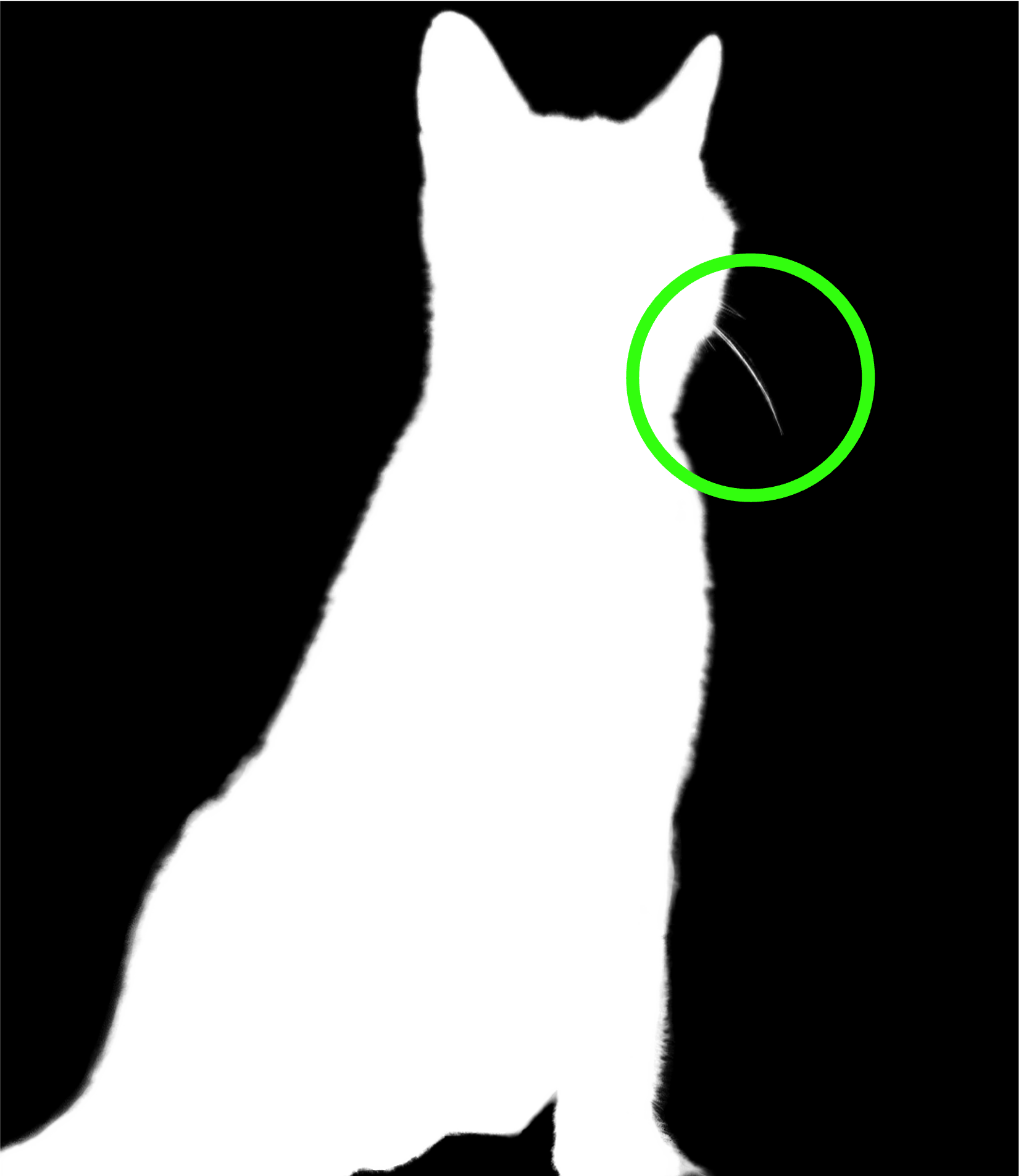}\\
P3M sample & Annotation & AM2K sample & Annotation \\
\end{tabular}
\caption{Imperfect annotations on P3M dataset and AM2K dataset.
The training data are usually either blurry or lacking in some details. Therefore, the regression-based model would overfit the imperfect ground truth.}
\Description{Imperfect human annotation.}
\label{fig:imperfect_label}
\vspace{-0.6cm}
\end{figure}

On one hand, the creation of manually annotated image matting datasets is not only labor-intensive but also often suffers from boundary inaccuracies. \Cref{fig:imperfect_label} showcases examples of such imperfect annotations from two widely used image matting datasets. This challenge is even more pronounced in the case of video matting datasets. Besides, most existing video matting datasets are composed using artificial backgrounds, leading to a noticeable gap between the foreground and background due to mismatched lighting conditions. As a result, the samples often appear unrealistic, making it challenging for models trained on these datasets to generalize effectively to real-world scenes. Thus, the absence of high-quality, accurately annotated data severely limits the performance of video matting methods. 

On the other hand, predicting video alpha mattes requires not only spatial consistency but also temporal coherence across all frames in the video sequence. A diverse range of approaches, each leveraging different types of priors, have been proposed to address this problem. Traditional methods have often relied on local or non-local affinities between pixel colors, as well as motion cues from the foreground~\cite{apostoloff2004bayesian,choi2012video,chuang2002video, li2013motion, zou2019unsupervised}. However, these methods still fall short, particularly when dealing with complex scenarios such as fast-moving objects or intricate background scenes. To address these issues, some methods~\cite{sindeev2013alpha} have incorporated optical flow to regularize the output and reduce artifacts. Yet, optical flow estimation remains highly unreliable within matting regions, where both foreground and background elements coexist. This is exacerbated by the challenge of accurately modeling large areas of semi-transparency, which current optical flow algorithms struggle to handle effectively. Some learning-based methods rely on per-frame trimaps to predict alpha mattes. While these approaches can yield good results, they are both resource-intensive and difficult to generalize across diverse, in-the-wild video datasets. To overcome the limitations of trimap-based methods, recent works like OTVM~\cite{seong2022one} have proposed trimap-free solutions, while BGM~\cite{sengupta2020background} has introduced a solution that only requires an additional background image captured without the foreground subject. However, while these approaches reduce the inference time compared to manual trimaps, they are less effective in dynamic or changing background environments. Moreover, the manual priors they require still hinder their application in real-time scenarios, such as video conferencing.
In recent years, some fully automatic solutions have been developed, such as MODNet~\cite{MODNet} and RVM~\cite{lin2022robust}, eliminating the need for manual input or auxiliary data. However, these methods remain challenged by diverse foreground categories, complex motion, and intricate physical conditions encountered in real-world video scenarios.

To tackle the above challenges, we propose a feasible solution for video matting of humans and animals from two complementary perspectives: (1) leveraging the existing large-scale synthetic segmentation datasets and pseudo labeling real-world video segmentation data for long video sequence pre-training, and constructing a synthetic video human matting dataset with fine-grained hair matte annotations. (2) presenting a novel generative video matting approach that utilizes the rich spatial and temporal priors from the pre-trained video diffusion model. 

\noindent\textbf{Synthetic Training Data.} With advancements in modern rendering engines such as Blender~\cite{blender} and Unreal Engine~\cite{unreal}, it has become feasible to generate a wide variety of photo-realistic foreground objects and background scenes. Recent datasets, like BEDLAM~\cite{bedlam}, offer high-quality video foreground segmentation masks. BEDLAM includes monocular RGB videos showcasing diverse human appearances (271 body shapes, 100 skin textures, and 27 types of hair), various motions, and a wide range of scenes. {The dataset provide rich and accurate annotations, including video human masks, camera intrinsic and extrinsic parameters, depth maps, and 3D human pose and shape parameters.} Another example, Dynamic Replica~\cite{karaev2023dynamicstereo}, comprises 524 videos featuring synthetic humans and animals performing actions in virtual environments, {with accurate video mask annotations of both humans and animals.} Although these datasets do not provide alpha matte annotations, they serve as valuable resources for large-scale pretraining of video matting models, as existing video matting datasets fall short in providing such complex, yet natural, video sequences. Besides, we use Blender~\cite{blender} to render 200 clips of human foregrounds, with fine-grained hairs and alpha matte annotations, to complement existing imperfect video matting datasets.

\noindent\textbf{Pseudo-labeled Training Data.} 
Due to the domain gap between synthetic datasets and real scenes, additional real-world datasets are still needed during the training of the matting network. However, large-scale annotation for video matting remains nearly impossible. With the emergence of large foundation segmentation models, such as SAM2~\cite{ravi2024sam2}, we can obtain temporally consistent video segmentation annotations from unlabeled videos. This allows us to annotate real-world videos and {obtain coarse video segmentation masks}, which are then used to perform pre-training for the video matting model.

\noindent\textbf{Network Architecture.} We reformulate the traditional regression-based video matting task as a conditional video \textit{generation} problem, where the alpha matte distribution is conditioned on the raw video input. This formulation offers several key advantages. First, generative models excel at handling uncertainties arising from imperfect video matte annotations. Training regression-based models with flawed data, such as the images shown in~\Cref{fig:imperfect_label}, often lead to over-smoothed matte outputs. In contrast, generative models are more robust to such imperfections in the training data. Second, this approach allows us to leverage existing pre-trained video generation models as priors. Modern video diffusion models are typically trained on vast amounts of video data, which allows them to encode rich fine-grained semantic and geometric priors, capturing both spatial and temporal details.
Our pivotal experiments demonstrate that the generative model shows exceptional generalization even when trained solely on synthetic segmentation datasets lacking semi-transparency regions and fine-grained annotations like hair and fur. It effectively predicts intricate fur details for animals and fine hair details for humans, showcasing the model's robustness and ability to handle complex real-world scenarios.
In practice, we employ an image-to-video diffusion model, specifically the Stable Video Diffusion model (SVD)~\cite{svd}, as our base model.\footnote{As a general framework, our method is adopted to any video diffusion models.} We adapt this model for video matting with minimal yet essential modifications: (1) Fine-tune the model using a flow-matching mechanism to reduce the number of inference steps, thereby accelerating prediction speed. Additionally, we introduce optional image-space supervision to enhance the fine-grained details of the matte predictions. (2) We propose a comprehensive three-stage training strategy designed to incorporate the synthetic data and real-world data for boosting the model performance.

In summary, our contributions are as follows:

\begin{itemize} \item We propose a comprehensive training strategy that effectively combines large-scale synthetic and pseudo-labeled video segmentation data with a smaller set of high-quality synthetic video matte data. This approach is inspired by the methodology used in recent large video foundation models to achieve strong generalization: first, leveraging extensive pre-training on large datasets, followed by fine-tuning on a more compact set of high-quality video matting data.
\item We create a high-quality synthetic video matting dataset, SynHairMan, with fine-grained annotations and diverse human motions to boost the performance of the portrait video matting.
\item We are the \textit{first} to incorporate video diffusion priors for video matting generation. We make several key modifications, e.g., (1) flow-matching supervision in latent space to accelerate inference speed. (2) Hybrid supervision in both latent space and image space for fine-grained matting. Our model demonstrates strong generalization capabilities, even for real-world nature scenes, and can predict alpha mattes for previously unseen animal categories. \end{itemize}

\section{Related Work}

\noindent\paragraph{Diffusion Models for Transparent Image Generation.} Recently, the computer vision and graphics community has witnessed the emergence of diffusion-based text-to-image  (T2I) models~\cite{rombach2022high, sdxl, sd3, flux}. However, existing diffusion-based image generators can only generate RGB images, while the vast majority of visual content editing workflows are layer-based, relying heavily on transparent layers to compose and create content. LayerDiffuse~\cite{layerdiffuse} collects  1M transparent image layer pairs and fine-tunes the pre-trained latent diffusion models to generate transparent images. 
Another line of research leverages diffusion priors to predict alpha mattes from input RGB images. A notable example is AlphaLDM~\cite{genmatte}, an image matting method that utilizes pre-trained priors from an image diffusion model, such as Stable Diffusion~\cite{rombach2022high}. However, adapting this framework to video matting presents several challenges. First, AlphaLDM~\cite{genmatte} requires at least 50 inference steps to obtain a rough alpha matte, followed by multi-scale inference to refine the prediction by cropping uncertain regions. This process is not acceptable for long-video matting. Second, the method is specifically designed for image matting, and its architecture does not inherently ensure temporal consistency across video frames. Lastly, AlphaLDM~\cite{genmatte} is trained on a small-scale image portrait matting dataset, and it does not address the challenges of scaling up the data for training temporal models.

\paragraph{Video Matting.} Traditional video matting can be performed through green-screen chroma keying, requiring complex filming setups and manual post-processing, or learning-based video matting expert models~\cite{video240k, lin2022robust,  Lin_2023_ICCV}. For example,
Background Matting~\cite{video240k} presents a high-speed, high-resolution background replacement technique with precise extraction of alpha mattes, supported by the VideoMatte240K dataset~\cite{video240k}. 
Robust Video Matting (RVM)~\cite{lin2022robust} proposes a high-quality human video matting method in real time with a recurrent architecture for improved temporal coherence, achieving state-of-the-art results without auxiliary inputs.
OmnimatteRF~\cite{Lin_2023_ICCV} introduces a video matting method that combines dynamic 2D foreground layers with a 3D background model, enabling more realistic scene reconstruction for real-world videos. Generally, previous models favor simple video backgrounds and are not robust to scenes with complicated backgrounds, fast-moving objects, or multiple unclear target subjects.

\paragraph{Video Generation.} Diffusion models trained on large-scale video datasets have shown exceptional performance in generating high-quality videos. For example, Stable Video Diffusion (SVD)\cite{svd} generates video sequences by extending the UNet~\cite{unet} architecture, which integrates temporal convolutions and cross-attention layers after each corresponding spatial layer. Sora~\cite{sora} is capable of generating 1080p resolution videos, lasting up to 60 seconds, with an emphasis on simulating real-world physical dynamics. These diffusion-based video generation models, which capture both rich spatial and temporal priors, present a compelling opportunity for adaptation to video matting tasks.

\paragraph{Matting Datasets.} High-quality matting datasets are essential for training learning-based matting models. However, the creation of these datasets is often labor-intensive, resulting in relatively small datasets. For example, Dinstinctions-646~\cite{qiao2020attention} contains 646 diverse foreground objects, while P3M-10K~\cite{Li-2021-P3M} offers 10,000 high-resolution face-blurred portrait images with corresponding alpha mattes. Video240K~\cite{video240k} includes 484 high-resolution green-screen videos, generating a total of 240,709 unique frames with alpha mattes and foregrounds. Despite its utility, Video240K suffers from coarse hair annotations and lacks guarantees of fine-grained temporal consistency, as it is generated using chroma-key software like Adobe After Effects. Additionally, due to the absence of eyebrow matting datasets, Wang et al.~\cite{eyebrow} introduced the first large-scale \textit{synthetic} eyebrow matting dataset, demonstrating the effectiveness of rendered datasets for the eyebrow matting task. In this work, we contribute a new synthetic video matting dataset that accurately captures the dynamics of hair matting, addressing the gap in high-quality hair annotations in existing video matting datasets.
\section{Method}

\begin{figure*}
    \centering
\includegraphics[width=1.00\linewidth]{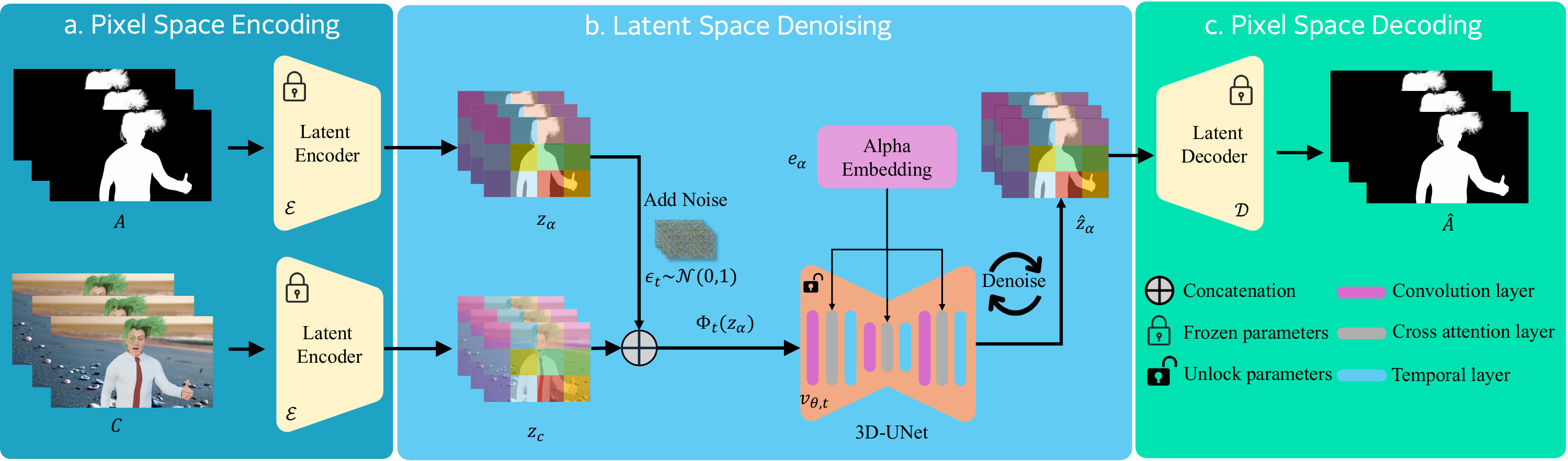}
    \caption{
        The whole pipeline of the proposed method. (a) The latent encoder processes video input $C$ and alpha mask $A$ into corresponding latents $z_{i}$ and $z_{\alpha}$. (b) In latent space, the sampled Gaussian noise $\epsilon_{t}$ is added to the clean alpha latent $z_{\alpha}$, producing the noisy alpha latent $\Phi_{t}(z_{\alpha})$. (c) During the inference stage, the alpha mask input $A$ and foreground is replaced with random sampled Gaussian noises, multi-step denoising is then performed, conditioned on the video input $C$, to estimate $\hat{z_{\alpha}}$. Finally, these latent estimates are passed through the frozen latent decoder to generate the predicted alpha masks. We drop the CLIP~\cite{clip} embedding and instead inject zero alpha embeddings ({replace the original SVD image embedding with the same size embedding with zero values}) into the 3D-UNet.
}
    \label{fig:pipeline}
    \Description{Method.}
\vspace{-0.55cm}
\end{figure*}

Our goal is to adapt pre-trained video generative models for video matting with minimal alterations, thereby retaining their extensive prior knowledge. In Section~\ref{sec:formulation}, we describe how the traditional video matting regression problem can be reformulated as a conditional generative denoising process. We also detail the architectural adjustments made for video matting. The overall pipeline is presented in ~\Cref{fig:pipeline}. The detailed training supervision is illustrated in~\Cref{sec:training_losses}. Beyond model architecture, we emphasize that the quality of fine-tuning data and the choice of fine-tuning strategy are critical for achieving fine-grained video matting. Accordingly, we outline the training datasets process in Section~\ref{sec:training_datasets} and provide a comprehensive training strategy in Section~\ref{sec:training}.

\subsection{Generative Formulation}
\label{sec:formulation}
Video matting is the task that given a video $\mathcal{V}=\left\{C_i\right\}_{i=1}^T$
of $N$ frames, the goal is to decompose each frame $C_t \in \mathcal{V}$ as:
\begin{equation}
C_i=\matte_i \cdot  F_i+\left(1-\matte_i\right) \cdot  B_i
\end{equation}
where $\matte_i$, $F_i$, and $B_i$ are the alpha matte, foreground color and background color, respectively. The symbol \text{*} means the Hadamard product. Video matting is a challenging task because it entails obtaining high-quality details of each frame while maintaining favorable temporal consistency across frames. 

We tackle this task by generating the alpha matte conditioned on the video diffusion model.
Specifically, we model the distribution of alpha matte $p(\matte)$ with the pre-trained Stable Video Diffusion model~\cite{svd}. To condition the matte generation on the input video $x_c$, we first apply a normalization to ensure it falls within the VAE's range of $[-1, 1]$, and then encode it using the pre-trained VAE encoder $\mathcal{E}$ into the latent space as $\latent_{c} = \mathcal{E}(x_c)$. Then, $\latent_{c}$ is concatenated with the latent matte code $\latent_{\matte}$ frame by frame, forming the input for the video denoiser ${v}_{\theta}$.
Note that some of the 3D VAE variants~\cite{zhao2024cvvae, cogvideox} compress spatial and temporal dimensions for more efficient video generation. Here we only compress the input spatial dimension into the latent space. This approach avoids motion blur artifacts that could arise while decoding latent mattes back into image space by using the VAE decoder $\mathcal{D}$.

The previous image matting method, AlphaLDM~\cite{genmatte}. which requires 50 denoising steps to generate the final alpha matte, is unsatisfactory in real-time application. In contrast to AlphaLDM, which maps Gaussian noise with image conditioning to matte using a diffusion process, we formulate matte estimation as a direct distribution transport between the image sequences feature $z_c$ and the matte features $z_{\alpha}$ by $p(z_{\alpha}|z_c)$, which can be effectively solved using conditional flow matching~\cite{lipman2023flowmatching}, achieving satisfactory results in just 1-3 steps. Concretely, the data corruption in our framework is formulated as a linear interpolation between Gaussian noise and matte data:
\begin{equation}
\phi_t(z_{\alpha}) = t z_{\alpha} + \left(1 - t\right) \epsilon,
\end{equation}
where $\phi_t(z_{\alpha})$ represents the corrupted data, with $t \in [0,1]$ as the time-dependent interpolation factor.

This formulation suggests a linear and uniform transformation with a constant velocity between the data and noise distributions. The corresponding time-dependent velocity field, which describes the motion from noise to data, is expressed as:
\begin{equation}
v_t(z_{\alpha}) = z_{\alpha} - \epsilon.
\end{equation}
The velocity field $v_t:[0,1] \times \mathbb{R}^d \rightarrow \mathbb{R}^d$ defines an ordinary differential equation (ODE):
\begin{equation}
d \phi_t(z_{\alpha})=v_t\left(\phi_t(z_{\alpha})\right)d t.
\end{equation}
By tracing the path of a noise vector through this field from $t = 0$ to $t = 1$, according to the ODE, we can transform noise into a matte latent code using the approximated velocity field $v_\theta$. Then, we can decode the matte latent to get the video matte prediction.

\subsection{Training Supervision}
\label{sec:training_losses}

During training, we utilize two types of losses: one in the latent space and another in the pixel space.
In the latent space, we replace the EDM noise scheduler~\cite{karras2022edm} used in SVD with a flow-matching scheduler~\cite{lipman2023flowmatching} to accelerate the inference speed. The flow matching objective directly predicts the target velocity to generate the desired probability trajectory:

\begin{equation}
\mathcal{L}_{latent}=\mathbb{E}_{t}\left\| v_\theta \left(\phi_t\left(z_{\alpha}\right), z_c, t\right) - v_t\left(z_{\alpha}\right)\right\|^2,
\end{equation}
where $z_{\alpha}$ and $z_c$ represent the latent alpha matte code and video sequence code, respectively.

Besides, we add auxiliary pixel space losses to enhance the alpha matte's fine-grained details, i.e., making the network more sensitive to boundaries and getting better performance in local smoothness. Specifically, we sample $\matte_t$ from $p(\matte_t|\matte_0, \matte_T)$ for $t\in[0, T]$ during the training. Due to the straight trajectories enforced through the flow-matching method, we can calculate the $\latent_{\matte_0}$ and then we decode with the pre-trained autoencoder to obtain an estimate of the alpha matte $\hat{\matte}$. Given the ground truth alpha matte $\matte$, we use the L1 loss, pyramid Laplacian loss~\cite{hou2019context} $\mathcal{L}_{lap}$, and gradient penalty loss~\cite{dai2022boosting} $\mathcal{L}_{gp}$ in the pixel space:
\begin{equation}
\mathcal{L}_{pixel} = \mathcal{L}_{1} + \mathcal{L}_{lap} + \lambda\mathcal{L}_{gp}.
\end{equation}

Combining the two parts of the loss, we can define our final training loss as a combination of the data-dependent flow-matching objective and
the pixel space losses:
\begin{equation}
\mathcal{L}_{total} = \mathcal{L}_{latent} + \lambda\mathcal{L}_{pixel},
\end{equation} where $\lambda$ is the loss weight between latent space loss and pixel space losses.

\subsection{Training Datasets}
\label{sec:training_datasets}
The scarcity of training data poses a significant challenge for deep learning-based video matting methods. For example, most existing video matting methods are trained on the composited VideoMatte240K dataset~\cite{video240k}, with 484 high-resolution green screen video clips, alpha mattes, and foregrounds. This video matting dataset has two significant limitations. First, the data is synthetically composited onto background images or videos, often resulting in unrealistic appearances due to lighting mismatches between the foreground and background. The foreground also lacks complicated human-human or human-object interactions. Training solely on this data can cause overfitting to the synthetic distribution. Second, the dataset lacks intricate details like hair or fur, which is crucial for achieving fine-grained portrait video matting.

\begin{figure*}
\small
\centering
\begin{tabular}{@{\hspace{0mm}}c@{\hspace{1.5mm}} @{\hspace{-1.mm}}c@{\hspace{1.5mm}}@{\hspace{-1.mm}}c@{\hspace{1.5mm}}@{\hspace{-1.mm}}c@{\hspace{1.5mm}}}
\includegraphics[width=0.5\columnwidth]{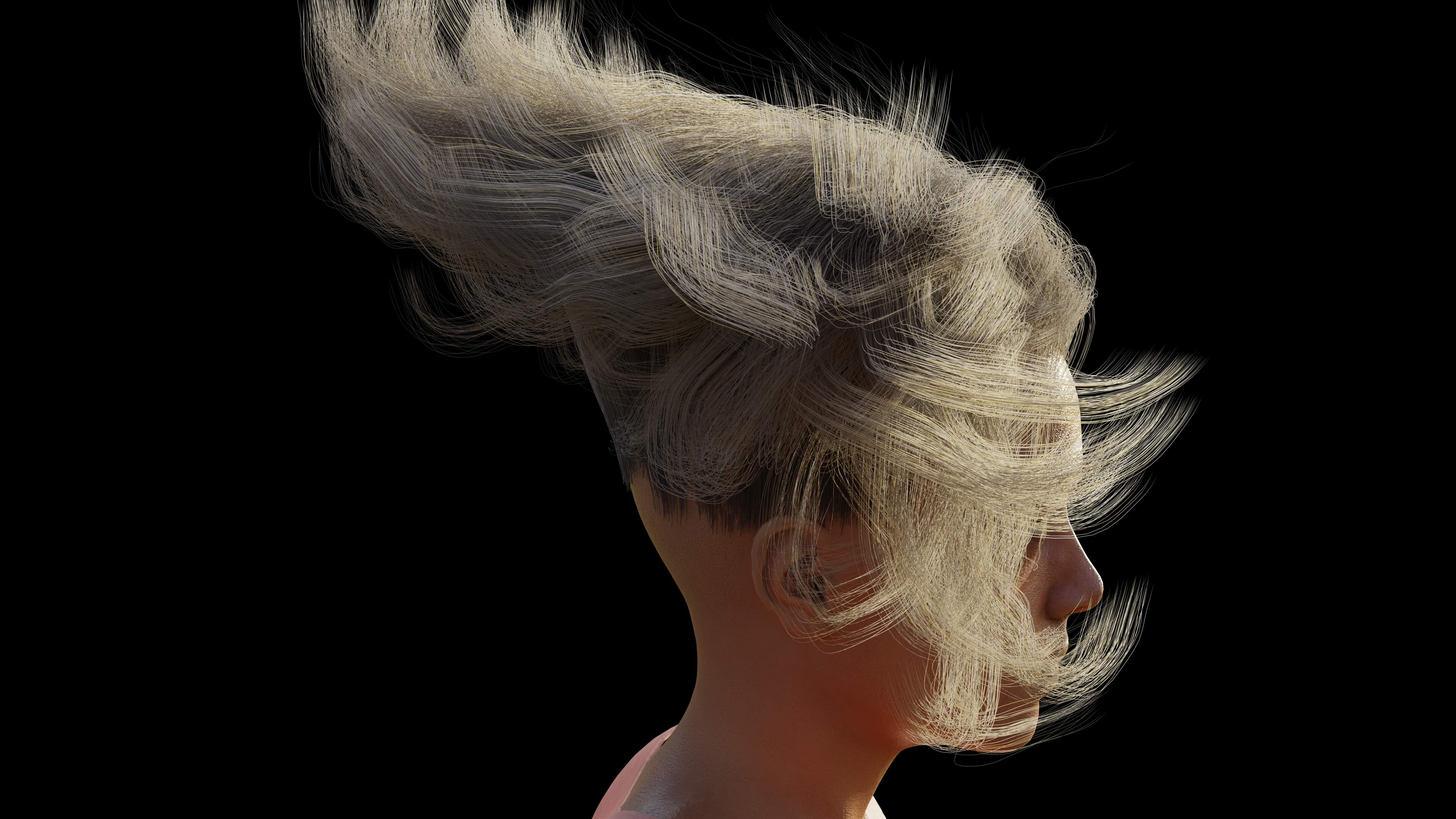} & 
\includegraphics[width=0.5\columnwidth]{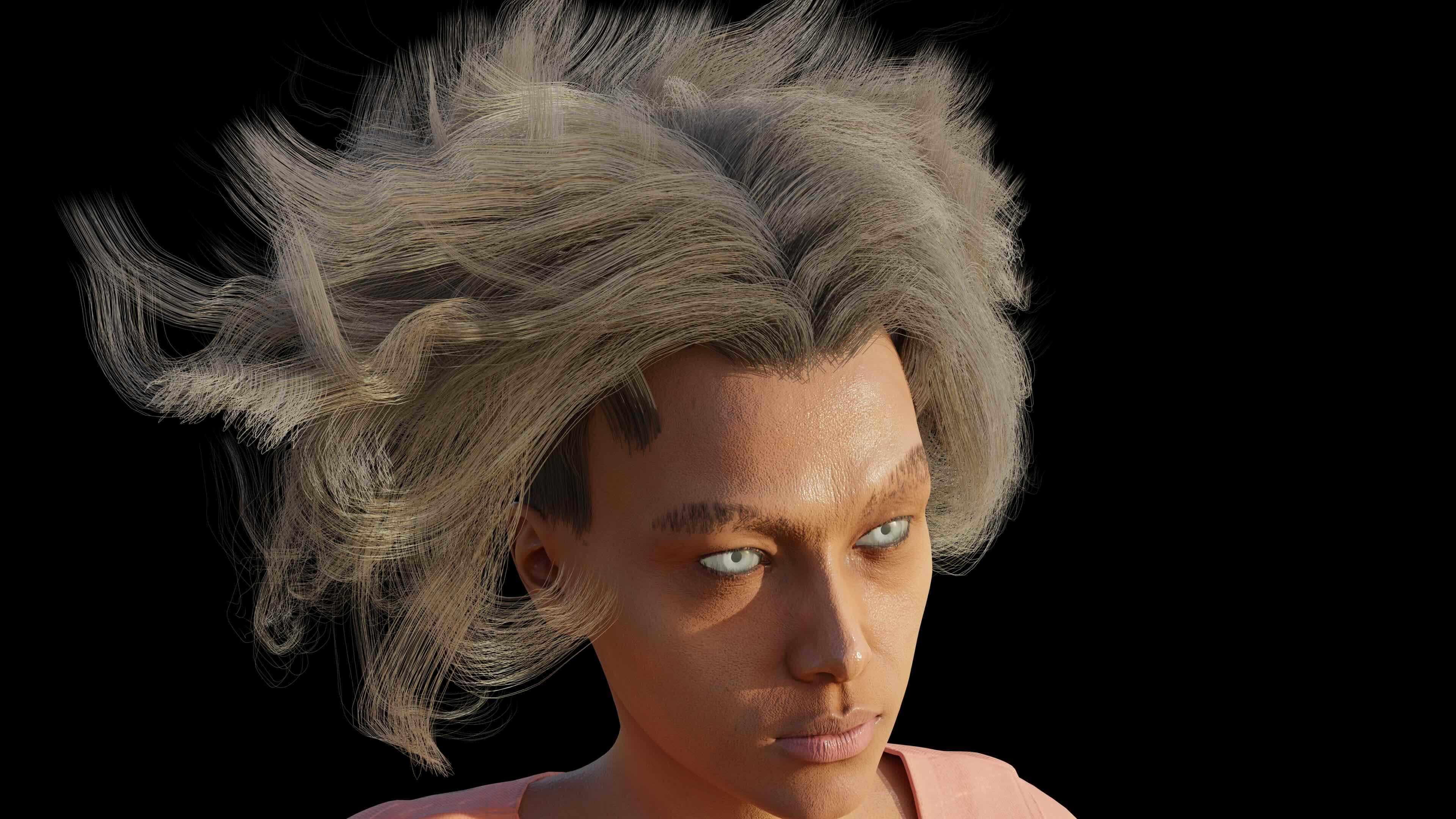} & 
\includegraphics[width=0.5\columnwidth]{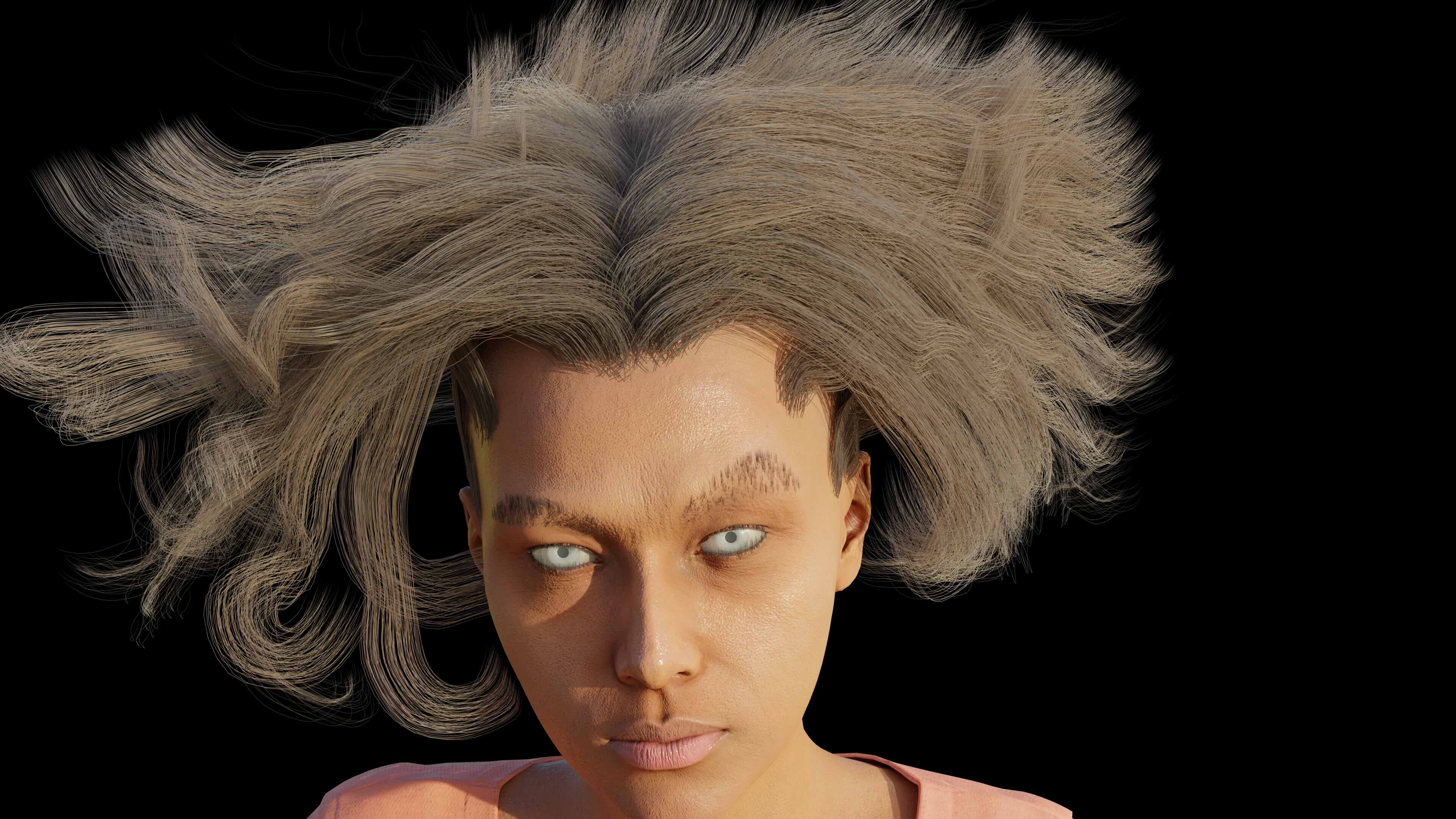} &  
\includegraphics[width=0.5\columnwidth]{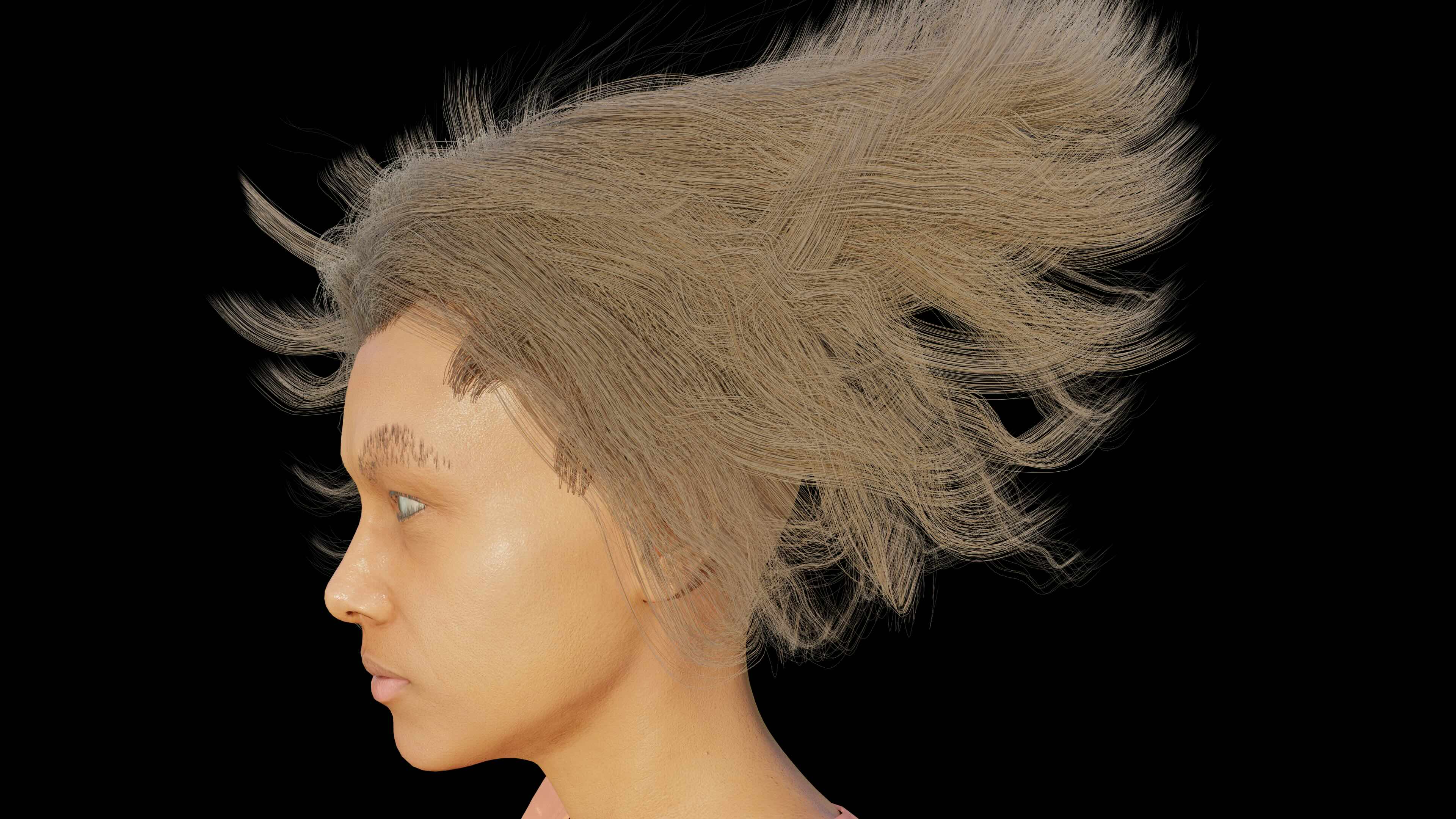}\\
\includegraphics[width=0.5\columnwidth]{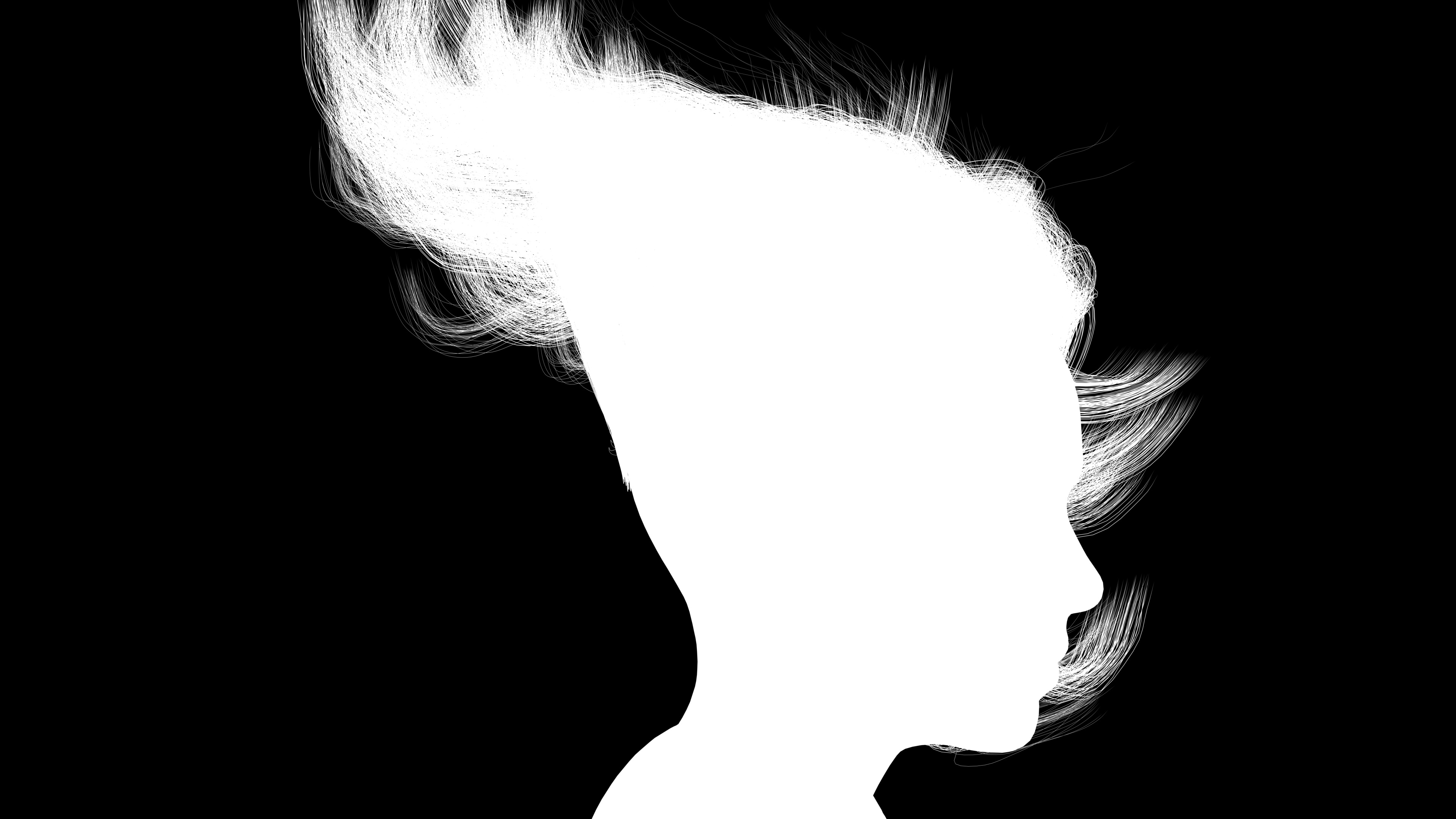} & 
\includegraphics[width=0.5\columnwidth]{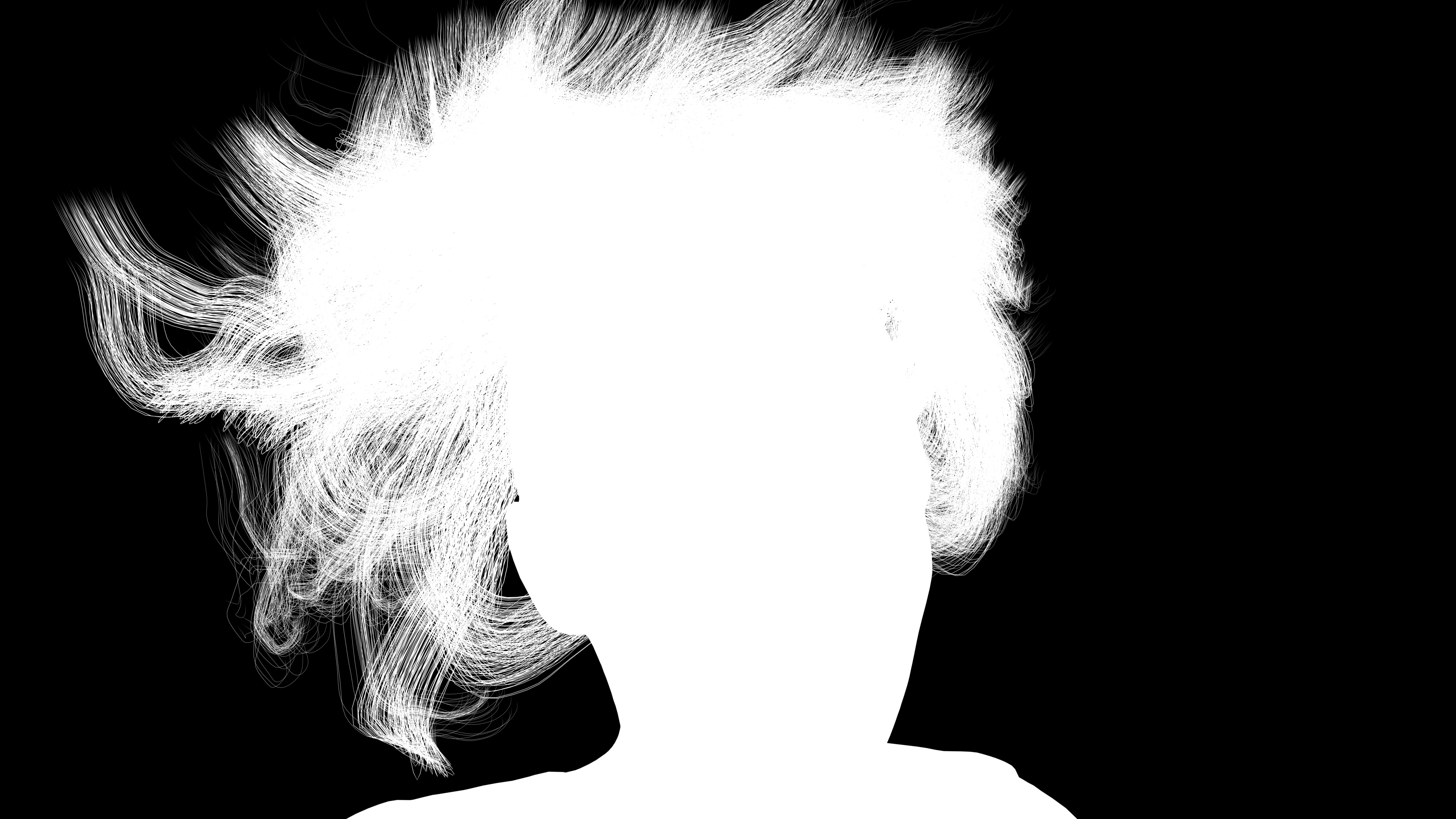} & 
\includegraphics[width=0.5\columnwidth]{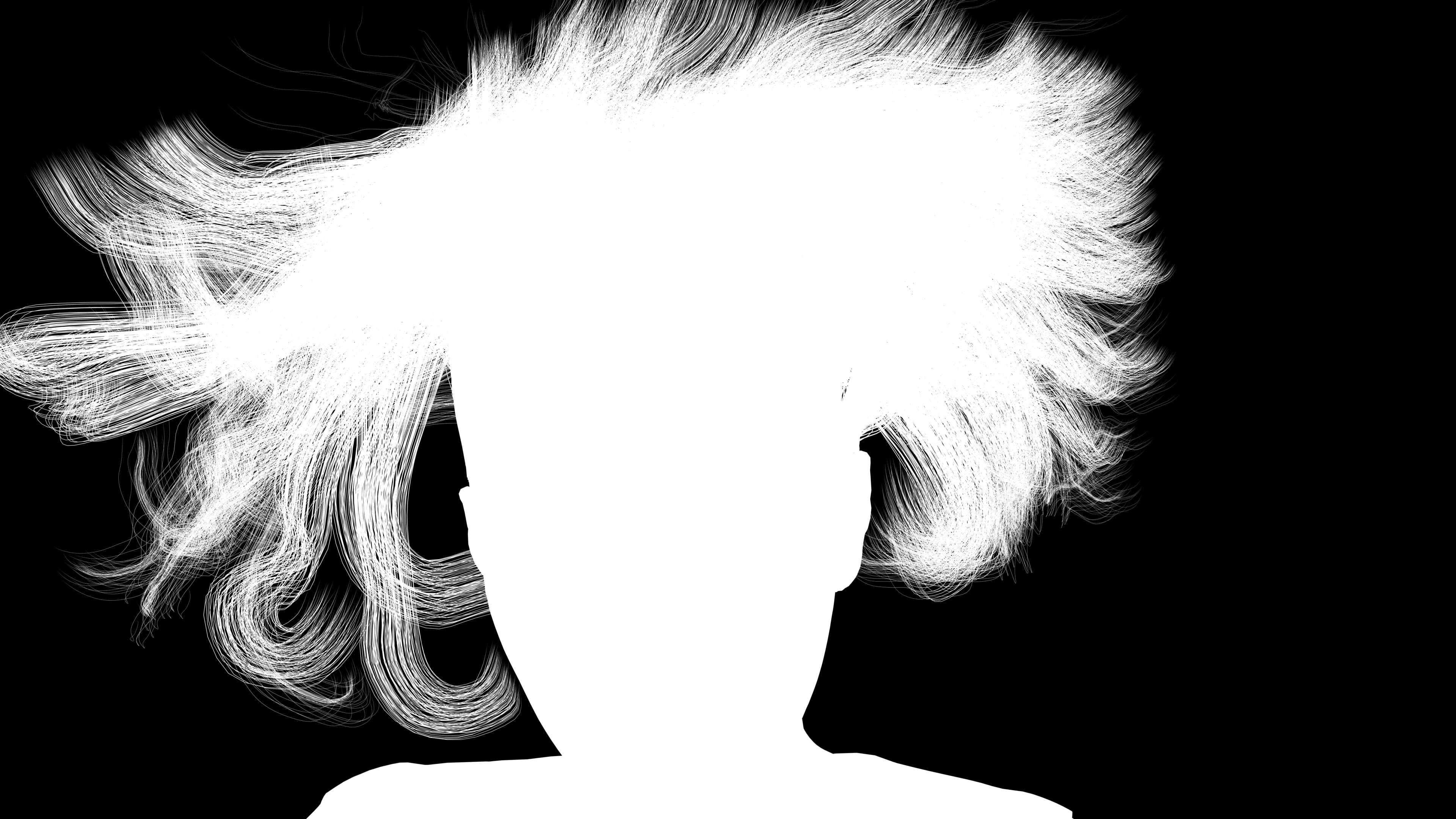} &  
\includegraphics[width=0.5\columnwidth]{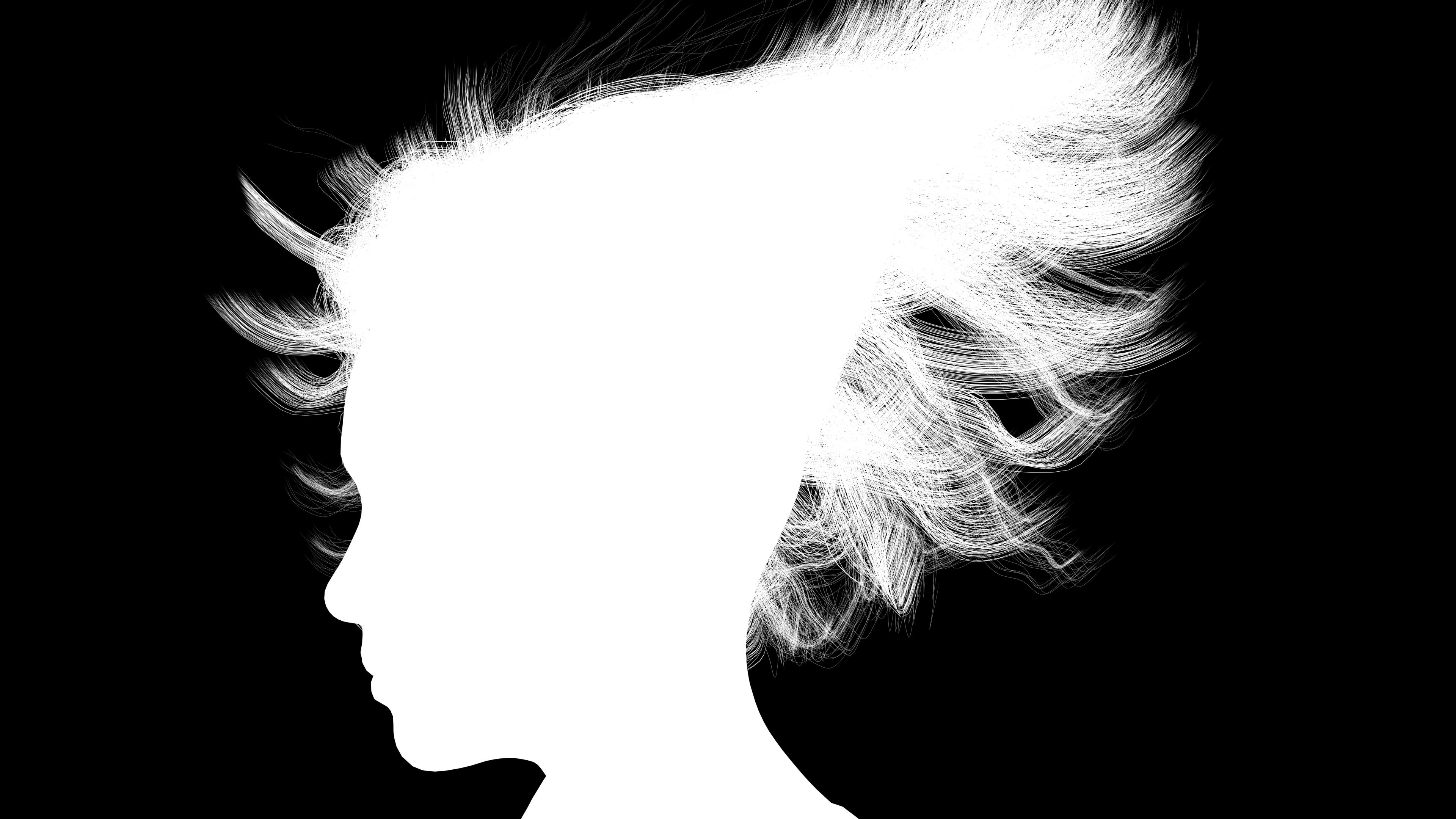} \\
\includegraphics[width=0.5\columnwidth]{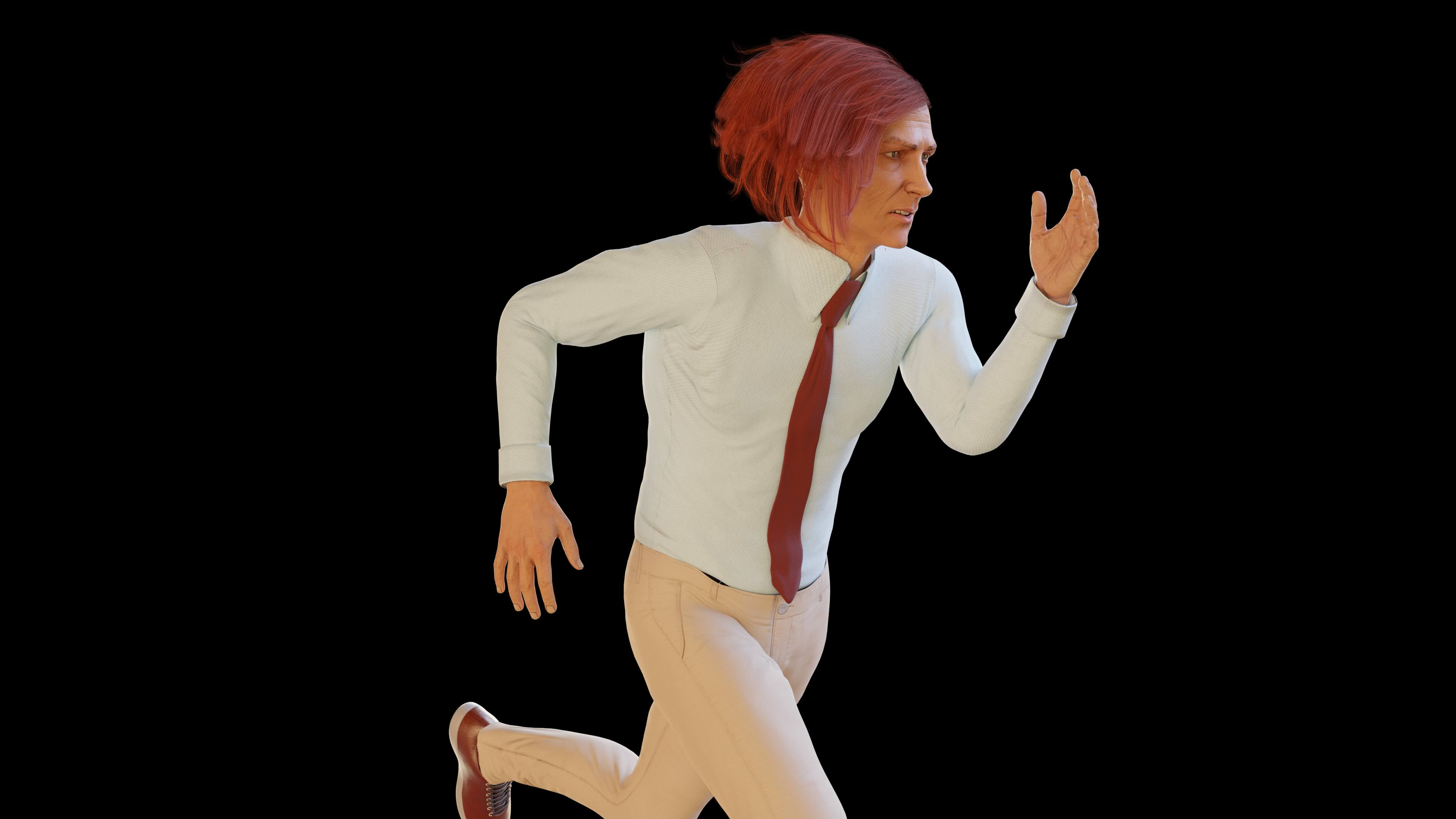} & 
\includegraphics[width=0.5\columnwidth]{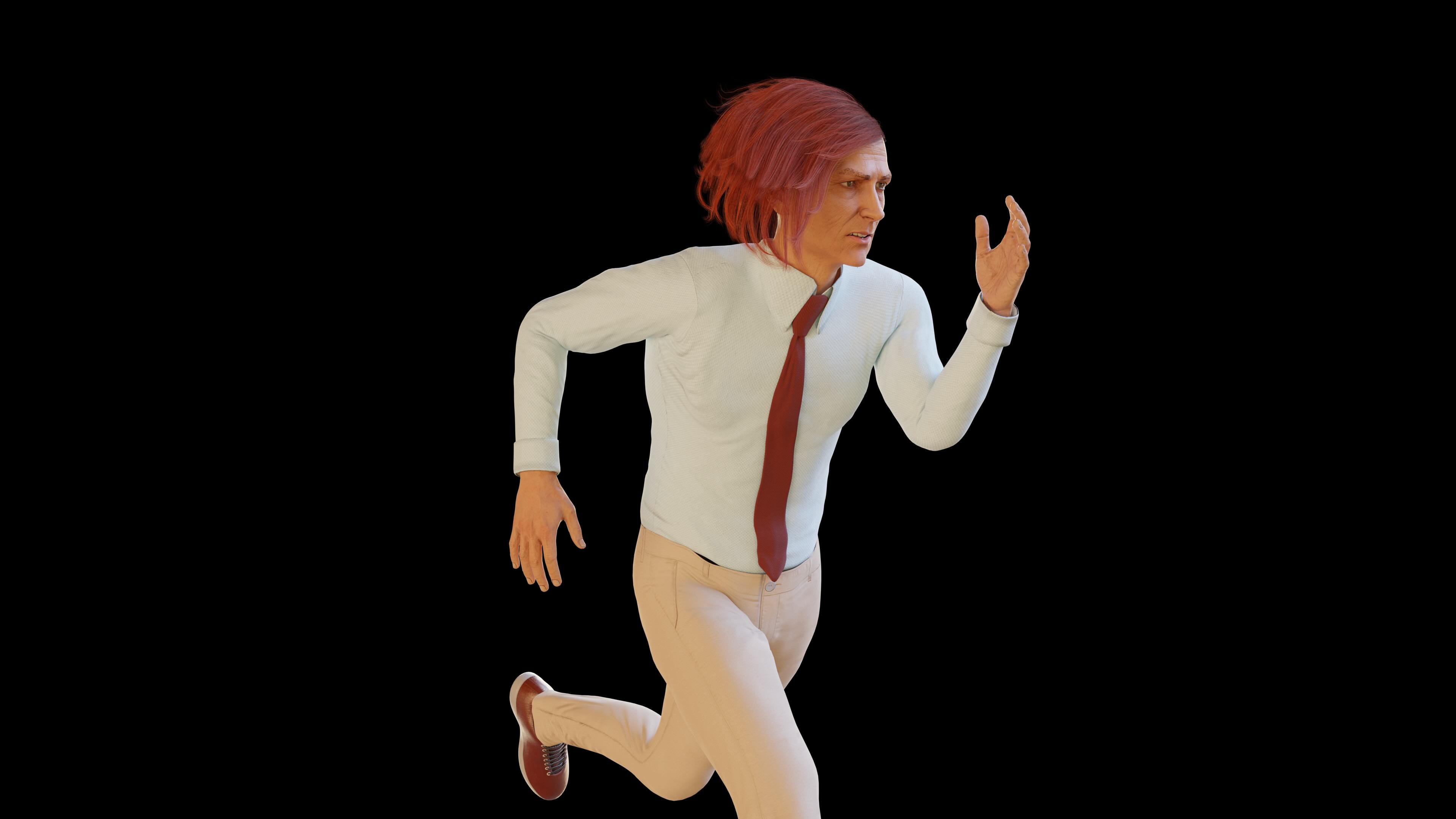} & 
\includegraphics[width=0.5\columnwidth]{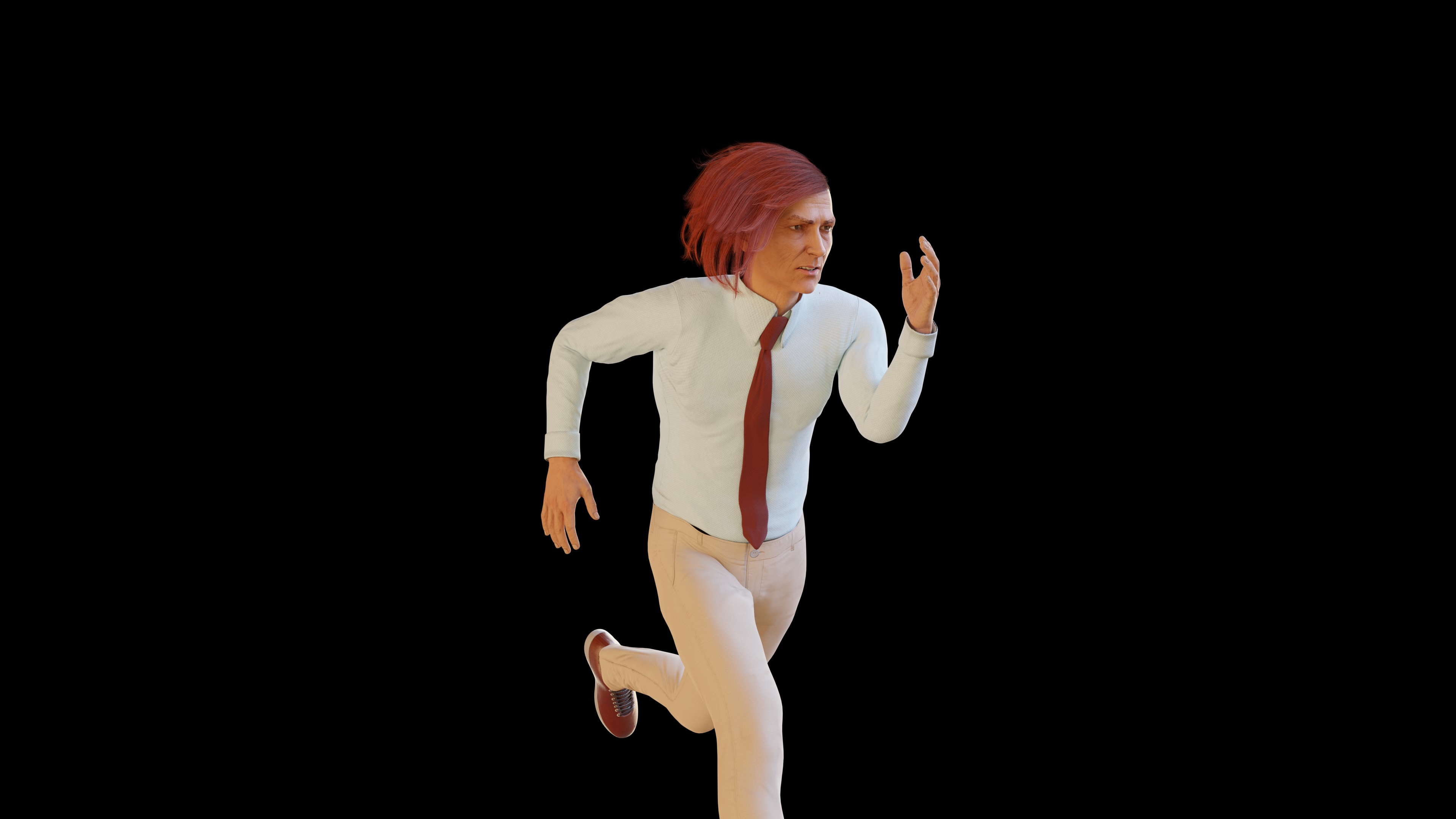} & 
\includegraphics[width=0.5\columnwidth]{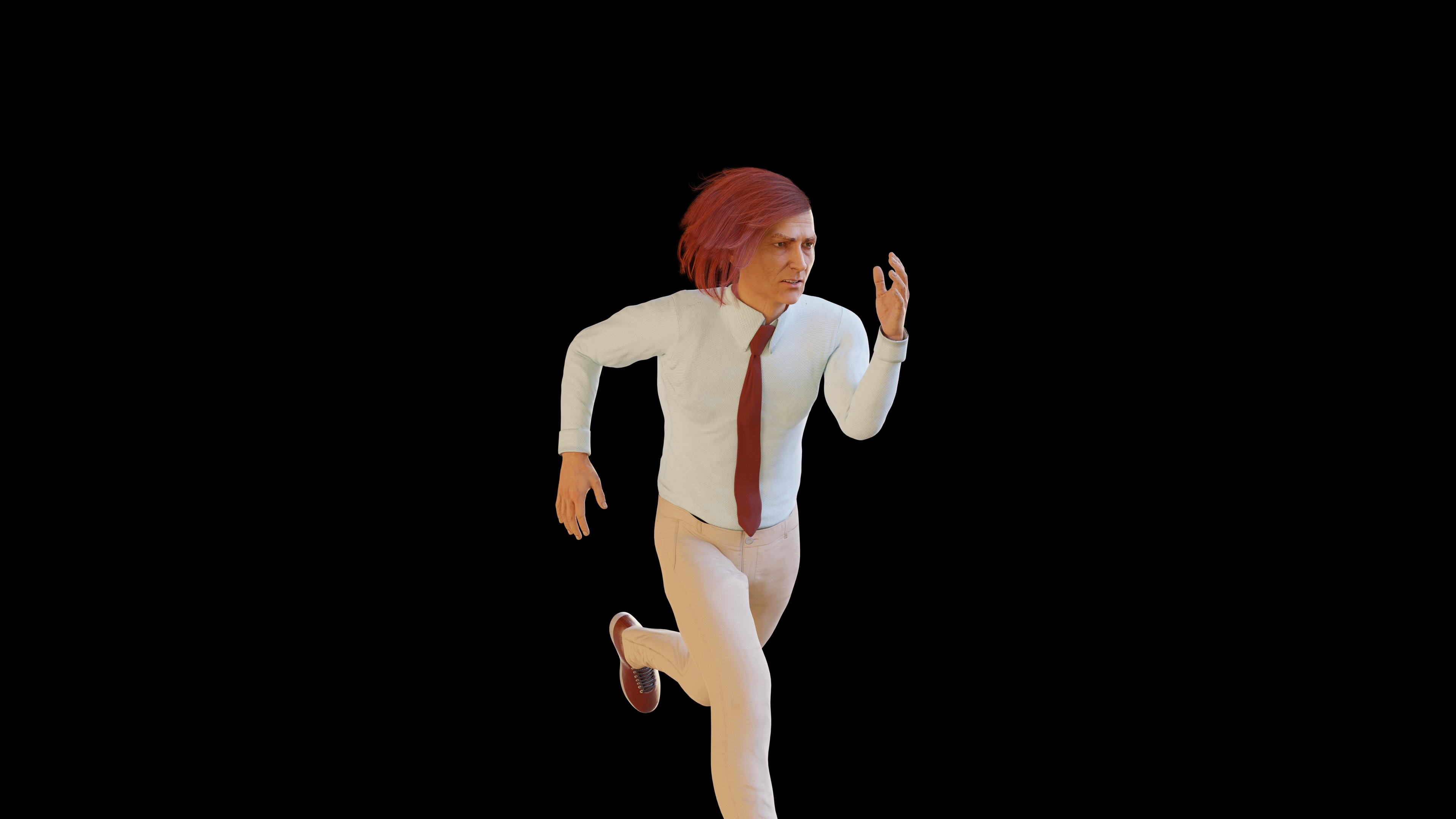}\\
\includegraphics[width=0.5\columnwidth]{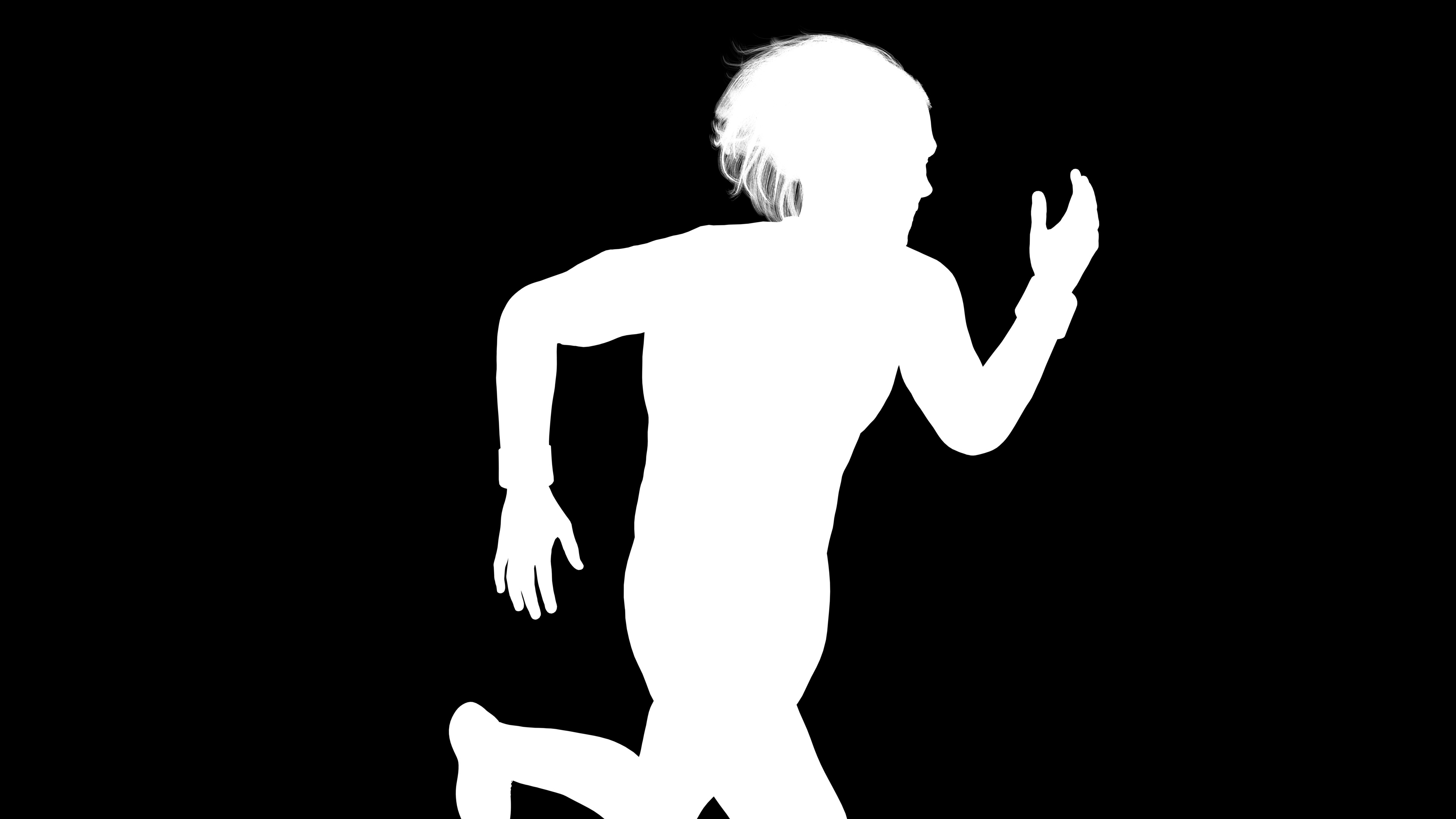} & 
\includegraphics[width=0.5\columnwidth]{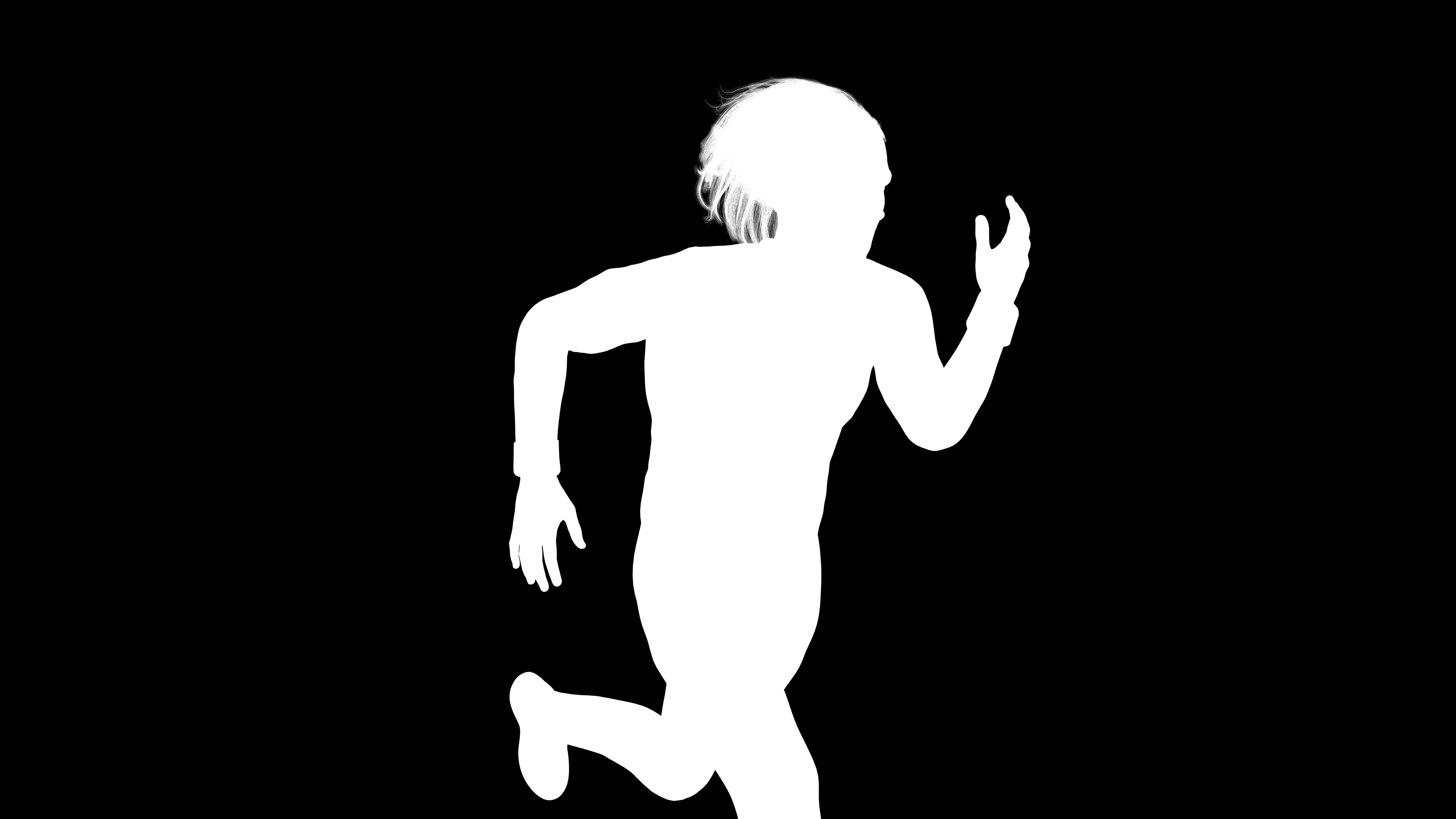} & 
\includegraphics[width=0.5\columnwidth]{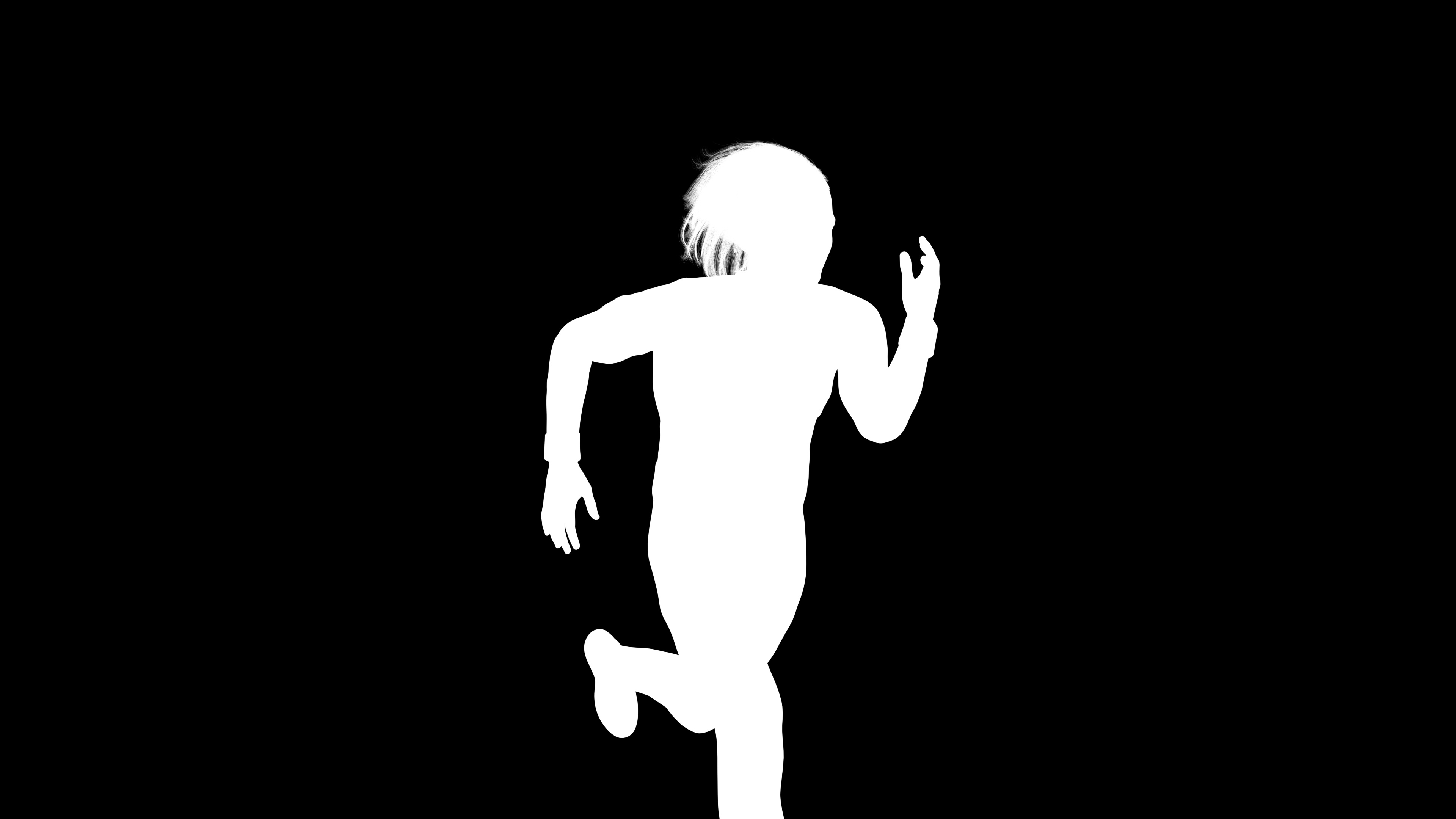} & 
\includegraphics[width=0.5\columnwidth]{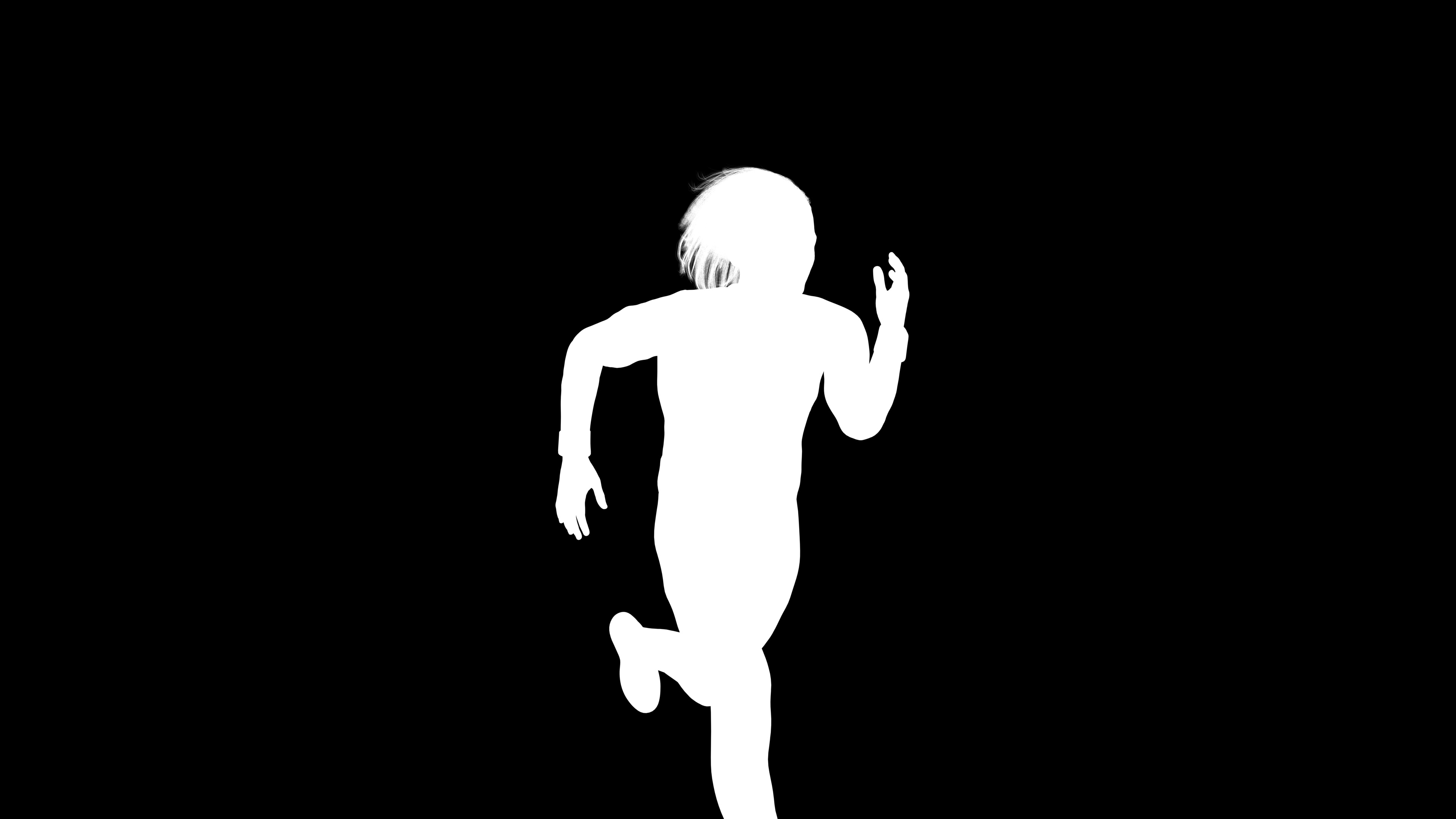}\\
Frame 0 & Frame 20 & Frame 40 & Frame 60 \\
\end{tabular}
\caption{Rendered portrait video matting dataset with fine-grained hairs.}
\Description{Rendered portrait video matting data.}
\label{fig:synhairhuman}
\end{figure*}

We address the challenges outlined above through a two-pronged approach. First, we leverage large-scale video segmentation data for pre-training. Specifically, we use two synthetic video datasets that feature humans and animals: BEDLAM~\cite{bedlam} and Dynamic Replica~\cite{karaev2023dynamicstereo}. These datasets provide a wide variety of foreground objects, background scenes, and accurate segmentation masks (as shown in~\Cref{tab:train_sources}). To bridge the domain gap between these synthetic datasets and real-world scenes, we also collect high-resolution videos with diverse human movements and background settings. We employ state-of-the-art video segmentation models, such as SAM2~\cite{ravi2024sam2} and Sapiens~\cite{khirodkar2024sapiens}, for pseudo-labeling, yielding 60 clips of varying lengths, which we name VideoHuman60. The combination of these datasets is used for video segmentation pre-training. However, both the synthetic and pseudo-labeled datasets lack fine-grained details like hair or fur. To address this issue, we render 200 4K video clips that feature detailed hair foregrounds, alpha mattes, and dynamic camera movements using Blender~\cite{blender}. As shown in~\Cref{fig:synhairhuman}, our rendered dataset provides high-quality paired video foregrounds and alpha matte annotations.

\paragraph{Details of generated synthetic dataset.}{The synthetic dataset has dynamic backgrounds by composing real-world video backgrounds with the synthetic foreground. For the camera and motion setting, we generate the camera path by adjusting the shape of the different trajectory curves in Blender, ensuring that the camera follows the path. We also simulate the effect of wind on hair, which creates a natural flow effect. We introduce a force field during the rendering process by adding turbulence to the scene, causing the subject's hair to exhibit realistic drifting effects similar to those seen in the real world.}

\subsection{Training Procedures}
\label{sec:training}
To fully utilize the prior knowledge from pre-trained video diffusion models and diverse training datasets, our video matting training is organized into three stages. All stages are trained using the AdamW optimizer across 4 Nvidia A100 GPUs. We employ the SVD model as the pre-trained video diffusion model and freeze the VAE throughout all stages. \Cref{tab:train_sources}
lists all the training datasets.

\noindent\textbf{Stage 1:} In this stage, we fine-tune the pre-trained SVD model on three datasets: BEDLAM, Dynamic Replica, and pseudo-labeled VideoHuman60. The model is trained for one epoch at multiple low resolutions, including $320\times320$, $480\times640$, $512\times512$, and $640\times480$. All parameters of the 3D-UNet are optimized during this phase. To accommodate GPU memory constraints, video lengths are randomly sampled between 2 and 8 frames. The datasets are sampled with the following ratios: BEDLAM ($0.4$), Dynamic Replica ($0.3$), and VideoHuman60 ($0.3$). The learning rate is set to $1e-5$, and only the flow-matching loss is applied in this stage.

\noindent\textbf{Stage 2:} In this stage, we freeze all temporal layers in the 3D-UNet and continue fine-tuning the remaining layers using the BEDLAM, Dynamic Replica, VideoHuman60, and VideoMatte240K datasets. The model is trained at higher resolutions of $768\times768$ and $1024\times1024$. The datasets are sampled with the following ratios: BEDLAM ($0.4$),  Dynamic Replica ($0.3$), VideoHuman60 ($0.1$), and VideoMatte240K ($0.2$). Only flow-matching loss is applied during this phase. The learning rate is maintained at $1e-5$.

\noindent\textbf{Stage 3:} To further improve the fine-grained details of the predictions, we freeze all layers of the 3D-UNet and introduce LoRA~\cite{lora} layers with a rank of 32. Additionally, we incorporate the VideoMatte240K image matting dataset, alongside SynthHairMan, for fine-tuning. The datasets are sampled with the following ratios: VideoMatte240K (0.4), and SynthHairMan (0.6). The batch size is set to 1 per GPU, and the training resolutions include $1024\times1024$, $512\times960$, and $576\times1024$. Both flow-matching and pixel-space losses are employed for rendering in this stage, with flow-matching loss used exclusively for VideoMatte240K. The learning rate is set to $1e$-$4$.

\paragraph{{Training details.}} {All stages are performed across 8 Nvidia A100 80G GPUs. VAE is fixed during all the stages, with a total training time of about two days. The sequence length is randomly sampled from [1, 12] frames, such that the model can learn to generate matte sequences with variable lengths. Note that due to the limitation of the GPU memory, we sample longer sequence lengths for low resolution sequences and shorter sequence lengths for high resolution sequences.
}

\section{Experiments}

\noindent\textbf{Evaluation metrics.} Following RVM \cite{lin2022robust}, we verify the pixel-wise accuracy of the alpha values in each frame by reporting ``MSE'' (mean squared error), ``SAD'' (sum of absolute difference), ``Grad'' (gradient error) and ``Conn'' (completeness) \cite{rhemann2009perceptually}. We employ the temporal coherence metric, namely dtSSD \cite{erofeev2015perceptually} (mean squared difference of direct temporal gradients) to evaluate the accuracy of the predicted video alpha mattes and their temporal coherence. For all metrics, lower is better.

\noindent\textbf{Evaluation datasets.} We evaluate the video matting results on two composition datasets, VideoMatte240K \cite{video240k} test set, and V-HIM60 \cite{huynh2024maggie}. For the VideoMatte240K \cite{video240k} test set, each test clip has 100 frames. Each test sample is composited with 5 video and 5 image backgrounds.

\noindent\textbf{Compared methods.} We compare our method against several state-of-the-art video matting approaches, including RVM \cite{lin2022robust}, MaGGle \cite{huynh2024maggie}, and SparseMat \cite{sun2023ultrahigh}. Like our method, RVM \cite{lin2022robust} performs video matting without requiring auxiliary inputs. However, RVM is primarily trained on real-world datasets, such as the YouTubeVIS (2985 clips) video segmentation dataset, COCO \cite{coco}, SPD \cite{spd} image segmentation datasets, VideoMatte240K, as well as two image matting datasets, Adobe Image Matting \cite{xu2017deep} and Distinctions-646 \cite{Qiao_2020_CVPR}. In contrast, MaGGle \cite{huynh2024maggie} and SparseMat \cite{sun2023ultrahigh} rely on additional guidance masks as input to assist with matting.

\begin{table*}[h!]
    \centering
    \caption{Zero-shot Video Matting Results on V-HIM60.}
    \vspace{-0.5em}
    \footnotesize
        \setlength{\tabcolsep}{3.0pt} %
        \renewcommand{\arraystretch}{1.2} %
        \resizebox{.99\linewidth}{!}{%
        \begin{tabular}{@{}l|rrrrr|rrrrr|rrrrr@{}}
        \toprule

        \multirow{2}{*}{Method} & \multicolumn{5}{c|}{V-HIM60 (Easy)} & \multicolumn{5}{c|}{V-HIM60 (Medium)} & \multicolumn{5}{c}{V-HIM60 (Hard)} \\
        \cline{2-16}
        & MAD $\downarrow$ & MSE $\downarrow$ & Grad $\downarrow$ & Conn $\downarrow$ & dtSSD $\downarrow$ & MAD $\downarrow$ & MSE $\downarrow$ & Grad $\downarrow$ & Conn $\downarrow$ & dtSSD $\downarrow$ & MAD $\downarrow$ & MSE $\downarrow$ & Grad $\downarrow$ & Conn $\downarrow$ & dtSSD $\downarrow$ \\
        \hline
        
        RVM \cite{lin2022robust} & 6.40 & 2.53 & 14.27 & 11.74 & 4.10 
            & 10.28 & 5.95 & 19.82 & 20.66 & 4.56 
            & 17.96 & 11.51 & 20.63 & 34.52 & 7.03 \\

        SparseMat \cite{sun2023ultrahigh} & 6.51 & 3.25 & 13.41 & 11.71 & 4.52
            & 10.08 & 6.07 & 16.91 & 19.79 & 5.07
            & 23.48 & 16.53 & 20.86 & 44.64 & 9.76 \\

        MaGGle \cite{huynh2024maggie} & 10.48 & 7.52 & 13.71 & 19.99 & 5.19
            & 26.16 & 22.57 & 19.61 & 71.29 & 7.14
            & 25.68 & 21.01 & 21.18 & 49.78 & 11.30 \\
        \hline
        Ours & 6.24 & 1.01 & 10.13 & 10.55 & 3.26
        & 10.13 & 1.68 & 15.00 & 13.61 & 3.55  
        & 11.04 & 1.55 & 13.52 & 13.16 & 4.11 \\
        \bottomrule

    \end{tabular}}
    
    \label{tab:vhim_video_matting}
    \vspace{-1.0em}
    
\end{table*}

\begin{table}[ht]
    \centering
    \caption{Our training data sources.}
    \label{tab:train_sources}
    \small
    \begin{tabular}{lccccc}
        \toprule
        Dataset & Cate & Type & Anno & Seq \\\midrule
        BEDLAM & human & Synthetic & Seg & 10450 \\
        Dynamic Replica  & human+animal & Synthetic & Seg & 524 \\
        VideoMatte240K & human & Composition & Matte & 478 \\
        Our VideoHuman60 & human & Pseudo & Seg & 60 \\
        Our SynHairMan & human & Synthetic & Matte & 200 \\
        \bottomrule
    \end{tabular}
    \vspace{-1.5em}
\end{table}

\begin{table}[ht]
    \centering
    \caption{Zero-shot evaluation on the AM-2k Dataset. We test trimap-free portrait matting models on the animal dataset AM-2K. All methods are not trained on the AM-2K training set.}
    \label{tab:zero_shot_am2k}
    \small
    \begin{tabular}{lccc}
        \toprule
         & MSE & MAD & SAD  \\\midrule
        MODNet \cite{MODNet} & 66.1 & 116.3 & 24.1 \\
        P3M \cite{Li-2021-P3M} & 50.7 & 80.1 & 16.7 \\
        ViTAE-S \cite{MA-2023-P3M-ViTAE} & 54.3 & 81.0 & 16.7 \\
        AlphaLDM \cite{genmatte} & 26.7 & 57.3 & 11.4 \\
        \midrule
        \Ours  & \textbf{20.4} & \textbf{24.3} & \textbf{10.2} \\
        \bottomrule
    \end{tabular}
    \vspace{-1.5em}
\end{table}

\begin{table}[t]
    \centering
    \caption{Zero-shot Evaluation on the P3M-500 Dataset. Our method achieves better performance when compared with other video matting methods. All the methods are not trained with P3M training set.}
    \label{tab:zero_shot_p3m_image_matting}
    \setlength{\tabcolsep}{2.0pt} %
    \resizebox{\linewidth}{!}{%
        \begin{tabular}{l|rrrrr|rrrrr}
        \toprule

        \multirow{2}{*}{Method} & \multicolumn{5}{c|}{P3M-500-NP} & \multicolumn{5}{c}{P3M-500-P} \\
        \cline{2-11}
        & MAD $\downarrow$ & MSE $\downarrow$ & SAD $\downarrow$ & Grad $\downarrow$ & Conn $\downarrow$ & MAD $\downarrow$ & MSE $\downarrow$ & SAD $\downarrow$ & Grad $\downarrow$ & Conn $\downarrow$ \\
        
        \hline

        RVM \cite{lin2022robust} & 10.89 & 6.57 & 18.91 & 14.97 & 18.81 & 12.34 & 7.84 & 21.13 & 17.93 & 21.26   \\
        
        SparseMat \cite{sun2023ultrahigh} & 11.14 & 6.36 & 19.14 & 13.47 & 18.13 & 13.31 & 8.39 & 22.91 & 16.94 & 22.11  \\
         
        MaGGle \cite{huynh2024maggie} & 99.76 & 95.21 & 174.67 & 26.73 & 72.83 & 77.76 & 73.76 & 134.11 & 27.82 & 61.18 \\

        \hline
        Ours & \textbf{8.39} & \textbf{4.89} & \textbf{14.58} & \textbf{12.89} & \textbf{14.21} & \textbf{9.48} & \textbf{8.51} & \textbf{19.2} & \textbf{15.24} & \textbf{16.12} \\
        \bottomrule
        \end{tabular}
    }
    \vspace{-1.5em}
\end{table}

\begin{table}[th!]
    \centering
    \caption{Video Matting on VideoMatte240K.}
    \vspace{-0.5em}
    \footnotesize
        \setlength{\tabcolsep}{3.0pt} 
        \renewcommand{\arraystretch}{1.2} 
        \resizebox{.99\linewidth}{!}{%
        \begin{tabular}{@{}l|rrrrr@{}}
        \toprule
        \multirow{2}{*}{Method} &
        \multicolumn{5}{c}{VM} \\
        \cline{2-6}
        & MAD $\downarrow$ & MSE $\downarrow$ & Grad $\downarrow$ & Conn $\downarrow$ & dtSSD $\downarrow$ \\
        \hline
        
        RVM \cite{lin2022robust} & 6.39 & 1.82 & 9.95 & 6.53 & 1.85  \\
        SparseMat \cite{sun2023ultrahigh} & 441.49 & 270.74 & 166.68 & 908.79 & 6.25 \\
    
        MaGGle \cite{huynh2024maggie} & 36.22 & 32.02 & 12.64 & 68.22 & 2.45 \\
        \hline
        Ours & 5.88 & 1.71 & 5.00 & 5.27 & 1.11 \\
        \bottomrule

    \end{tabular}}
    \label{tab:vm_video_matting}
    \vspace{-1.5em}
\end{table}

\begin{table}[]
    \centering
    \caption{Data ablation study. We train our method on different training sets and evaluate the performance on VideoMatte240: (1) Training only on the SynHairMan matting dataset; (2) Training only on the BEDLAM segmentation dataset; (3) Training on both the BEDLAM segmentation dataset and SynHairMan matting dataset; (4) Training on Pseudo-labeled VideoHuman60, BEDLAM and SynHairMan datasets.
    }
    \label{tab:ablation}
    \small
    \begin{tabular}{lrrrrr}
        \toprule
         & MAD$\downarrow$ & MSE$\downarrow$ & Grad$\downarrow$ & Conn$\downarrow$ & dtSSD$\downarrow$ \\\midrule
         (1) & 44.61 & 41.2 & 19.58 & 72.14 & 5.13 \\
         (2) & 10.24 & 4.87 & 12.9 & 8.95 & 1.68 \\
         (3) & 7.12 & 1.97 & 10.46 & 6.95 & 1.30 \\
         (4) & 7.04 & 1.93 & 10.25 & 6.39 & 1.24 \\
        \bottomrule
    \end{tabular}
    \vspace{-1.5em}
\end{table}

\begin{table}[]
    \centering
    \caption{Training strategy ablation study. We implement four variants of our method and conduct the ablation study on VideoMatte240: (1) Our model without using the pre-trained SVD weights; (2)  Our model without using image-space losses; (3) Our model finetuned with LoRA.
    }
    \label{tab:training_ablation}
    \small
    \resizebox{\linewidth}{!}{%
    \begin{tabular}{lrrrrr}
        \toprule
         & MAD$\downarrow$ & MSE$\downarrow$ & Grad$\downarrow$ & Conn$\downarrow$ & dtSSD$\downarrow$ \\\midrule
        Base model & 7.04 & 1.93 & 10.25 & 6.39 & 1.24 \\
        - (1) \emph{w/o} diffusion prior & - & - & - & - & -\\
        - (2) \emph{w/o} Image-space losses & 7.36 & 2.04 & 10.59  & 7.04 & 1.39 \\
        + (3) LoRA & 6.13 & 1.74 & 8.69 & 5.62 & 1.23 \\
        \bottomrule
    \end{tabular}
    }
\vspace{-1.5em}
\end{table}

\subsection{Evaluation on Composition Datasets}

\Cref{tab:vhim_video_matting} presents the video matting results on the V-HIM dataset, which is divided into three subsets based on varying levels of difficulty. \Cref{tab:vm_video_matting} shows the video matting results on the VideoMatte240K dataset. When compared to both mask-free methods, e.g., RVM \cite{lin2022robust}, and mask-guided matting methods, our approach consistently outperforms the competition, reducing errors across all the evaluation metrics. Notably, our model excels in capturing fine details, as evidenced by its top scores in the Grad metric across all test sets. Our method achieves the best temporal consistency by getting the lowest dtSSD metric across all the evaluation subsets. These findings highlight the robustness of our model, particularly in complex scenarios that demand both high temporal consistency and precise detail preservation, making it a strong contender for advanced video matting tasks.

\subsection{Evaluation on Image Datasets}
\begin{figure*}
\small
\centering
\begin{tabular}{c@{\hspace{2.mm}} @{\hspace{-1.5mm}}c@{\hspace{2.mm}}@{\hspace{-1.5mm}}c@{\hspace{2.mm}}}
\includegraphics[width=0.67\columnwidth]{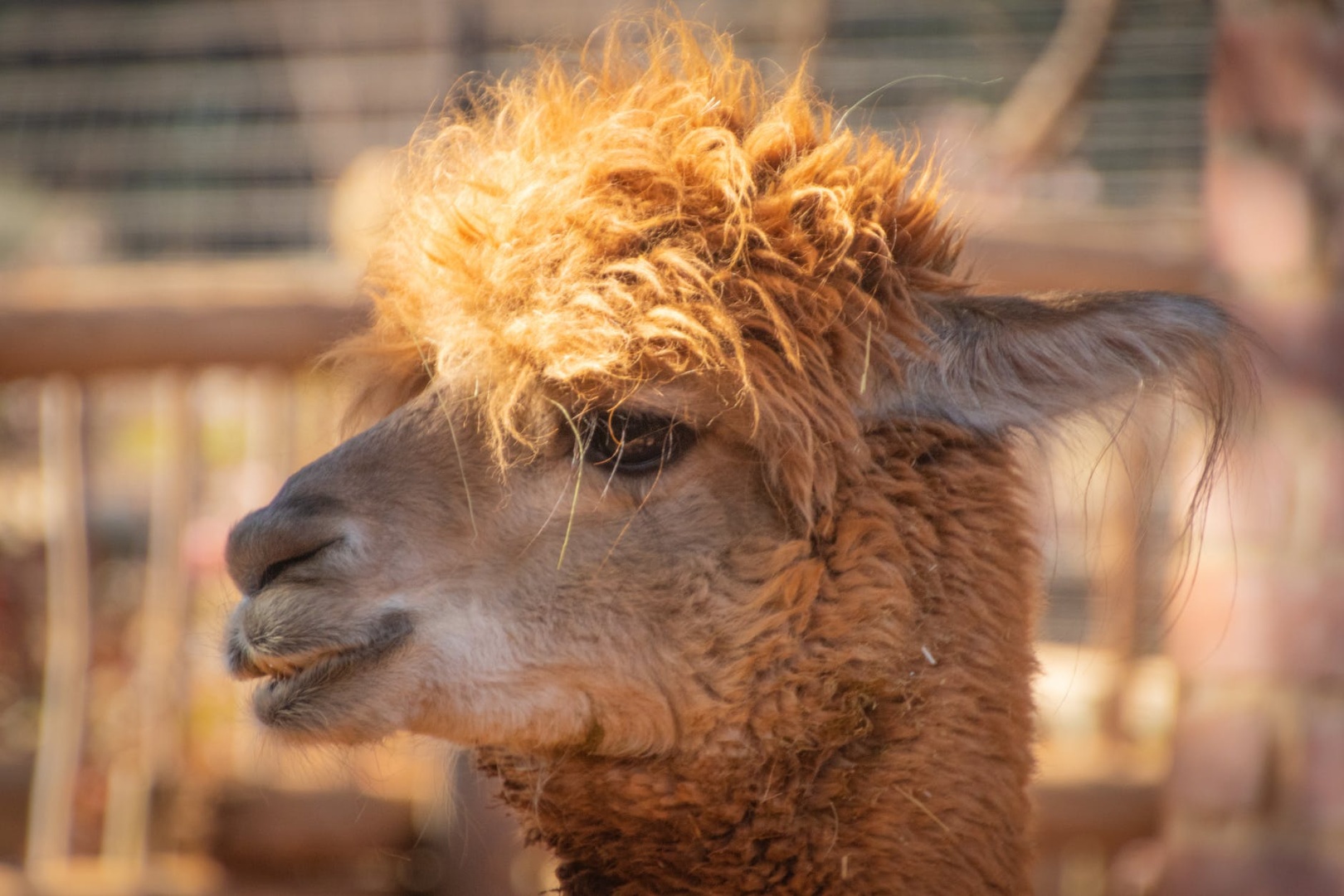} & 
\includegraphics[width=0.67\columnwidth]{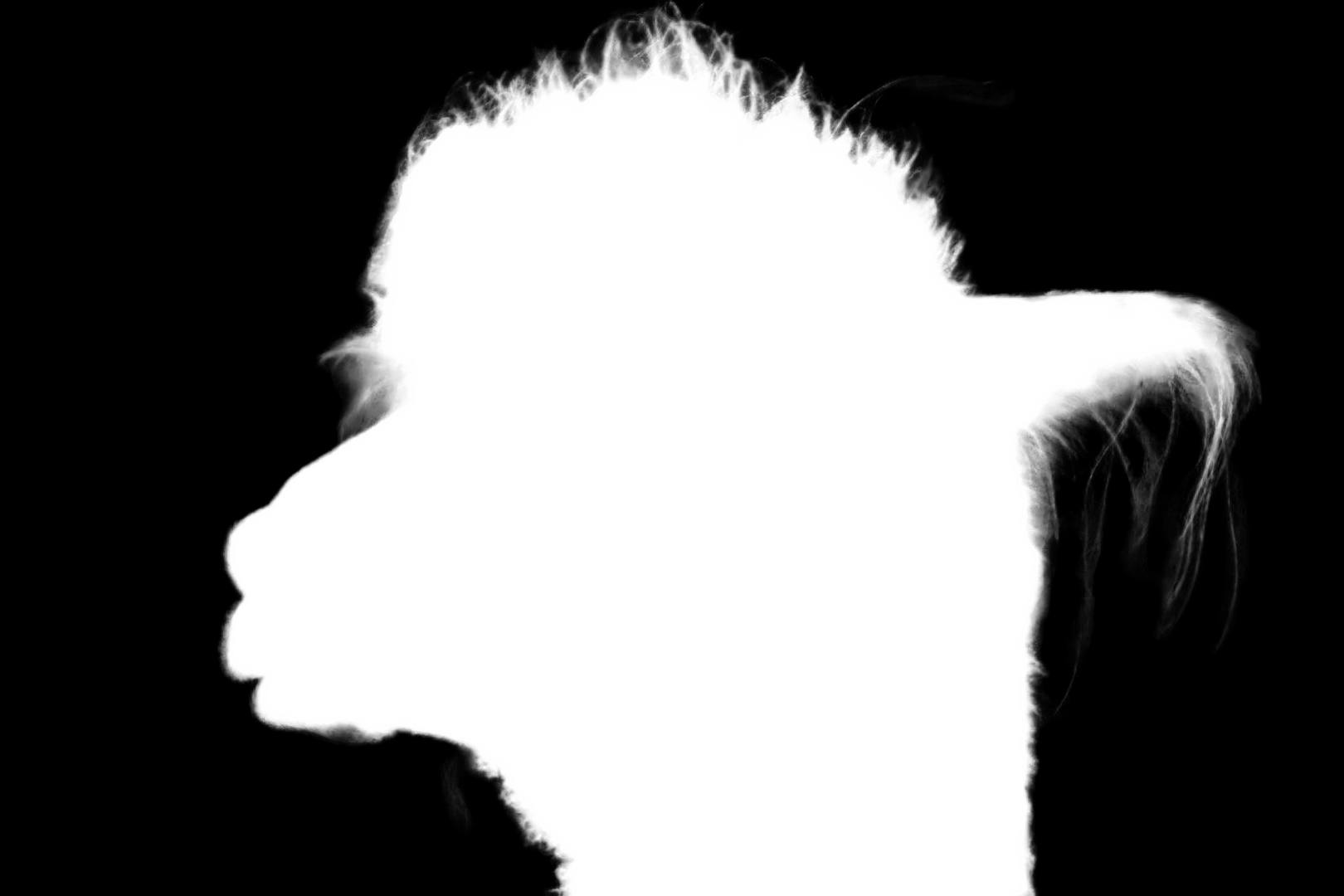} & 
\includegraphics[width=0.67\columnwidth]{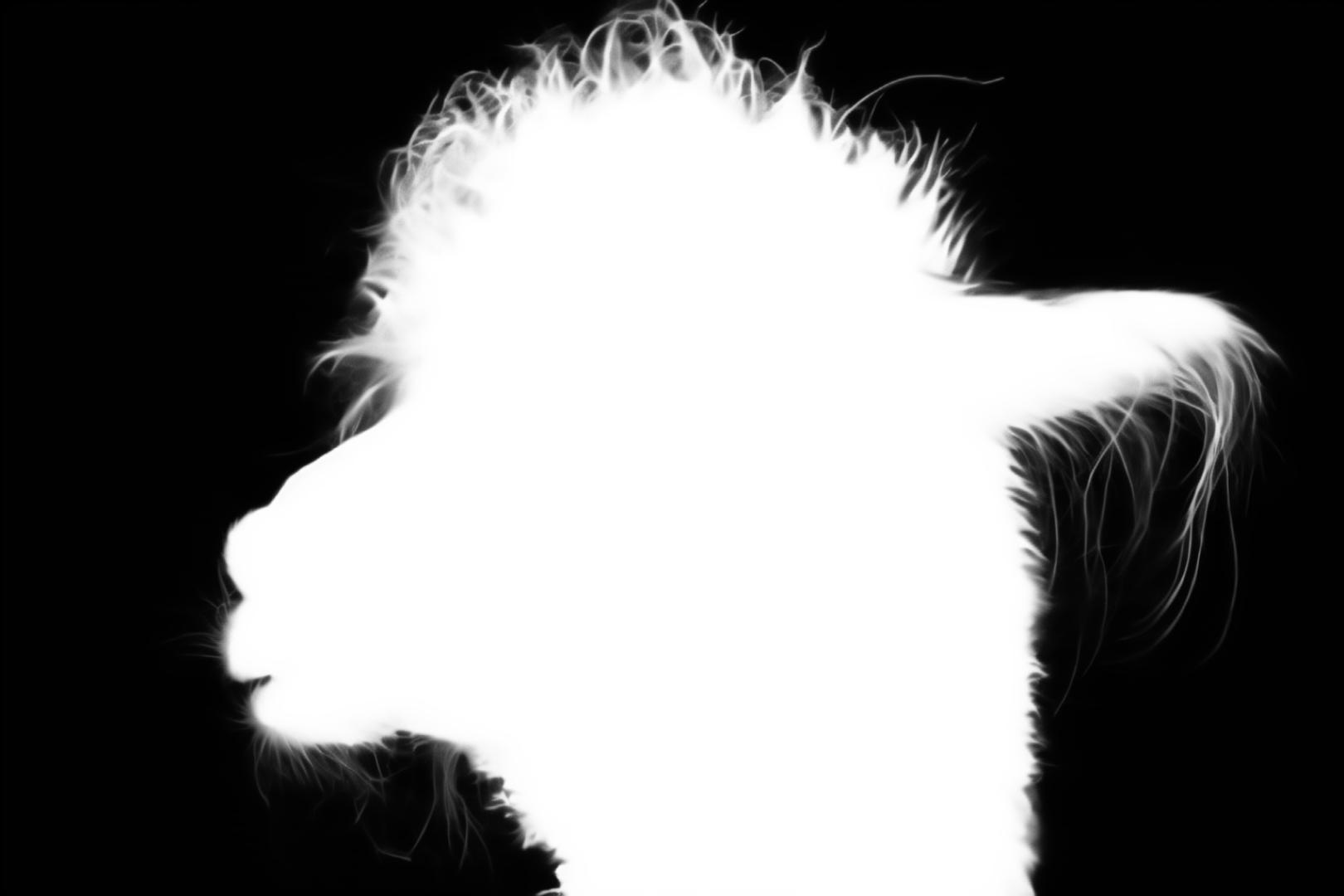} \\
\includegraphics[width=0.67\columnwidth]{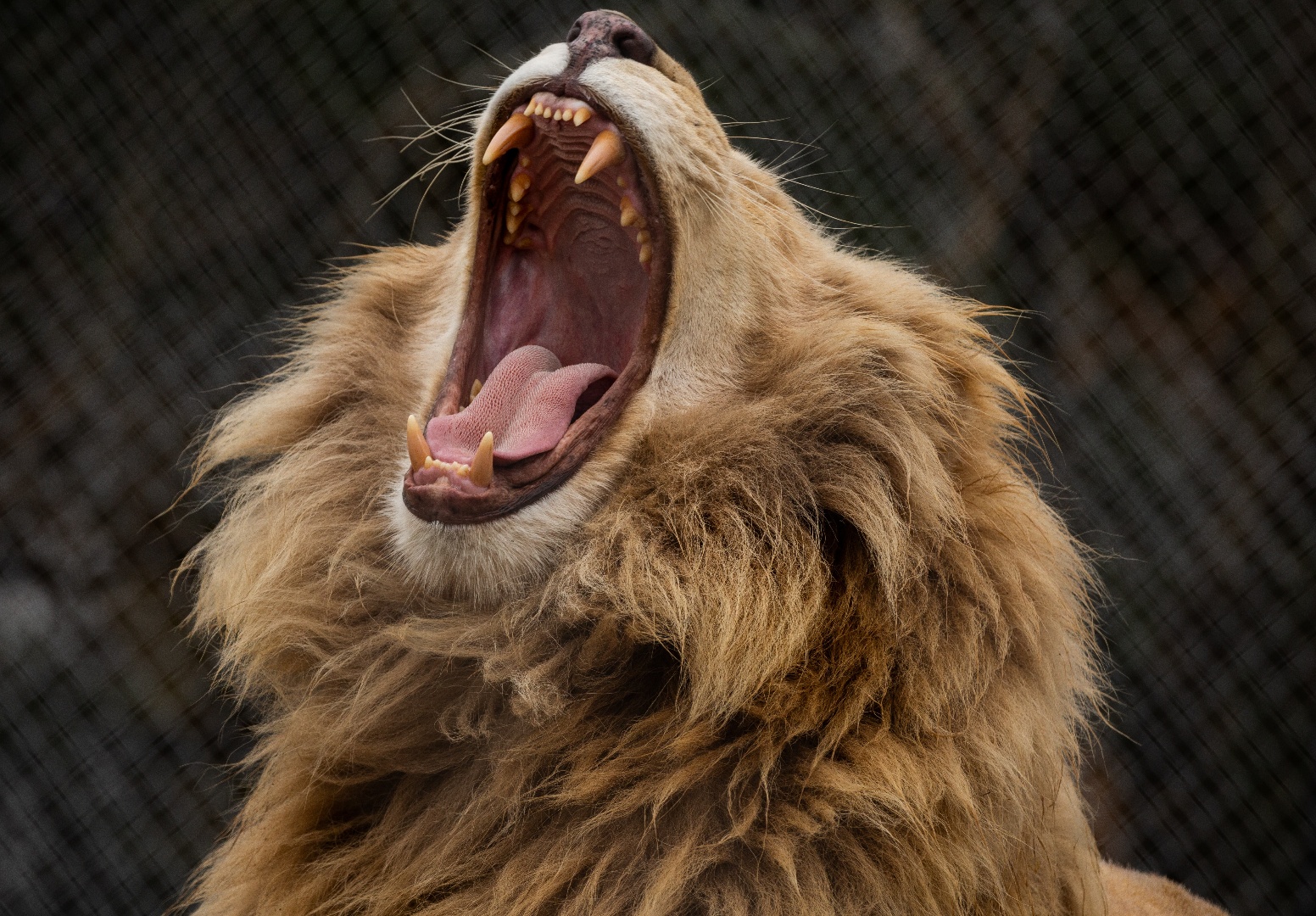} & 
\includegraphics[width=0.67\columnwidth]{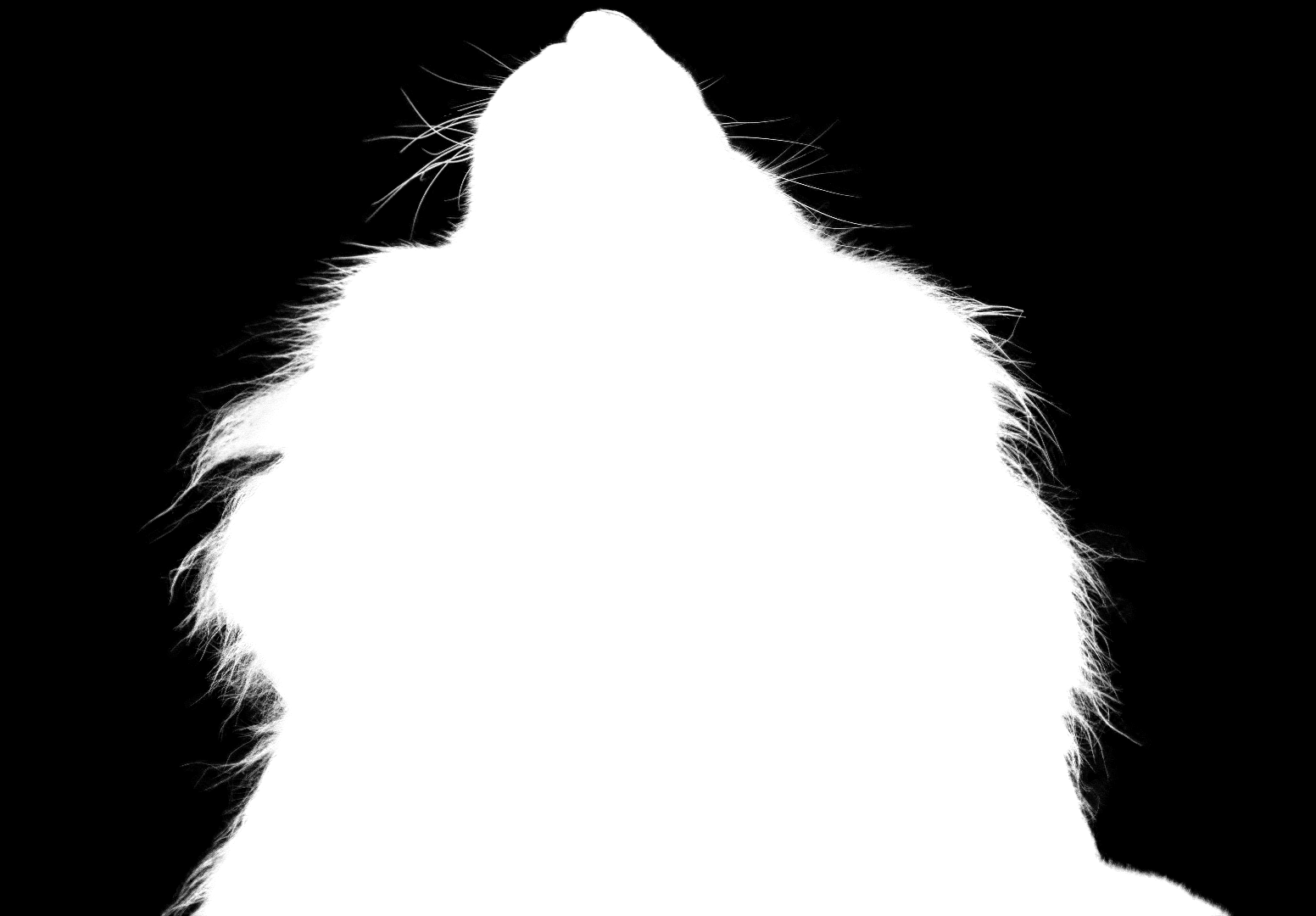} & 
\includegraphics[width=0.67\columnwidth]{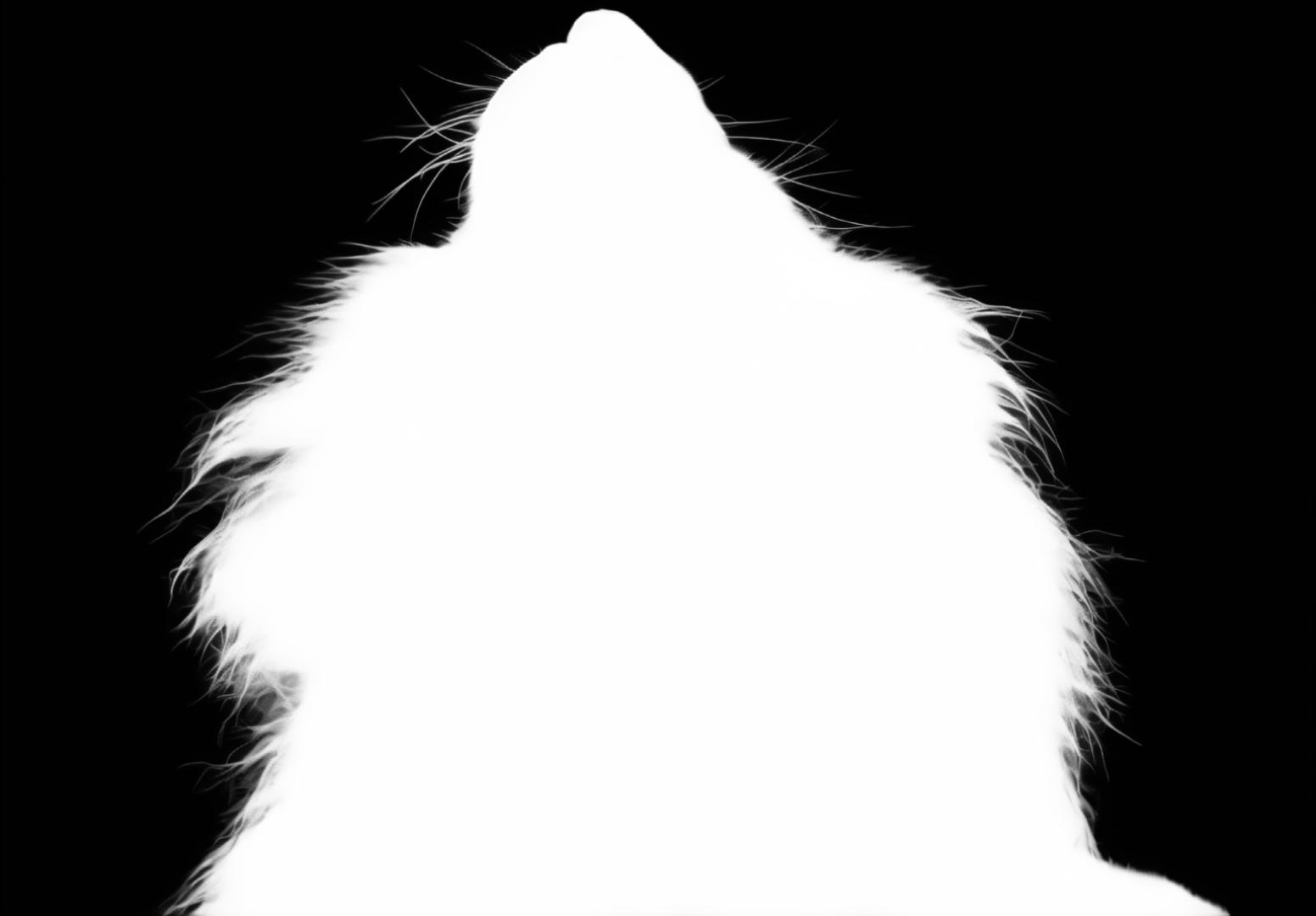} \\
Input & Human annotation & Ours  \\
\end{tabular}
\caption{Visualization on in-the-wild animals. \Ours has great generalization capability even if it is not trained on any animal matting dataset. It sometimes generates more fine-grained details than human annotations (the first row).
}
\label{fig:zero_shot_am2k}
\end{figure*}

In this section, we conduct a zero-shot evaluation on two image matting datasets: the P3M \cite{Li-2021-P3M} portrait matting dataset and the AM-2K \cite{am2k} animal matting dataset. Our method has not been trained on either of these datasets. As shown in~\Cref{tab:zero_shot_p3m_image_matting}, our approach achieves state-of-the-art performance when compared to other video matting methods. Furthermore, as illustrated in~\Cref{tab:zero_shot_am2k}, although our model is not specifically tailored for single-image animal matting, it still outperforms image matting methods that were not trained on the AM-2K training set. These results highlight the impressive generalization ability of our method across various scenes and domains. We also visualize the AM-2K validation results in~\Cref{fig:zero_shot_am2k}.

\subsection{Ablation Studies}
In this section, we choose VideoMatte240K as the zero-shot evaluation dataset.
\paragraph{Dataset ablation.} We perform an ablation study to evaluate the impact of different training datasets on model performance. The model is trained under identical configurations (20,000 iterations with a fixed learning rate of $1e-5$) across various datasets. The results reveal the following key observations: (1). Domain Gap of SynHairMan: Training solely on SynHairMan demonstrates a noticeable domain gap when evaluated on real-world test datasets, highlighting the limitations of synthetic matting data in capturing real-world complexities. (2). Importance of High-Quality Video Segmentation Data: Incorporating high-quality video segmentation datasets, such as BEDLAM, proves essential for achieving robust video matting performance. (3). Effectiveness of Mixed Synthetic Matting Data: Combining synthetic matting datasets with BEDLAM significantly enhances the zero-shot generalization ability of the model, bridging the gap between synthetic and real-world scenarios. (4). Benefit of Mixing Pseudo-Labeled and Synthetic Data: Integrating pseudo-labeled datasets with synthetic data further boosts the model's zero-shot performance, demonstrating the complementary nature of these datasets in improving generalization.

\paragraph{Training strategy ablation.} In~\Cref{tab:training_ablation}, we conduct an ablation study to evaluate the impact of different training strategies on the performance of our model. The results highlight the following key findings: (1). Importance of Pre-trained Diffusion Priors: Pre-trained diffusion priors play a critical role in video matting fine-tuning. Without these priors, the model fails to converge, underscoring their significance in guiding the training process. (2). Effectiveness of Image-Space Decoder Losses: Incorporating image-space decoder losses improves the model's performance compared to using only latent losses. This suggests that directly supervising the reconstructed output in image space helps capture finer details and enhances matting quality. (3). LoRA for Performance Enhancement: The integration of LoRA significantly boosts the model's final performance, demonstrating its utility in refining and optimizing the network for video matting tasks. These findings underscore the necessity of leveraging robust priors, effective loss functions, and targeted architectural enhancements for achieving high-quality video matting results.

\section{Discussion and Conclusion}
\noindent\paragraph{Limitations} While our method demonstrates strong generalization capabilities for human and animal matting, its performance on a broader range of objects remains uncertain. Specifically, we have not trained our model on diverse materials such as glasses, water, fire, or dust. As a result, the model's ability to handle these complex objects with different optical properties and transparency behaviors is still an open question and requires further exploration. 

Another potential limitation is the inference speed, although our generative pipeline only requires 1-3 inference steps for denoising, it is still not comparable with the traditional regression-based video matting methods.

\noindent\paragraph{Video Synthetic Defocus} There are many potential applications for our video matting pipeline. One example is its use in creating synthetic defocus (bokeh) effects in videos. As demonstrated in~\Cref{fig:defocus}, our foreground matte can be utilized to apply a synthetic blur to the background. To achieve this effect, we first blur the background based on a depth map generated by Marigold~\cite{ke2023repurposing}, then composite the foreground RGBA layer onto the blurred background. However, a single-layer depth map alone is insufficient for achieving this effect, as it lacks the necessary accuracy around the subject's boundary, where fine details are essential for a natural-looking result.

\noindent\paragraph{Conclusion.} In this work, we reformulate the traditional regression-based video matting as a video generation task. The key innovation of this paper lies in leveraging the rich generative priors from pre-trained video diffusion models to perform video matting through scalable training datasets. Our approach demonstrates robust performance across a range of challenging video scenarios, showcasing its effectiveness in real-world applications.

\noindent\textbf{Acknowledgment:} C.\ Shen is the corresponding author. C.\ Shen's participation was in part supported by Ningbo Science and Technology Bureau (Grant Number 2024Z291).

\begin{figure*}
\small
\centering
\begin{tabular}{@{\hspace{0mm}}c@{\hspace{1.2mm}} @{\hspace{-0.5mm}}c@{\hspace{1.2mm}} @{\hspace{-0.5mm}}c@{\hspace{1.2mm}}}
\includegraphics[width=0.36\columnwidth]{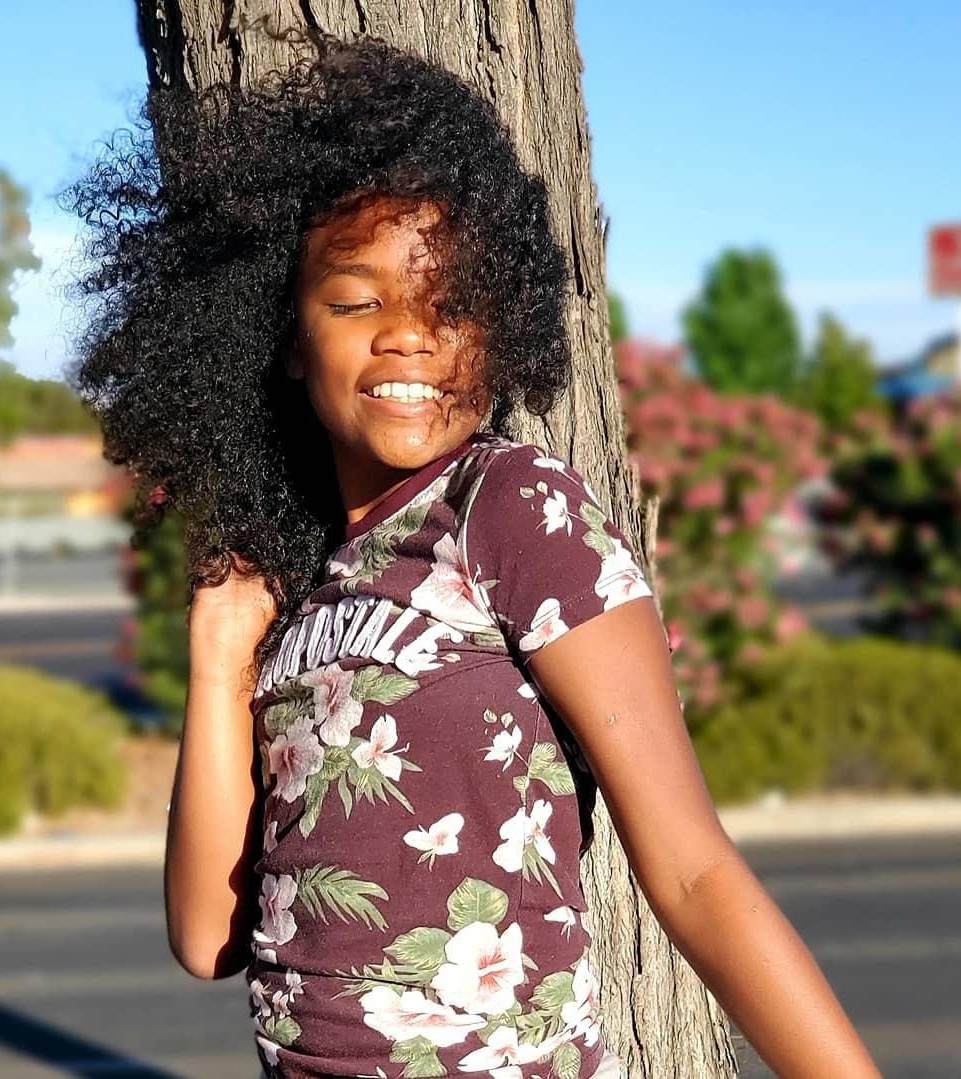} & 
\includegraphics[width=0.36\columnwidth]{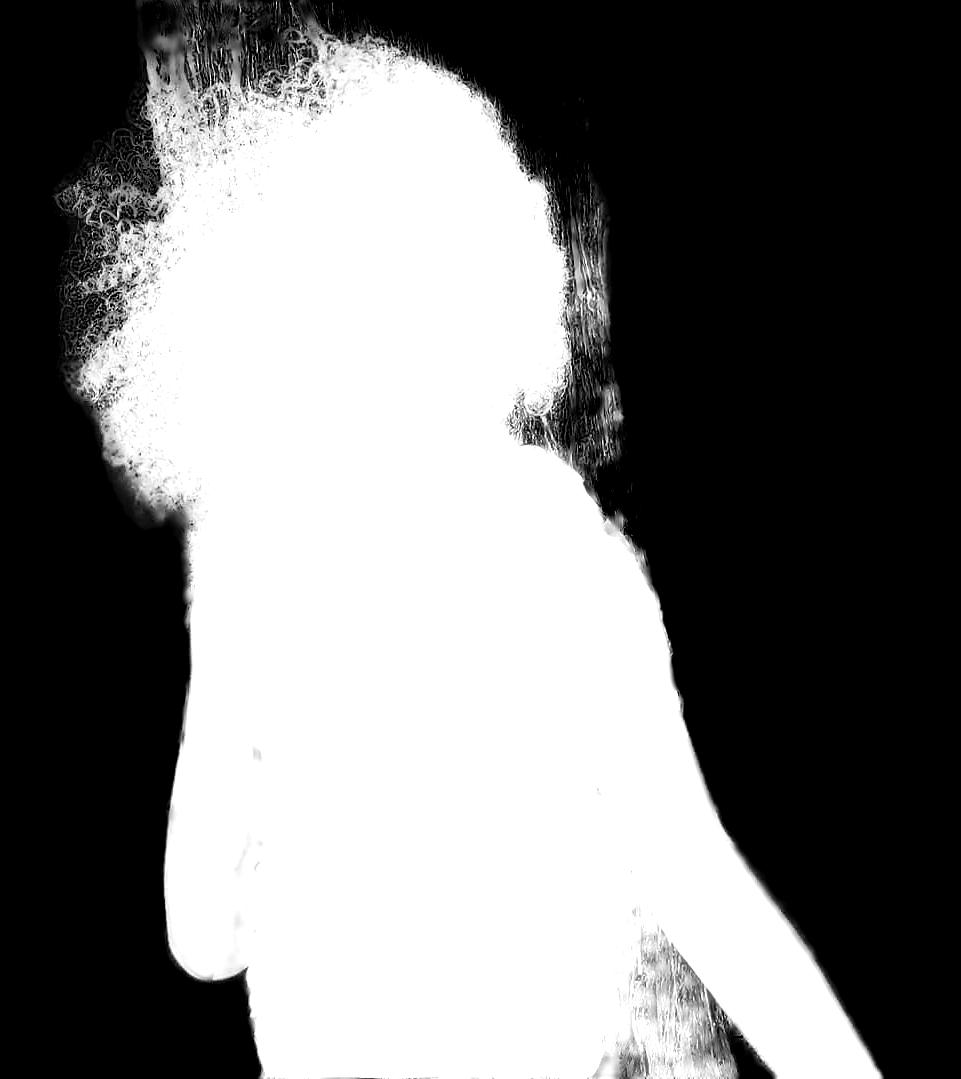} &
\includegraphics[width=0.36\columnwidth]{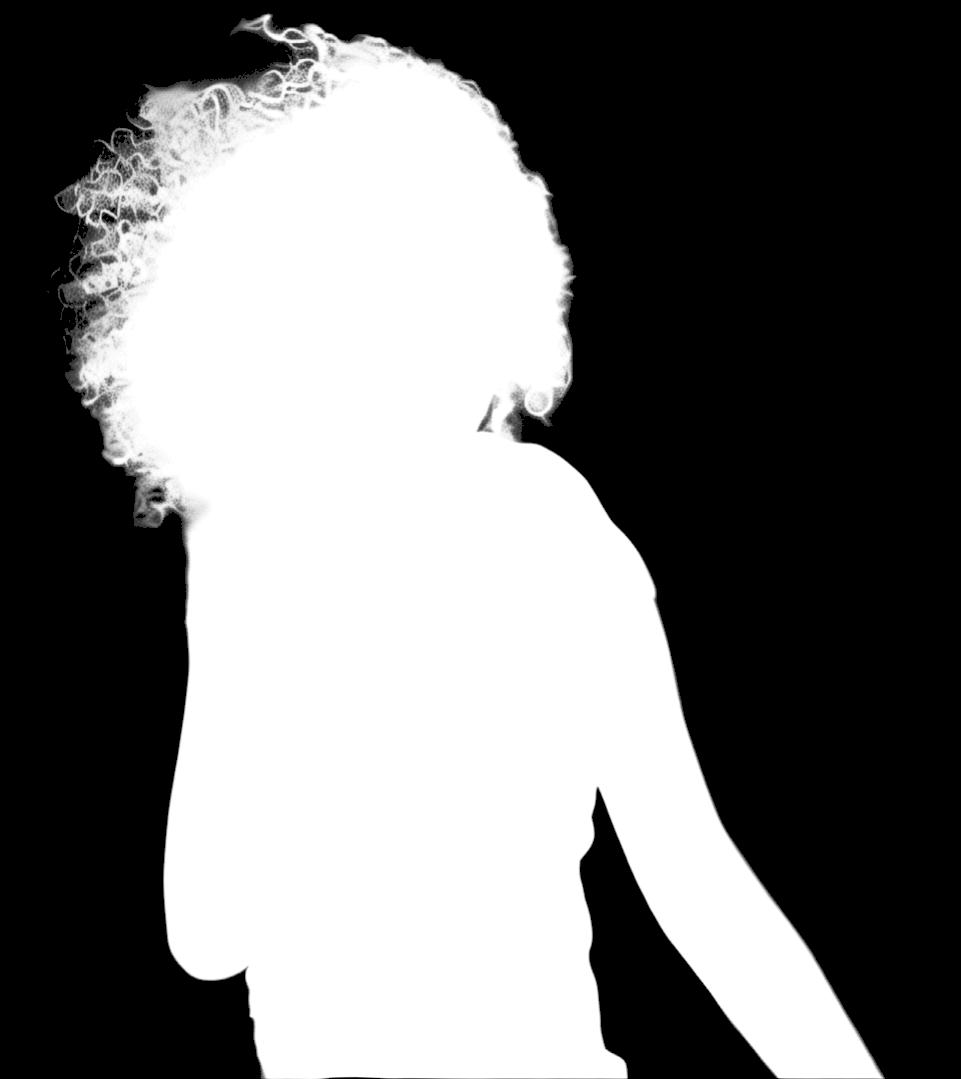} \\
\includegraphics[width=0.36\columnwidth]{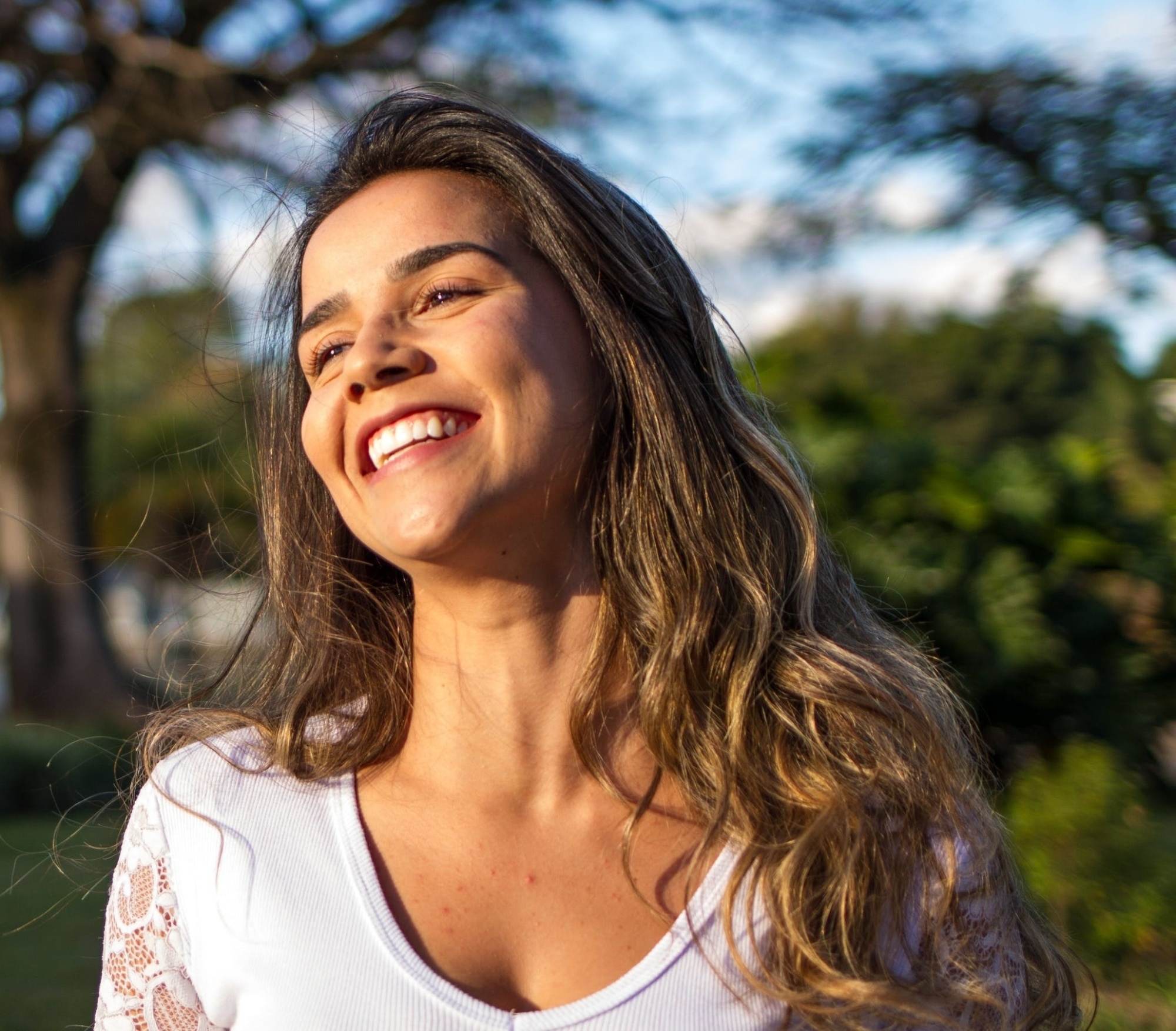} & 
\includegraphics[width=0.36\columnwidth]{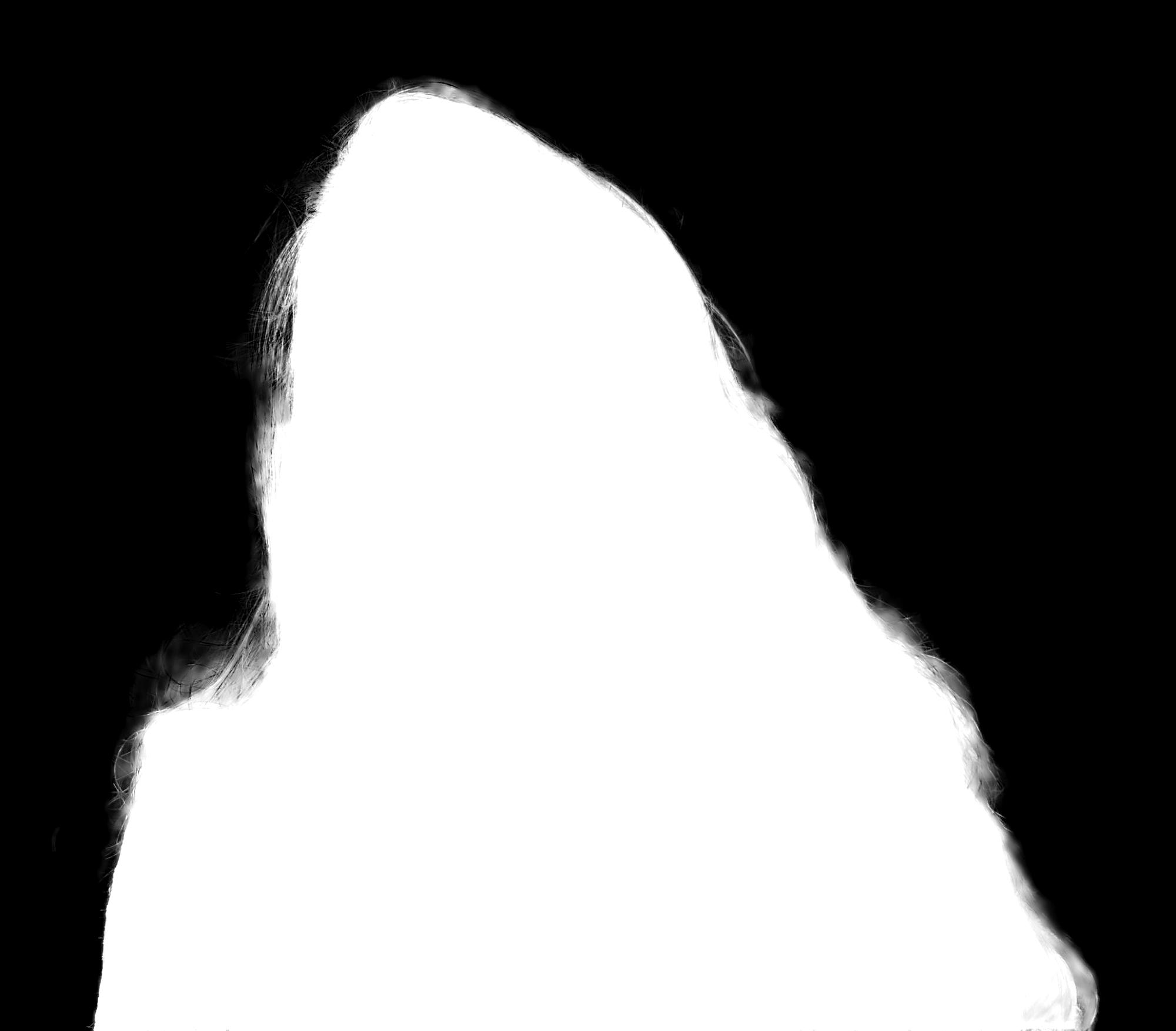} &
\includegraphics[width=0.36\columnwidth]{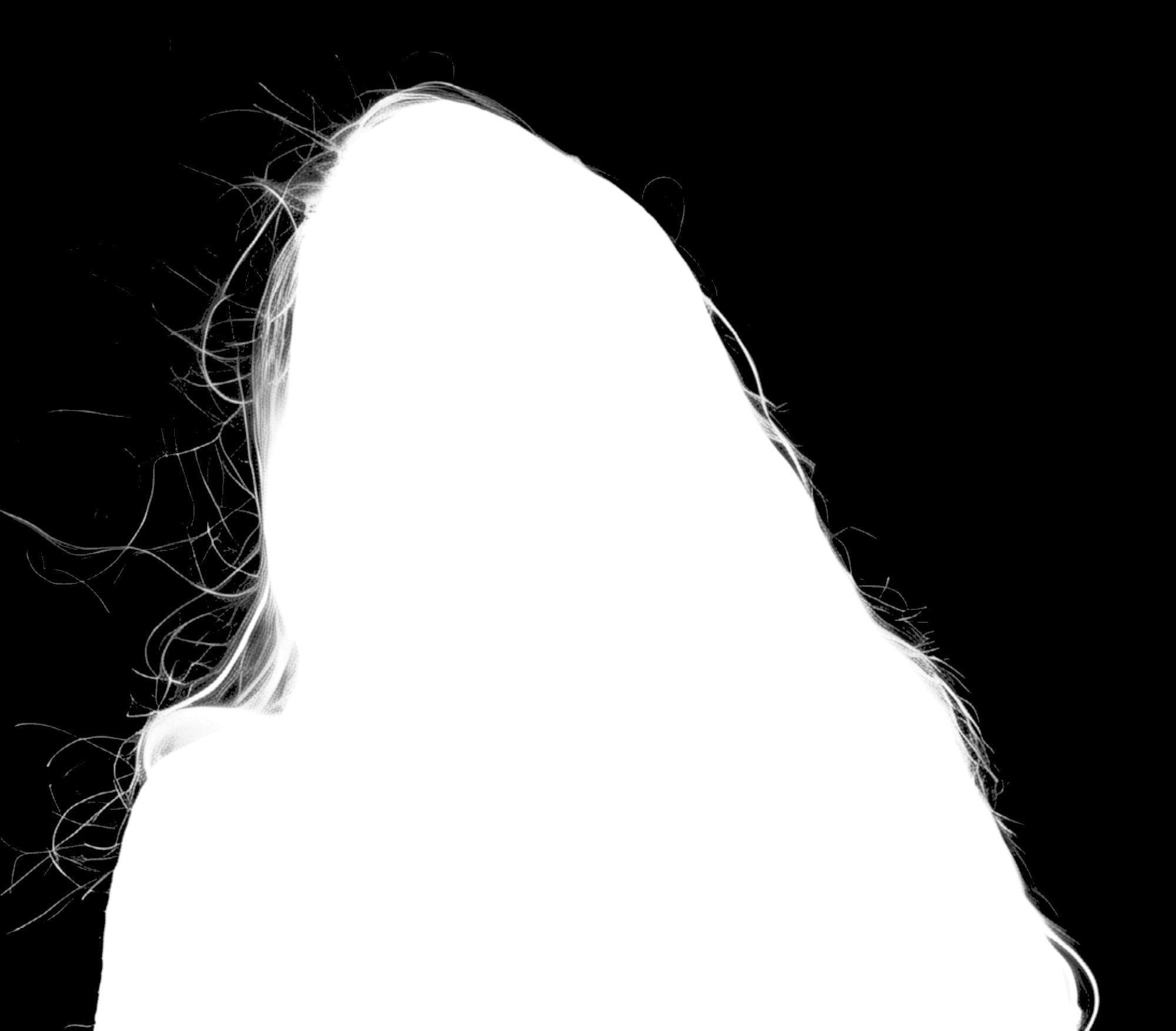} \\
\includegraphics[width=0.36\columnwidth]{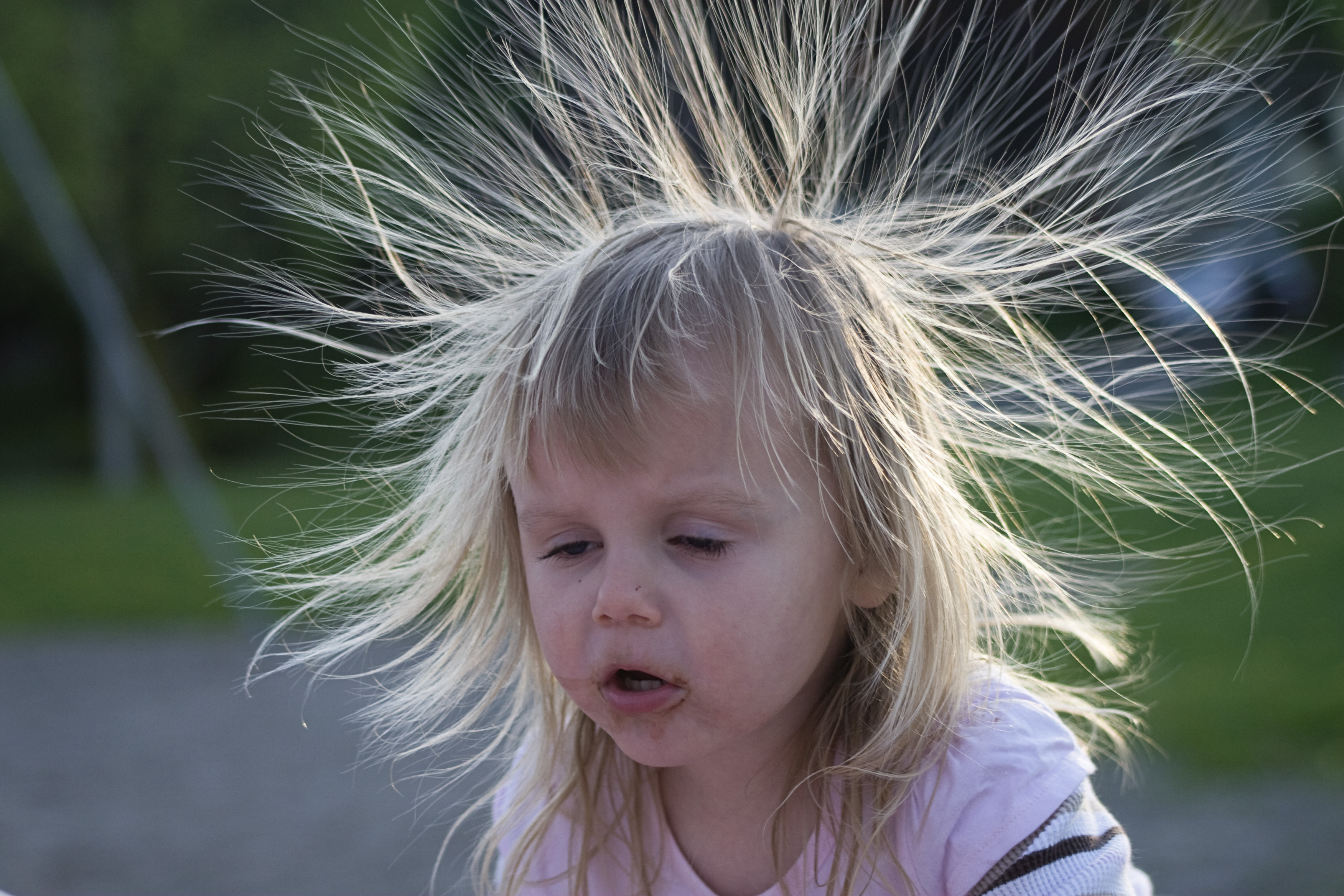} & 
\includegraphics[width=0.36\columnwidth]{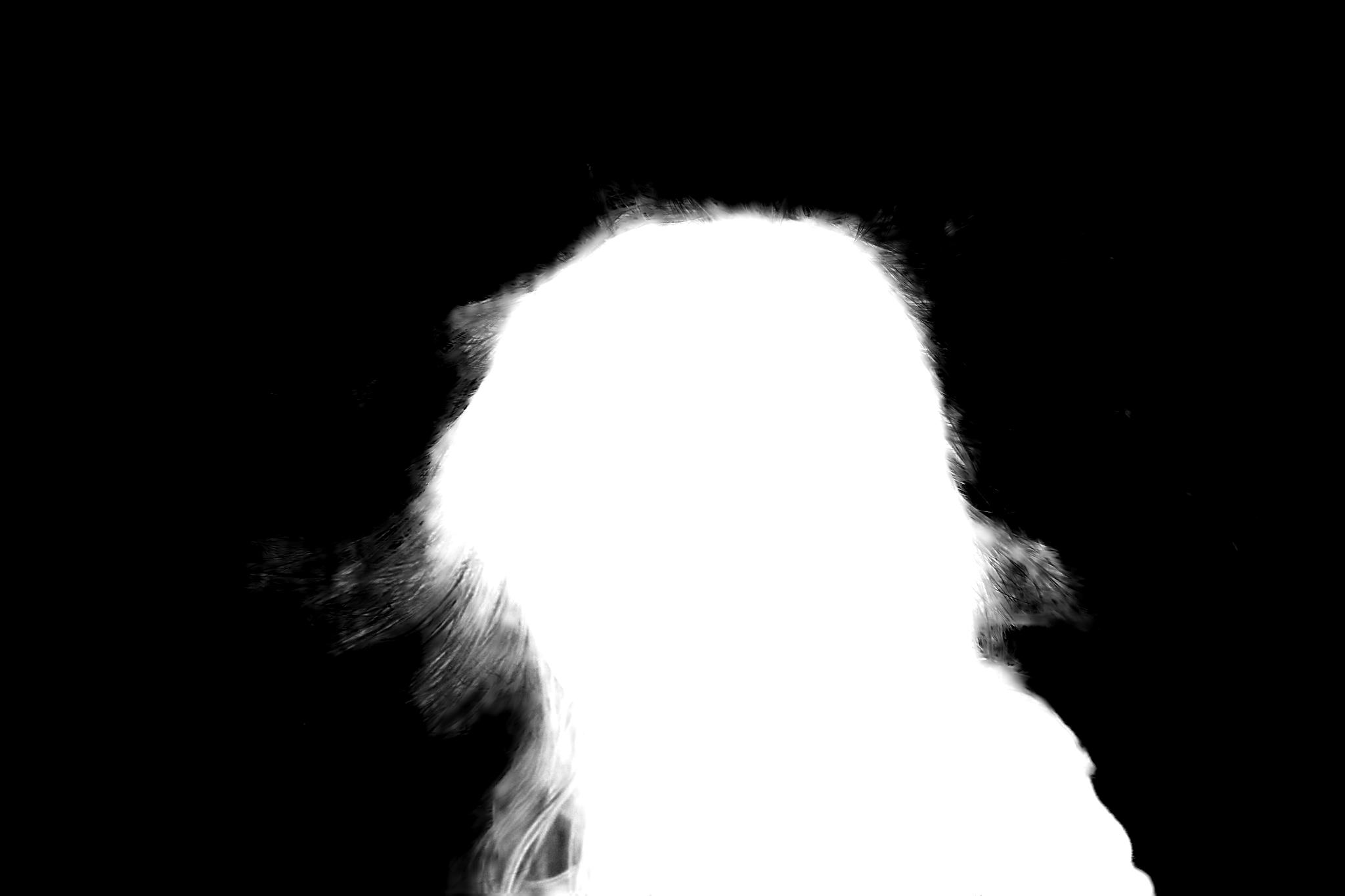} &
\includegraphics[width=0.36\columnwidth]{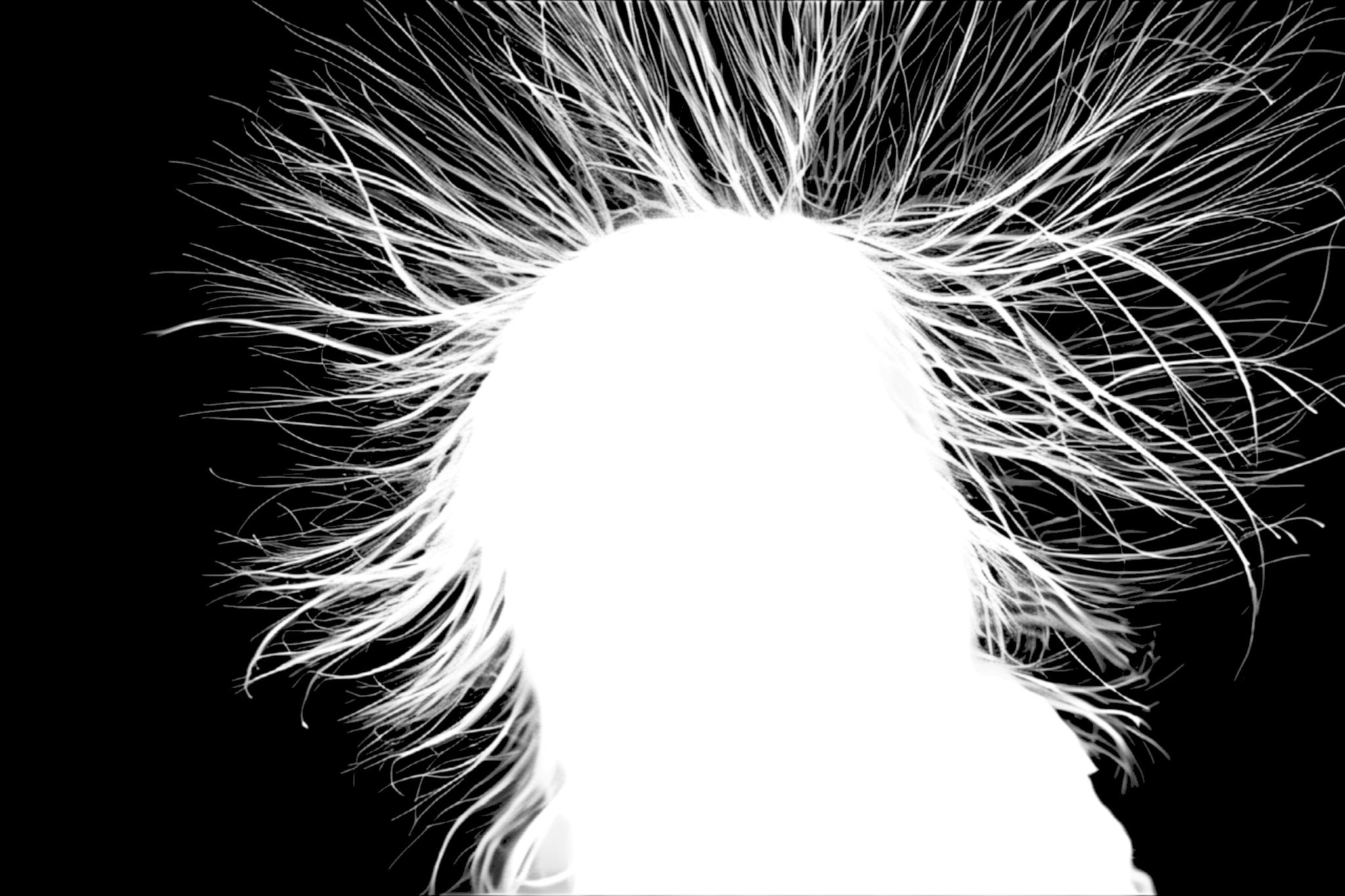} \\
Input frame & RVM prediction & \Ours prediction \\
\end{tabular}
\caption{Qualitative comparison between RVM~\cite{lin2022robust} and \Ours on in-the-wild images with complicated background and human hairs.}
\Description{Comparsion between RVM and \Ours.}
\label{fig:in_the_wild_hair}
\end{figure*}

\begin{figure*}
\small
\centering
\begin{tabular}{
c@{\hspace{1mm}}@{\hspace{0.mm}}c@{\hspace{1mm}}@{\hspace{0.mm}}c@{\hspace{1mm}}@{\hspace{0.mm}}c@{\hspace{1mm}}@{\hspace{0.mm}}c@{\hspace{1mm}}@{\hspace{0.mm}}c@{\hspace{1mm}}
}
\includegraphics[width=0.3\columnwidth]{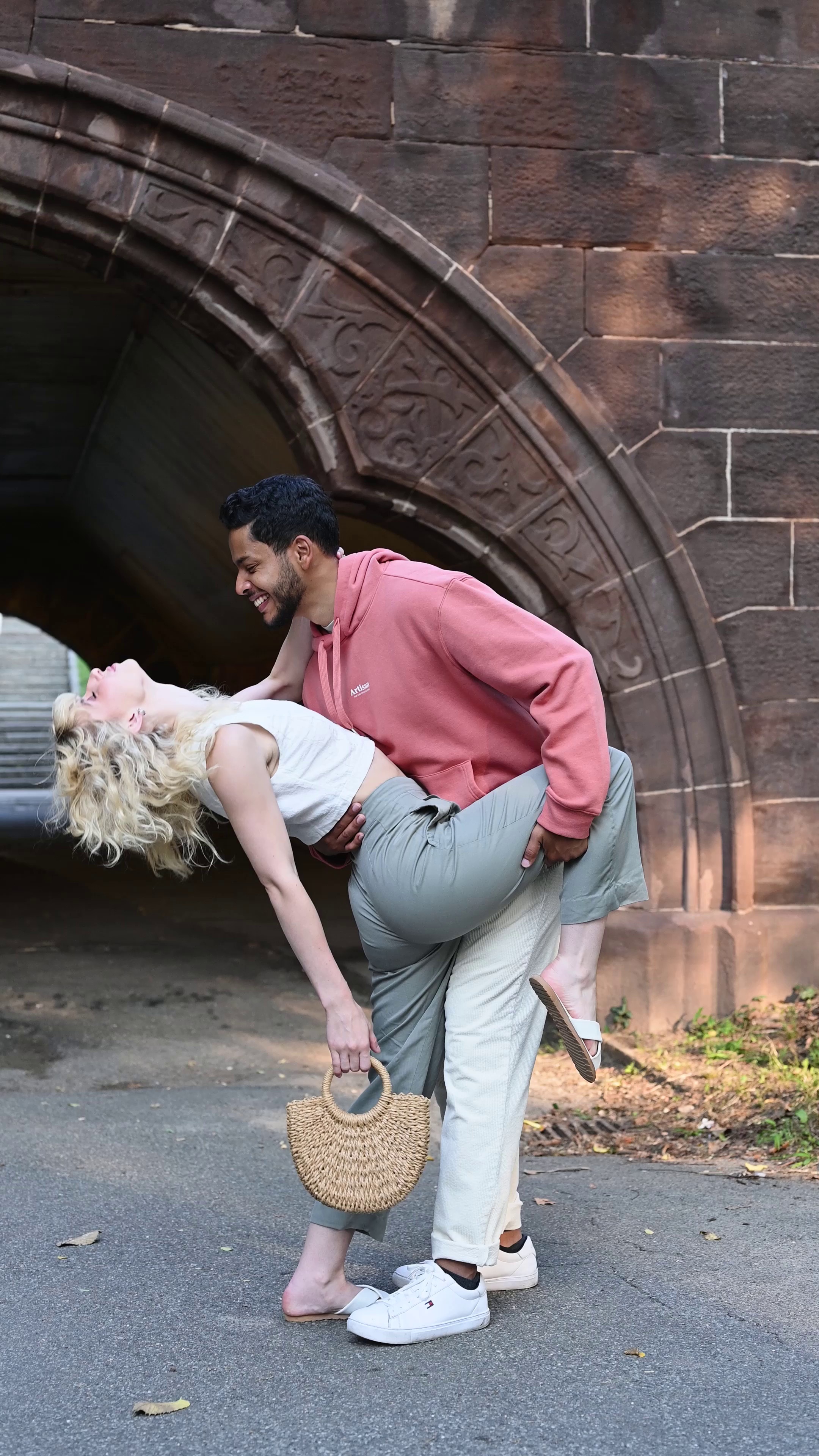} & 
\includegraphics[width=0.3\columnwidth]{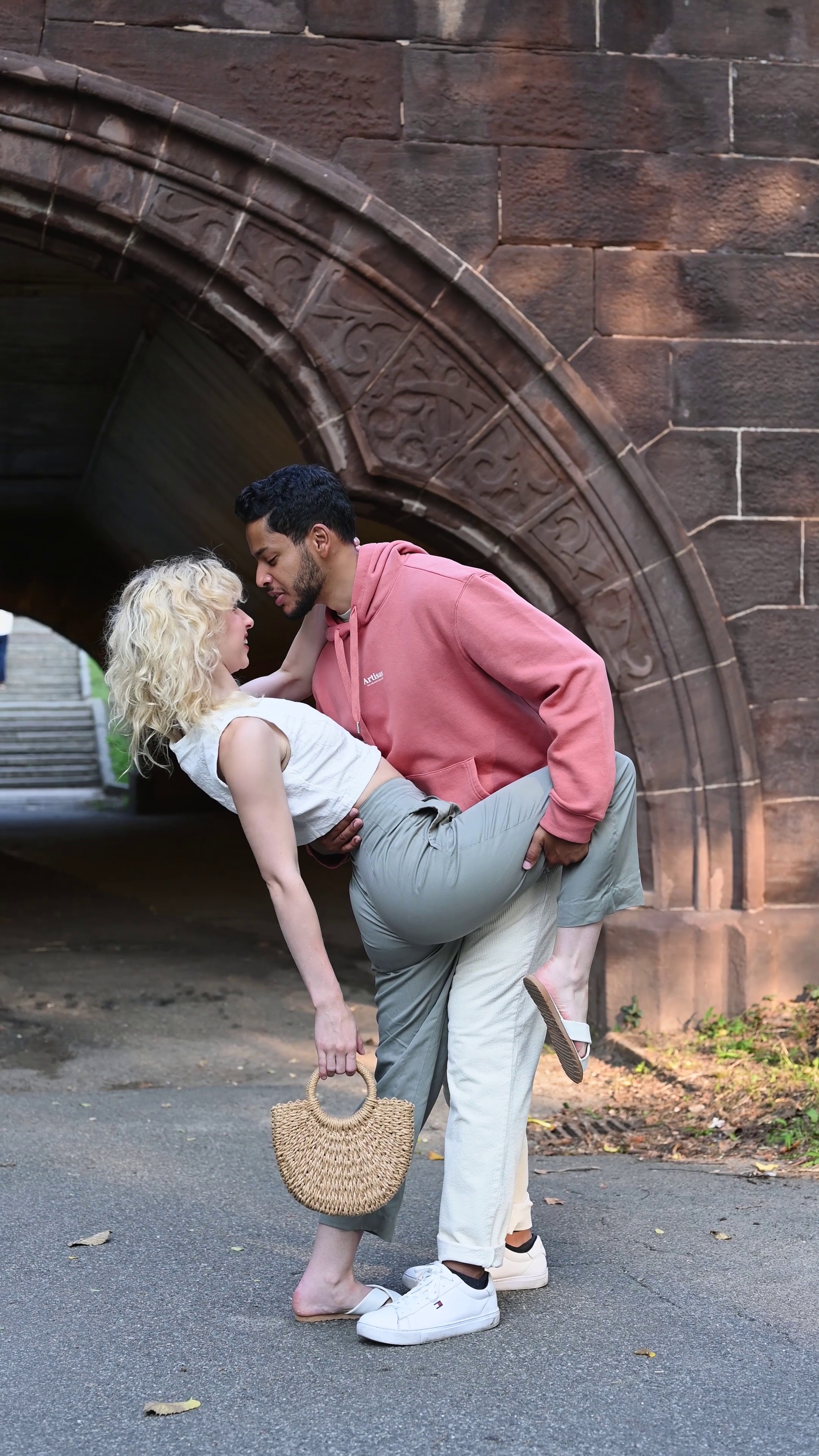} & 
\includegraphics[width=0.3\columnwidth]{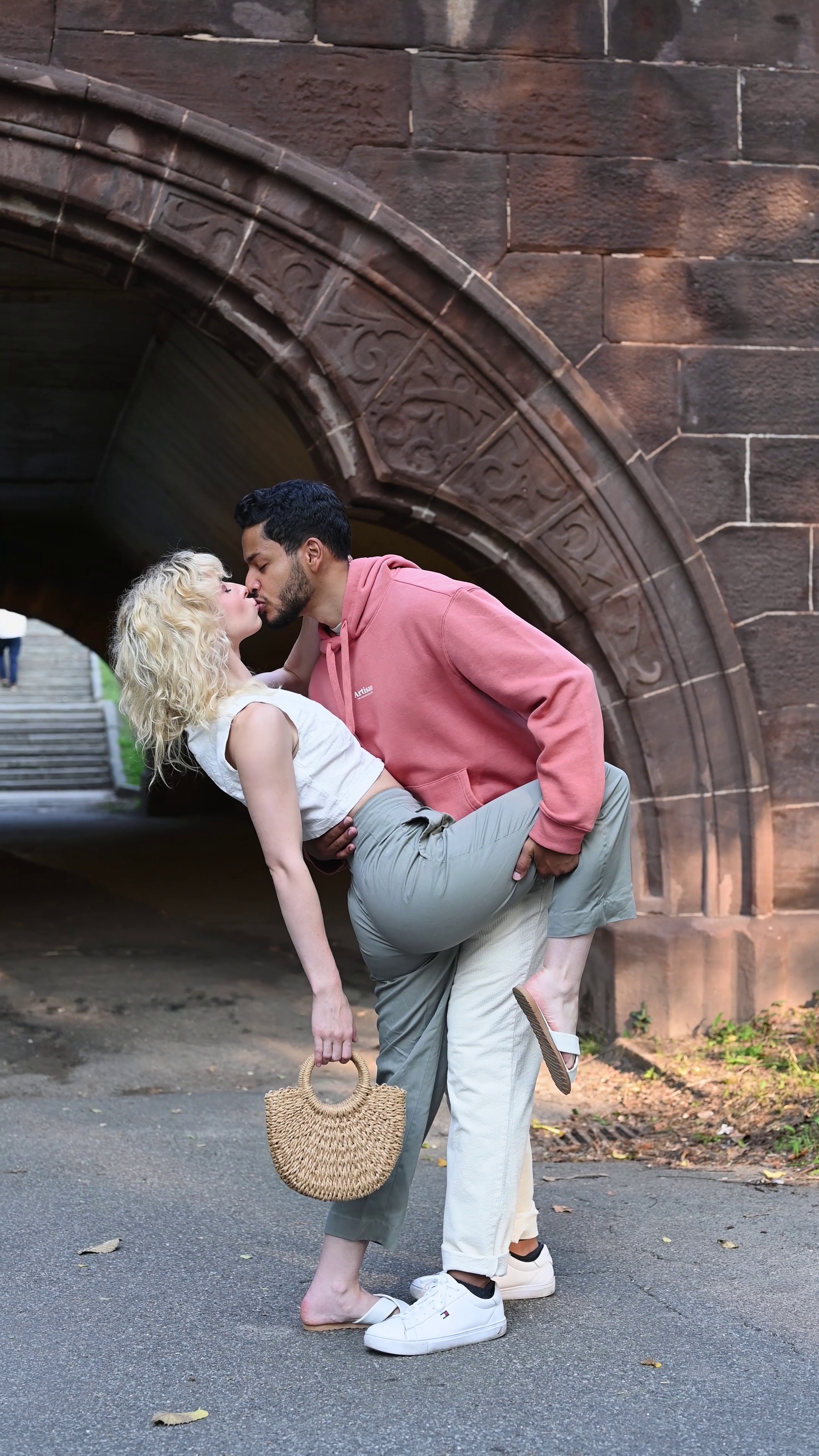} & 
\includegraphics[width=0.3\columnwidth]{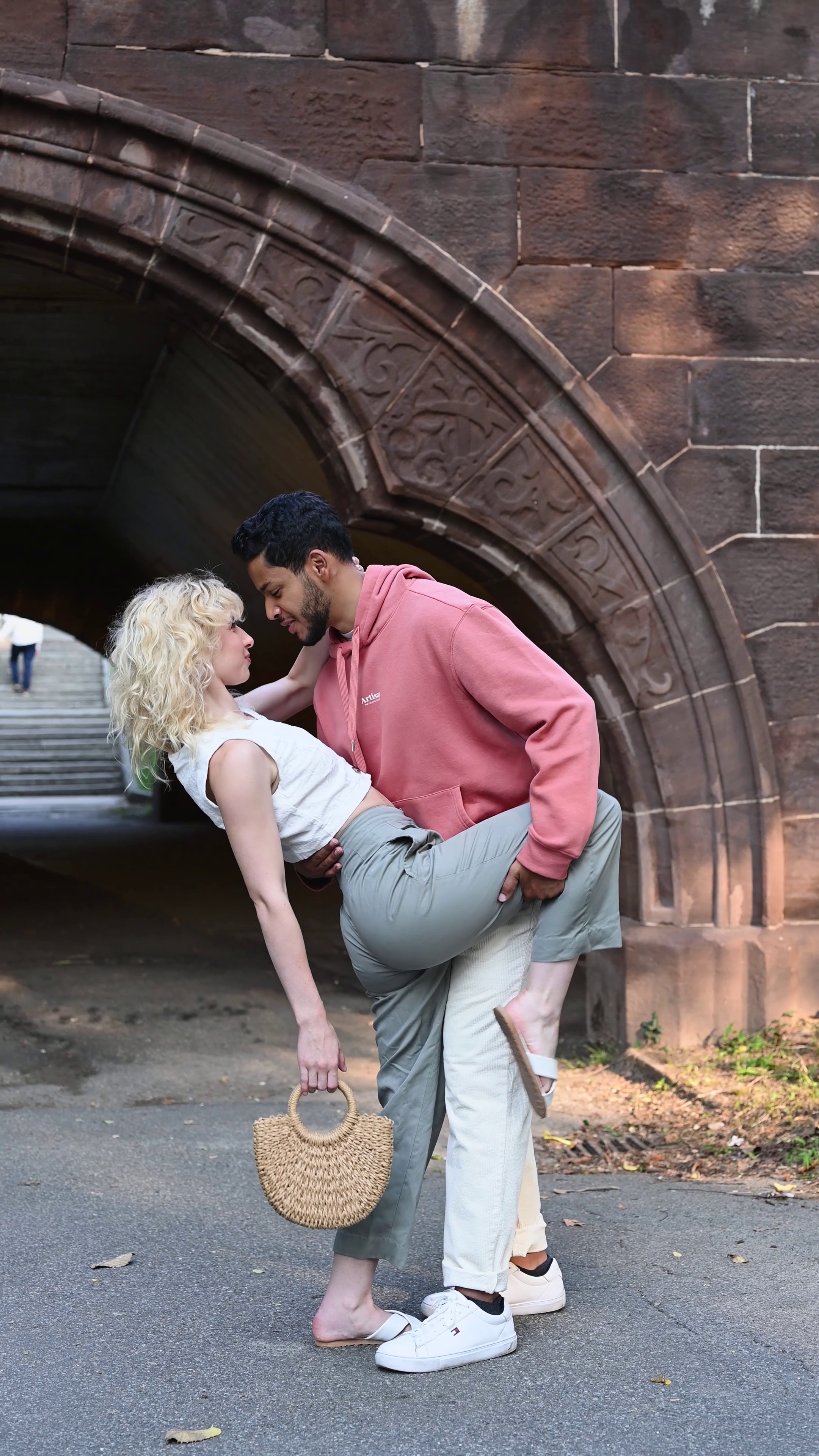} &
\includegraphics[width=0.3\columnwidth]{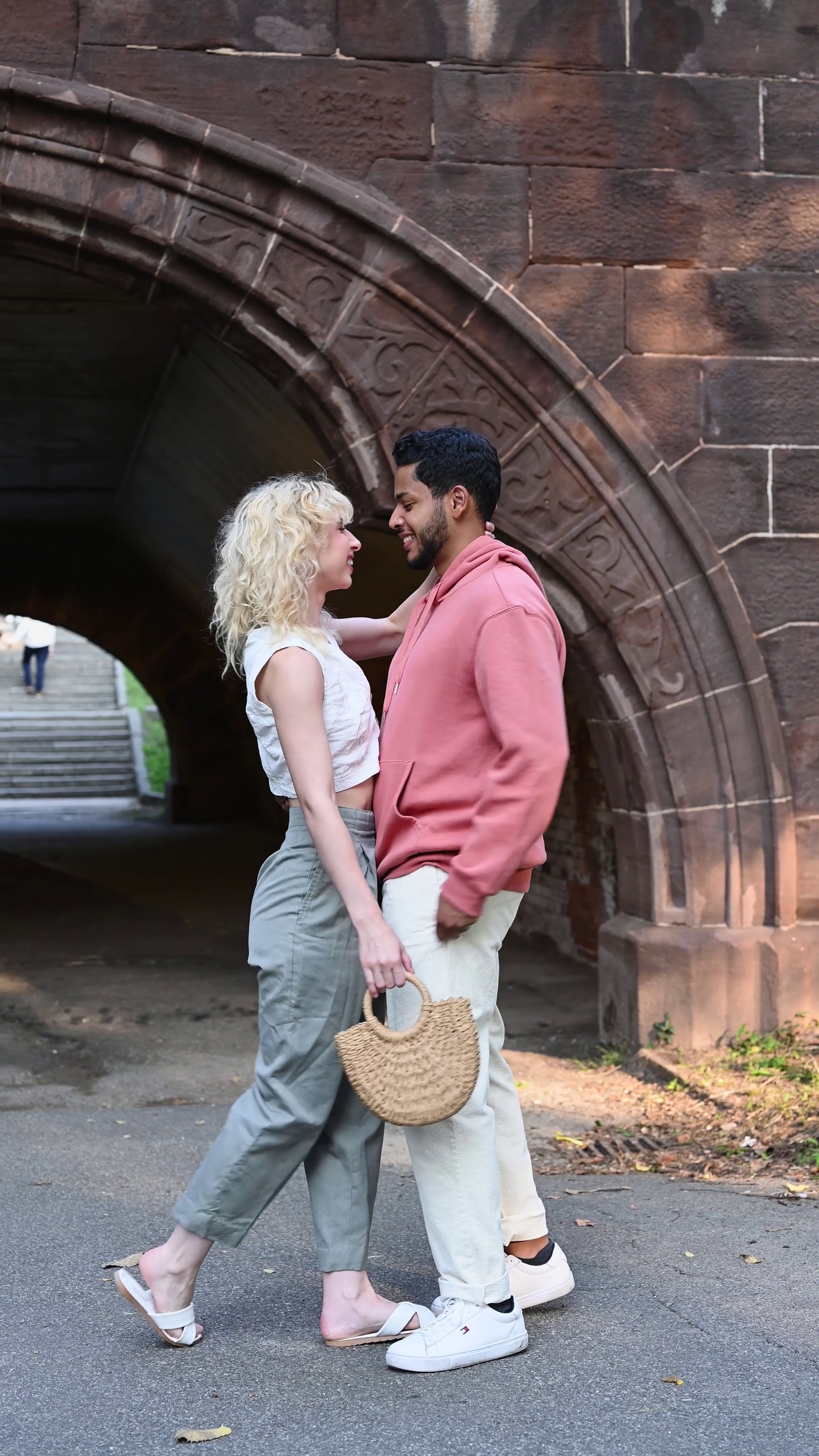} &
\includegraphics[width=0.3\columnwidth]{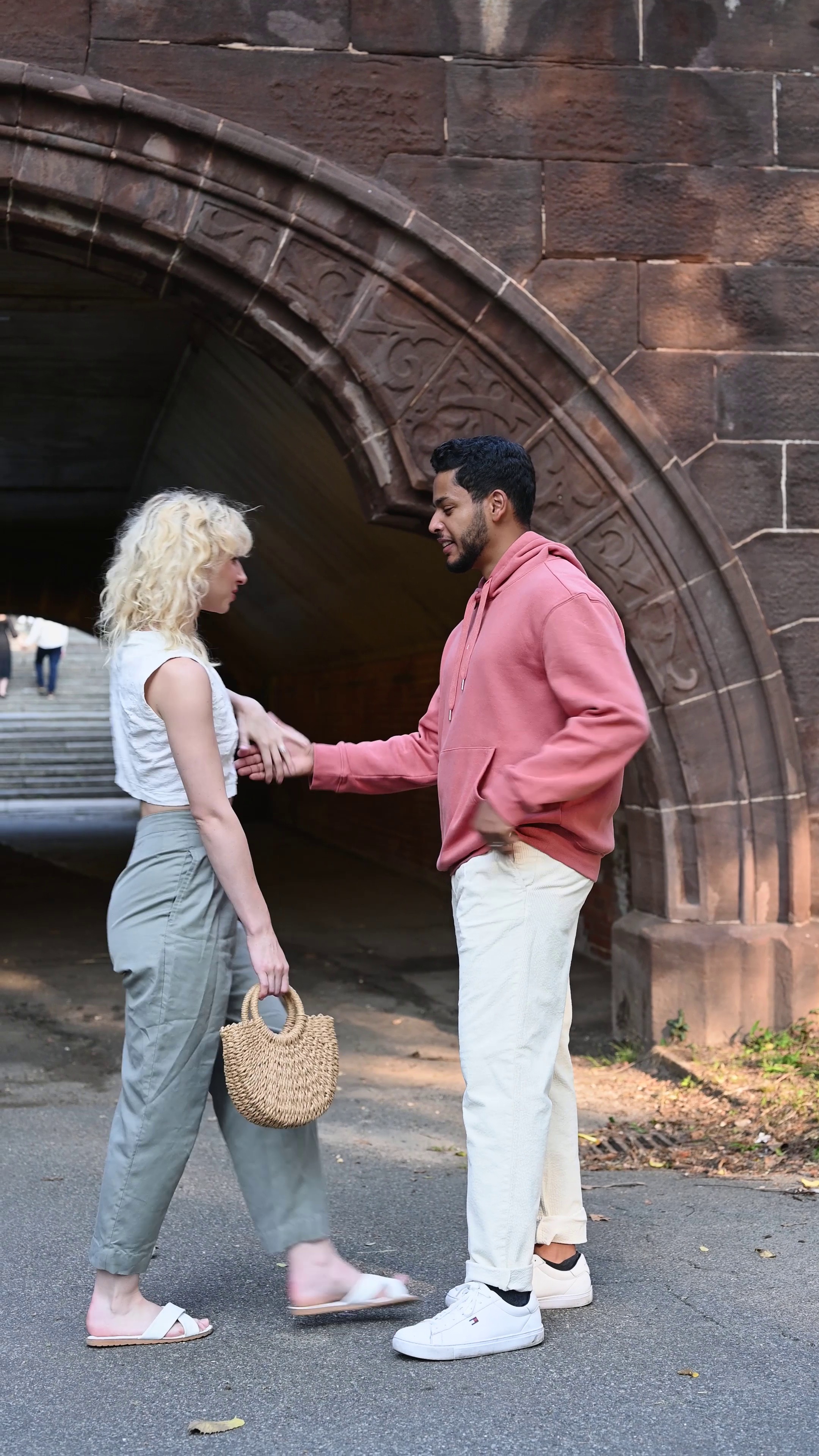} \\ 
\includegraphics[width=0.3\columnwidth]{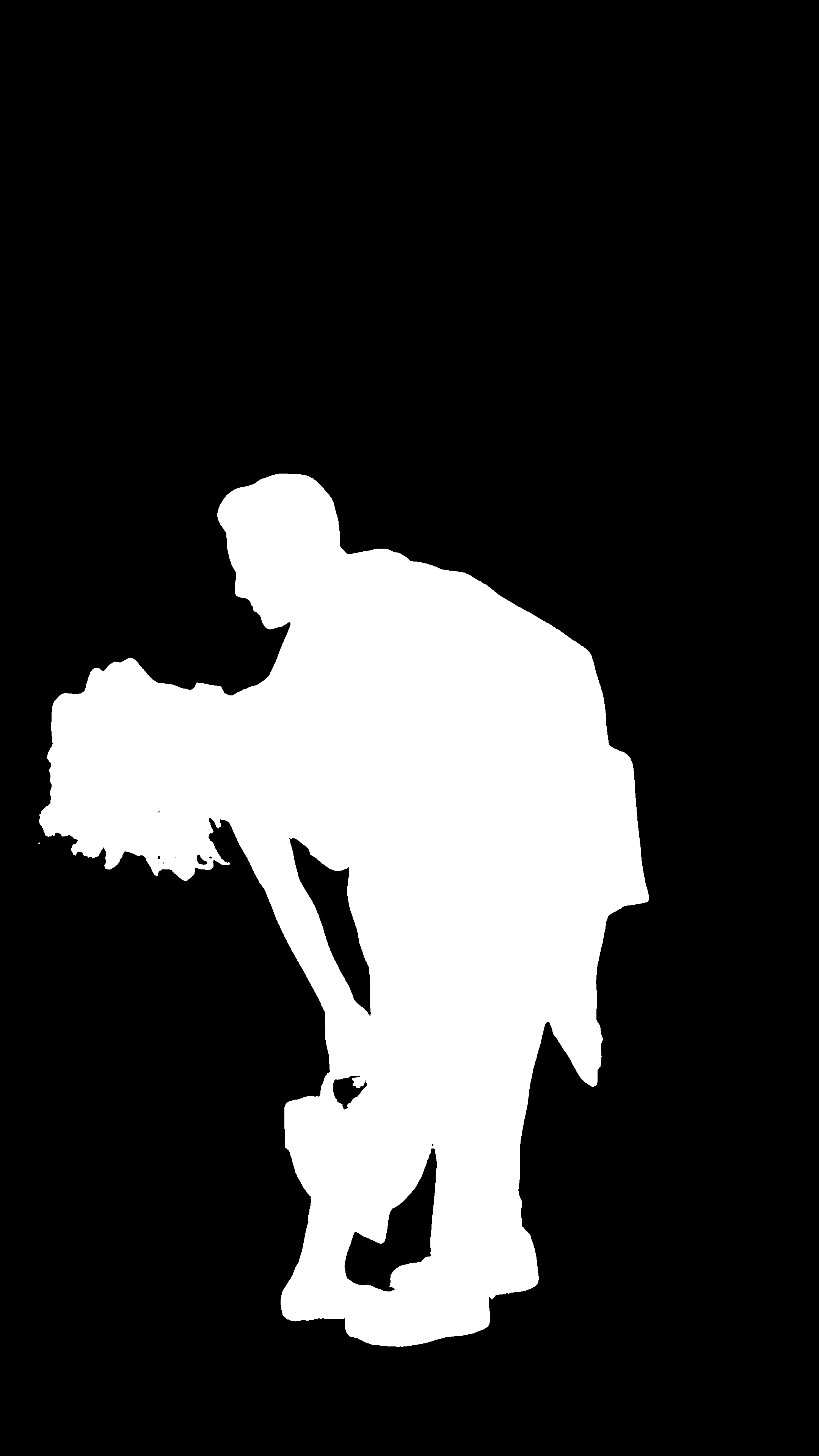} & 
\includegraphics[width=0.3\columnwidth]{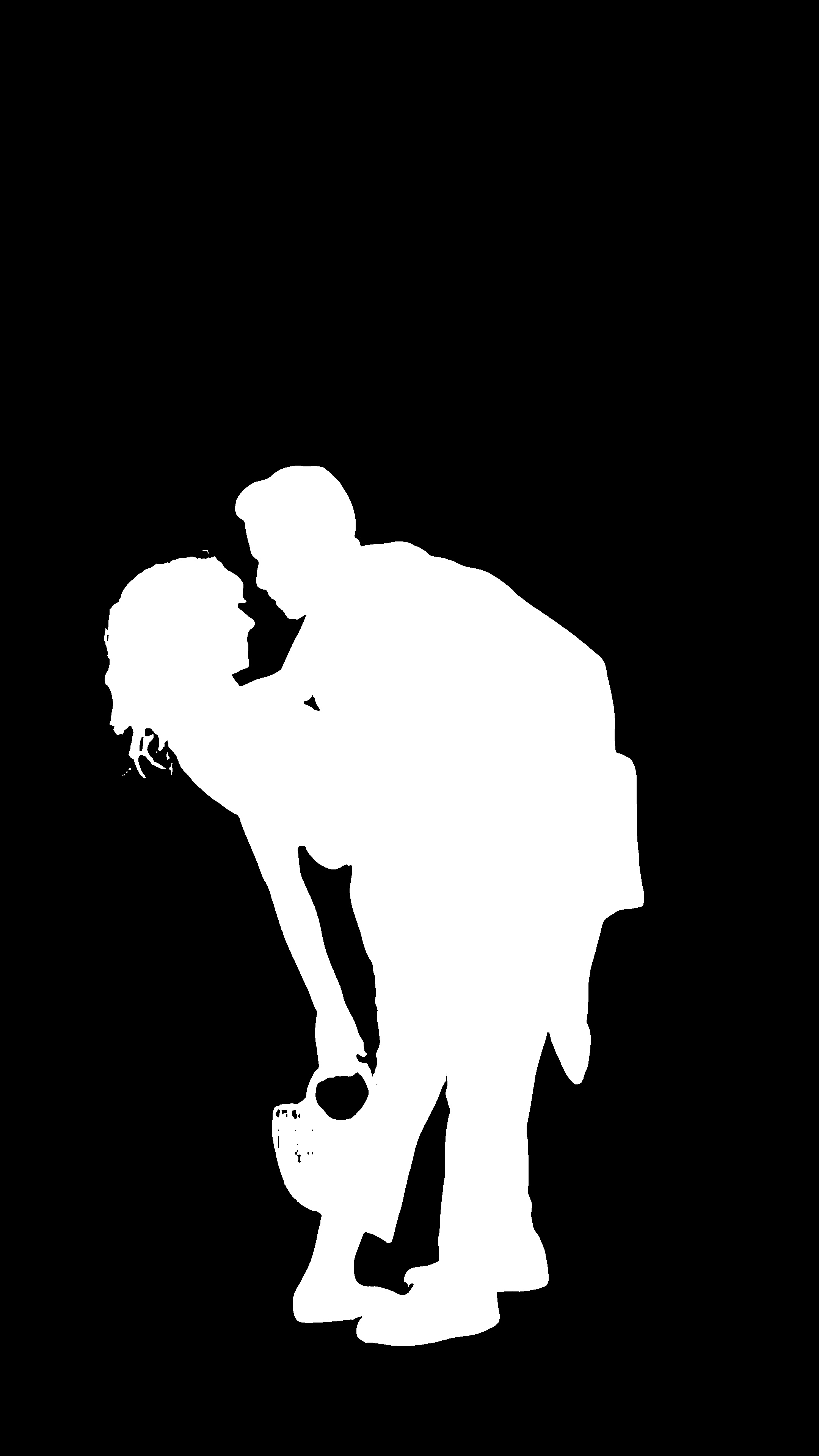} & 
\includegraphics[width=0.3\columnwidth]{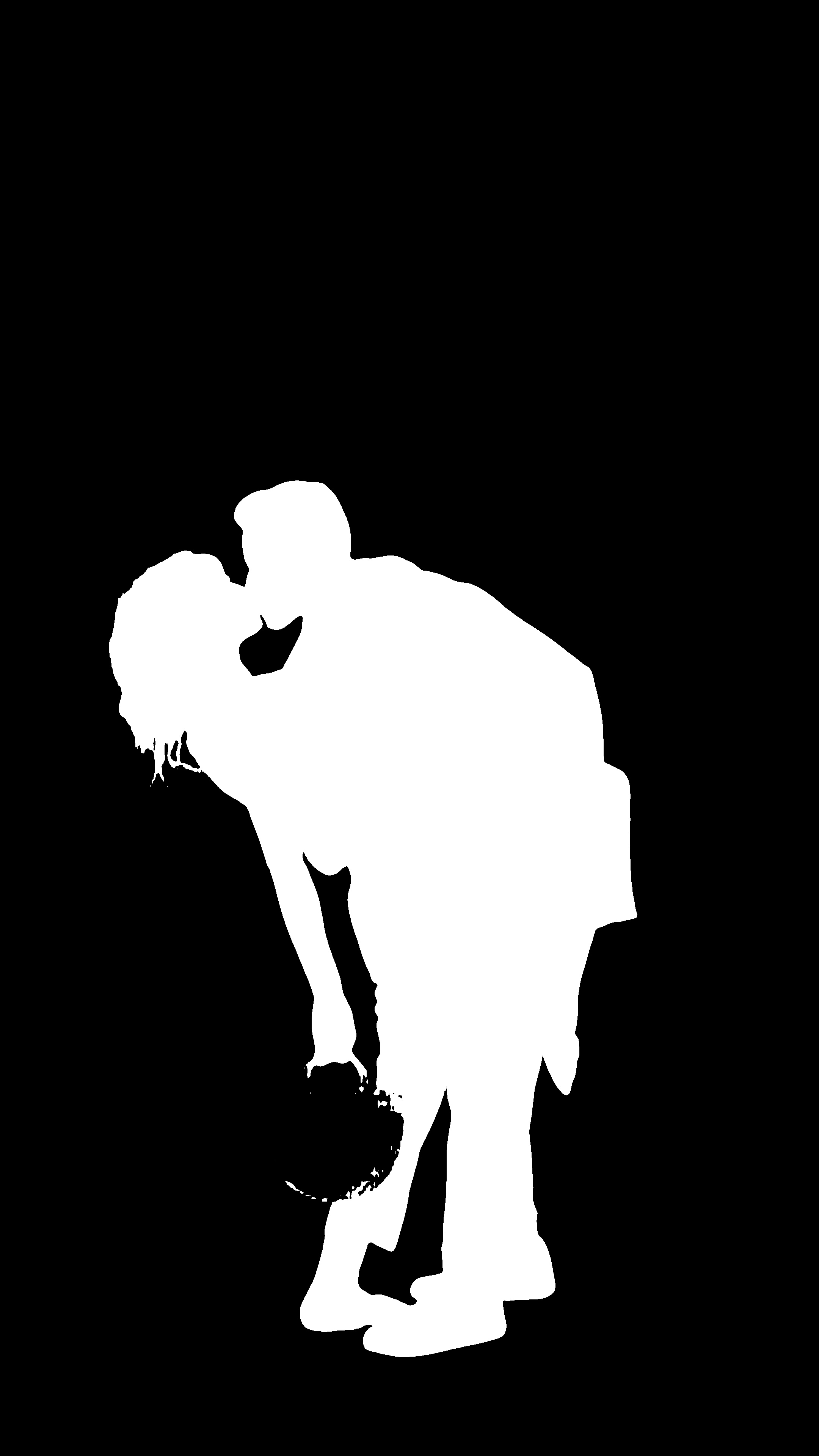} & 
\includegraphics[width=0.3\columnwidth]{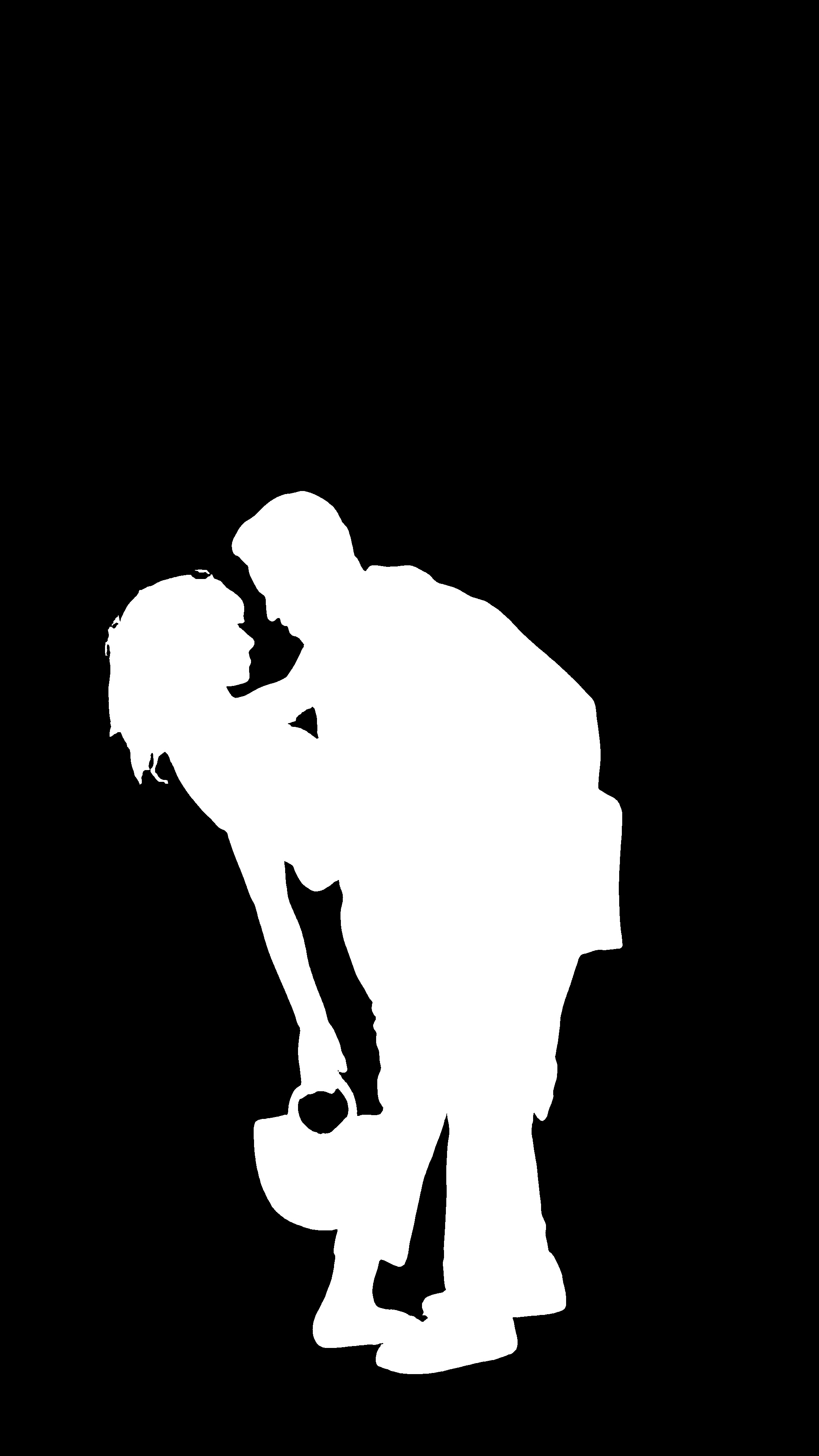} &
\includegraphics[width=0.3\columnwidth]{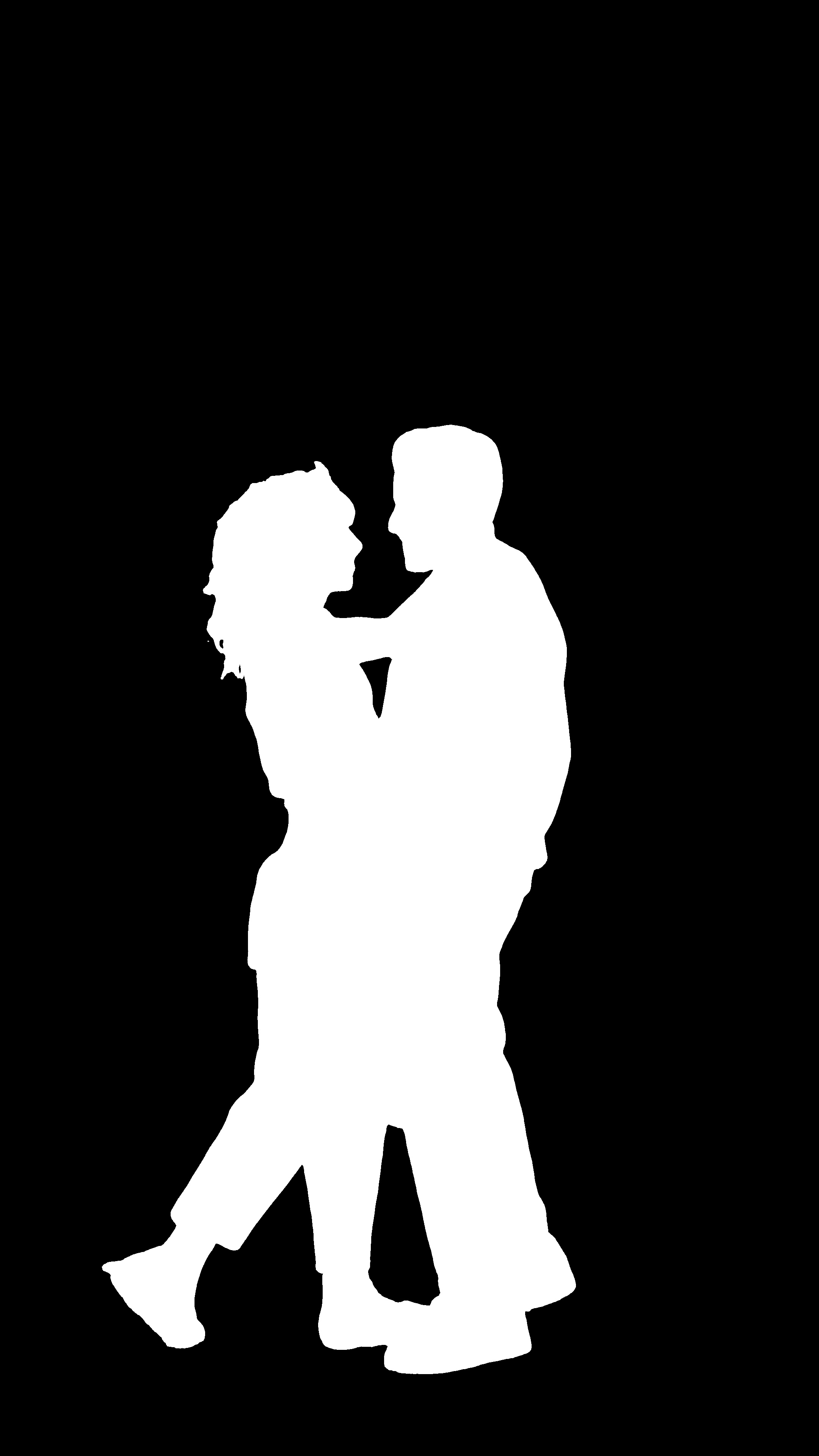} &
\includegraphics[width=0.3\columnwidth]{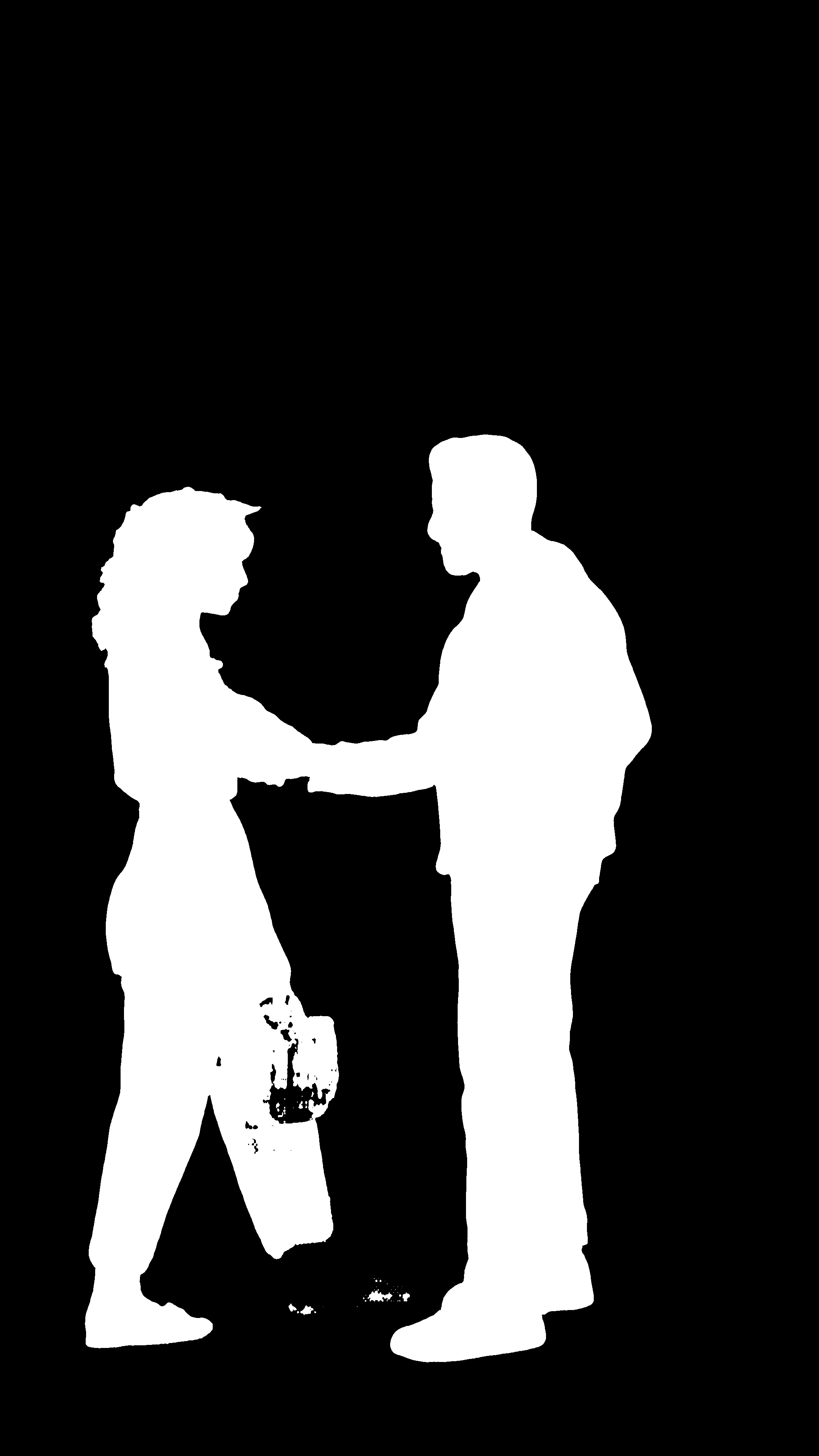} \\
Frame 0 & Frame 30 & Frame 60 & Frame 90 & Frame 120 & Frame 150 \\
\end{tabular}
\caption{Pseudo-labeled video segmentation annotations: There exists noise in the label, e.g. the bag disappears in Frame 30, and the woman's left foot disappears in Frame 150. We only use it for pre-training.}
\Description{Pseudo video segmentation data.}
\label{fig:videohuman60}
\end{figure*}

\bibliographystyle{ACM-Reference-Format}
\bibliography{egbib}

\clearpage
\appendix
\setcounter{page}{1}
\renewcommand{\thesection}{\Alph{section}}%
\label{sec:appendix}

\section{More Details About Rendered Dataset} 
\label{sup-sec: render_dataset}
The synthetic video matting dataset is rendered using Blender~\cite{blender}, with 3D characters sourced from the Human3D plugin and FaceBuilder plugin. During the rendering stage, the physics-based renderer Blender Cycles is employed. To ensure diversity, random aspect ratios of 1:1, 16:9, and 9:16 are used, corresponding to resolutions of $3840\times3840$, $3840\times2160$, and $2160\times3840$, respectively. To achieve photorealistic and high-quality rendering, particularly for hair and annotations, the sampling rate is set to 2048. Rendering a single image on a 4090 GPU takes approximately one minute. The final outputs, including rendered images and their corresponding alpha masks, are saved in PNG format. Each sequence in the dataset consists of approximately 90 frames, ensuring sufficient temporal continuity and diversity for tasks such as video matting and motion analysis.

To enhance realism, realistic light sources such as point lights and area lights are added to simulate real-world lighting conditions. Additionally, camera movement is simulated by positioning the camera at various angles and distances from the character, enabling dynamic and diverse rendering perspectives. This approach ensures the generation of high-quality, realistic data suitable for video matting applications.
To compose the rendered foreground human with diverse backgrounds, we collected approximately 200 background images encompassing both indoor and outdoor wild scenes. These backgrounds are used to create realistic and varied compositions, enhancing the diversity and applicability of the dataset for training and evaluation purposes.

\section{Analysis of Synthetic Segmentation Datasets}
\begin{figure}[h]
\small
\centering
\begin{tabular}{@{\hspace{0mm}}c@{\hspace{1.5mm}} @{\hspace{-1.mm}}c@{\hspace{1.5mm}}}
\includegraphics[width=0.5\columnwidth]{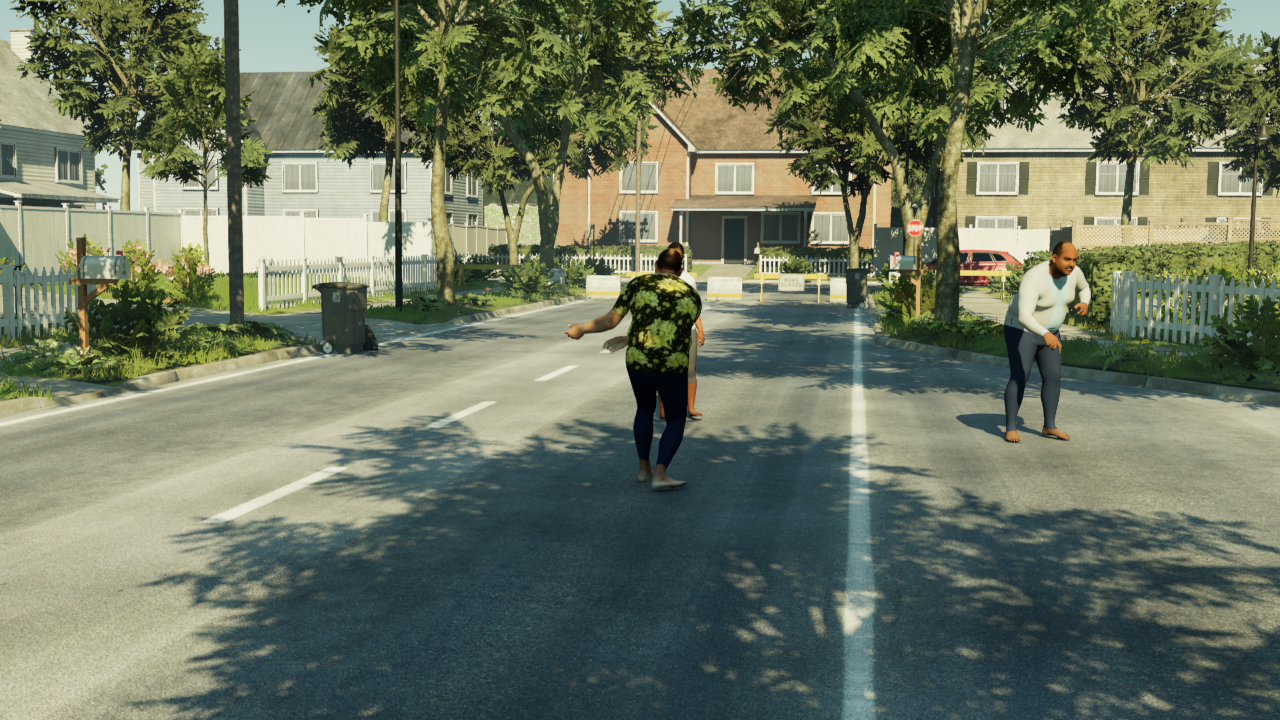} & 
\includegraphics[width=0.5\columnwidth]{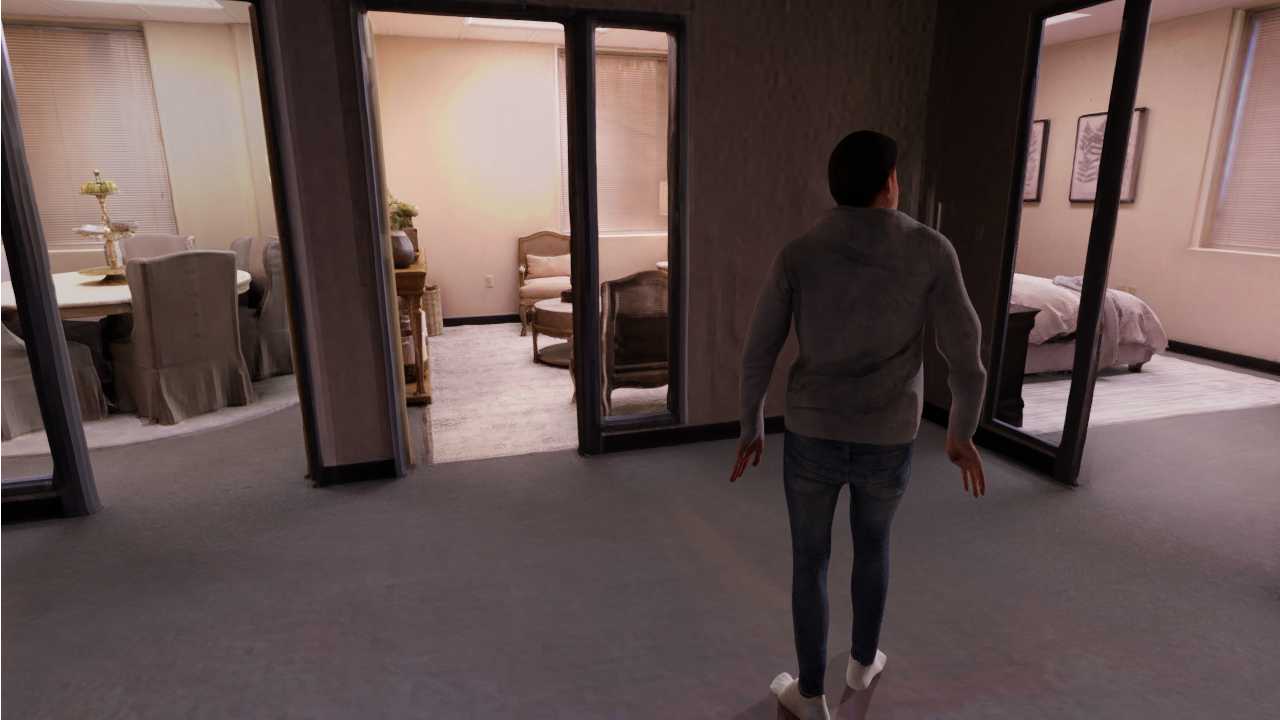} \\
\includegraphics[width=0.5\columnwidth]{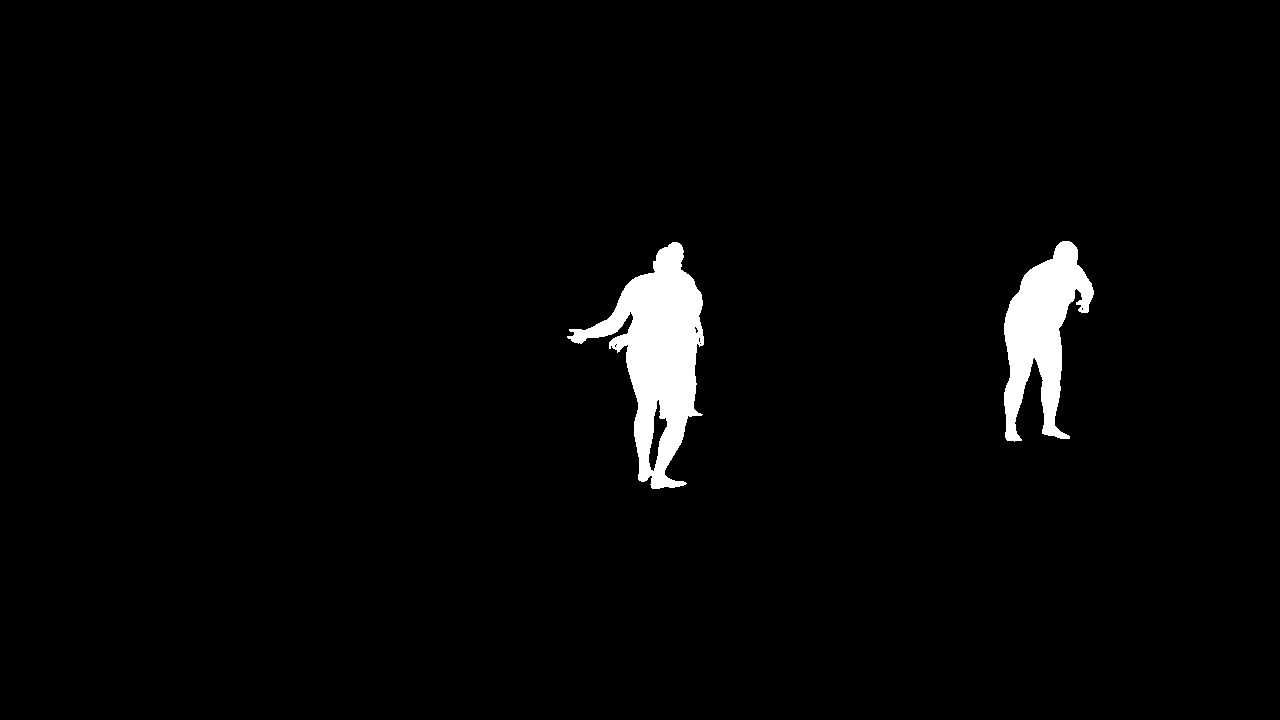} &  
\includegraphics[width=0.5\columnwidth]{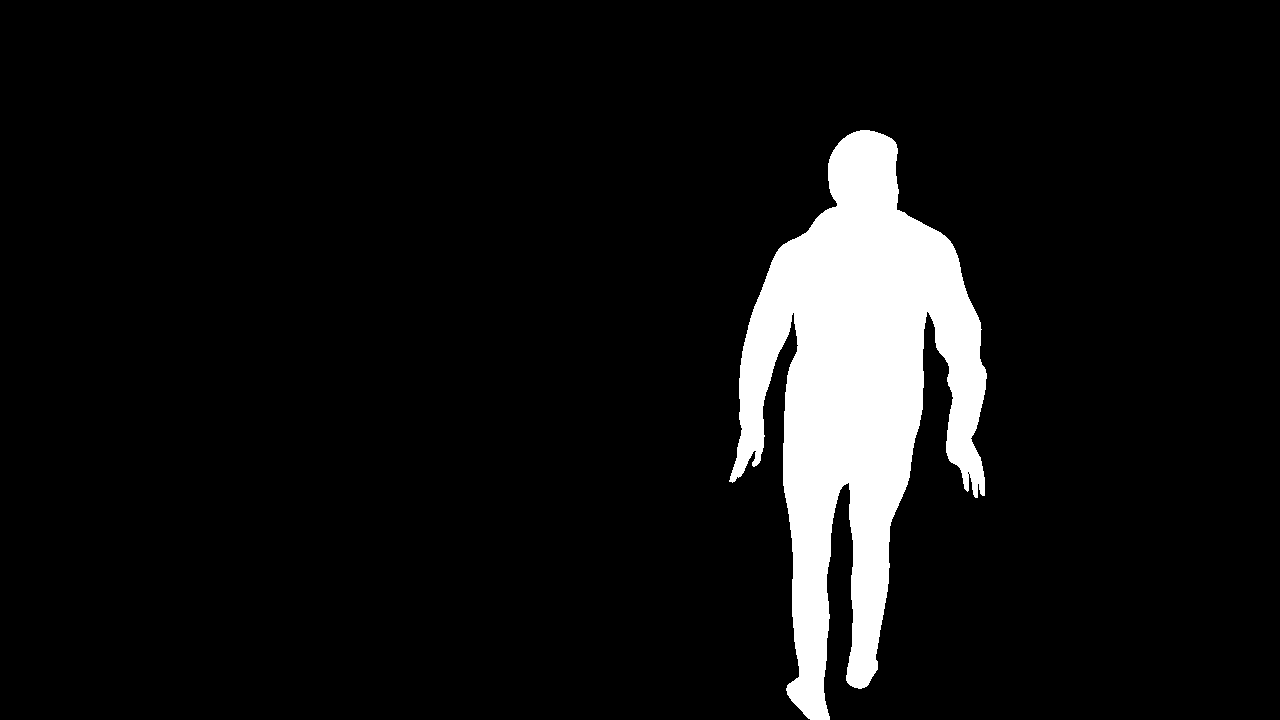}\\
BEDLAM sample & Dynamic Replica sample \\
\end{tabular}
\caption{Syntheic segmentation datasets. left is BEDLAM~\cite{bedlam}, right is Dynamic Replica~\cite{karaev2023dynamicstereo}.}
\Description{Syntheic segmentation datasets.}
\label{fig:synthetic_seg_data}
\end{figure}

As illustrated in~\Cref{fig:synthetic_seg_data}, the ground truth semantic segmentation annotations from BEDLAM~\cite{bedlam} and Dynamic Replica~\cite{karaev2023dynamicstereo} are displayed. Both synthetic datasets provide high-quality human mask annotations; however, they lack fine-grained annotations for hair or fur. Consequently, models trained exclusively on these synthetic segmentation datasets often fail to accurately predict fine-grained hair details in most cases.

\section{Reconstruction Quality of the Variational Autoencoder}
The reconstruction performance of the pre-trained Variational Autoencoder (VAE) significantly influences the final performance of matting generation. To evaluate this, we conducted an experiment to measure the reconstruction accuracy of the VAE from SVD~\cite{svd} using PSNR and SSIM metrics across various matting datasets. For both metrics, higher values indicate better performance. As shown in~\Cref{tab:vae_reconstruction}, there is substantial variance in reconstruction quality across different datasets, highlighting the limitations imposed by the default VAE. This suggests that the upper bound of video matting performance is constrained by the reconstruction capability of the VAE. Future work could explore more advanced 3D VAEs as replacements to achieve superior video matting results.

\begin{table}[ht!]
    \centering
    \caption{Quantitative Evaluation of Image Reconstruction Performance of 3D VAE.}
    \vspace{-0.5em}
    \footnotesize
    \setlength{\tabcolsep}{3.0pt} %
    \renewcommand{\arraystretch}{1.2} %
    \begin{tabular}{@{}c|cccccc@{}}
    \toprule
    \multirow{2}{*}{Metric} & \multirow{2}{*}{VM} & V-HIM60 & V-HIM60 & V-HIM60 & \multirow{2}{*}{P3M-500-NP} & \multirow{2}{*}{P3M-500-P} \\
     & & Easy & Medium & Hard  & & \\
    \hline
    PSNR $\uparrow$ & 0.87 & 0.82 & 0.86 & 0.85 & 0.89 & 0.86 \\
    SSIM $\uparrow$ & 30.31 & 28.8 & 29.27 & 29.62 & 32.56 & 31.92 \\
    \bottomrule
    \end{tabular}
    \label{tab:vae_reconstruction}
\end{table}

\section{Comparison of inference speed.}
In~\Cref{tab:infer_speed}, {we report the single-image inference speed of our model, Stable Video Diffusion, and RVM, with two input resolutions, $512\times512$ and $1024\times1024$. The experiment is performed on an A100 GPU machine. Our method has an inferior speed with RVM due to our model is much larger than RVM, but much faster than the 50-step Stable Video Diffusion model due to the introduced flow matching scheduler.}

\begin{table}[ht]
    \centering
    \caption{Inference speed of different methods (Time in seconds).}
    \label{tab:infer_speed}
    \small
    \begin{tabular}{lccc}
        \toprule
        Method & steps & $512\times512$ & $1024\times1024$ \\\midrule
        RVM (ResNet-50)	& - &  0.01	& 0.02 \\
        \Ours	& 1 & 0.32	& 0.65 \\
        \Ours	& 3 & 0.53	& 0.9 \\
        SVD 	& 50 & 5.44	& 7.33\\
        \bottomrule
    \end{tabular}
\end{table}

\section{Visualization of Video Synthetic Defocus}
\begin{figure}
\small
\centering
\begin{tabular}{@{\hspace{0mm}}c@{\hspace{1.2mm}} @{\hspace{-0.5mm}}c@{\hspace{1.2mm}}}
\includegraphics[width=0.5\columnwidth]{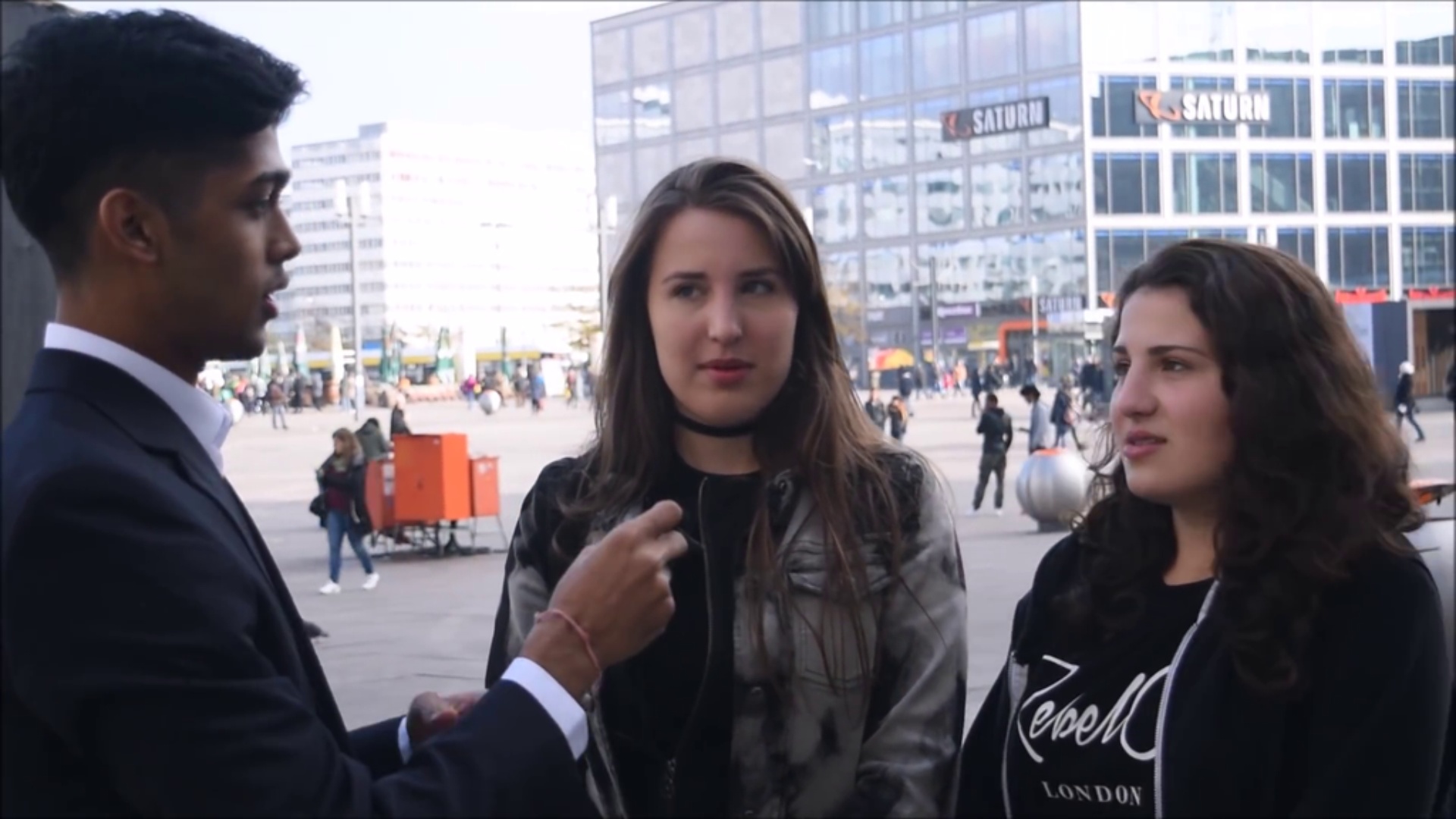} & 
\includegraphics[width=0.5\columnwidth]{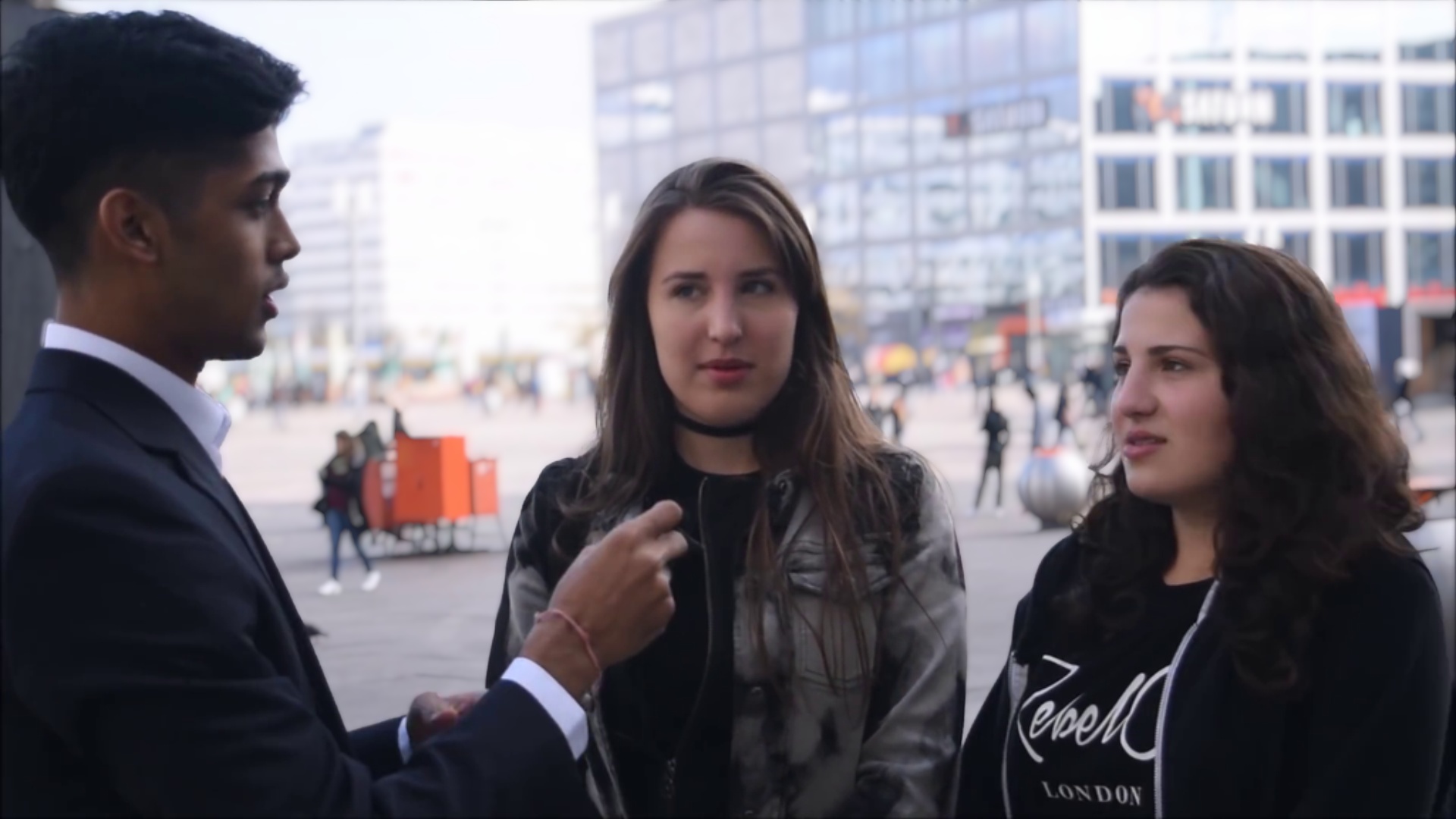} \\
Input frame & Frame effect \\
\end{tabular}
\caption{Synthetic defocus application.}
\Description{Synthetic defocus application.}
\label{fig:defocus}
\end{figure}

There are many potential applications for our video matting pipeline. One example is its use in creating synthetic defocus (bokeh) effects in videos. To achieve this effect, we first blur the background based on a depth map generated by Marigold~\cite{ke2023repurposing}, then composite the foreground RGBA layer onto the blurred background. We visualize the video synthetic defocus in~\Cref{fig:defocus}. A single-layer depth map alone is insufficient for achieving this effect, as it lacks the necessary accuracy around the subject's boundary, where fine details are essential for a natural-looking result.

\section{More Visualizations}
\begin{figure*}
\small
\centering
\begin{tabular}{c@{\hspace{2.mm}} @{\hspace{-1.5mm}}c@{\hspace{2.mm}}@{\hspace{-1.5mm}}c@{\hspace{2.mm}}}
\includegraphics[width=0.67\columnwidth]{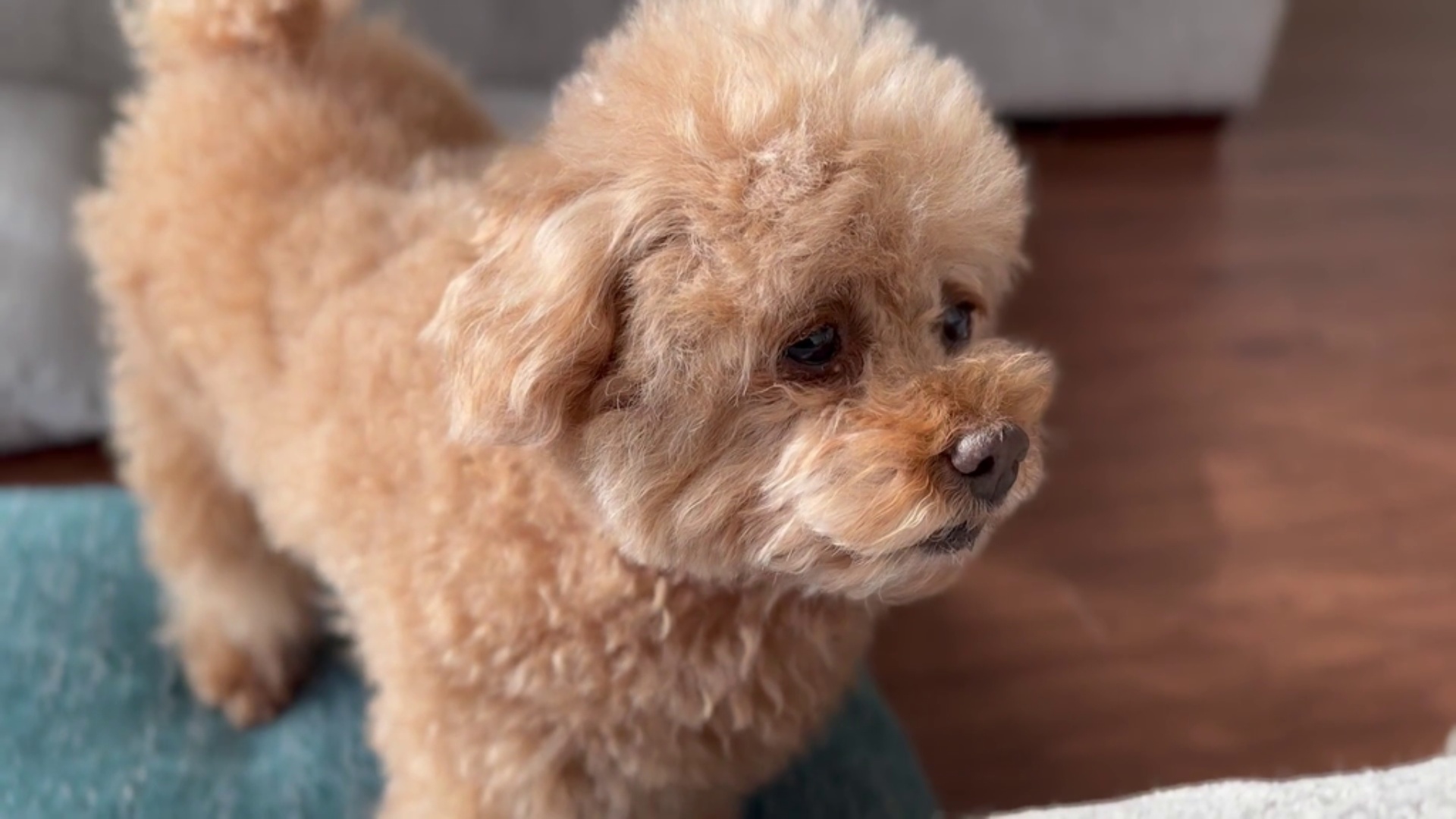} & 
\includegraphics[width=0.67\columnwidth]{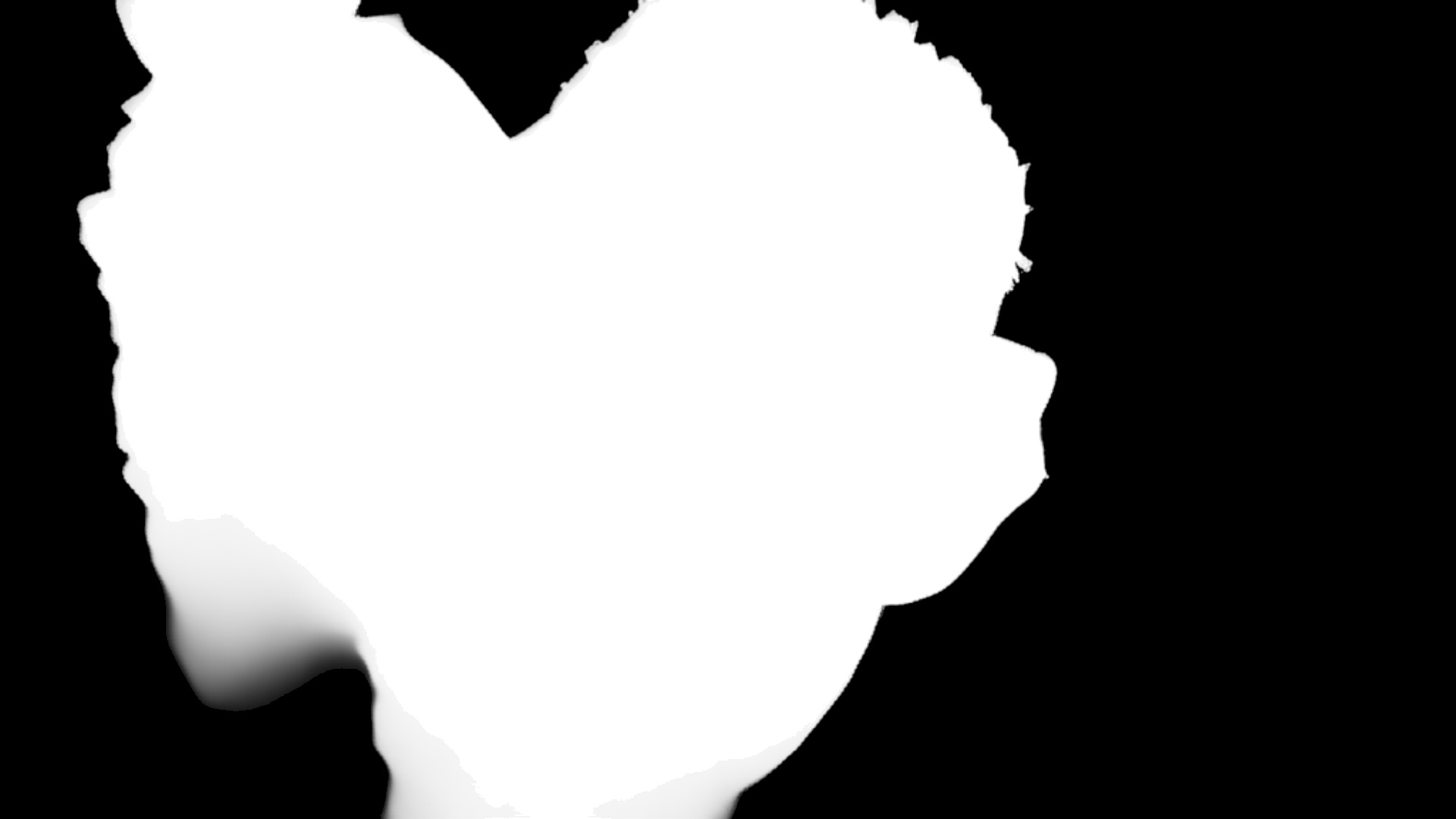} & 
\includegraphics[width=0.67\columnwidth]{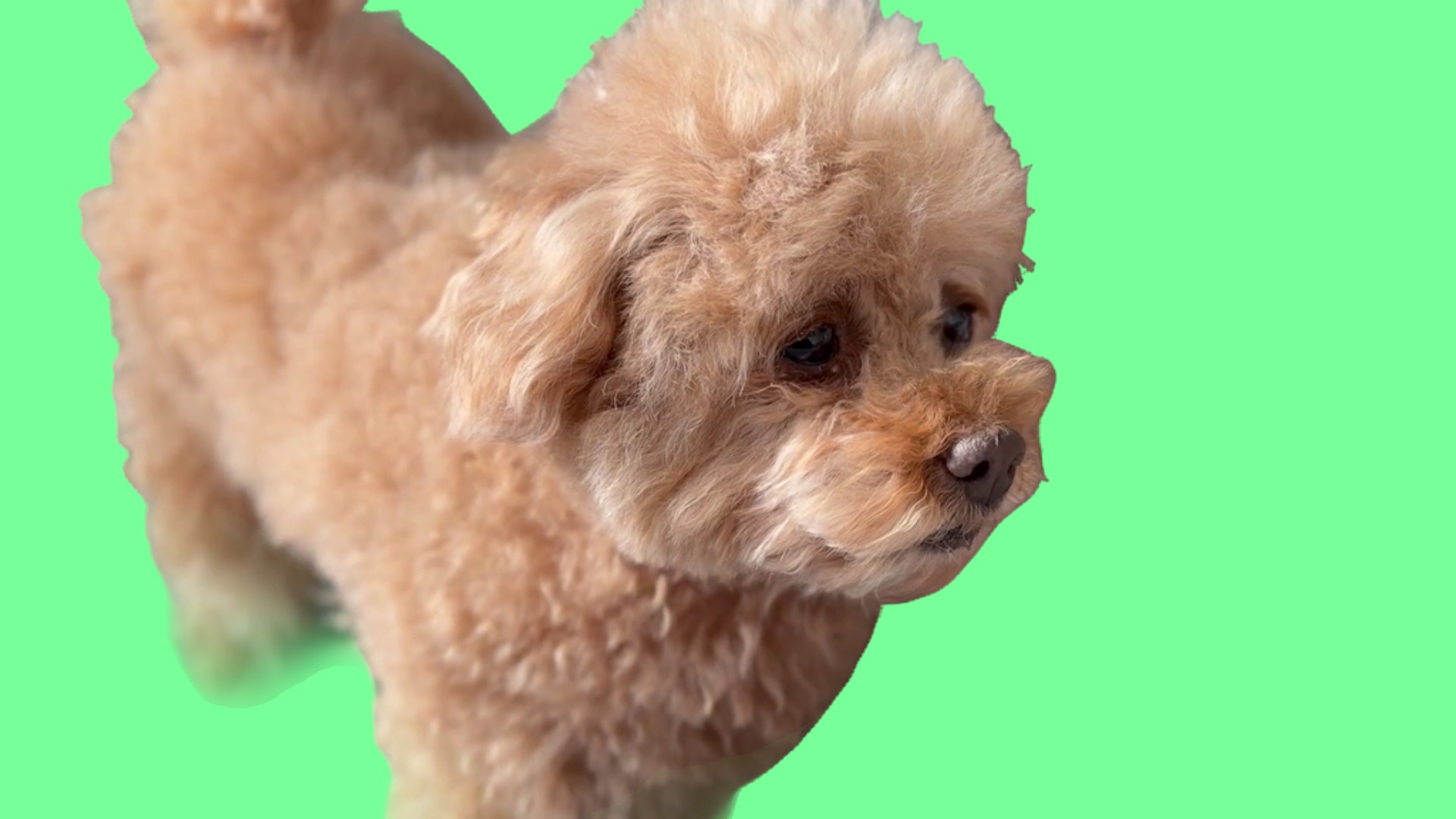} \\
\includegraphics[width=0.67\columnwidth]{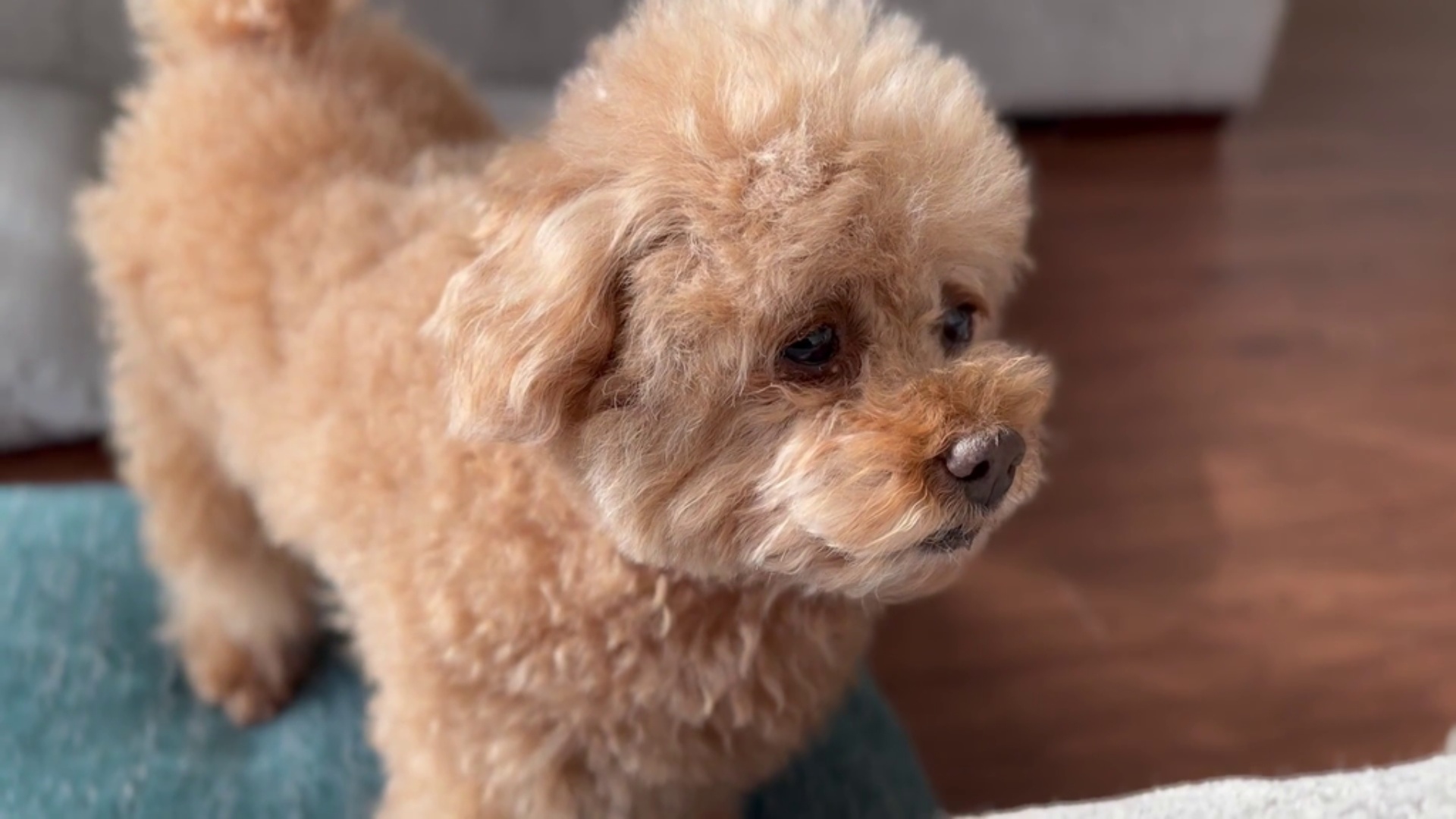} & 
\includegraphics[width=0.67\columnwidth]{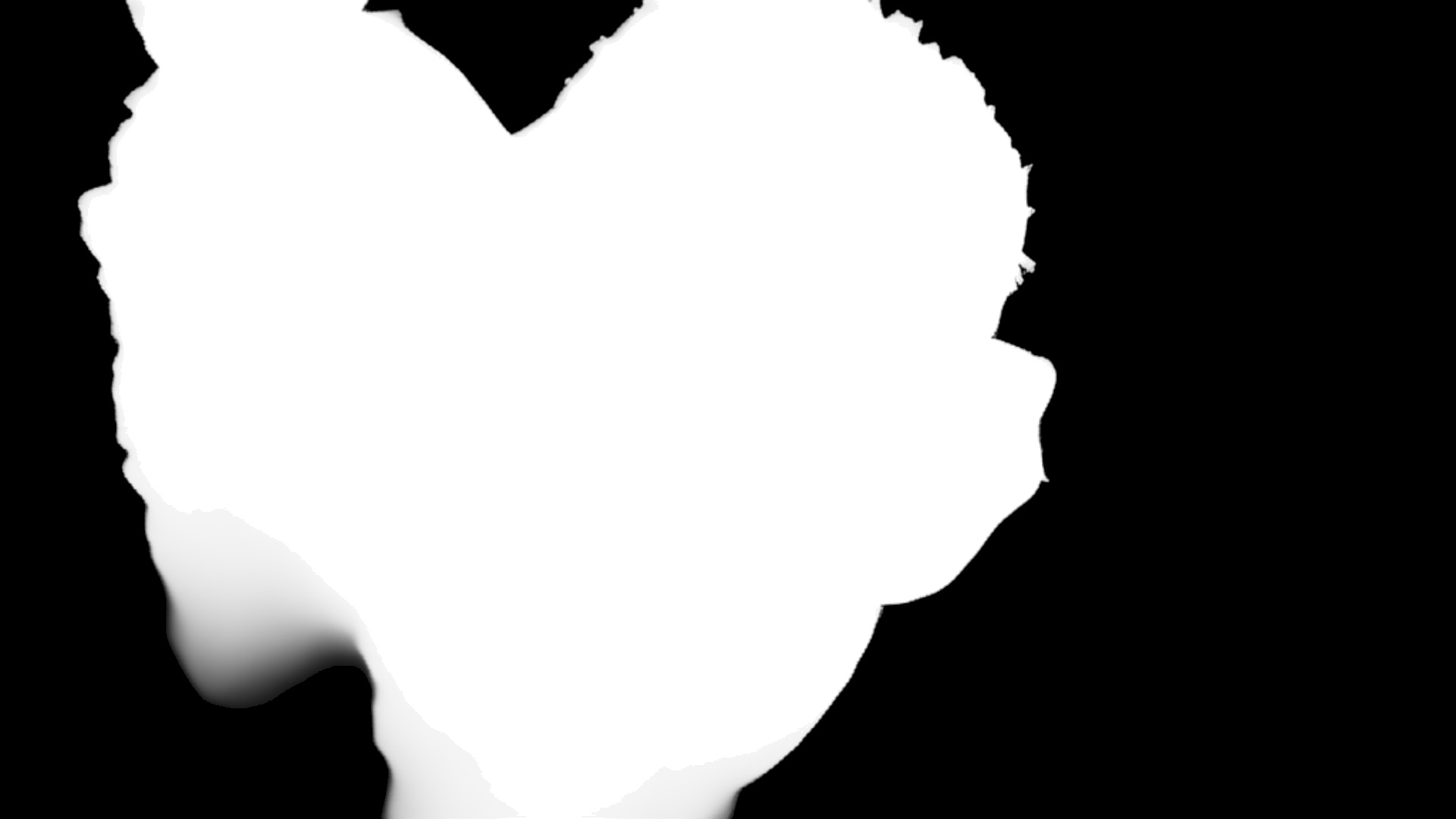} & 
\includegraphics[width=0.67\columnwidth]{figure/dog/104_green.jpg} \\
\includegraphics[width=0.67\columnwidth]{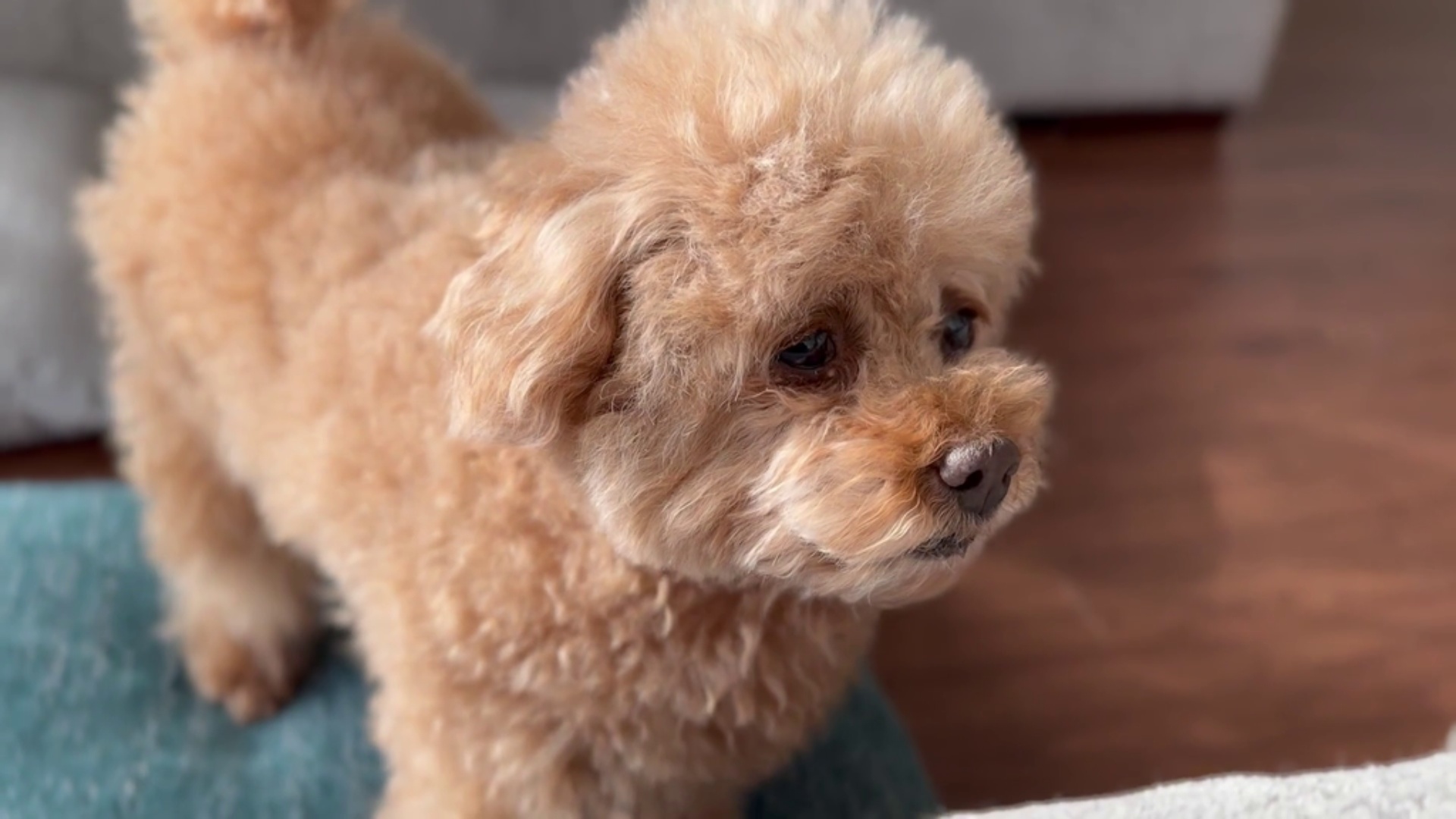} & 
\includegraphics[width=0.67\columnwidth]{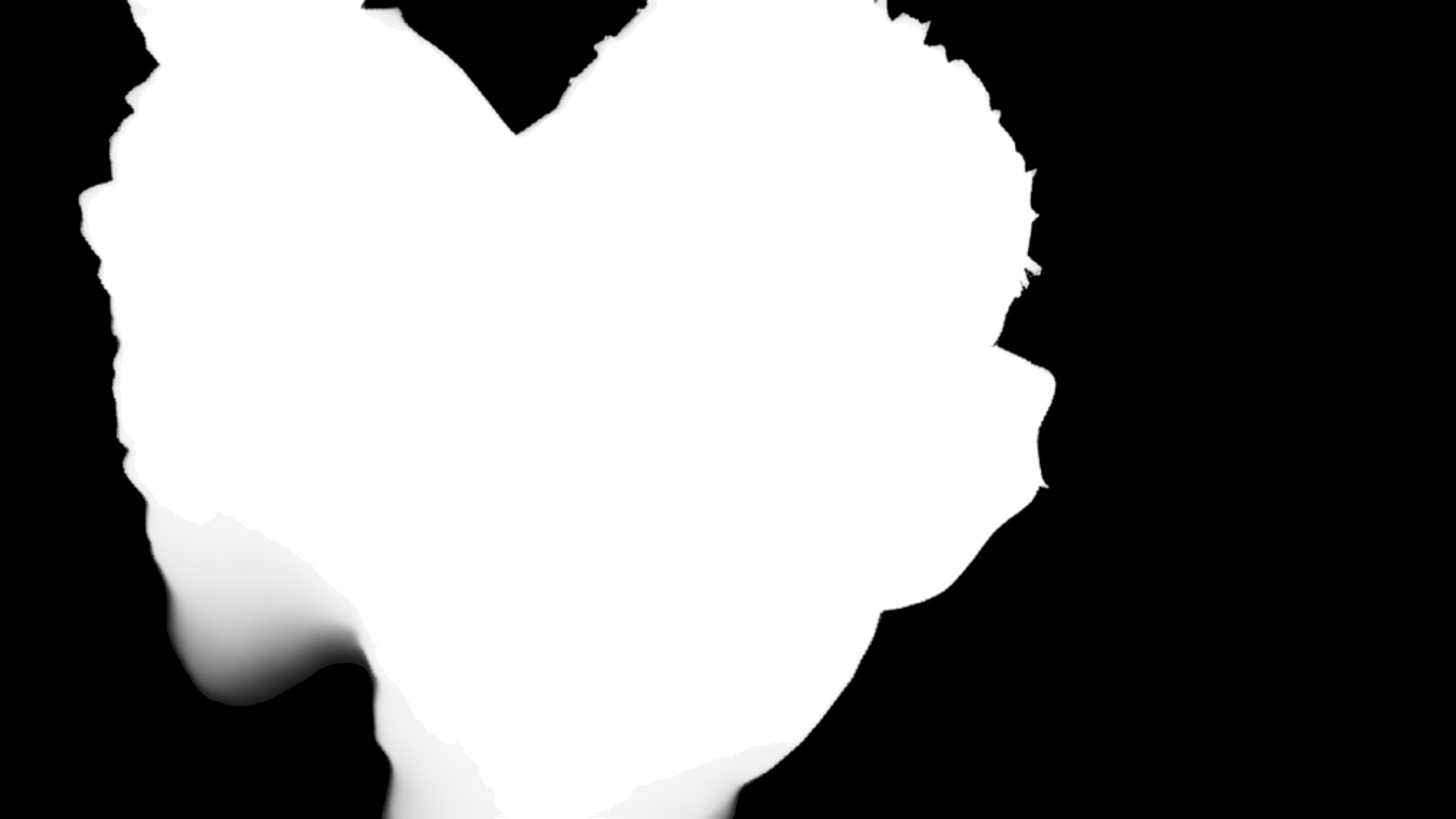} & 
\includegraphics[width=0.67\columnwidth]{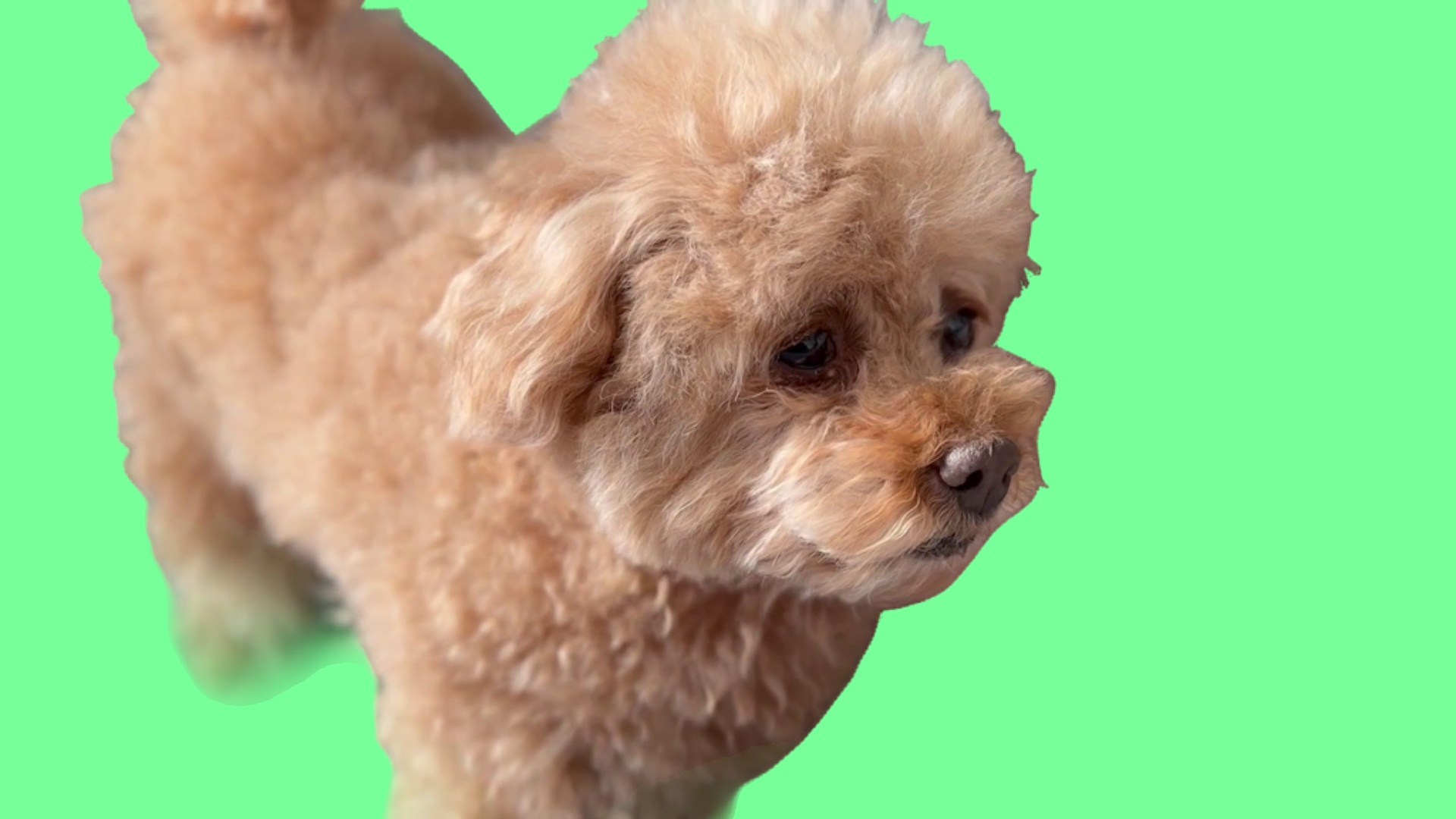} \\
\includegraphics[width=0.67\columnwidth]{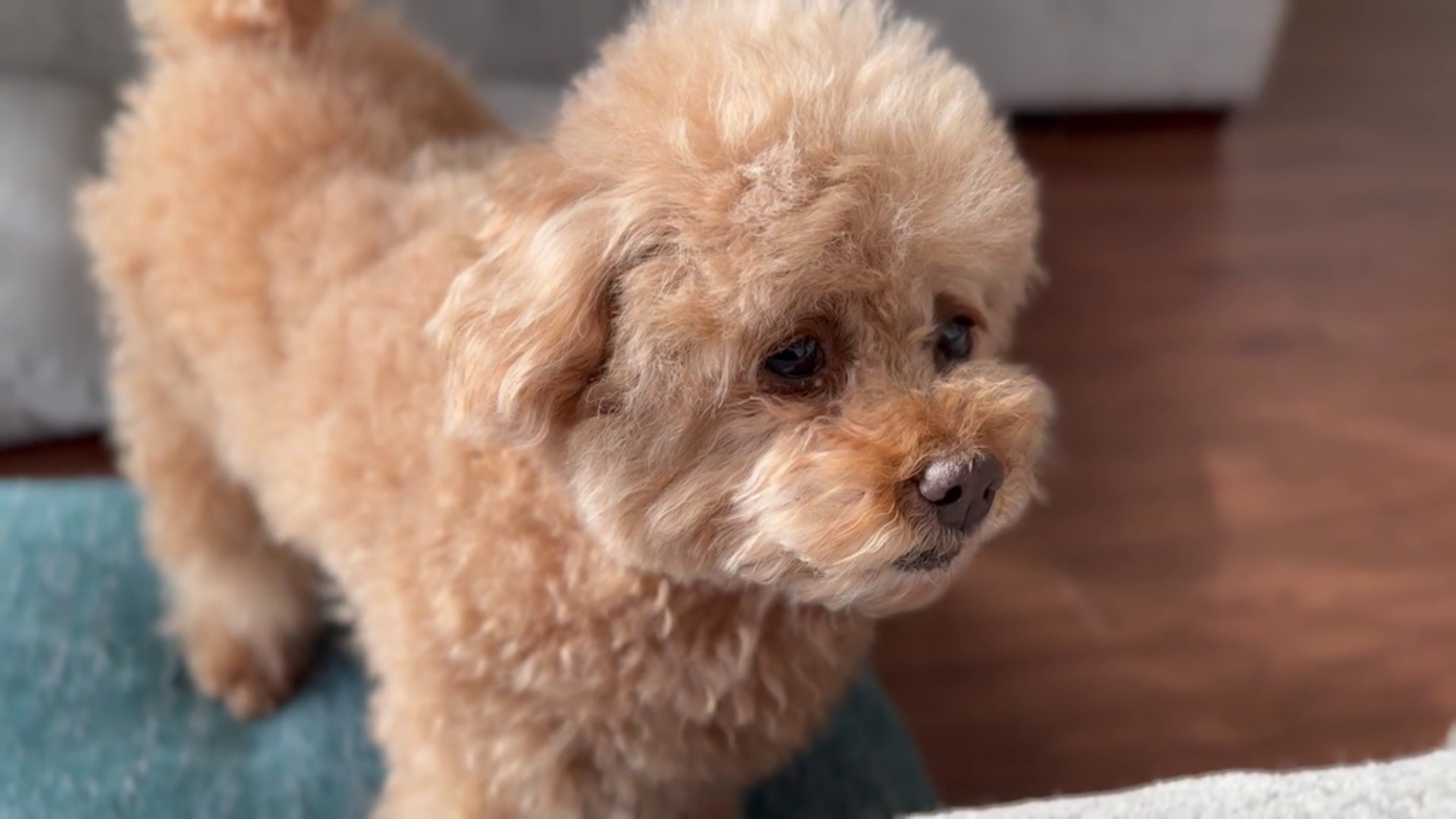} & 
\includegraphics[width=0.67\columnwidth]{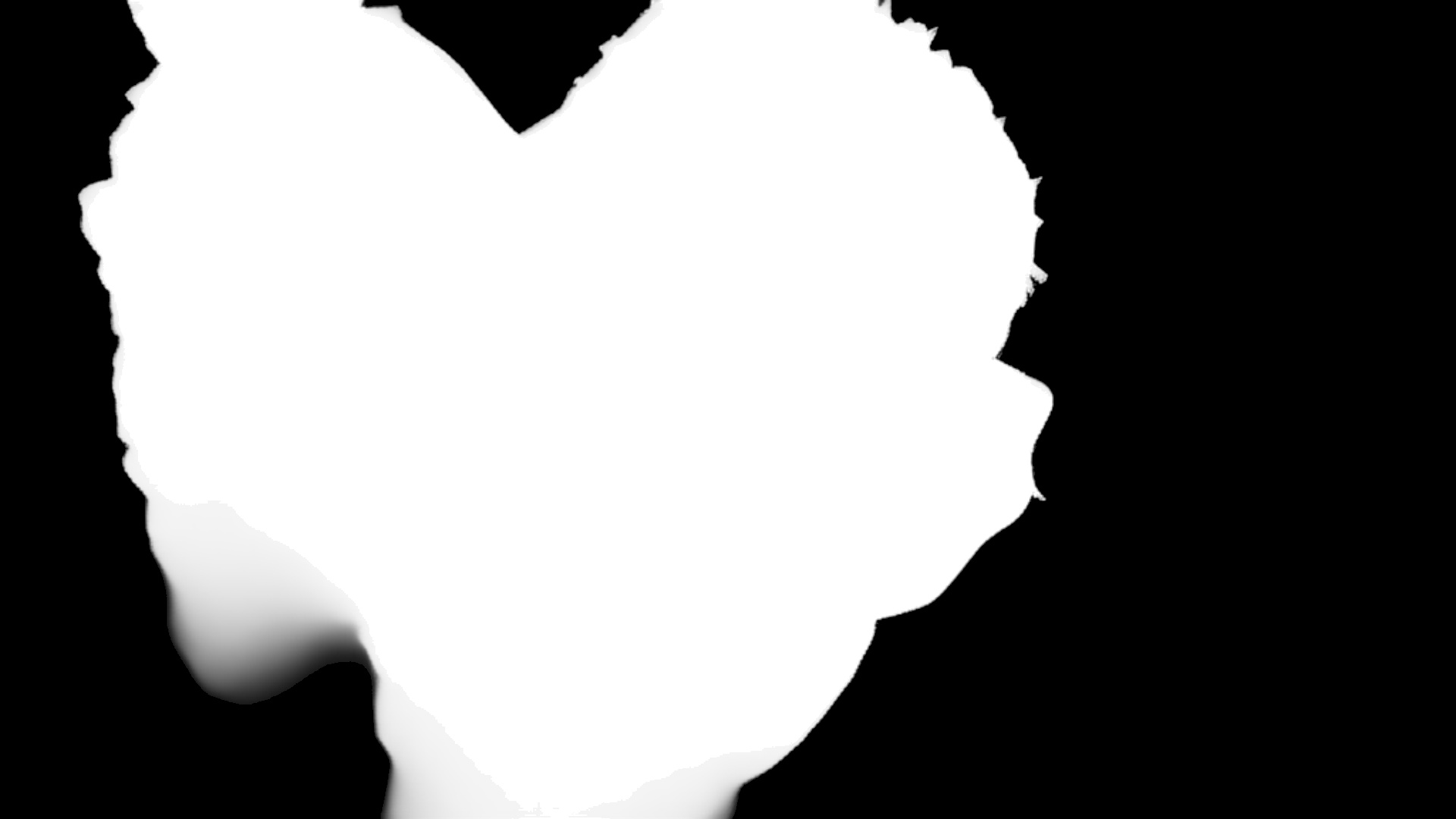} & 
\includegraphics[width=0.67\columnwidth]{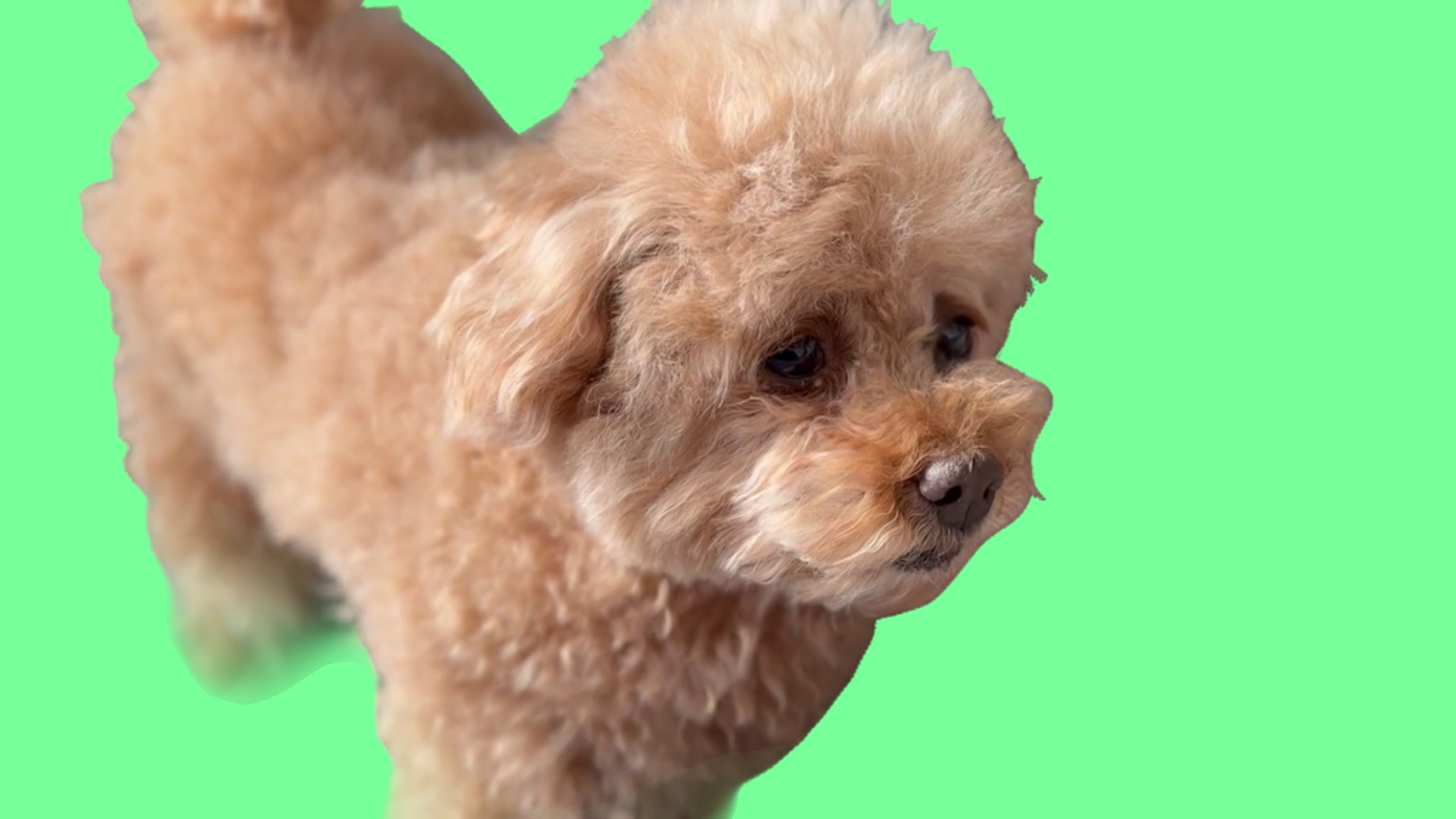} \\
\includegraphics[width=0.67\columnwidth]{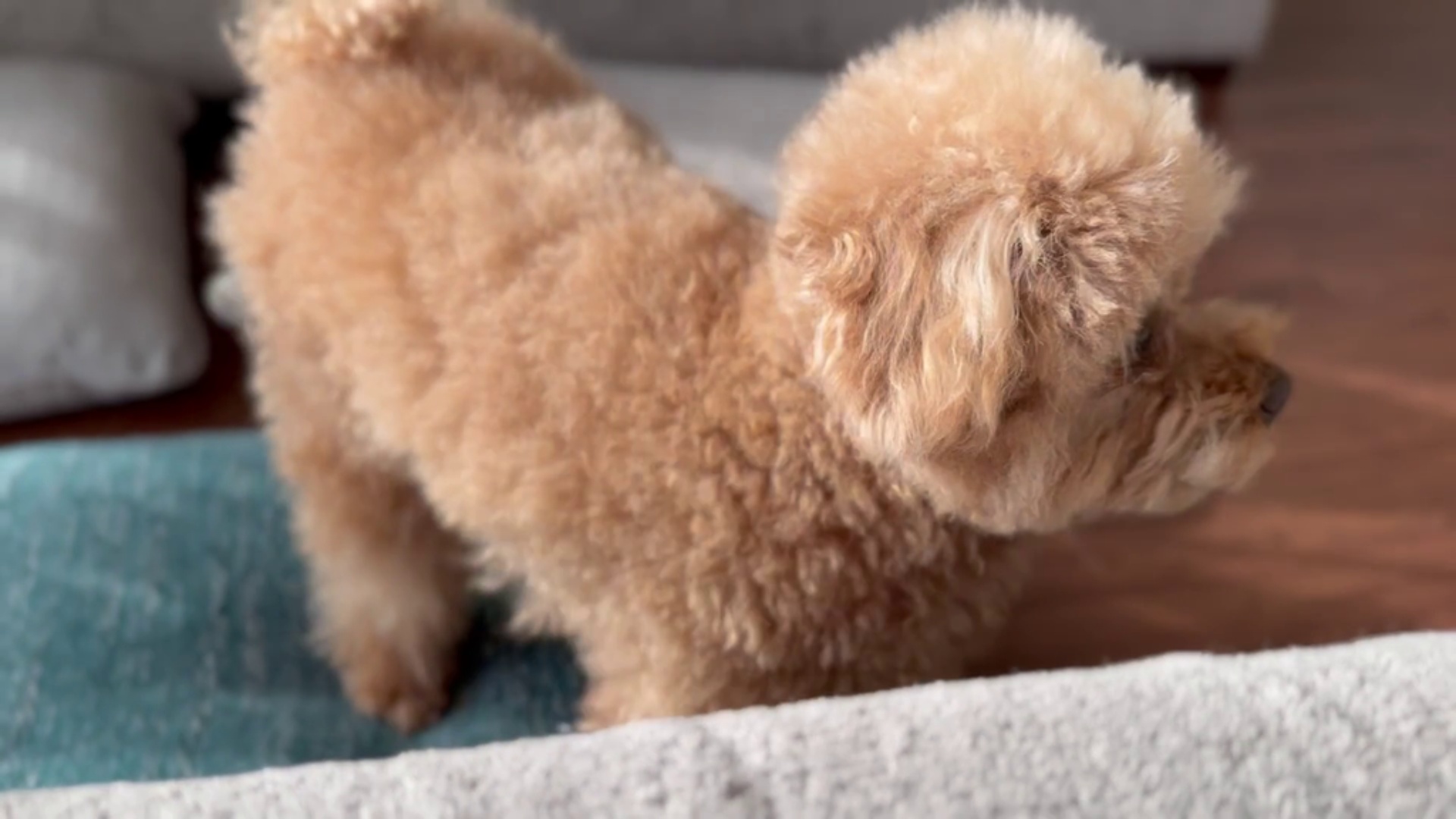} & 
\includegraphics[width=0.67\columnwidth]{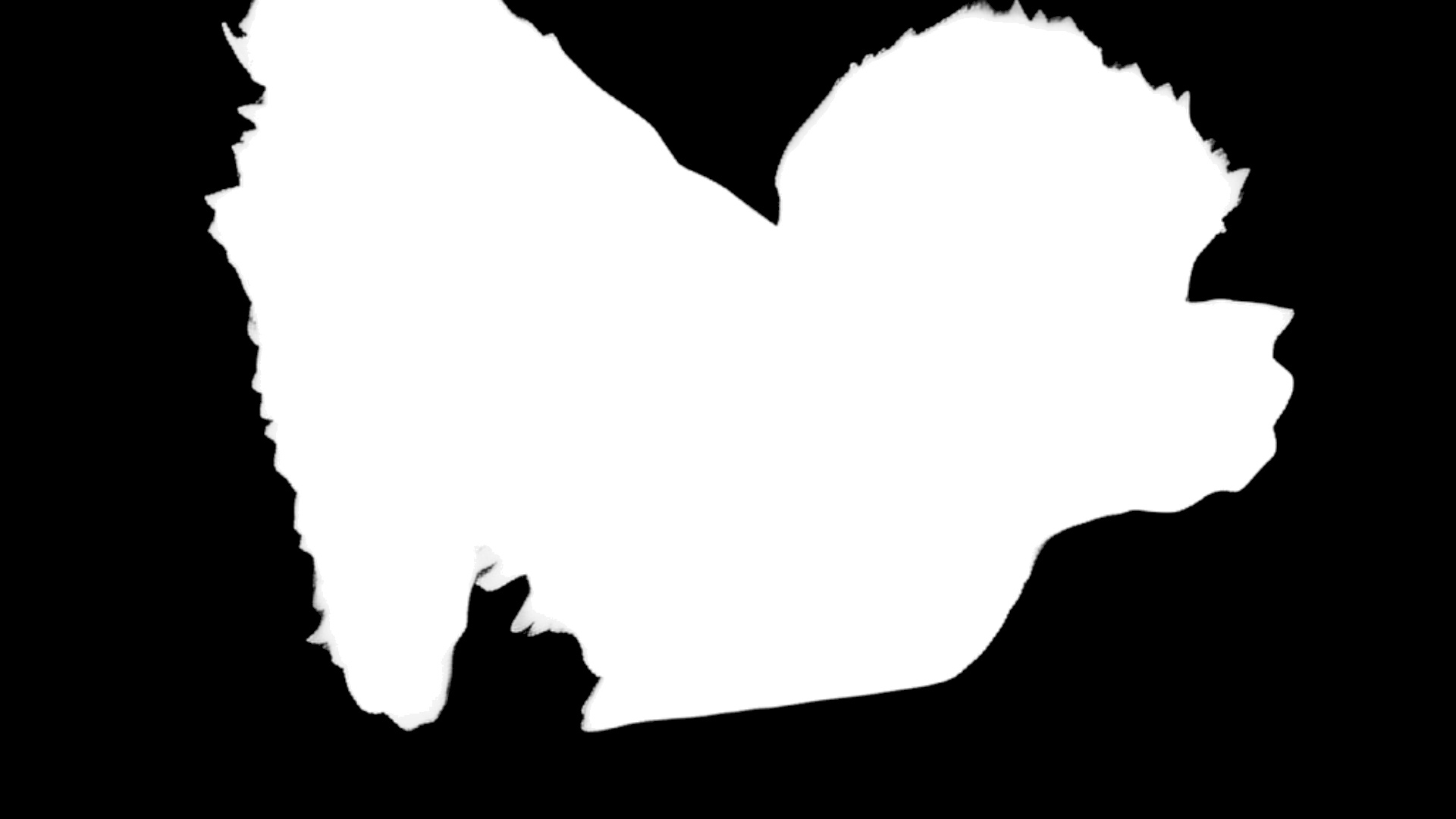} & 
\includegraphics[width=0.67\columnwidth]{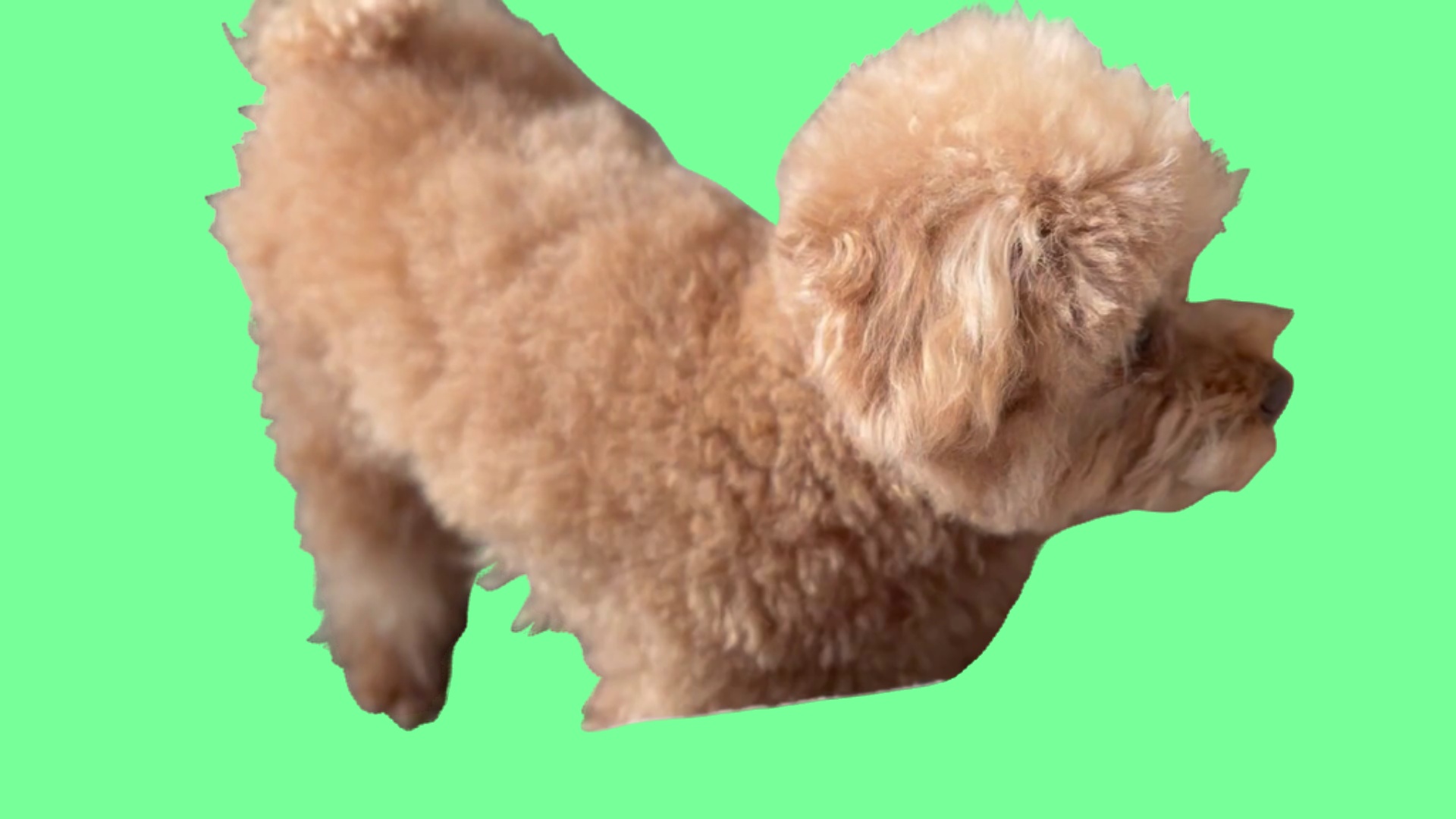} \\
\includegraphics[width=0.67\columnwidth]{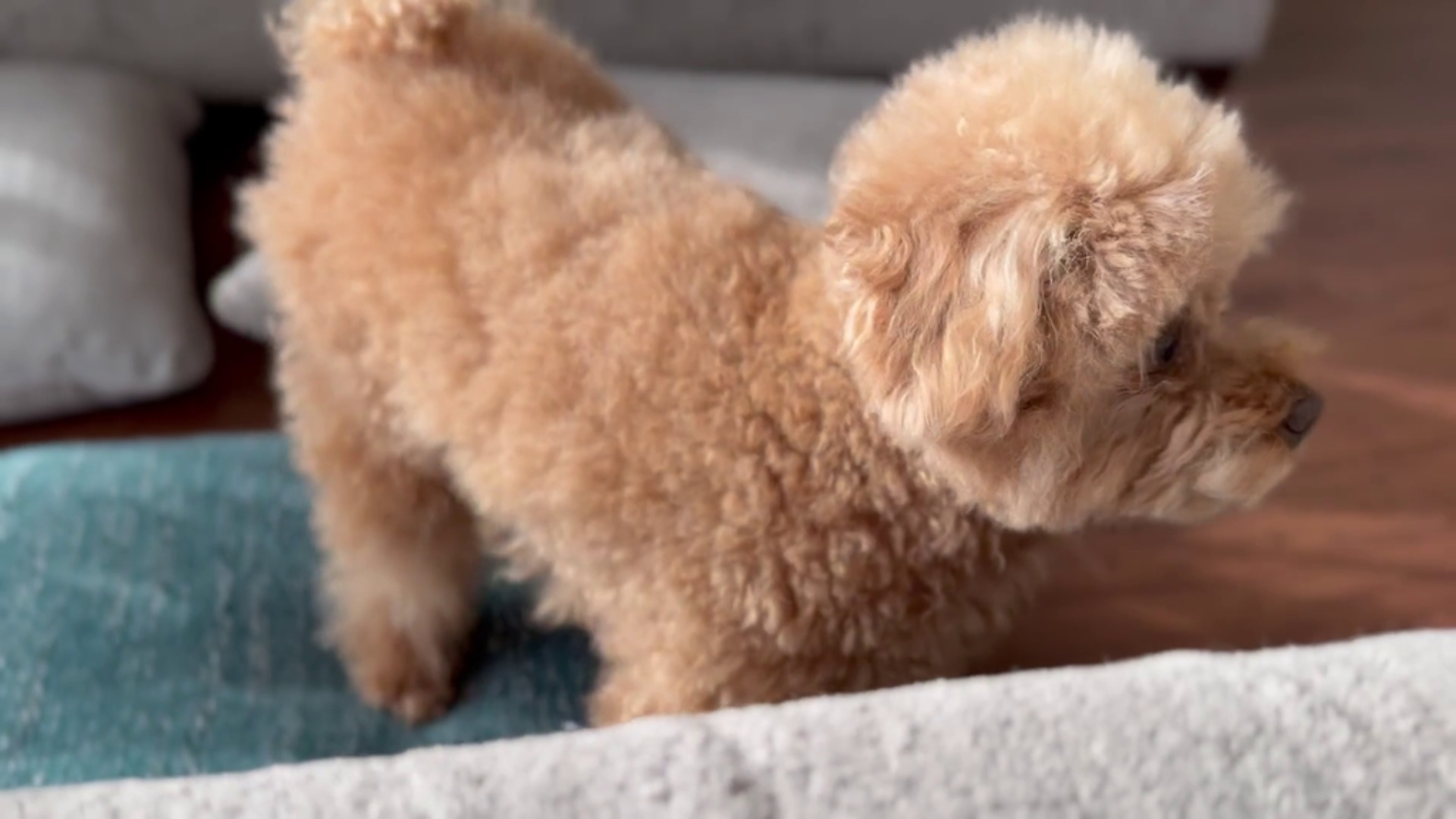} & 
\includegraphics[width=0.67\columnwidth]{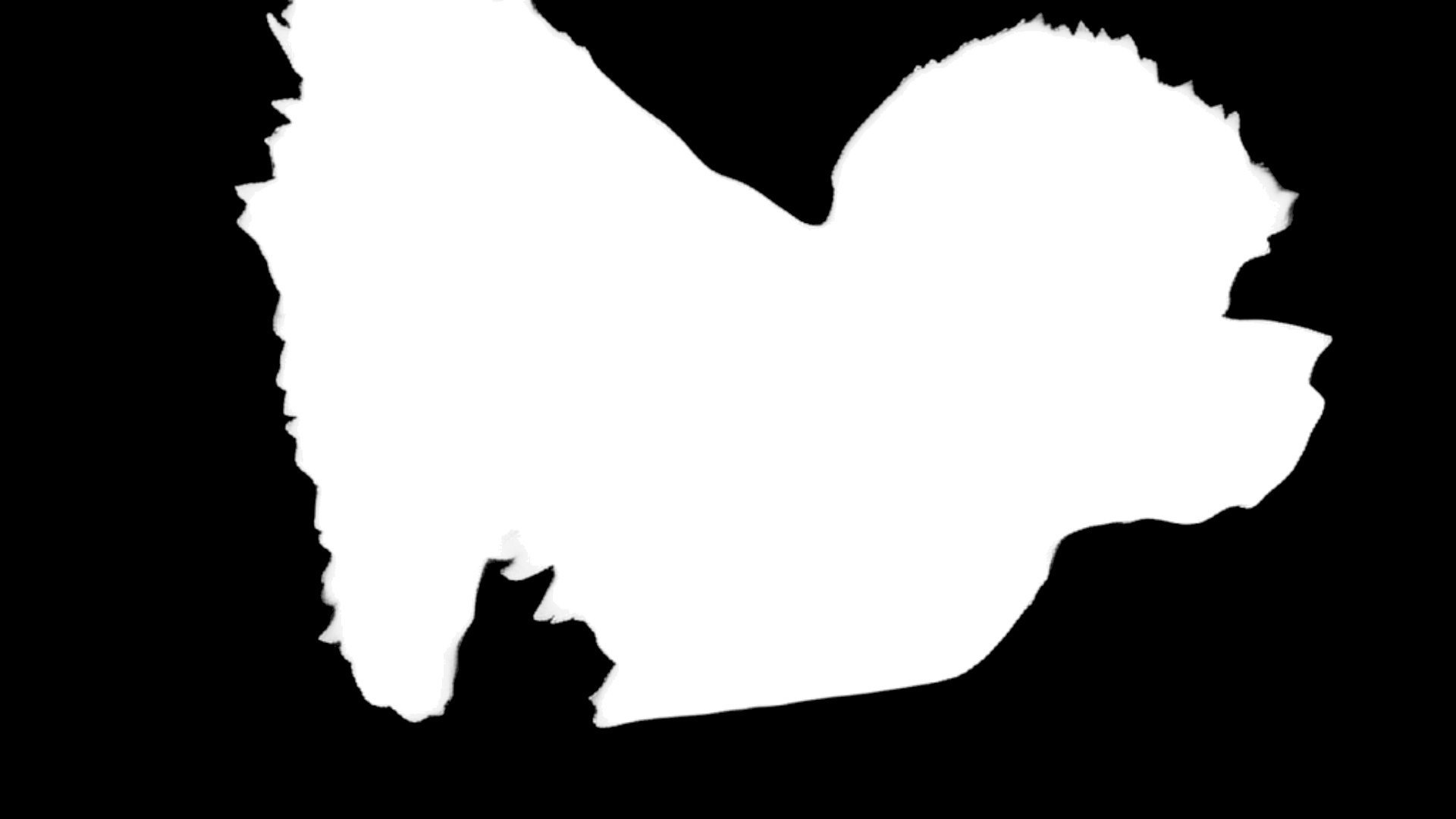} & 
\includegraphics[width=0.67\columnwidth]{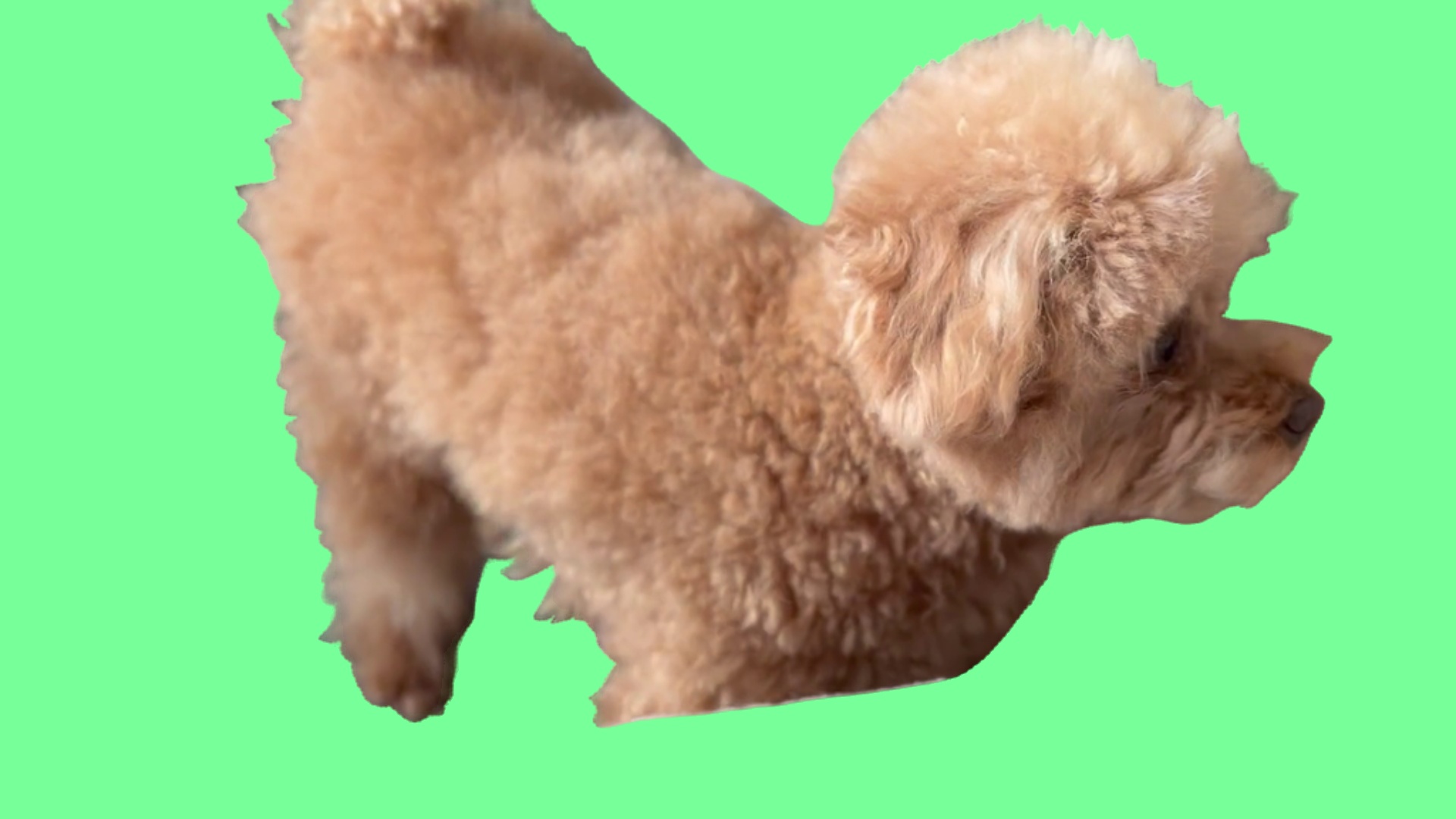} \\
Input & Ours Matte & Ours Composition  \\
\end{tabular}
\caption{Visualization on in-the-wild animals. \Ours has great generalization capability even if it is not trained on any animal matting dataset. It sometimes generates more fine-grained details than human annotations (the first row). RVM~\cite{lin2022robust} fails in this scenario.
}

\label{fig:supp_video_2}
\end{figure*}

\begin{figure*}
\small
\centering
\begin{tabular}{c@{\hspace{2.mm}} @{\hspace{-1.5mm}}c@{\hspace{2.mm}}@{\hspace{-1.5mm}}c@{\hspace{2.mm}}}
\includegraphics[width=0.67\columnwidth]{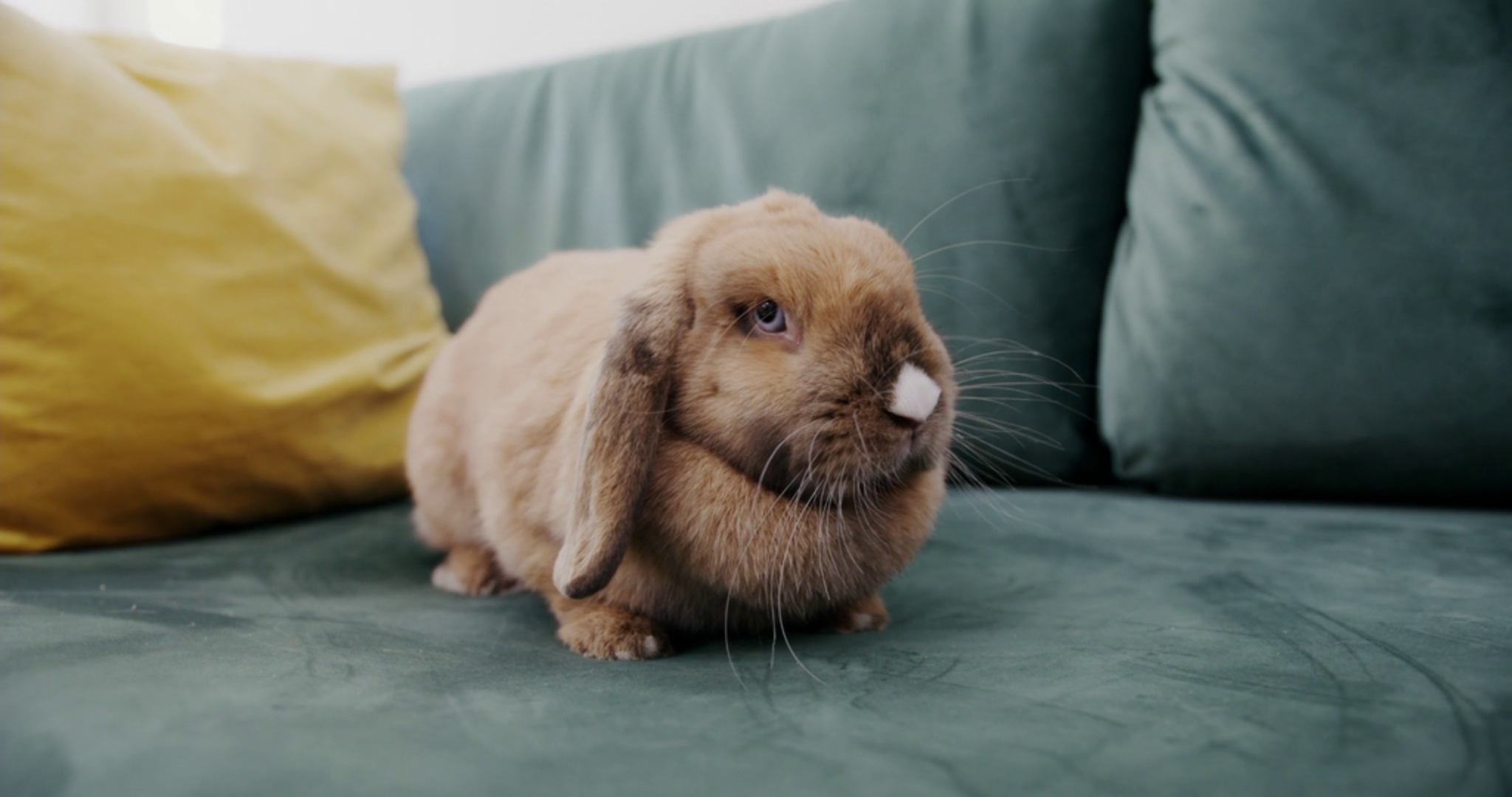} & 
\includegraphics[width=0.67\columnwidth]{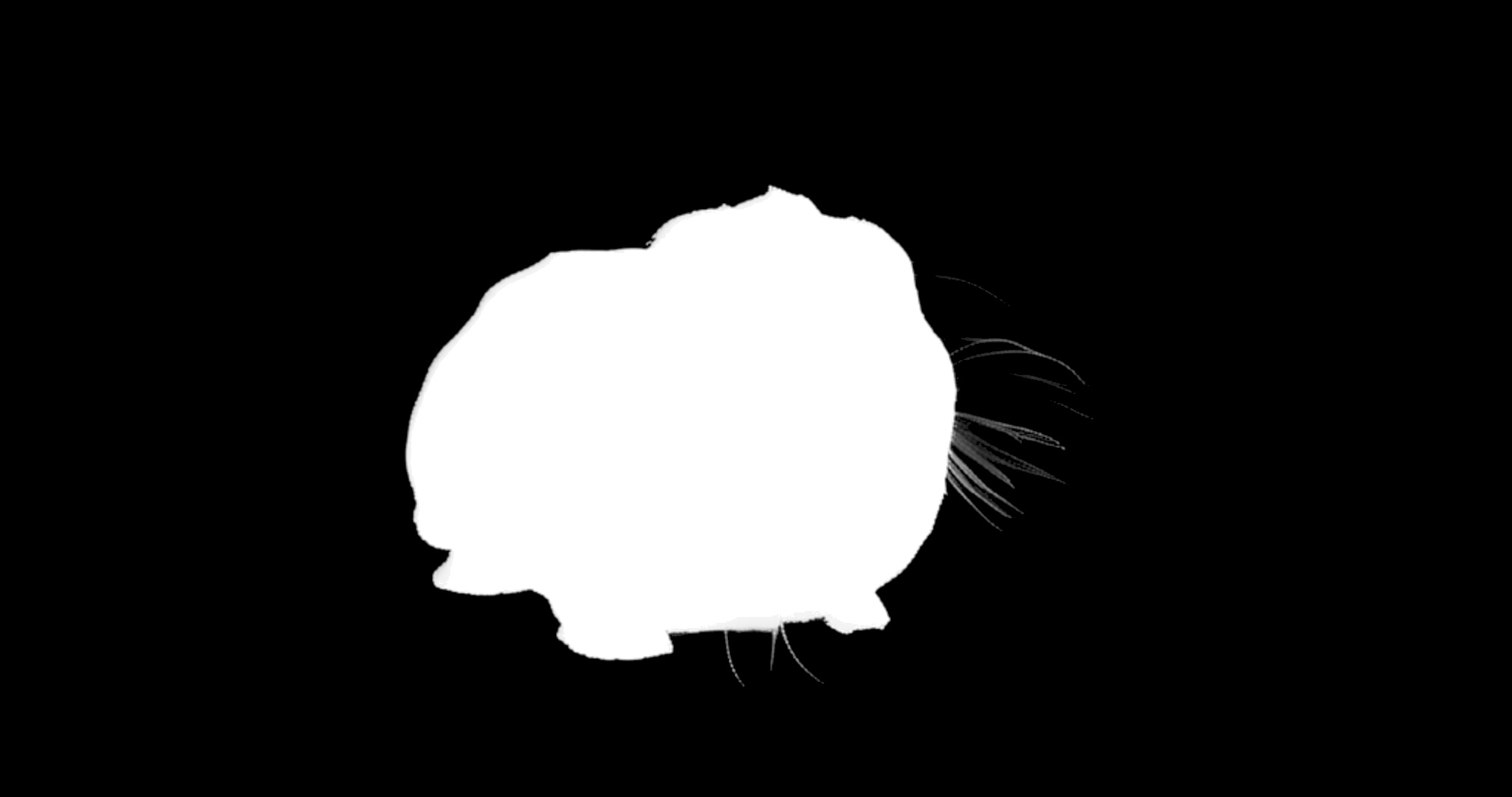} & 
\includegraphics[width=0.67\columnwidth]{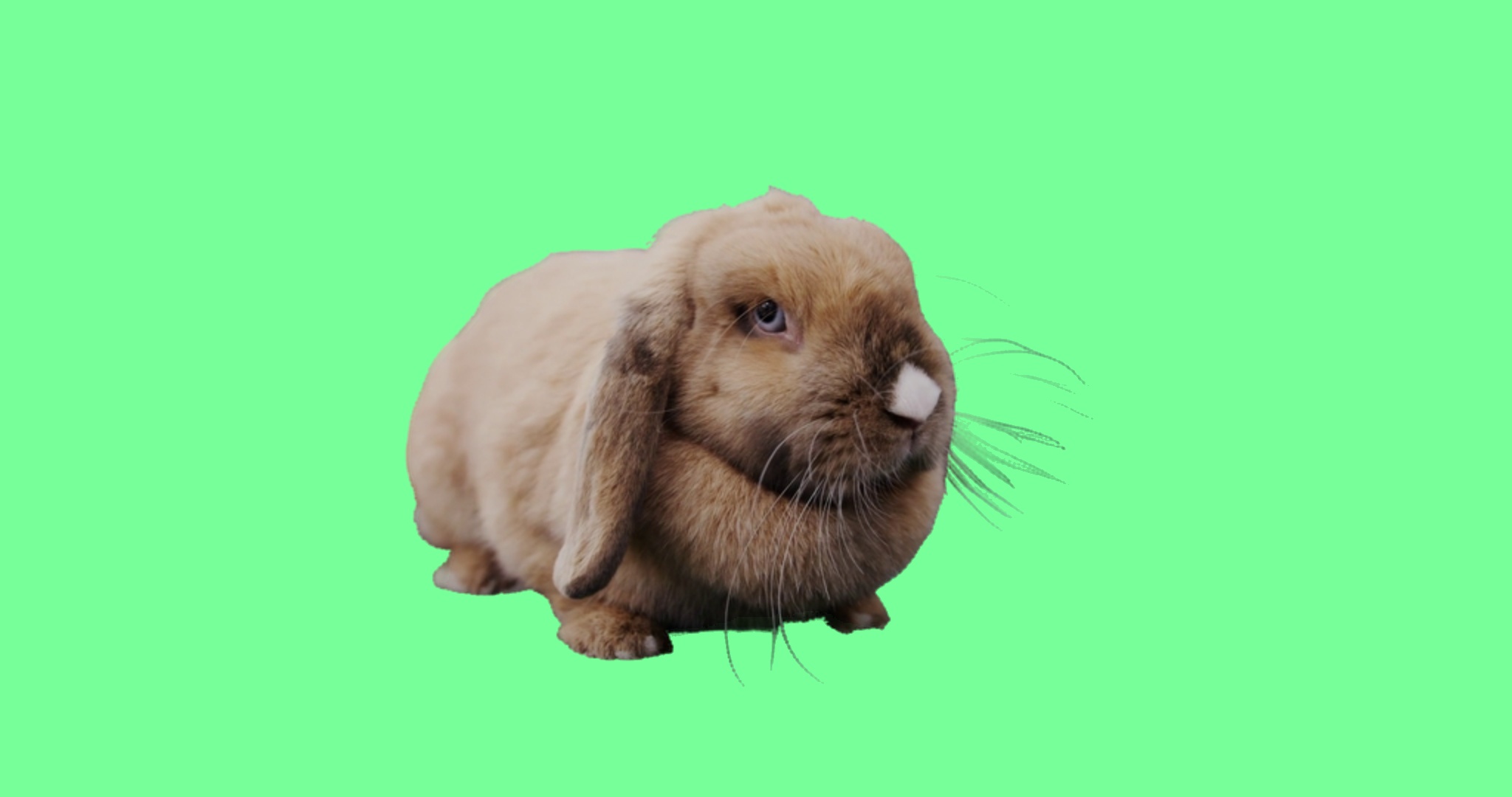} \\
\includegraphics[width=0.67\columnwidth]{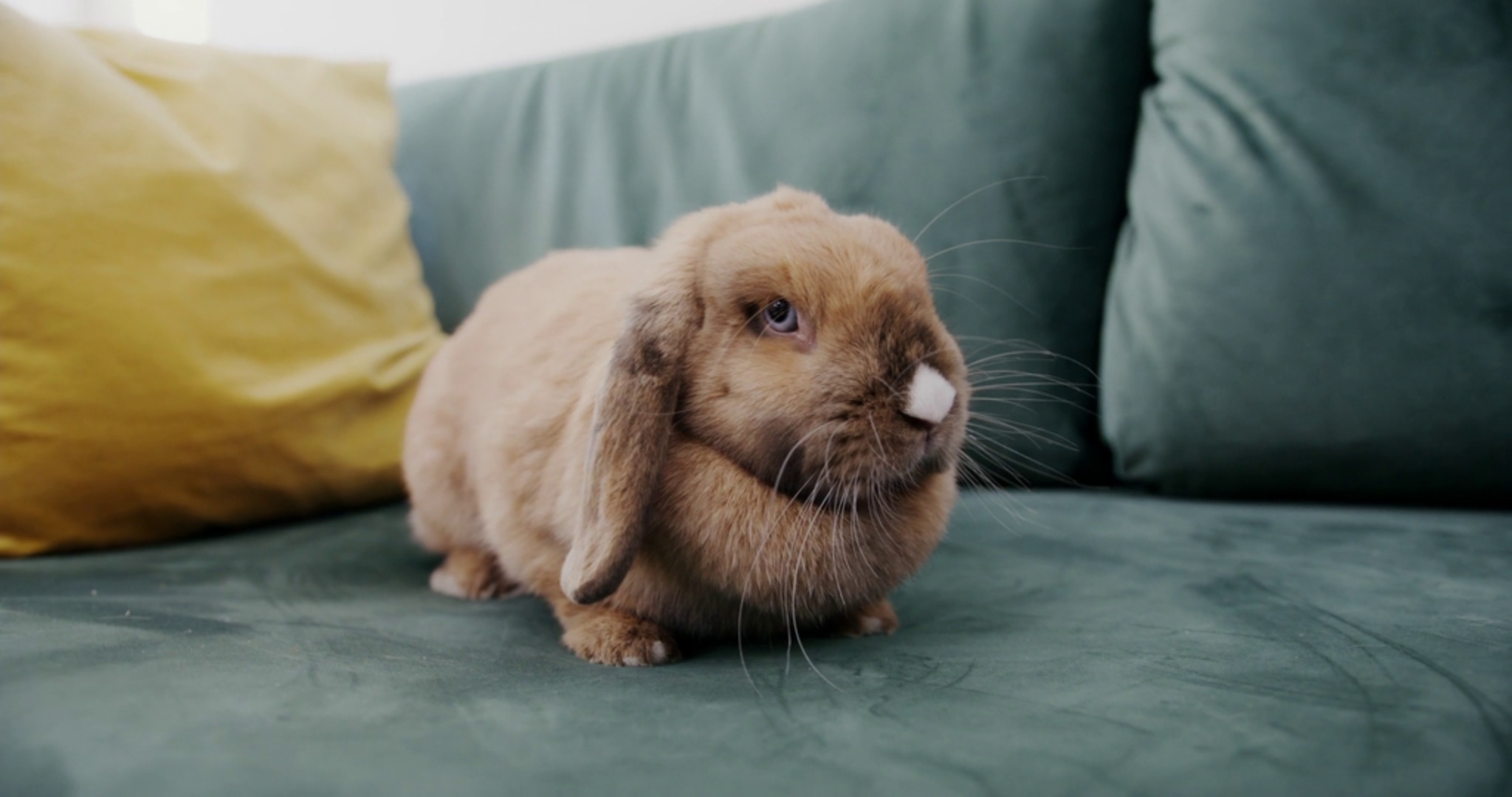} & 
\includegraphics[width=0.67\columnwidth]{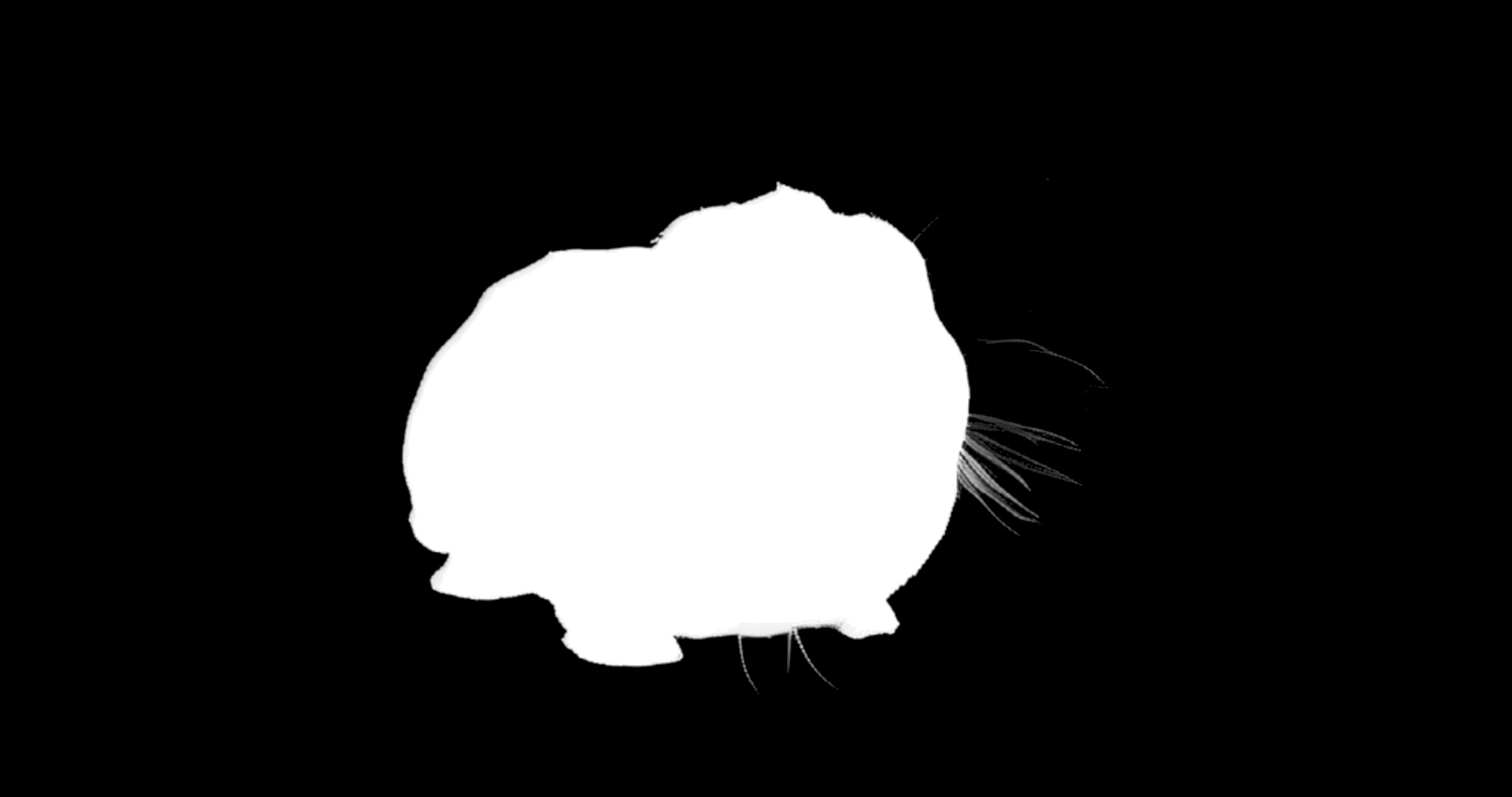} & 
\includegraphics[width=0.67\columnwidth]{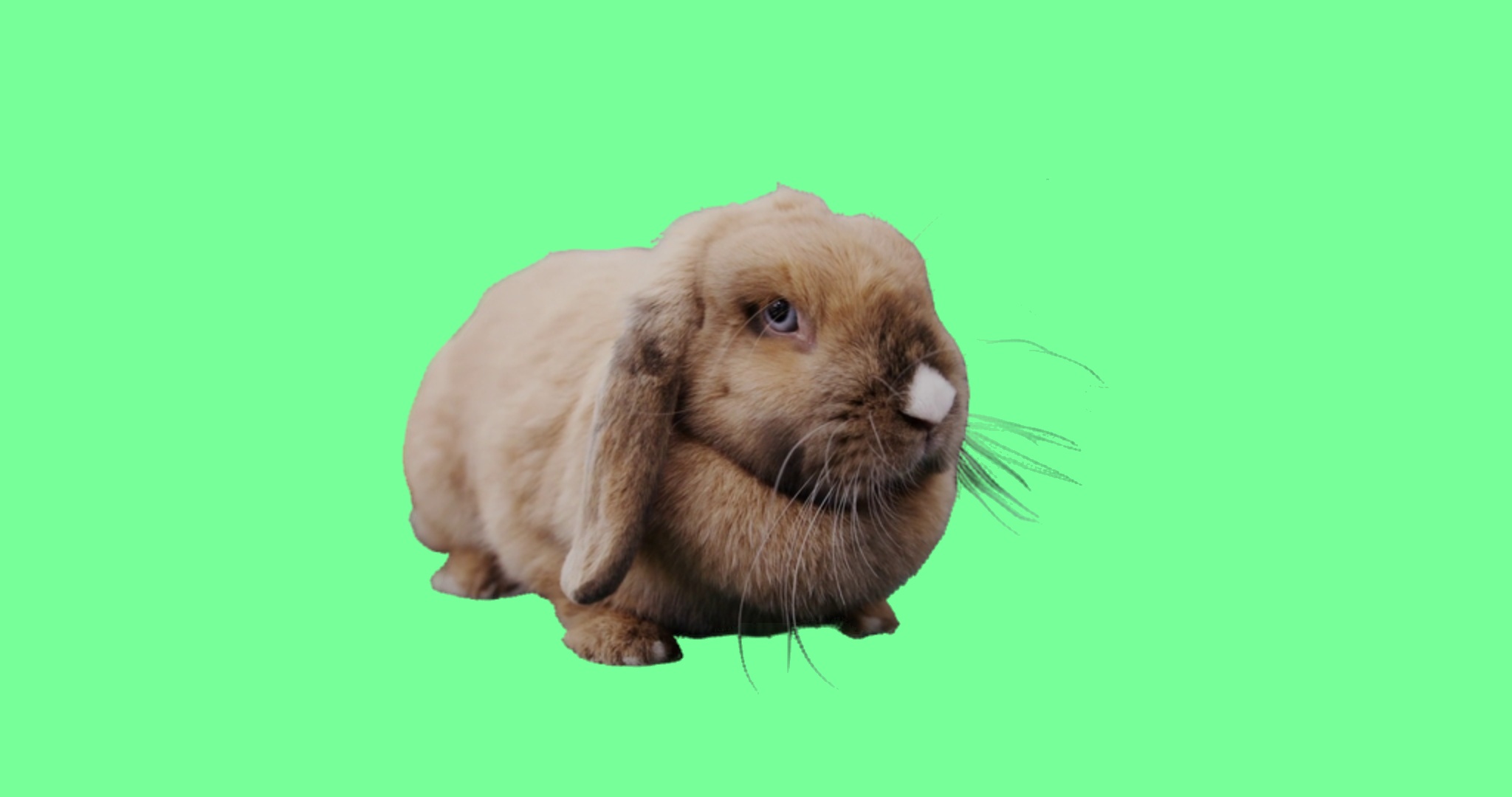} \\
\includegraphics[width=0.67\columnwidth]{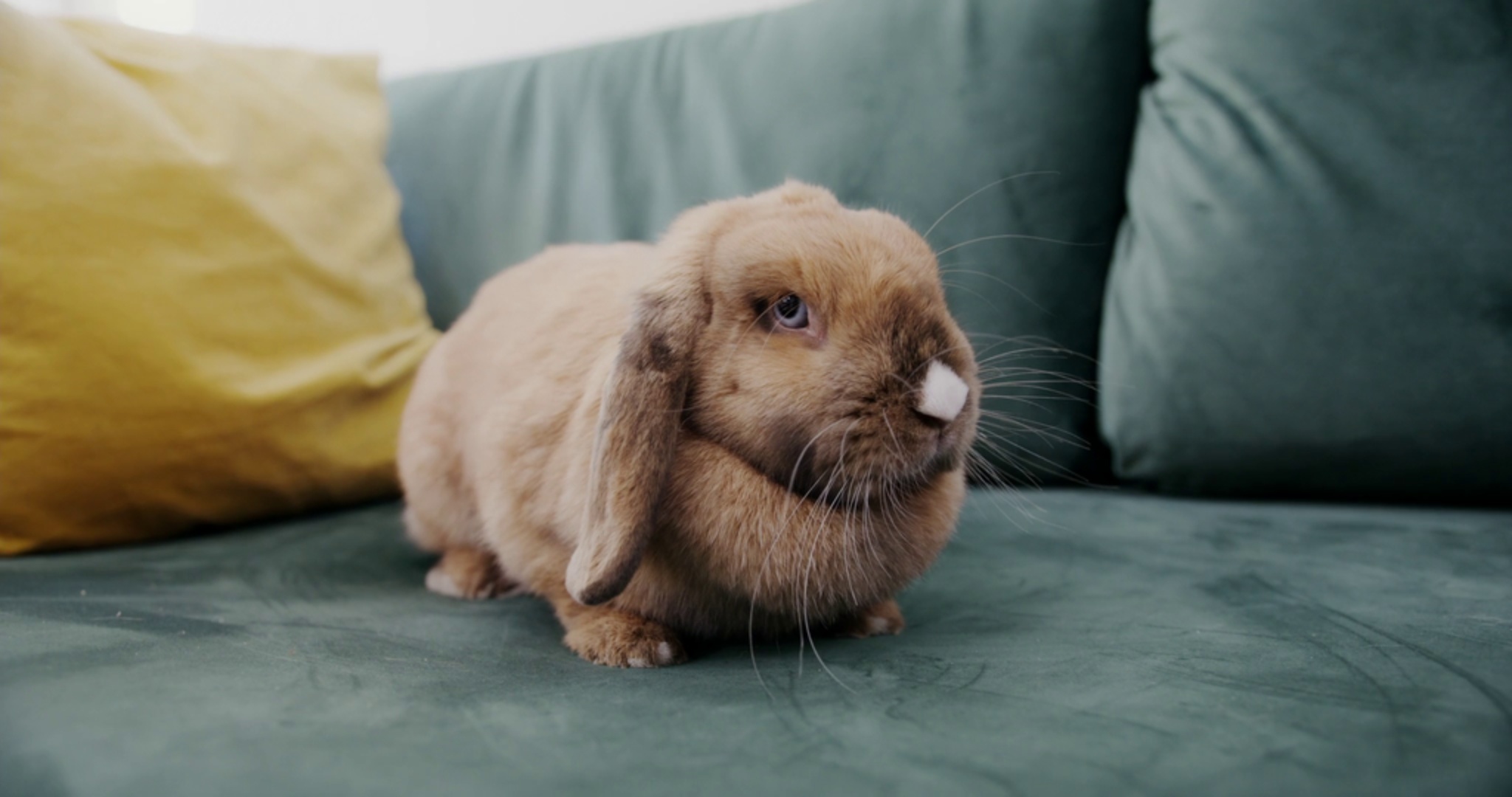} & 
\includegraphics[width=0.67\columnwidth]{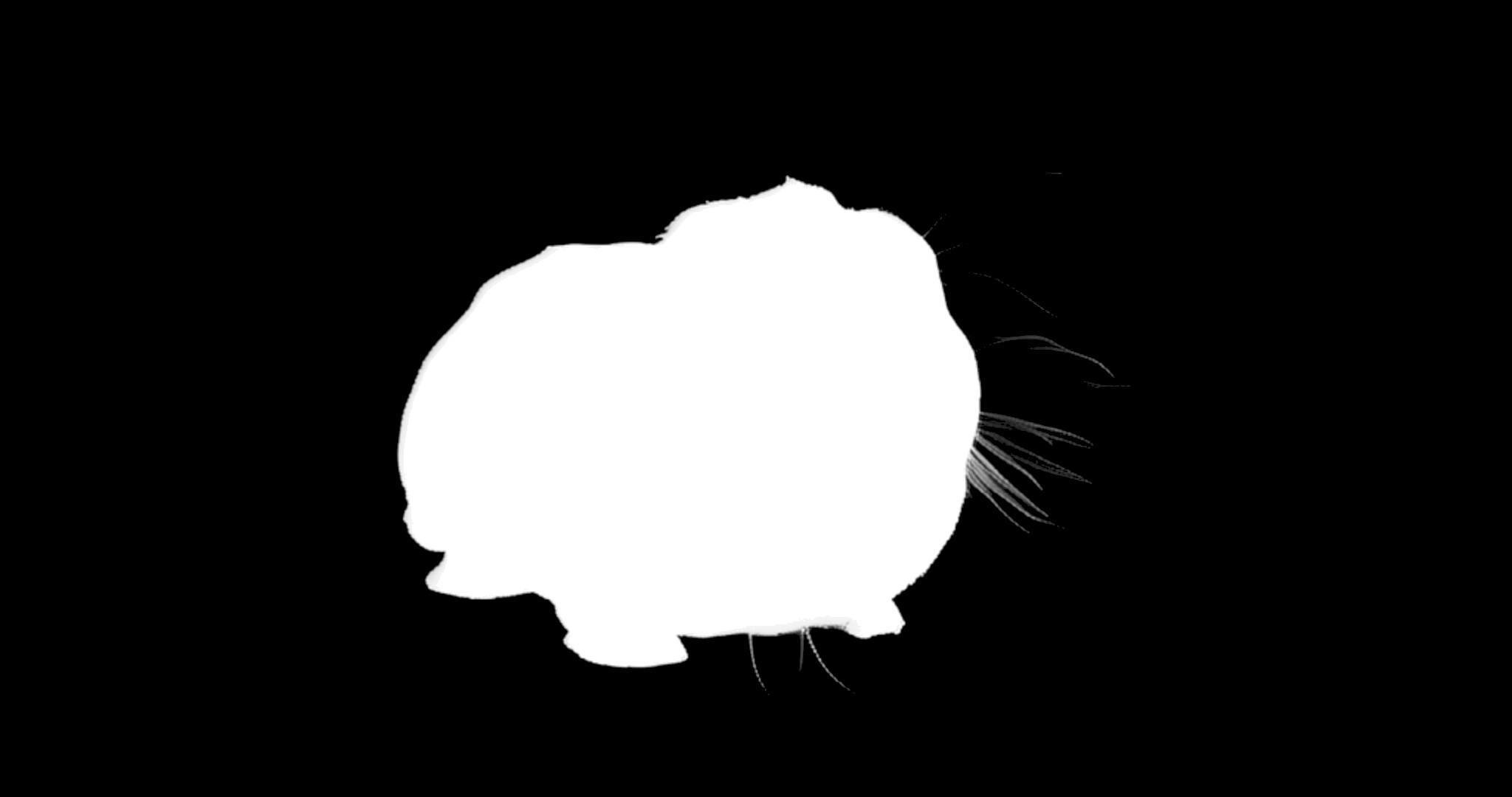} & 
\includegraphics[width=0.67\columnwidth]{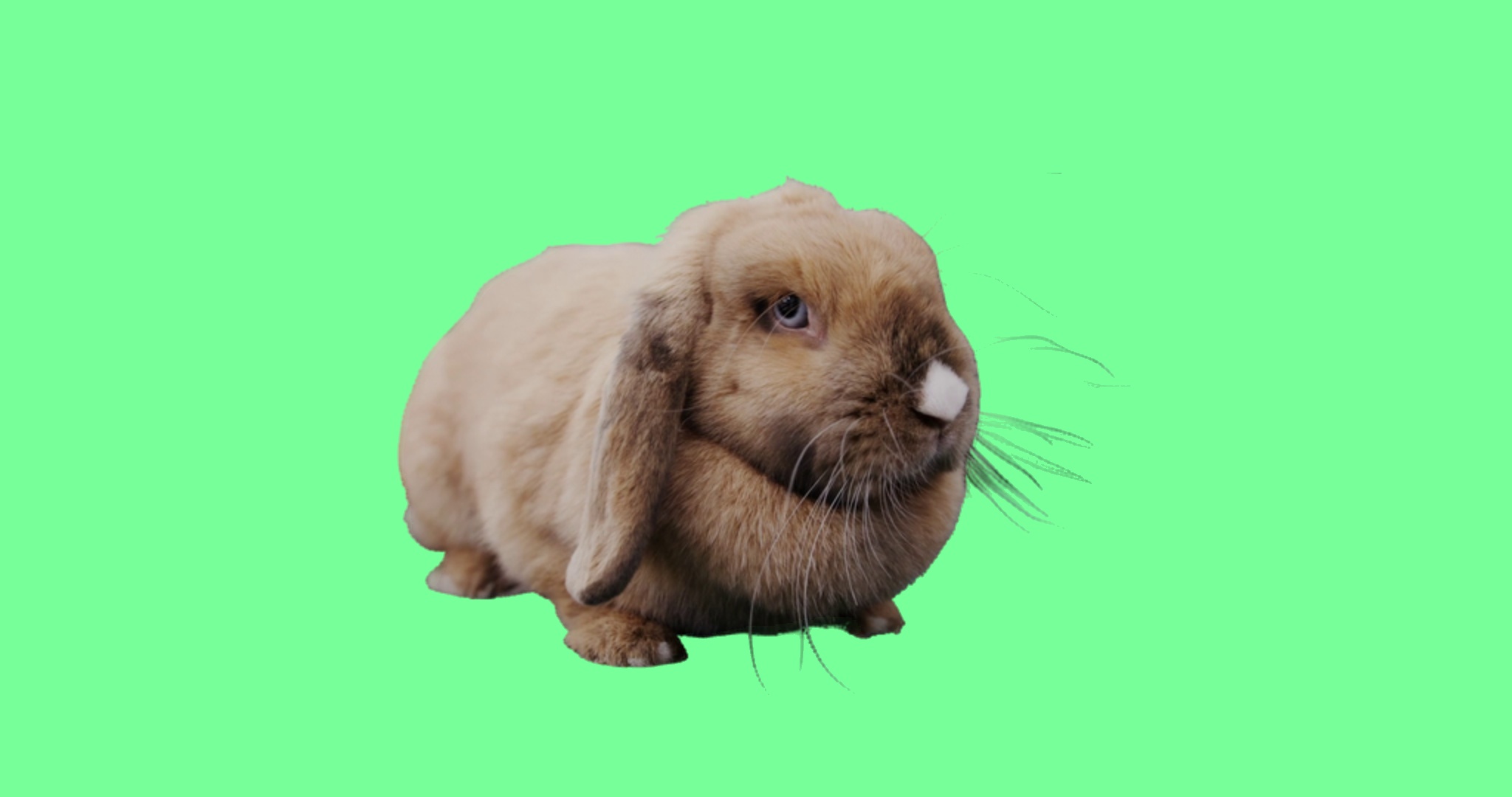} \\
\includegraphics[width=0.67\columnwidth]{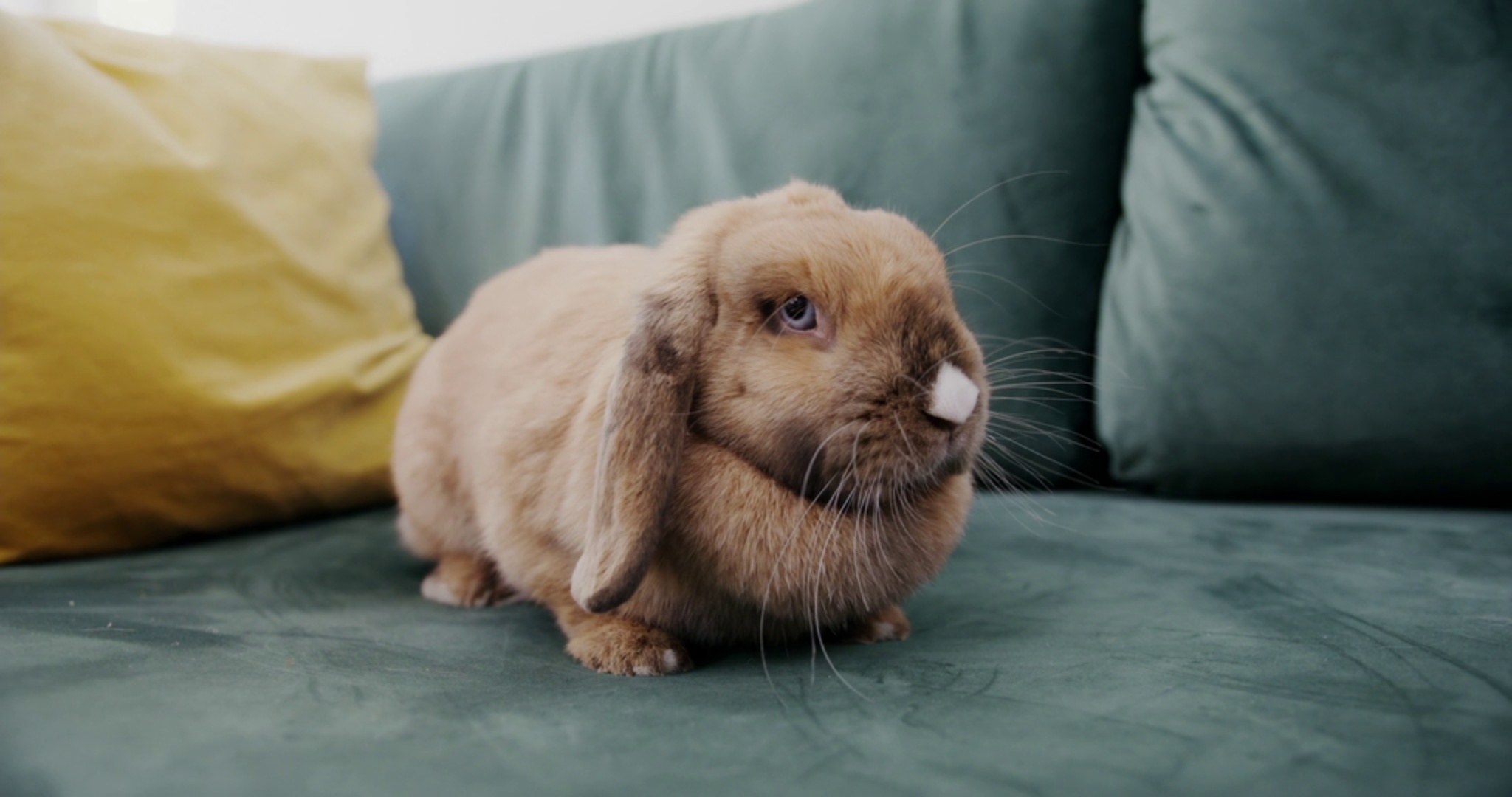} & 
\includegraphics[width=0.67\columnwidth]{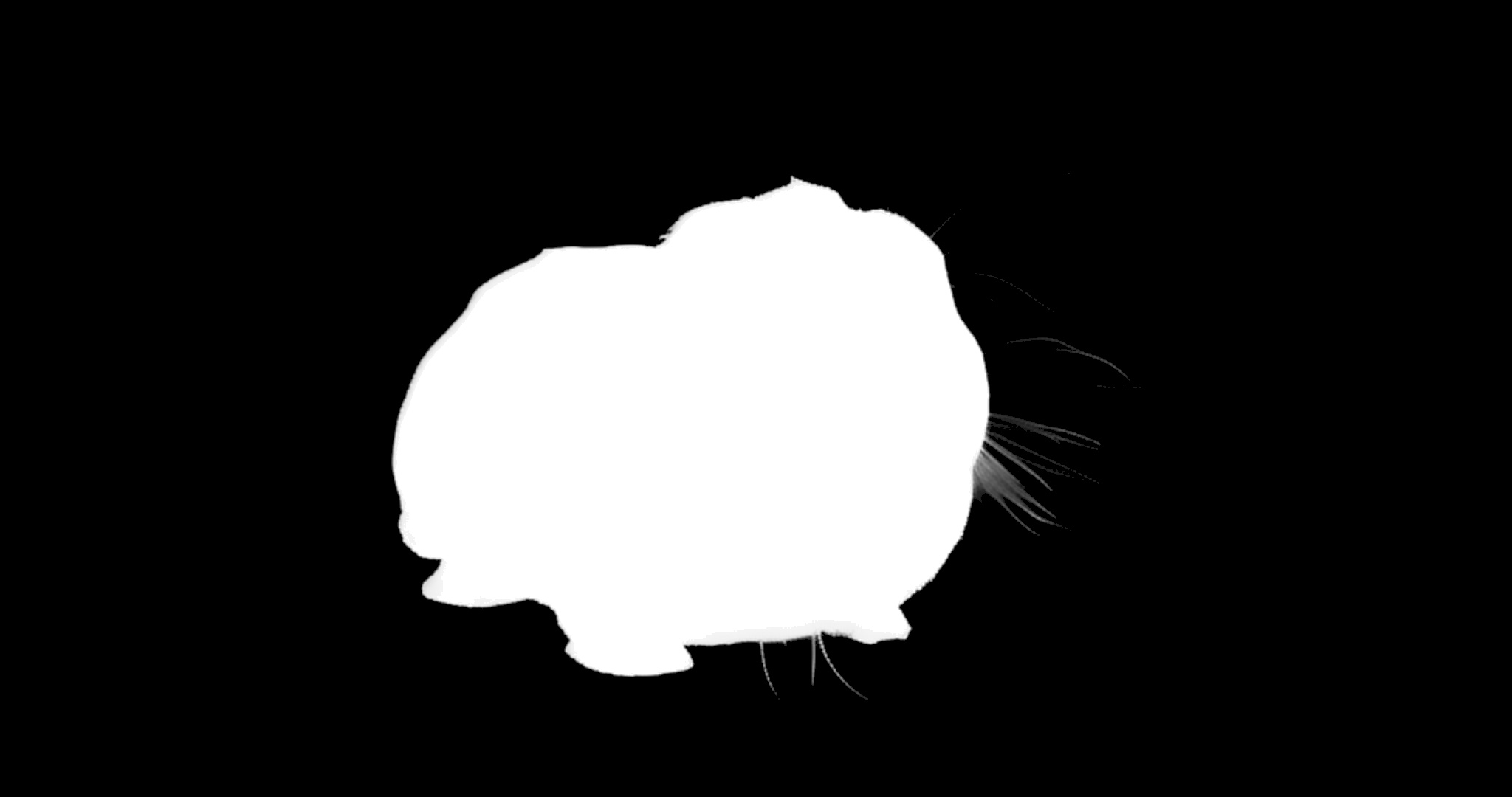} & 
\includegraphics[width=0.67\columnwidth]{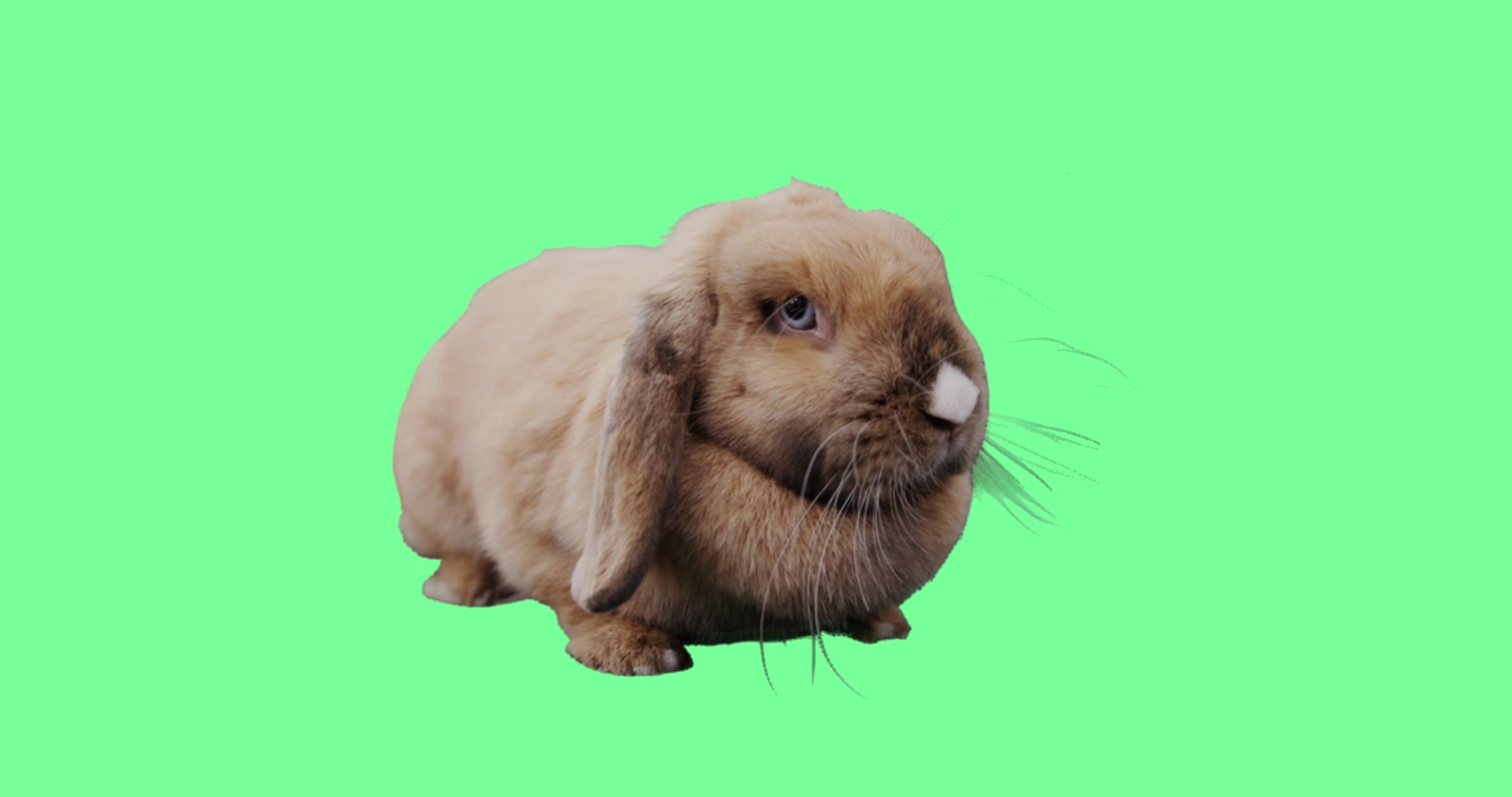} \\
\includegraphics[width=0.67\columnwidth]{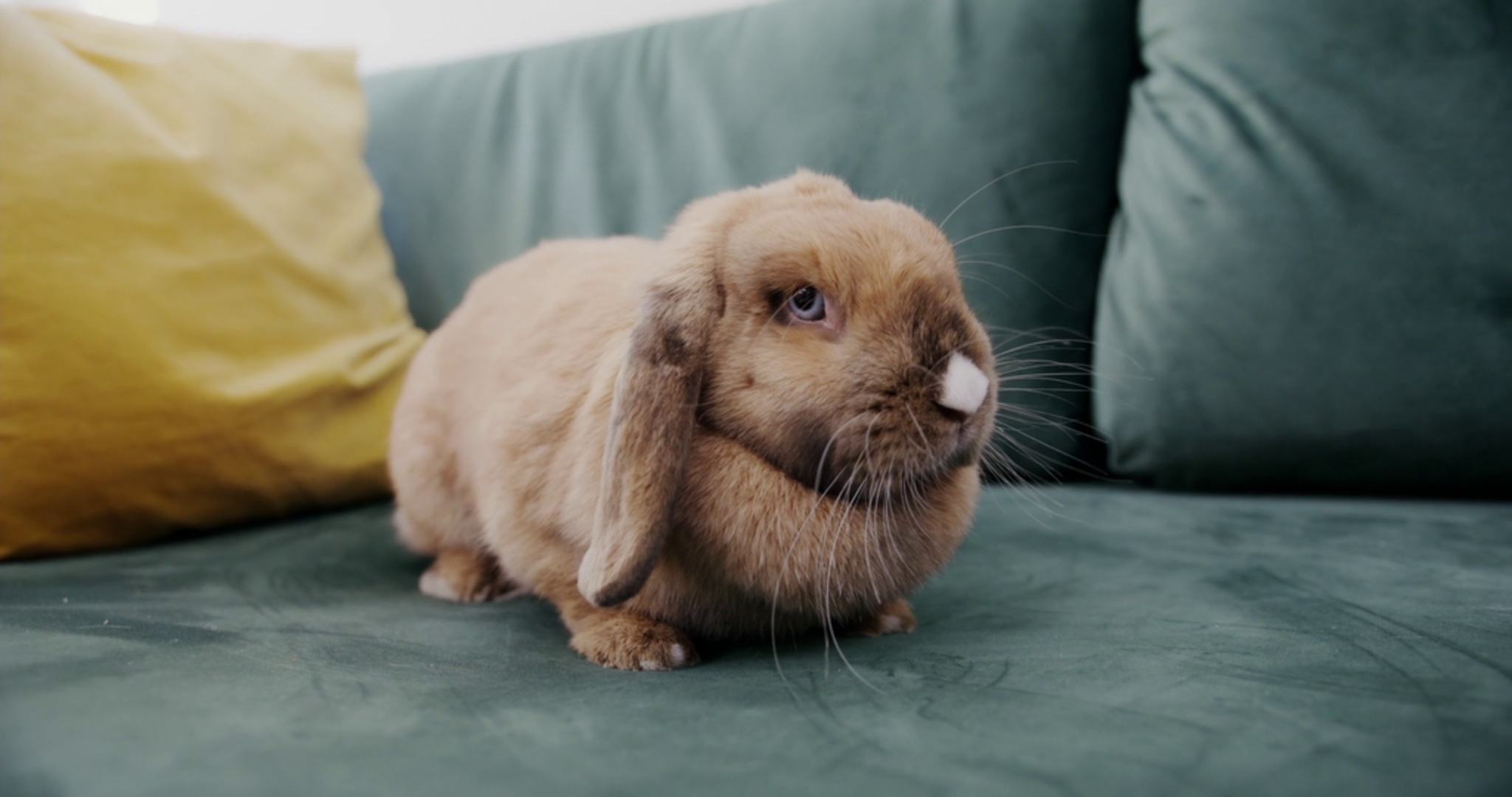} & 
\includegraphics[width=0.67\columnwidth]{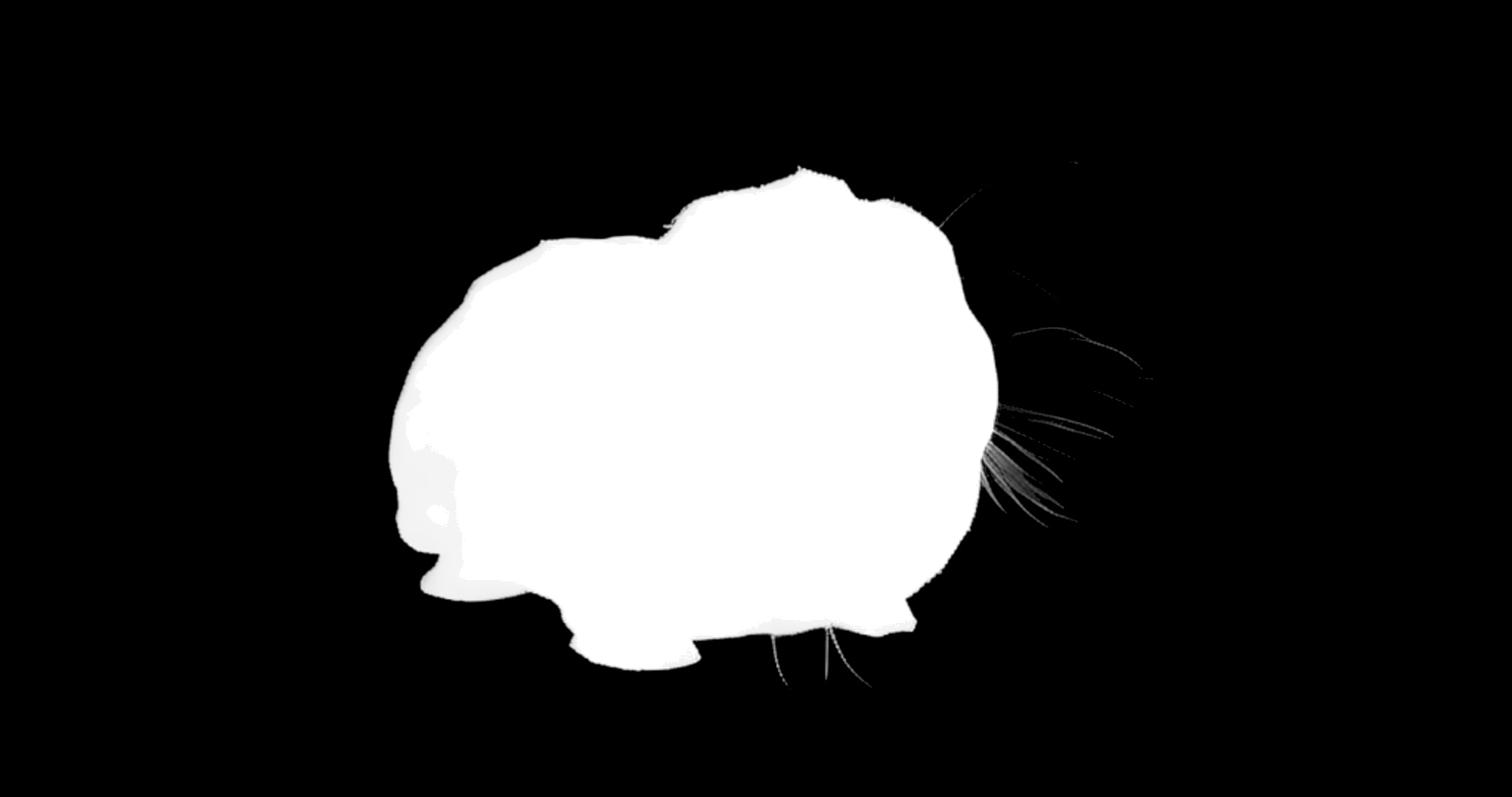} & 
\includegraphics[width=0.67\columnwidth]{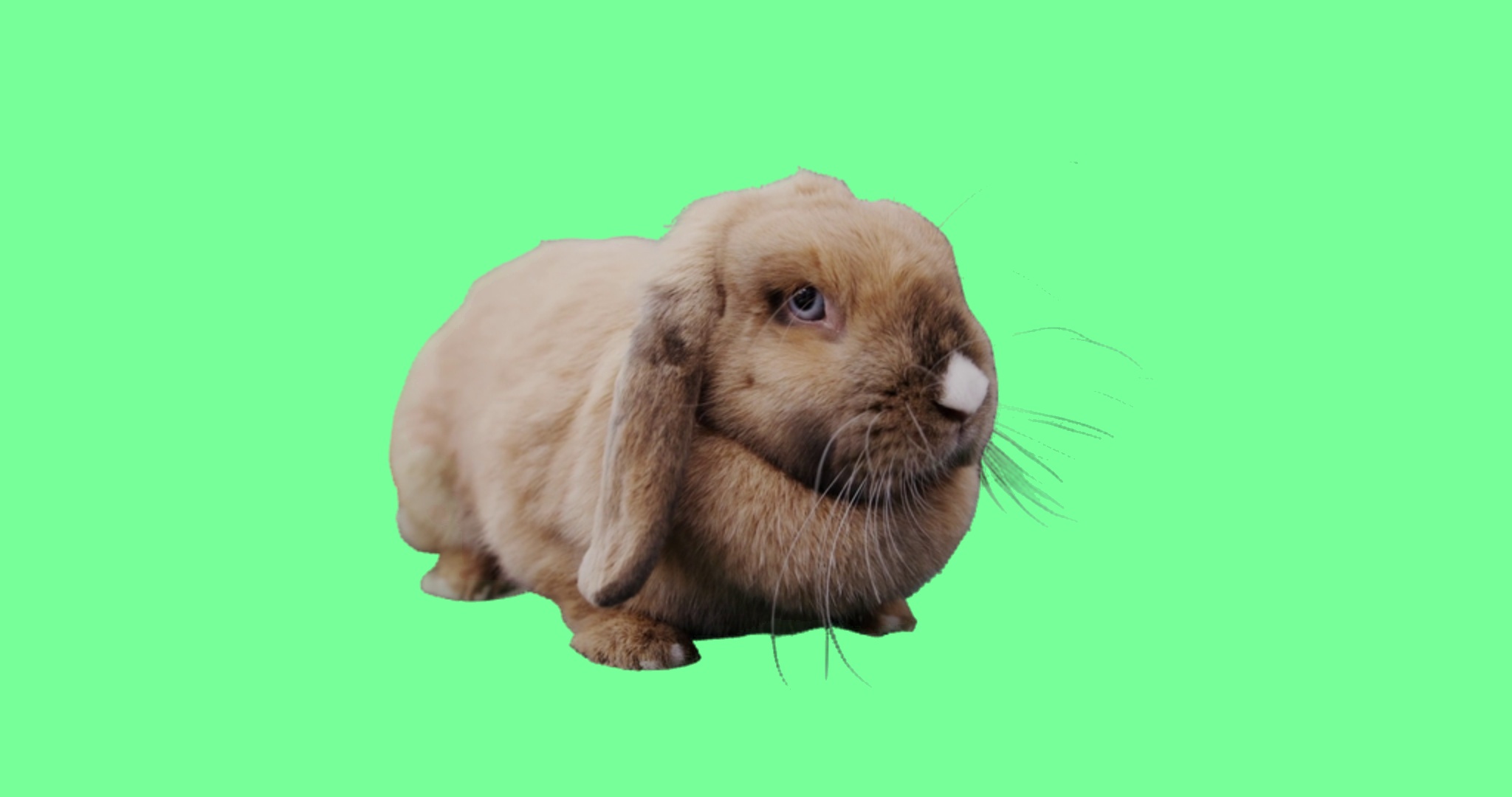} \\
\includegraphics[width=0.67\columnwidth]{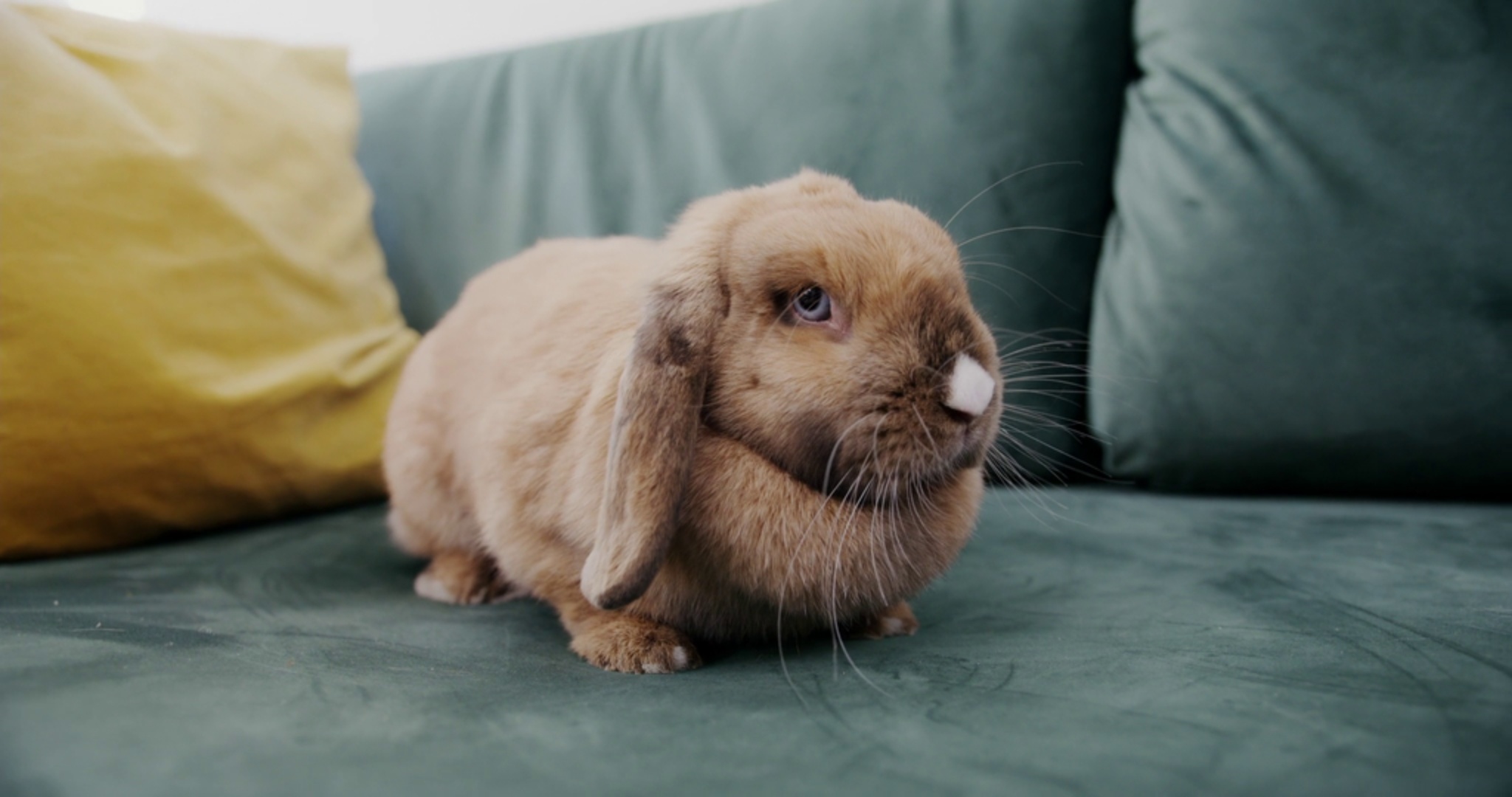} & 
\includegraphics[width=0.67\columnwidth]{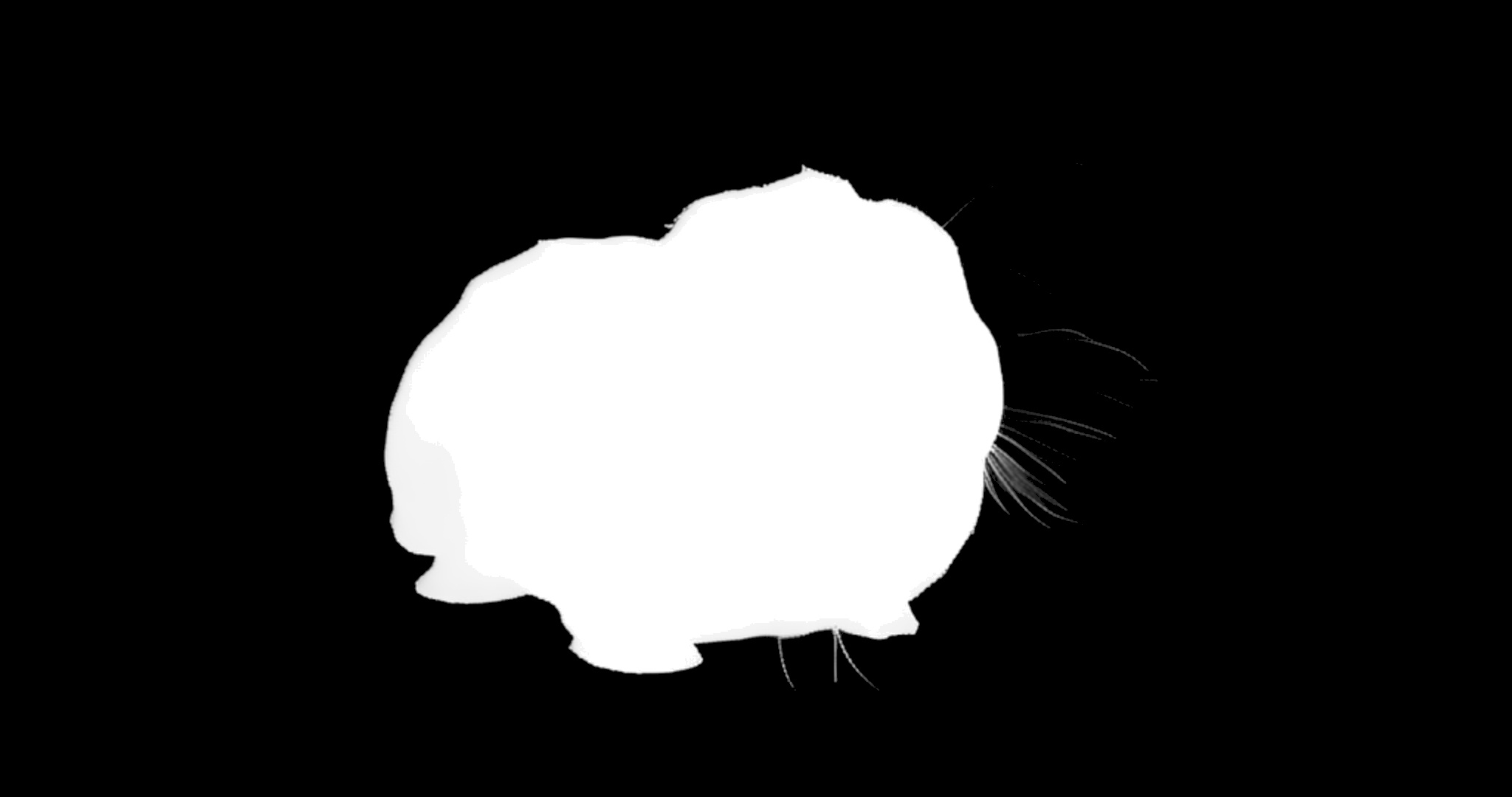} & 
\includegraphics[width=0.67\columnwidth]{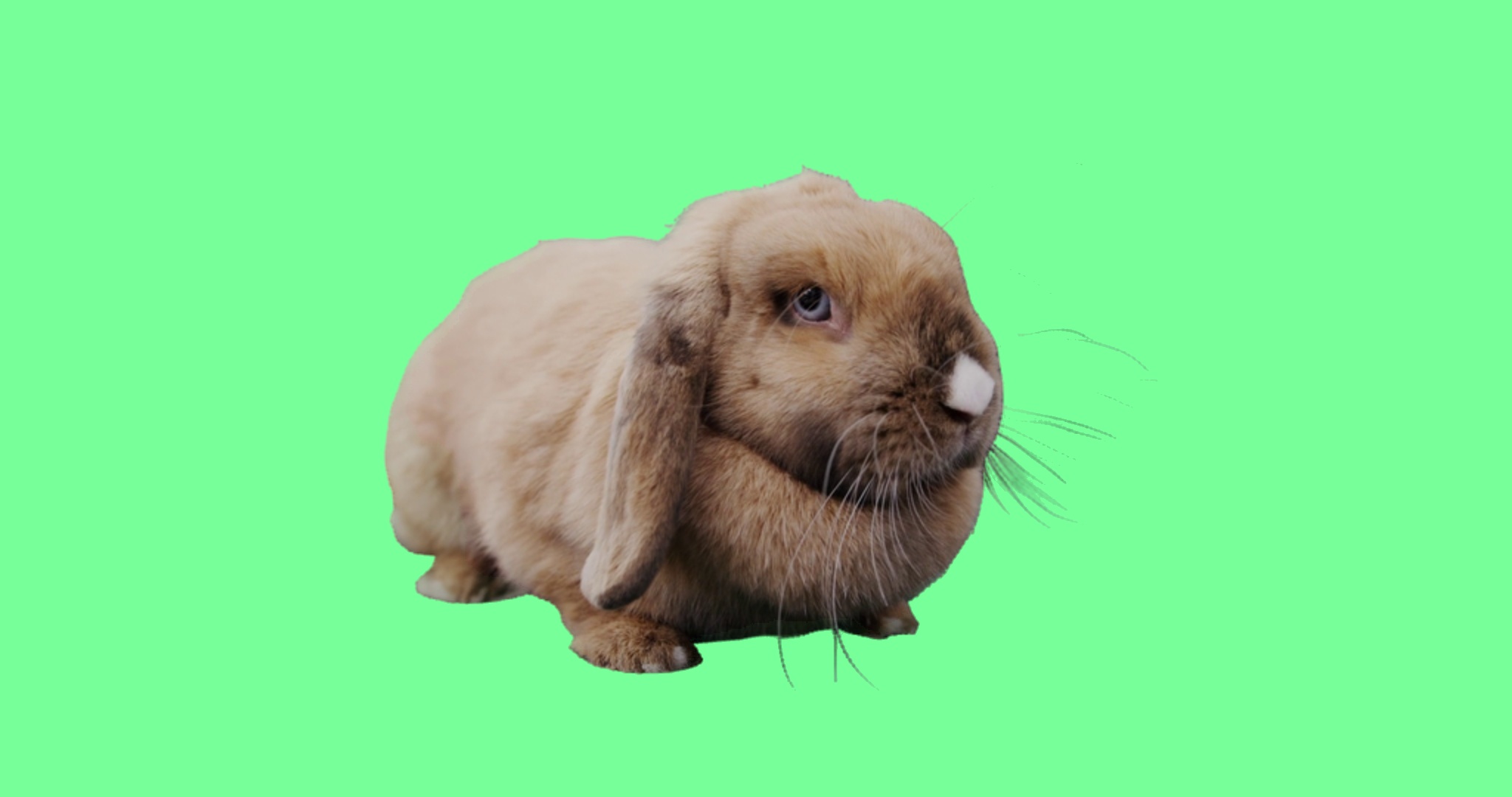} \\
Input & Ours Matte & Ours Composition  \\
\end{tabular}
\caption{Although our model was not trained on any animal datasets containing fur, it demonstrates strong generalization capabilities, successfully predicting fine details such as bunny whiskers. (RVM~\cite{lin2022robust} fails in this scenario)}

\label{fig:supp_video_3}
\end{figure*}

 We provide more video results in~\Cref{fig:supp_video_0}, ~\Cref{fig:supp_video_1}, ~\Cref{fig:supp_video_2} and ~\Cref{fig:supp_video_3} by comparing our method with RVM~\cite{lin2022robust}. Our method generates more fine-grained and temporal consistent alpha mattes in all the scenes. Notably, RVM is not generalizable to scenes containing animals, while our method successfully predicts furs and whiskers of the animals.

\begin{figure*}
\centering
\begin{tabular}{c@{\hspace{8.mm}} @{\hspace{-7mm}}c@{\hspace{8.mm}}}
\includegraphics[width=1.\columnwidth]{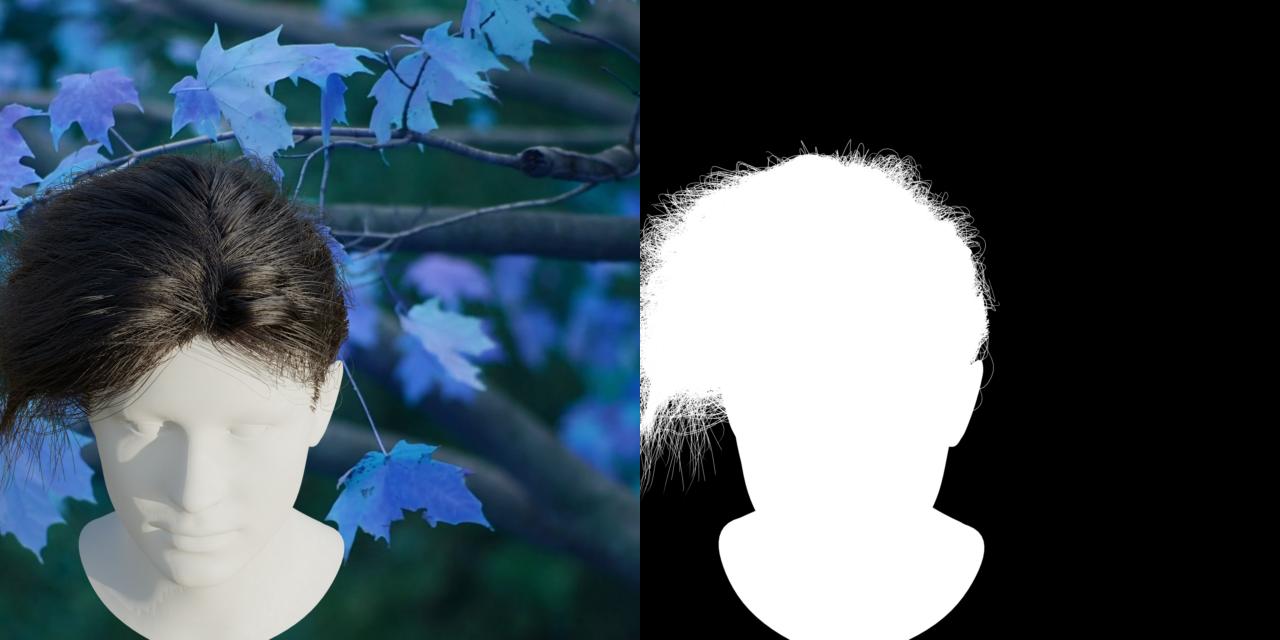} & 
\includegraphics[width=1.\columnwidth]{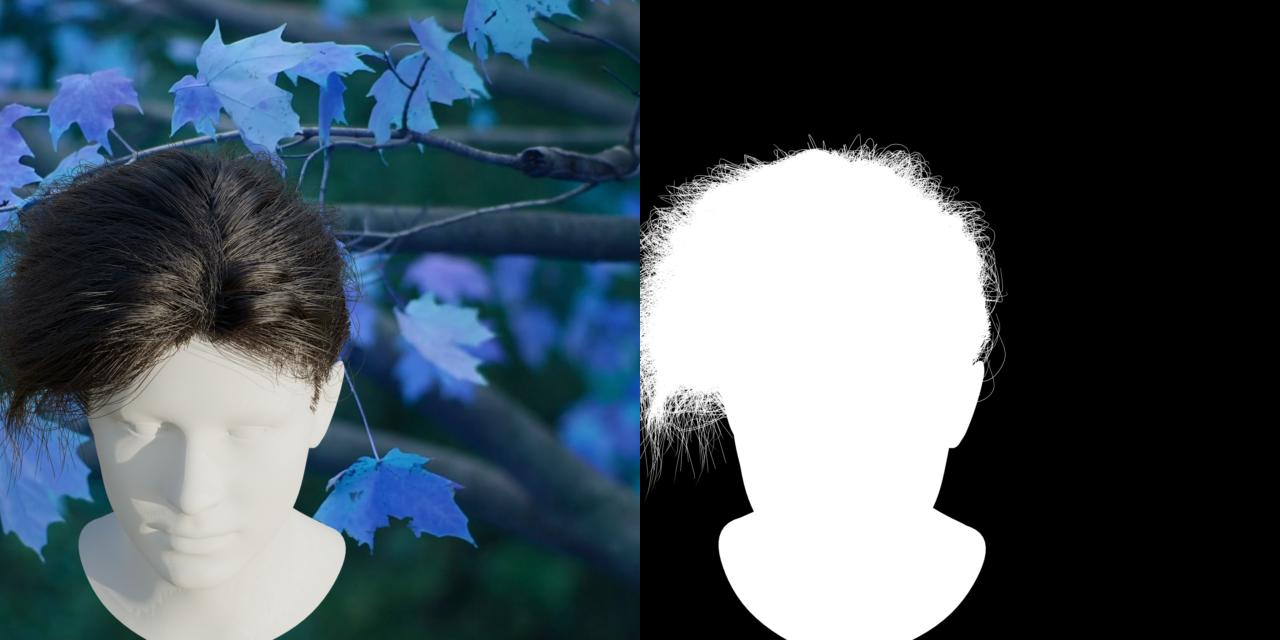}  \\
Sample 0, Frame 0 & Sample 0, Frame 1  \\
\includegraphics[width=1.\columnwidth]{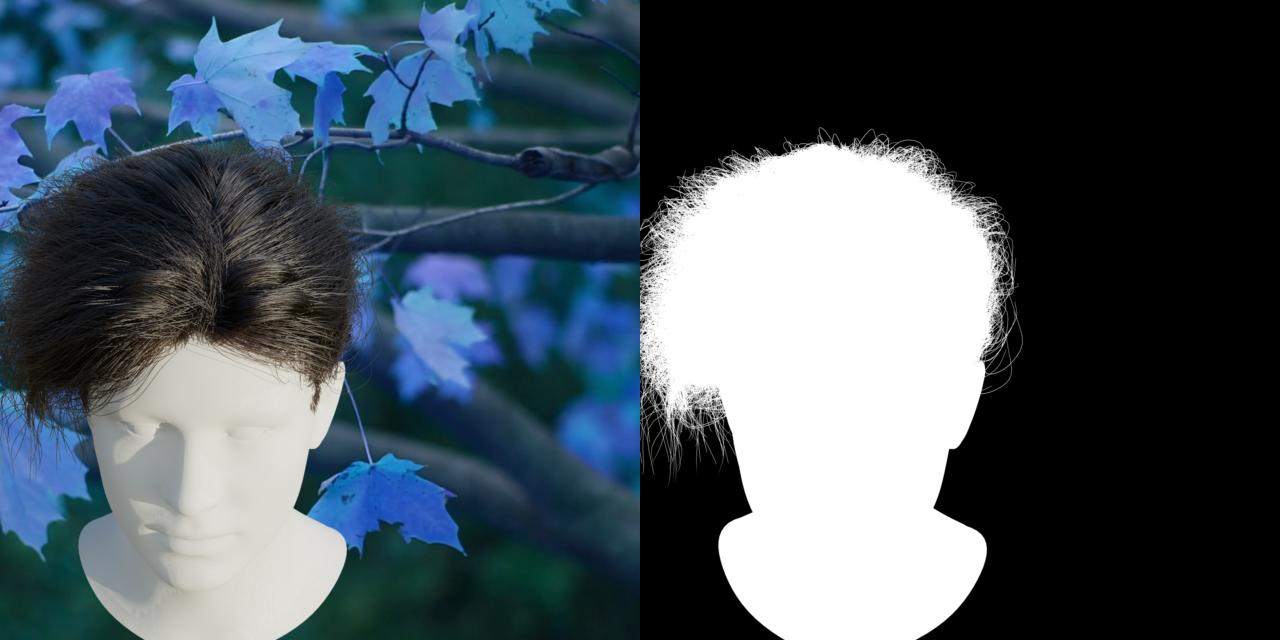} & 
\includegraphics[width=1.\columnwidth]{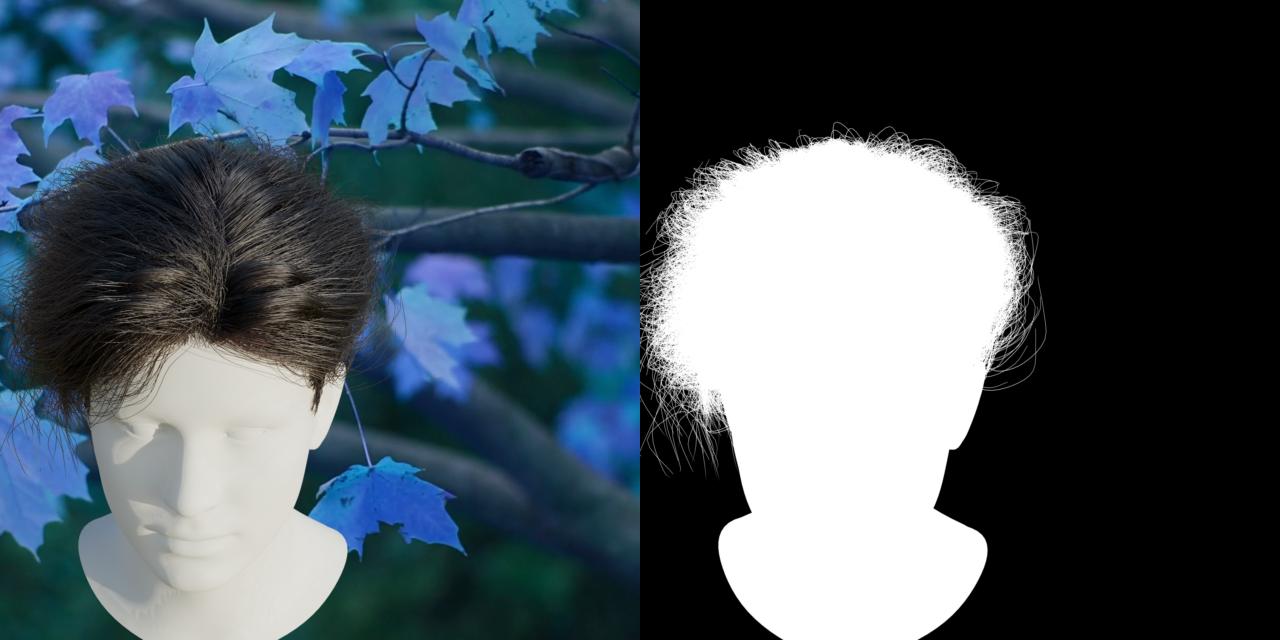} \\
Sample 0, Frame 2 & Sample 0, Frame 3  \\

\includegraphics[width=1.\columnwidth]{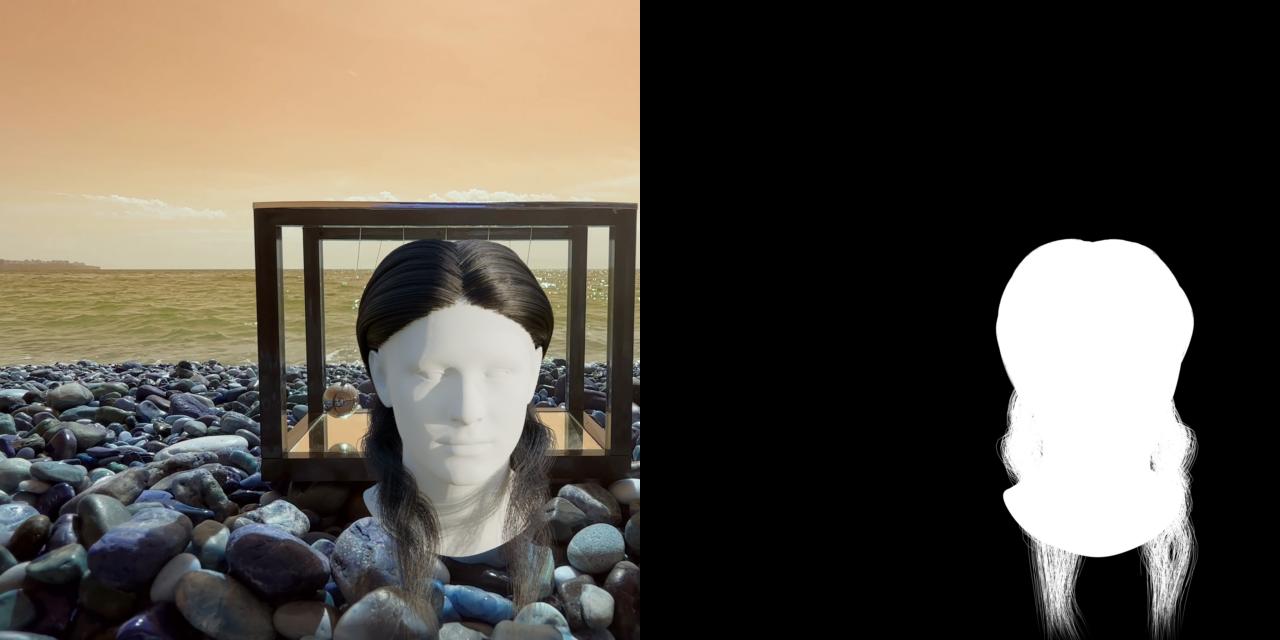} & 
\includegraphics[width=1.\columnwidth]{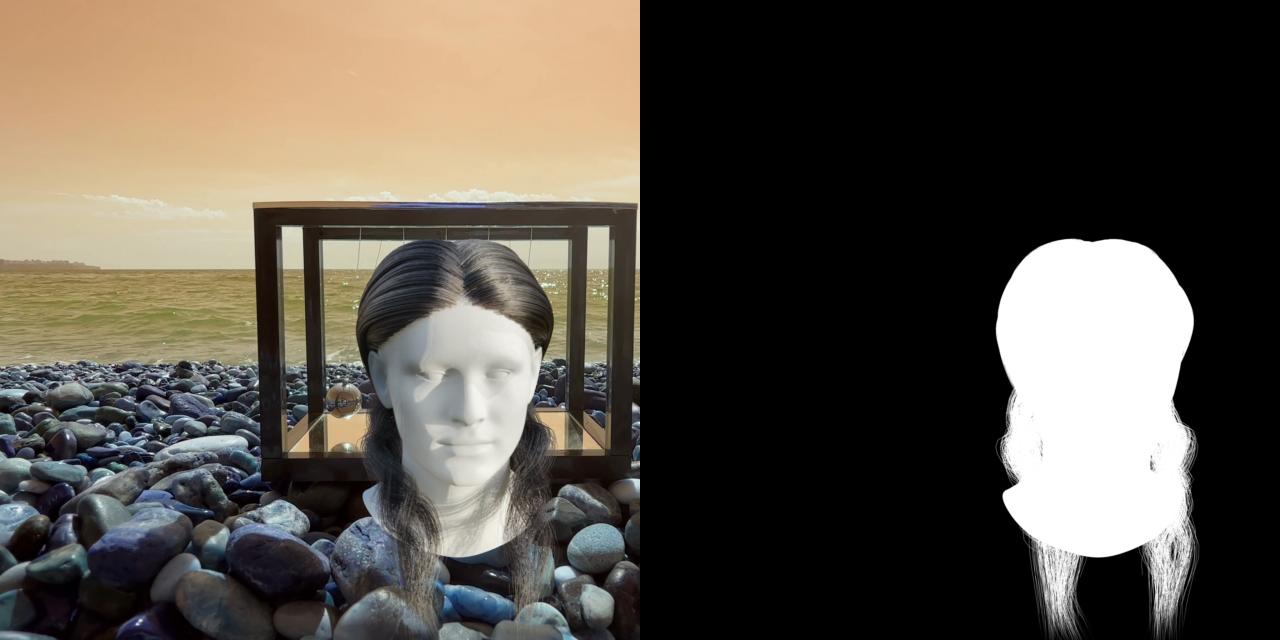}  \\
Sample 1, Frame 0 & Sample 1, Frame 1  \\
\includegraphics[width=1.\columnwidth]{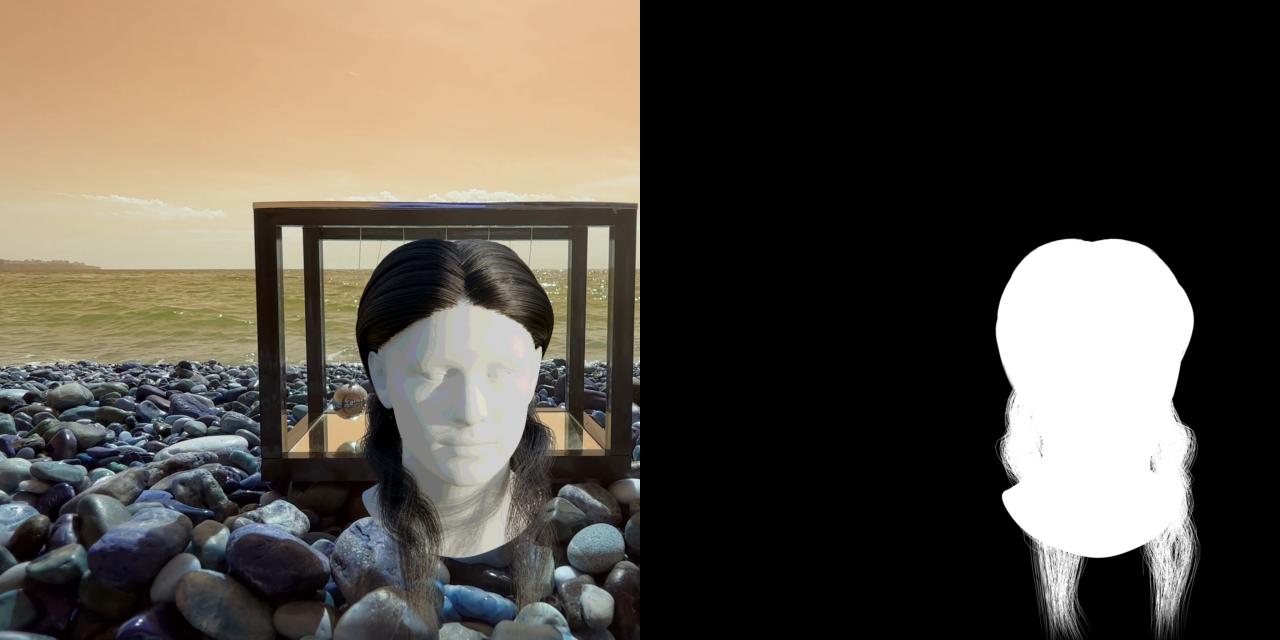} & 
\includegraphics[width=1.\columnwidth]{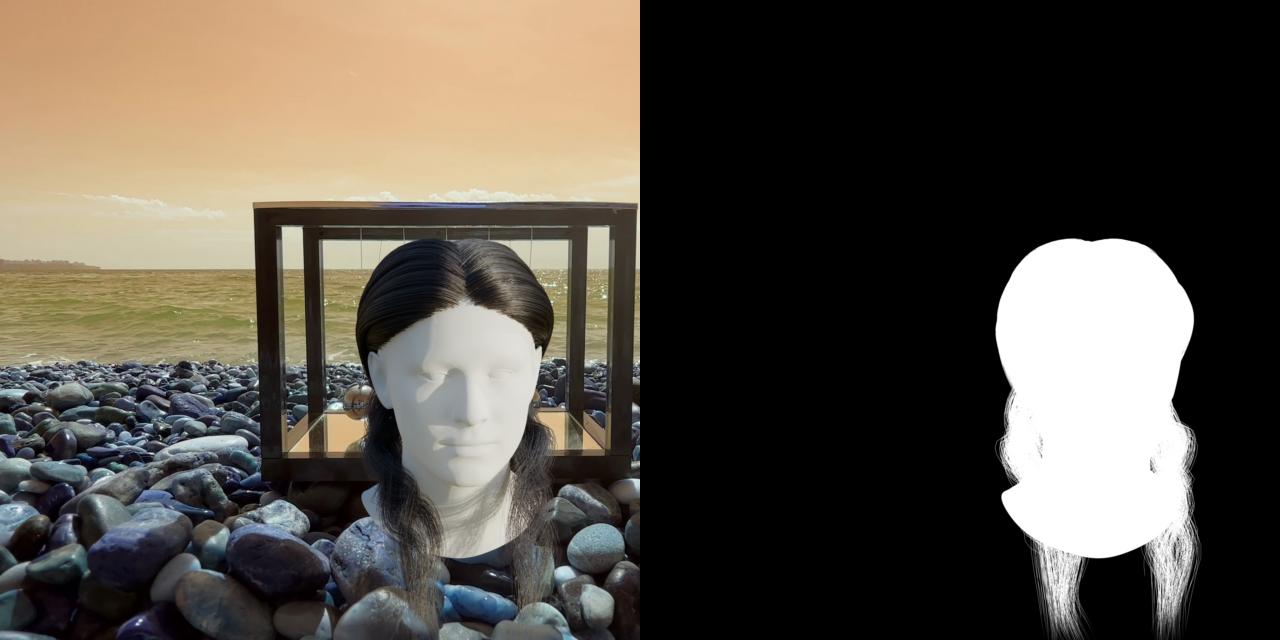} \\
Sample 1, Frame 2 & Sample 1, Frame 3  \\

\end{tabular}

\caption{Visualization of the rendered video matting dataset. We composite the rendered foreground with diverse background scenes, resulting in RGB frames paired with their corresponding alpha mattes. The face model is created by the FaceBuilder plugin.
}
\end{figure*}

\end{document}